\documentclass[runningheads]{llncs}

\PassOptionsToPackage{table,dvipsnames}{xcolor}
\usepackage{eccv}

\usepackage{eccvabbrv}

\usepackage{graphicx}
\usepackage{booktabs}

\usepackage[accsupp]{axessibility}

\usepackage{hyperref}

\usepackage{orcidlink}

\usepackage{booktabs}
\usepackage[utf8]{inputenc}
\usepackage{float}
\usepackage{afterpage}
\usepackage{amsmath}
\usepackage{color}
\usepackage{algorithm}
\usepackage{multicol}
\usepackage{makecell}
\usepackage{multirow}
\usepackage{array}
\usepackage{stfloats}
\usepackage{graphicx}
\usepackage{tikz}
\usepackage{enumitem}
\usepackage{fontawesome5}
\usepackage{manfnt}
\usepackage[font=small,skip=5pt]{caption}
\usepackage{algpseudocode}
\usepackage[export]{adjustbox}
\usepackage{soul}
\usepackage{xcolor}
\usepackage{ulem}

\usepackage[textsize=tiny]{todonotes}

\usepackage{amsfonts}

\makeatletter
\DeclareRobustCommand\onedot{\futurelet\@let@token\@onedot}
\def\@onedot{\ifx\@let@token.\else.\null\fi\xspace}

\makeatother

\usepackage{xcolor}
\definecolor{darkred}{rgb}{0.7,0.2,0.1}
\definecolor{darkgreen}{rgb}{0.2,0.7,0}
\definecolor{orange}{RGB}{255,127,0}
\definecolor{ourpurple}{RGB}{127,127,204}
\definecolor{ourteal}{RGB}{60,160,160}
\definecolor{palgreen}{RGB}{51,179,179}
\definecolor{magenta}{RGB}{199,21,133}
\definecolor{olive}{RGB}{100,150,85}
\definecolor{bestgreen}{RGB}{198,239,206}   
\definecolor{secondyellow}{RGB}{255,235,156} 

\newcommand{\ourmodel}{Axolotl3D}

\begin{document}

\title{{\ourmodel}: a Unified Framework \\
for Faithful 3D Shape Completion}

\titlerunning{{\ourmodel}}

\author{Anita Hu\inst{1}\orcidlink{0009-0004-2275-4580} \and
Maria Shugrina\inst{1}\orcidlink{0000-0002-7583-6772}}

\authorrunning{A.~Hu and M.~Shugrina}

\institute{NVIDIA \\ \url{https://research.nvidia.com/labs/sil/projects/axolotl3d}}

\addtocontents{toc}{\protect\setcounter{tocdepth}{-10}}
\let\oldaddcontentsline\addcontentsline
\def\addcontentsline#1#2#3{}
\maketitle
\def\addcontentsline#1#2#3{\oldaddcontentsline{#1}{#2}{#3}}

\begin{figure}[h!]
  \centering
  \includegraphics[width=0.9\linewidth]{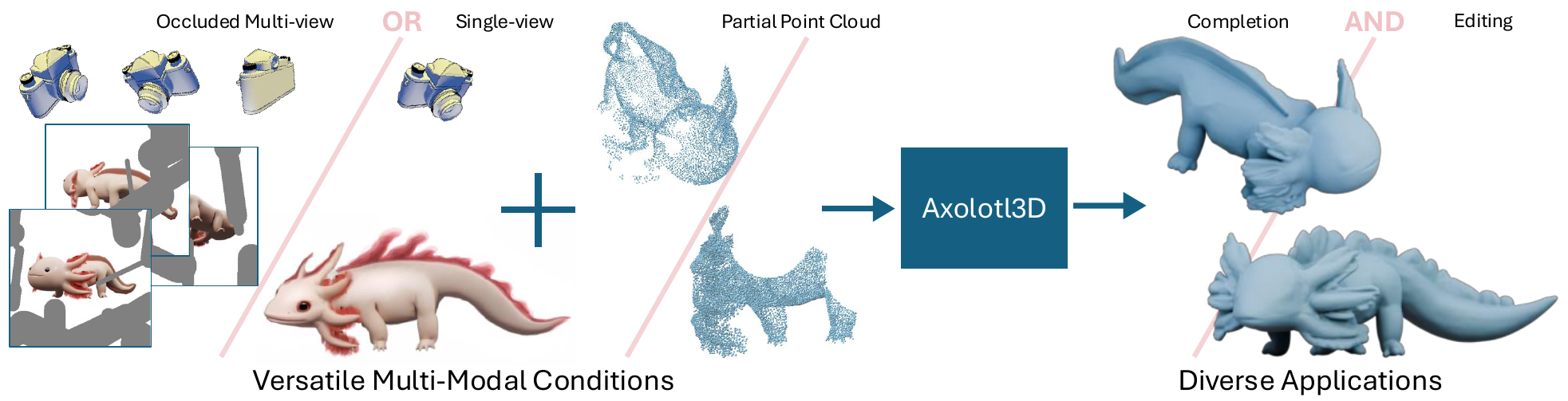}
  \caption{\textbf{{\ourmodel}} enables controllable, occlusion-aware 3D generation across diverse applications with consistent and faithful shape completion.}
  \label{fig:teaser}
\end{figure}

\begin{abstract}
Recent 3D generative models produce high-quality geometry from a single image using large-scale priors and diffusion architectures. However, they assume complete visibility and single-view inputs, limiting applicability in multi-view, occluded, or editing scenarios. Although prior works address these challenges individually, they lack a unified framework for controllable 3D completion under diverse conditioning signals.

We present {\ourmodel}, a multi-modal and occlusion-aware 3D generation model that jointly conditions on images, visibility masks, camera parameters, and a partial point cloud. The point cloud serves as a geometric anchor promoting faithful shape completion, while camera parameters ensure consistent multi-view alignment in a shared 3D coordinate system. A unified training strategy synthesizes diverse conditioning regimes from large-scale 3D data, enabling robust cross-modal reasoning.

Experiments on Toys4K and OmniObject3D demonstrate state-of-the-art performance under both clean and occluded settings, as well as strong results in real-world reconstruction and geometry-consistent editing.

\end{abstract}

\section{Introduction}
\label{sec:intro}

Recent advances in 3D generative modeling, leveraging large-scale 3D datasets and diffusion-based architectures (e.g., Hunyuan3D 2.1 \cite{hunyuan3d2025hunyuan3d}, Trellis \cite{xiang2025structured,xiang2025trellis2}), have enabled detailed and visually compelling geometry synthesis from a single image. These approaches demonstrate the power of learned 3D priors trained on large collections of synthetic and scanned objects. However, despite their strong generative capabilities, such methods remain limited in real-world scenarios, where objects are often captured under sparse multi-view setups, partially occluded, or require geometry-aware editing of existing meshes. By assuming complete visibility and single-view inputs, existing approaches are unable to reason under occlusion, integrate multiple views, or enforce geometric constraints, restricting their applicability to realistic capture and content creation pipelines.

Several research directions address these challenges in isolation. Works such as Amodal3R \cite{wu2025amodal3r} focus on recovering occluded geometry, while ReconViaGen \cite{chang2026reconviagen} emphasizes cross-view consistency. Editing methods including Instant3dit \cite{barda2025instant3dit}, PrEditor3D \cite{erkoc2025preditor3d} enable precise user-guided shape modification. More recent approaches, such as AmodalGen3D \cite{zhou2025amodalgen3d}, begin to combine multi-view reasoning with occlusion handling. While effective within their respective scopes, these approaches remain fragmented, limiting their applicability across broad conditioning and editing scenarios. We argue instead that these problems share a common underlying structure: each requires completing a 3D object from incomplete observations while enforcing consistency with explicit geometric constraints. From this perspective, they can be viewed as different manifestations of constrained 3D completion, motivating a unified and controllable generation framework.

We introduce {\ourmodel}, a multi-modal and occlusion-aware 3D generation model that conditions jointly on images, per-view visibility masks, camera parameters, and points. To promote geometric faithfulness, a partial point cloud---sampled from mesh surfaces or projected from predicted depths---acts as a geometric anchor. Camera parameters help align these 3D points and multi-view images within a shared coordinate system, allowing the model to integrate visual and geometric evidence consistently. To enable robust cross-modal reasoning, we further propose a unified training strategy that synthesizes diverse conditioning regimes from large-scale 3D mesh datasets. By simulating partial observations, occlusions, and mixed image–geometry inputs, we train the model to generalize across single-view, sparse multi-view, and editing scenarios.

We evaluate our approach on 3D object datasets including Toys4K \cite{stojanov2021toys4k} and OmniObject3D \cite{wu2023omniobject3d} under both single-view and multi-view settings, with and without synthetic occlusions. Across all scenarios, {\ourmodel} outperforms state-of-the-art methods in geometric accuracy and reconstruction fidelity. Furthermore, we demonstrate its applicability in real-world pipelines by reconstructing objects from images using predicted points and cameras from MapAnything \cite{keetha2026mapanything} and performing geometry-consistent object editing using views inpainted by Stable Diffusion Inpainting \cite{rombach2022stablediffusion}. 

In summary, our contributions are threefold:
\begin{itemize}
\item We introduce a unified multi-modal, occlusion-aware 3D generation framework that supports single-view generation, sparse multi-view reconstruction, occlusion completion, and geometry-aware editing within a single model.
\item We propose a cross-modal training strategy that simulates diverse conditioning scenarios from large-scale 3D data, enabling robust reasoning under partial and heterogeneous observations.
\item We achieve state-of-the-art performance on benchmark datasets, producing faithful geometry and effectively handling occlusions, while demonstrating effectiveness in real-world applications.
\end{itemize}

\section{Related Work}
\label{sec:related}

\subsubsection{Multi-View 3D Generation:}
While recent image-to-3D generative models \cite{hunyuan3d2025hunyuan3d,xiang2025structured,xiang2025trellis2} achieve impressive mesh quality from a single image, fewer works address sparse, unposed multi-view shape generation. Although single-view models can be adapted to multi-view inputs---for example, by cycling views during denoising as in Trellis \cite{xiang2025structured}---without jointly modeling cross-view interactions, geometric consistency remains limited. 
An alternative two-stage strategy first performs novel view synthesis (NVS) before 3D generation: EscherNet \cite{kong2024eschernet} produces consistent novel views from one or more reference images for subsequent reconstruction with NeuS \cite{wang2023neus}, while SpaRP \cite{xu2024sparp} outputs six fixed views for 3D generation with One-2-3-45++ \cite{liu2025one2345pp}. These pipelines can be brittle, as small inconsistencies in the synthesized views may compromise fidelity and detail. 
With the emergence of large-scale 3D foundation models \cite{wang2025vggt, wang2026pi3, keetha2026mapanything}, we can now directly predict explicit 3D attributes from images---such as camera parameters, depth maps, and point maps---providing strong priors for downstream tasks.
VGGT \cite{wang2025vggt} demonstrates impressive cross-view consistency in novel view synthesis, with its features leveraged by ReconViaGen \cite{chang2026reconviagen} for accurate multi-view generation.
Motivated by this progress, we condition our generative model on explicit geometric signals---point clouds and camera parameters---allowing flexible application to posed and unposed multi-view scenarios. A concurrent work ShapeR \cite{siddiqui2026shaper} also generates 3D shapes from multi-view images, sparse point clouds, and camera poses, but is tailored to SLAM \cite{engel2018slam} outputs.

\subsubsection{Geometry-Controlled Generation:}
Some recent 3D generation methods incorporate geometric cues to guide the generated shape.
3DShape2VecSet \cite{zhang20233dshape2vecset} introduces a fixed-length latent representation (VecSet) for shapes and demonstrates applications including 3D generation from partial point clouds. 
CLAY \cite{zhang2024clay} supports diverse conditioning modalities, including sparse points, voxels, bounding boxes, and multi-view images, but it is not designed to handle partial point clouds or occluded views. PoseMaster \cite{yan2025posemaster} generates articulated 3D characters conditioned on skeletons and images. 
More recently, Hunyuan3D-Omni \cite{hunyuan3d2025hunyuan3domni} unifies geometry-controlled generation across point clouds, bounding boxes, voxels, and skeleton inputs, though it only accepts a single image and cannot robustly handle incomplete observations. In contrast, our method handles both multi-view images and point cloud inputs, supporting occlusions in both modalities.

\subsubsection{Amodal Object Completion:}  
Amodal object completion aims to recover the full 3D shape from partial observations, typically images and segmentation masks. 
A related line of work focuses on single-image generation, producing complete 3D scenes and objects from monocular inputs. Methods such as CAST \cite{yao2025cast} and SAM 3D \cite{chen2025sam3d} generate object-level reconstructions and align them within the scene, using monocular depth estimators for partial point clouds and orientation cues. MIDI \cite{huang2025midi} generates all objects in a scene jointly, allowing the model to capture inter-object relationships and maintain spatial coherence without requiring a separate layout optimization step. At the object level, Amodal3R \cite{wu2025amodal3r} reconstructs occluded geometry from a single image by leveraging occlusion and visibility masks, using mask-weighted multi-head cross-attention and occlusion-aware conditioning layers. 
Occlusion-aware view synthesis methods, such as EscherNet++ \cite{zhang2025eschernetpp} and DeOcc-1-to-3 \cite{qu2025deocc}, generate novel views from occluded inputs to facilitate amodal completion. However, the reconstructed geometry often lacks fine details, revealing limitations of NVS-based pipelines.
A recent work, AmodalGen3D \cite{zhou2025amodalgen3d}, extends Amodal3R \cite{wu2025amodal3r} to unposed multi-view input and incorporating partial points from $\pi^3$ \cite{wang2026pi3}, producing more geometrically faithful reconstructions. In comparison, our method targets the posed setting, yet show strong results in an unposed ablation.

\subsubsection{Editing:}
Methods perform 3D editing using different strategies, from multi-view inpainting to latent-space mesh manipulation. Instant3dit \cite{barda2025instant3dit} and PrEditor3D \cite{erkoc2025preditor3d} perform text-guided edits through multi-view image inpainting from four fixed viewpoints, followed by 3D reconstruction. ObjFiller-3D \cite{feng2025objfiller3d} extends this paradigm using video diffusion, removing the restriction on viewpoints and improving cross-view consistency.
More recently, concurrent work VecSet-Edit \cite{hsiao2026vecsetedit} enables localized mesh editing guided by a single image directly in the VecSet latent space, better preserving appearance in untouched regions.
While effective for targeted edits, these methods cannot handle general object completion without an initial object or guidance signal.

\section{Method}
\label{sec:method}

We consider the task of 3D object completion from partial and heterogeneous observations, common in scenarios like 3D capture or editing. For multi-view reconstruction, the known information typically contains multiple occluded views and partial geometry, e.g.\ after processing an image sequence with techniques like \cite{wang2025vggt,kerbl3Dgaussians,tsang2026artisangs}.
For targeted shape editing, completion should respect fixed geometry regions and take additional guidance, e.g.\ an edited view of the object. To support such applications, our goal is to predict a complete and geometrically consistent 3D shape given: a collection of images $\{I_i\}_{i=1}^N$, corresponding masks $\{M_i\}_{i=1}^N$ and camera parameters $\{C_i\}_{i=1}^N$, and a partial point cloud $P$. Each mask $M_i$ specifies regions of $I_i$ that provide valid observations. For amodal completion, $M_i$ encodes visibility, whereas for object editing, it indicates regions to preserve while users provide modified views with fully valid $M_i$ to drive geometry updates. By treating all visual, geometric, and user-guided signals as partial observations of the object, our formulation unifies single- and multi-view generation, amodal completion, and geometry-aware editing under a single framework. 

\subsection{Model Architecture}\label{ssec:model}
\begin{figure}[tb]
  \centering
  \includegraphics[width=0.99\linewidth]{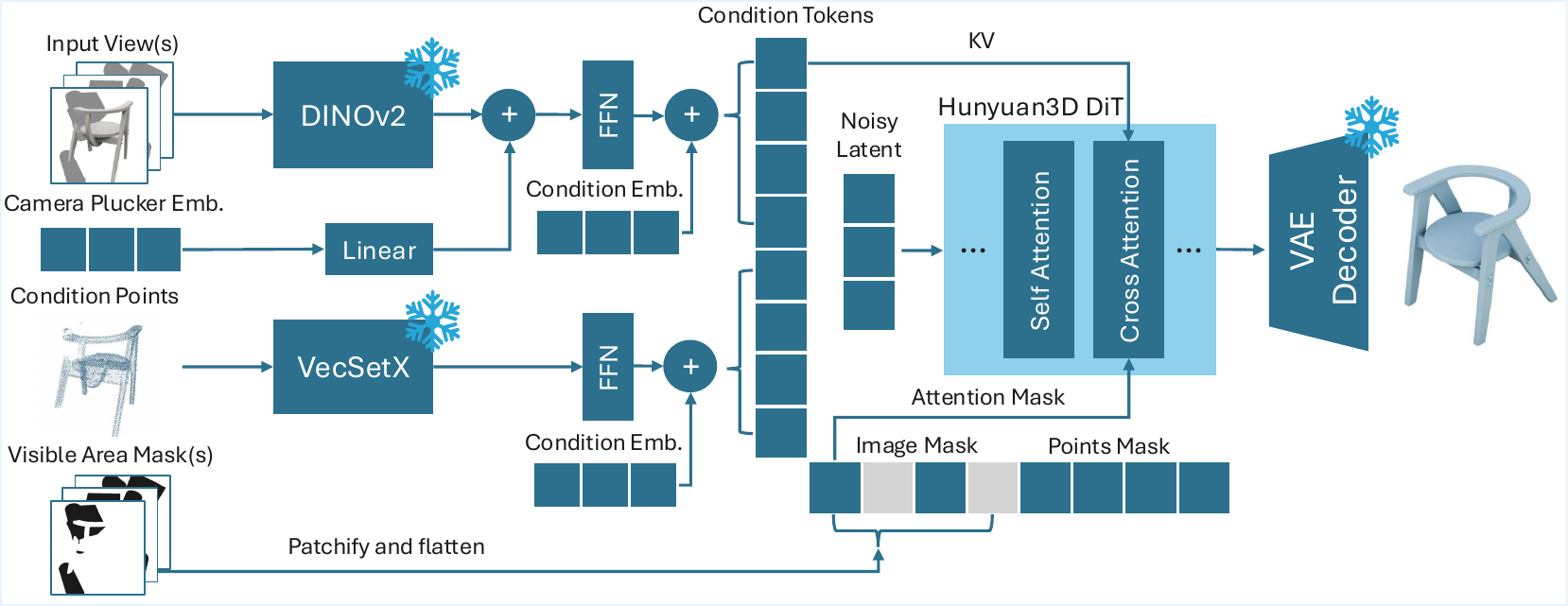}
  \caption{\textbf{Overview of {\ourmodel}.} Given multi-modal inputs---posed images, visibility masks, and partial points---our method encodes each modality and fuses them into multi-modal condition tokens. We fine-tune Hunyuan3D-DiT on these tokens to produce completed shape latents, which are decoded by the ShapeVAE.}
  \label{fig:method}
\end{figure}

{\ourmodel} is built upon recent state-of-the-art image-to-3D diffusion model, Hunyuan3D 2.1 \cite{hunyuan3d2025hunyuan3d}. Its shape generation model comprises a shape autoencoder (Hunyuan3D-ShapeVAE) that encodes meshes into a compact latent representation, and an image-conditioned flow-based diffusion transformer (Hunyuan3D-DiT) trained to predict shape tokens in this latent space, which are then decoded into meshes. We adapt this DiT backbone to jointly process all heterogeneous inputs, replacing the original single-image cross-attention with a multi-modal cross-attention mechanism that integrates visual and geometric cues simultaneously. An overview of the proposed architecture is shown in Fig.~\ref{fig:method}.

\subsubsection{Visual Encoding:} Following original architecture \cite{hunyuan3d2025hunyuan3d}, 
each input image $I_i$ is encoded into $\mathtt{enc}(I_i)$ using DINOv2 \cite{oquab2023dinov2}, resulting in rich spatially-aligned visual features suitable for multi-view reasoning. Our model takes up to $N=6$ images, where missing views are zero-padded and masked. To enable cross-view alignment of visual features in a shared 3D coordinate system, our model also ingests corresponding camera parameters $C_i$, represented as $\mathtt{pl}(C_i)$ using Pl\"ucker embeddings \cite{sitzmann2021plucker}.
Specifically, for each pixel $(u,v)$, its Pl\"ucker embedding is:
\begin{equation}
\mathbf{p}_{u,v} = \left( \mathbf{o} \times \mathbf{d}_{u,v}, \; \mathbf{d}_{u,v} \right) \in \mathbb{R}^6,
\end{equation}
where $\mathbf{o}$ is the camera center and $\mathbf{d}_{u,v}$ is the corresponding ray direction from camera center to the pixel.
These $\mathtt{pl}(C_i)$ are first average-pooled to match the spatial resolution of the image features $\mathtt{enc}(I_i)$, and then projected via a linear layer to match their channel dimension, resulting in $\mathtt{enc}(C_i)$. We
obtain final visual embedding $\mathtt{enc}_V(I_i, C_i)$ by adding $\mathtt{enc}(C_i)$ to $\mathtt{enc}(I_i)$, enriching image features with camera information. Each mask $M_i$ is partitioned into non-overlapping $14 \times 14$ patches (patch size of DINOv2), where each patch corresponds to a pixel in the downscaled mask $B_i$ matching the resolution of $\mathtt{enc}(I_i)$. A pixel in $B_i$ is set to 1 if it is completely unoccluded and 0 otherwise. Note that the background is treated as unoccluded to enforce free-space constraints. These masks $B_i$ bias the attention toward unoccluded regions (see later), similar to Amodal3R \cite{wu2025amodal3r}.

\subsubsection{Geometry Encoding:} To provide additional geometric grounding and ensure faithful completions, partial point cloud $P$ forms an additional conditioning input. We encode $P$ into $\mathtt{enc}(P)$ using VecSetX \cite{github2025vecsetx}, a follow-up to \cite{zhang20233dshape2vecset}, which maps input points into a compact, fixed-length set of latent codes. Pretrained on ShapeNet-v2 \cite{chang2015shapenet}, it yields high fidelity reconstruction retaining local details. 

\subsubsection{Multi-Modal Fusion:} We now process geometric features $\mathtt{enc}(P)$ and camera-enriched multi-view features $\{\mathtt{enc}_V(I_i, C_i)\}_{i=1}^N$ into multi-modal feature tokens $F$. Each modality's features are flattened into point tokens $F_p \in \mathbb{R}^{T_p \times C_p}$
and combined visual tokens $F_v \in \mathbb{R}^{N T_v \times C_v}$, where $N$ is number of views (6), $T_v$ and $T_p$ are the token lengths and $C_v$ and $C_p$ are the channel dimensions of each modality.
Then, each token in $F_p$ and $F_v$ is processed with its respective per-modality gated feedforward network (FFN) of similar design used in \cite{chen2025sam3d,grattafiori2024llama3} (see Supplement), mapping tokens to a unified channel dimension $C=1024$.
This produces richer, modality-specific features that emphasize important signals and prepare tokens for multi-modal fusion.
Two learnable embeddings, one per modality (visual and geometric) of shape $C$ are added element-wise to each token of the corresponding modality to differentiate between visual and geometric inputs, resulting in final tokens $F'_p \in \mathbb{R}^{T_p \times C}$ and $F'_v \in \mathbb{R}^{N T_v \times C}$.

\subsubsection{Mask-biased Cross-Attention:}
We concatenate $F'_p$ and $F'_v$ along the token dimension, forming a set of multi-modal condition tokens $F$ of length $T = T_p + NT_v$ and channel dimension $C$, which are used in the multi-head cross-attention modules of our DiT, trained to denoise the shape latent $Z \in \mathbb{R}^{L \times C'}$ across timesteps (See \cite{ho2020ddpm} for background). 
Let $Q = Z W_Q$, $K = F W_K$, $V = F W_V$ be the query, key, and value projections of the latent and condition tokens.  
For multi-head cross-attention with $H$ heads, the attention is computed as:
\begin{equation}
\mathrm{CrossAttn} = \mathrm{Softmax}\Big( \frac{Q K^\top}{\sqrt{D}} + B \Big) V,
\end{equation}
where each head has dimension $D = C' / H$, 
and $B \in \mathbb{R}^{L \times T}$ is an attention bias derived from the patchified image masks $B_i$.
Specifically, entries corresponding to occluded visual tokens are set to $-\infty$, preventing them from contributing to attention, while unoccluded tokens are assigned $0$ (unbiased). Point tokens, which are unmasked due to their fixed-length representation, are also set to $0$.

\subsection{Synthetic Data Augmentation}\label{ssec:augment}

We train our model on a large collection of textured 3D meshes, synthesizing diverse conditioning regimes for robust cross-modal reasoning. During dataset preprocessing, we normalize each mesh $\mathcal{M}_k$ to a unit cube $[-1,1]^3$ and sample a dense surface point cloud $P^k_\text{surface}$ (see Supplement). We also sample 150 cameras $C^k$  on a sphere and render corresponding images $I^k$ following \cite{xiang2025structured}. During training, we sample and augment this data to produce input partial point cloud $P$ and multi-view $\{I_i\}_{i=1}^N$, $\{M_i\}_{i=1}^N$, $\{C_i\}_{i=1}^N$, $N=6$, according to the following scenarios chosen with equal probability (if desired, this can be adjusted to prioritize specific regimes). See Fig.~\ref{fig:data_aug} for an example of each augmentation.

\subsubsection{Sparse Views \& Occlusions:}
To support amodal completion and promote learning of cross-modal correspondences, we simulate sparse view observation settings with varying levels of occlusion. 
We select one random view $I_0 \in I^k$ as the unobserved view, and sample up to $N$ conditioning cameras $C_i \in C^k$ from non-nearby views defined based on the angle between camera backward vectors (with $90^\circ$ angle threshold). 
Each sampled view $I_i$ is occluded with probability $p_{\text{occl}}$. 
When occlusion occurs, we generate a random mask $\tilde{M}_i$ using brush strokes or bounding boxes following LaMa \cite{suvorov2022lama}. 
We then remove points in $P^k_\text{surface}$ that are not visible in the sampled views or project into regions covered by occlusion masks (see Supplement for details), resulting in realistic patchy point clouds.
In addition, we further introduce a masking probability $p_{\text{mask}}$, to avoid over-reliance on point-based cues and promote complementary multi-modal learning. 
With this probability, the same occlusion mask $\tilde{M}_i$ is applied to the multi-view inputs $I_i$ and their masks $M_i$.
Using this procedure, we can further divide this task into single- and multi-view settings, sampled equally during training. 
We set $p_\text{mask}=0.5$ for both settings. In the single-view setting, where many points are missing due to limited visibility, we use $p_\text{occl}=0.25$, while in the multi-view setting we set $p_\text{occl}=1.0$.

\newcommand{\qcwidth}{0.09\linewidth}
\newcommand{\tmpwidth}{0.23\linewidth}

\begingroup
\setlength{\tabcolsep}{1pt}
    
\renewcommand{\arraystretch}{0}

\begin{figure*}[tb]
	\centering

\resizebox{\linewidth}{!} {
\begin{tabular}{@{}c|ccc|cc|c@{}}

\scalebox{0.5}{Surface Points} &
\multicolumn{3}{c|}{\scalebox{0.5}{Sparse Views \& Occlusions}} &
\multicolumn{2}{c|}{\scalebox{0.5}{Large Area Point Dropout}} & 
\scalebox{0.5}{Editing} \\ 

\includegraphics[width=\qcwidth, valign=m]{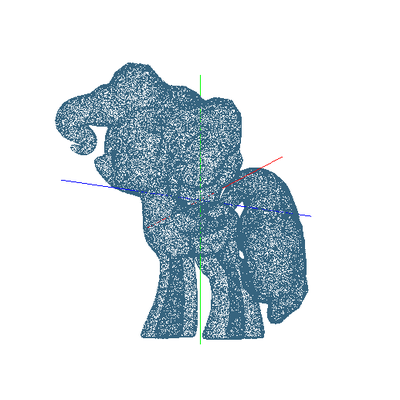} & 
\includegraphics[width=\qcwidth, valign=m]{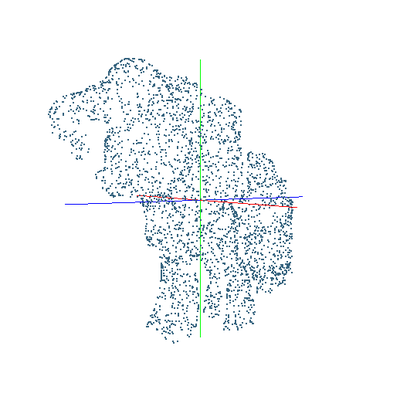} & 
\includegraphics[width=\qcwidth, valign=m]{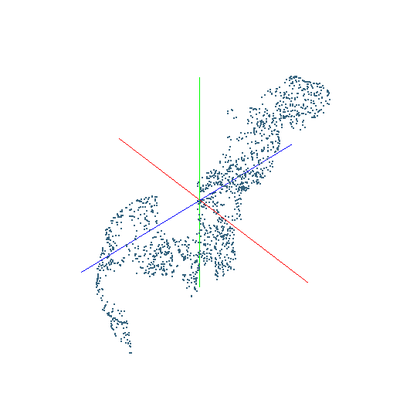} & 
\includegraphics[width=\qcwidth, valign=m]{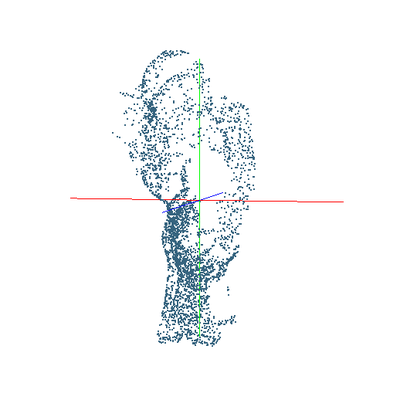} & 
\includegraphics[width=\qcwidth, valign=m]{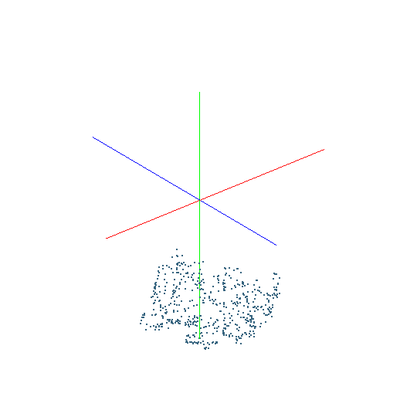} & 
\includegraphics[width=\qcwidth, valign=m]{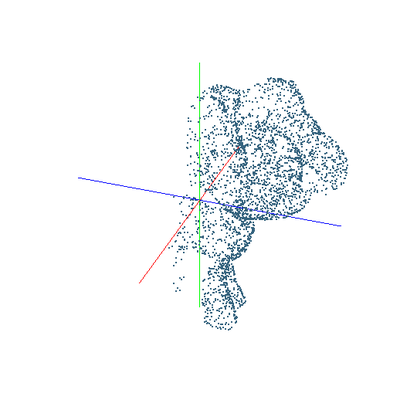} & 
\includegraphics[width=\qcwidth, valign=m]{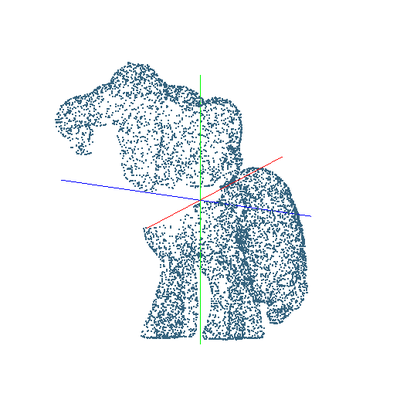} \\

\scalebox{0.5}{Original Image} &
\multicolumn{3}{c|}{\scalebox{0.5}{Condition Images}} & 
\multicolumn{2}{c|}{\scalebox{0.5}{Condition Images}} &
\scalebox{0.5}{Condition Images} \\ 

\includegraphics[width=\qcwidth, valign=m]{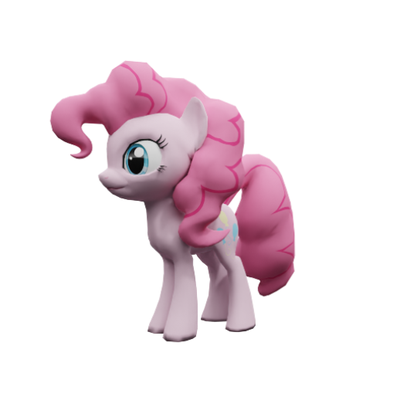} & 
\includegraphics[width=\qcwidth, valign=m]{img/data_aug_single_view_0_input_view.png} & 
\includegraphics[width=\qcwidth, valign=m]{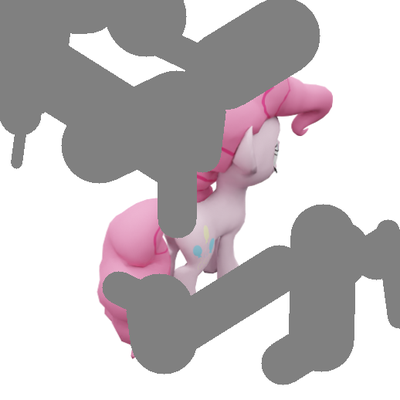} & 
\includegraphics[width=\tmpwidth, valign=m]{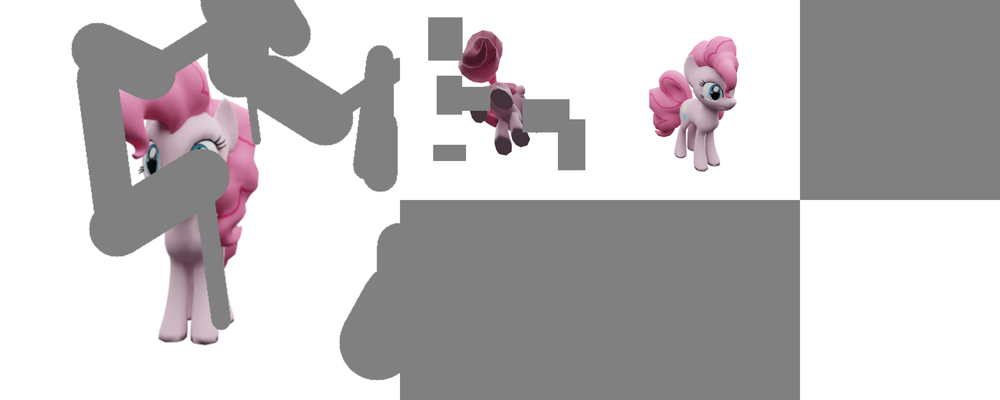} & 
\includegraphics[width=\qcwidth, valign=m]{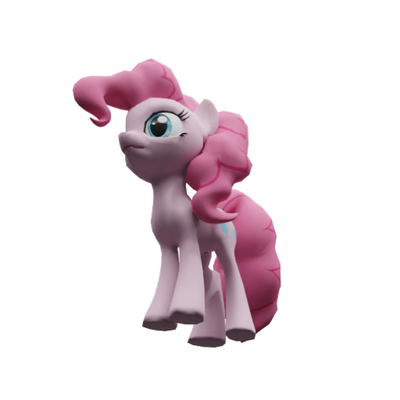} & 
\includegraphics[width=\tmpwidth, valign=m]{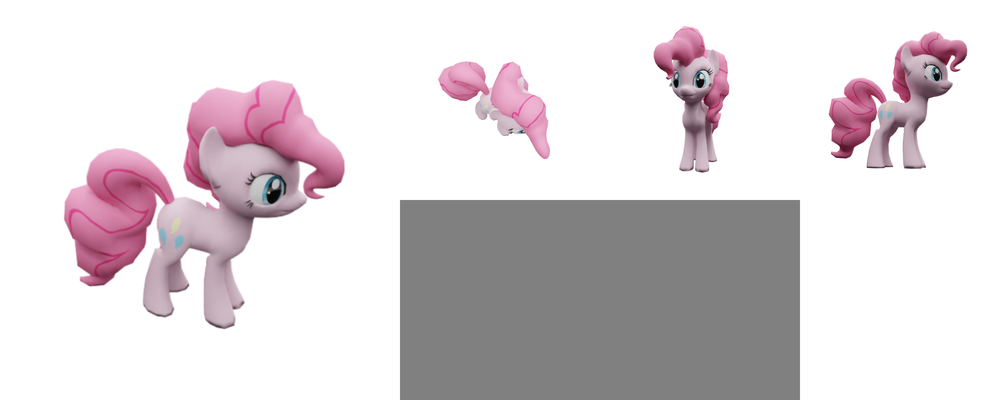} & 
\includegraphics[width=\tmpwidth, valign=m]{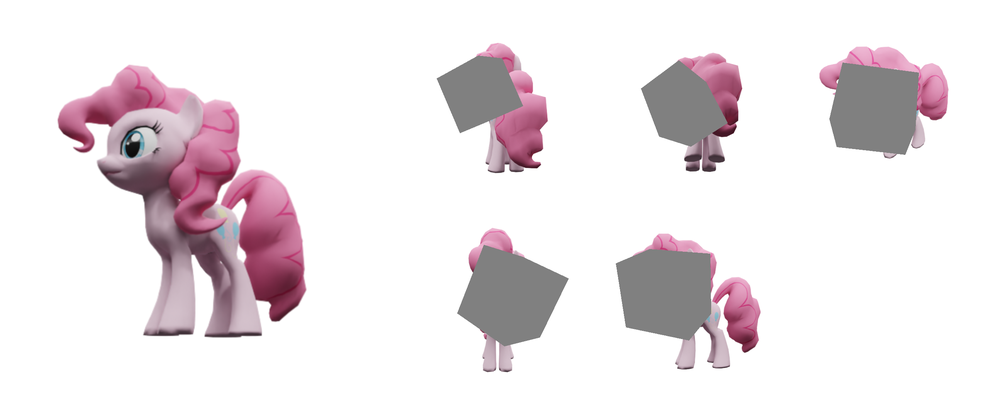} \\

  \end{tabular}
}
	\caption{\textbf{Data Augmentation}: Examples of our on-the-fly data augmentation.}
	\label{fig:data_aug}
\end{figure*}

\endgroup

\subsubsection{Large Area Point Dropout:}
This augmentation simulates scenarios where large regions of the object are missing and unobserved from camera viewpoints, requiring the model to infer missing geometry from limited points and indirect image observations. This provides more extreme point dropout compared to the previous, training the model to not overrely on available points.
First, we randomly select a dropout axis ($x$, $y$, or $z$) and a direction (positive or negative), then sample a threshold along this axis. 
Points in $P^k_\text{surface}$ beyond the threshold in the chosen direction are removed, with the drop probability decreasing linearly with distance from the threshold, creating a smooth gradient of missing geometry and more natural partial observations.
Afterwards, points are further filtered based on the alignment of their normals with the dropout direction mimicking realistic visibility effects, with the drop probability proportional to the cosine similarity between the normal and the downward direction.
To prevent the missing region from being directly observed, which is more realistic in real-world sparse-view scenarios, we filter out camera views whose backward vectors are within $90^\circ$ of the dropout direction. The final set of conditioning views are then sampled from the remaining cameras. Points not visible in any conditioning camera view are removed using their projected locations and depth comparisons. 

\subsubsection{Editing:}
In local editing scenarios, a bounding box or other simple geometric primitives are often used to specify the region to be modified, as in the 3D interface of Instant3dit \cite{barda2025instant3dit},
and edits are typically guided by a reference image or a text prompt \cite{barda2025instant3dit,erkoc2025preditor3d,hsiao2026vecsetedit}. In this work, we focus on image-based guidance, where additional views may be provided to help preserve fine details in the unedited regions and maintain cross-view consistency.
We emulate such user-guided shape modification by selecting a random camera view as the "modified" view. We then place a 3D bounding box centered on a randomly chosen point in $P^k_\text{surface}$ that is visible in this view. All points inside the bounding box are removed from $P^k_\text{surface}$, creating a region for inpainting.
Additional camera views are then randomly sampled to provide observations of the unmodified regions, with the edited region masked in all $I_i$, $M_i$ by projecting the bounding box using camera $C_i$ (see Supplement for details). 
This creates a task where only the modified view shows the full object with the edit, while the remaining views reveal only the surrounding geometry outside the edited region.

\section{Experiments}
\label{sec:eval}

\subsection{Data and Setting}

\subsubsection{Training Data:} 
We train on 407k shapes from TRELLIS-500K \cite{xiang2025structured}, containing aesthetics-filtered objects from ObjaverseXL \cite{deitke2023objaverseXL}, ABO \cite{collins2022abo}, 3D-FUTURE \cite{fu20213dfuture}, and HSSD \cite{khanna2024hssd}. We remove buildings and scenes from the ObjaverseXL subset using a classifier trained on G-Objaverse labels \cite{zuo2024gobjaverse}, as these often depict multi-object scenes (see Supplement). Shapes are scaled and rendered with turn-around cameras at fixed distance and intrinsics \cite{xiang2025structured}. The geometry is preprocessed following Hunyuan3D 2.1 \cite{hunyuan3d2025hunyuan3d}, including normalization, watertight processing, SDF and surface sampling. Train-time augmentations are described in \S\ref{ssec:augment}.

\subsubsection{Implementation:} For our backbone we adopt Hunyuan3D 2.1 \cite{hunyuan3d2025hunyuan3d} and finetune from pre-trained weights utilizing their strong 3D generation priors. 
Additional layers and embeddings in \S\ref{ssec:model} are trained from scratch: linear layers use Kaiming uniform initialization \cite{he2015delving}, while embeddings use normal initalization $\mathcal{N}(0, 1/\sqrt{n})$ where $n$ is the condition embedding dimension (1024).
We use classifier-free guidance (CFG) \cite{ho2021cfg} with drop rate of 0.1 and AdamW \cite{loshchilov2018adamw} optimizer with learning rate of $1e-5$. The model is trained for 500k steps with batch size of 16 on 8 A100 GPUs.
At inference, the CFG guidance scale is set to 5 and sampling steps is 50. 

\subsubsection{Evaluation Data:} For quantitative evaluation (\S\ref{ssec:quant}), we use Toys4K~\cite{stojanov2021toys4k} and OmniObject3D~\cite{wu2023omniobject3d}. Toys4K contains 4k synthetic objects in 105 categories, and OmniObject3D has 6k scanned objects in 190 categories. To simulate realistic 3D-consistent occlusions, we add random occluders from a set of shape primatives (see Supplement). Different from training, we obtain point condition by backprojection from rendered depth maps of each sampled view, excluding occluded regions. This is downsampled to 8,192 points for ShapeR and our method while Hy3D-Omni supports a max of 2,048 points and SAM3D uses pointmaps of shape 518x518x3. We further provide comparisons using 2,048 input points and evaluate robustness using predicted depth from Depth Anything v3~\cite{lin2026depthanything}, both with and without camera perturbations. Predicted depth is aligned to ground-truth depth via per-scene scale and shift, then backprojected using ground-truth camera parameters. Camera extrinsics are perturbed with Gaussian noise (10° rotation, 10\% translation scale), while the focal length is jittered by ±5\%. For qualitative visualization (\S\ref{ssec:qual}), we focus on Toys4K’s geometrically challenging objects with richer appearance details (OmniObject3D results in Supplement). For applications (\S\ref{ssec:applications}), we show single-view generation with MapAnything \cite{keetha2026mapanything} using MVImgNet images \cite{yu2023mvimgnet}, and editing results on Toys4K objects.

\subsection{Baselines and Metrics}\label{ssec:eval:baselines}

\subsubsection{Baselines:} We compare our method to the recent state-of-the-art methods \cite{wu2025amodal3r}, \cite{hunyuan3d2025hunyuan3domni}, \cite{chen2025sam3d},  and to one concurrent work \cite{siddiqui2026shaper}. We omit AmodalGen3D \cite{zhou2025amodalgen3d} as its code is not publicly available. Amodal3R \cite{wu2025amodal3r} is capable of single- and multi-view completion and is the only method that does not accept point condition. Hunyuan3D-Omni (Hy3D-Omni) \cite{hunyuan3d2025hunyuan3domni} and SAM 3D \cite{chen2025sam3d} are both single-view methods that accept point condition, where SAM 3D pointmaps are in camera space. Concurrent ShapeR \cite{siddiqui2026shaper} is the only multi-view method that also accepts points. 
Because Hy3D-Omni is not trained for occlusions, we complete partial views with Pix2Gestalt \cite{ozguroglu2024pix2gest} before inference. While it accepts only a single image, points from multiple views can still be used.
Thus, we evaluate all models, except SAM 3D, on multi-view and single-view conditioned generation. In the multi-view scenario, Hy3D-Omni is given the best view out of the available views as image input (selected based on max object coverage; see Supplement). 
Finally, to provide required prompts to ShapeR, we use captions included with TRELLIS-500K \cite{xiang2025structured} for Toys4k \cite{stojanov2021toys4k} and object category for OmniObject3D \cite{wu2023omniobject3d}.

\subsubsection{Metrics:}
To assess the fidelity and completeness of the reconstructed shapes, we choose complementary geometric metrics.
F-score (at 0.05 threshold) measures point-level correspondence, balancing precision and recall to assess accuracy and coverage. Voxel IoU (vIoU) at $64^3$ resolution evaluates coarse volumetric agreement, sensitive to major occupancy errors. Chamfer Distance (CD) captures fine-grained geometric deviations and penalizes missing or distorted regions.
Because not all methods use world-space points as input, we first align generated and groundtruth meshes, normalized to $[-1,1]$, using ICP \cite{besl1992method}.
1M points are uniformly sampled from both meshes before computing the metrics.

\begin{table*}[!tb]

\caption{\textbf{Quantitative Evaluation,} marking
\begingroup
\setlength{\fboxsep}{0pt}
\colorbox{bestgreen}{\strut best} and
\colorbox{secondyellow}{\strut second-best.}
\endgroup
Our method shows superior performance against state-of-the-art and performs comparably to ShapeR, with lower variance and stronger single-view results. 
}

\centering 
\label{tab:quant}

\begingroup
\setlength{\tabcolsep}{3pt} 

\definecolor{pmGray}{HTML}{6B6B6B}

\newcommand{\pmitem}[1]{%
    \textcolor{pmGray}{\scalebox{0.7}{$\pm$ #1}}%
}

\newcommand{\pmrow}[9]{%
\pmitem{#1} & 
\pmitem{#2} & 
\pmitem{#3} & 
\pmitem{#4} & 
\pmitem{#5} & 
\pmitem{#6} & 
\pmitem{#7} & 
\pmitem{#8} & 
\pmitem{#9} & 
\pmrowcontinued 
}

\newcommand{\pmrowcontinued}[3]{%
\pmitem{#1} & 
\pmitem{#2} & 
\pmitem{#3}
}

\resizebox{\linewidth}{!} {
\begin{tabular}{@{}l||lll|lll|lll|lll@{}}
\Xhline{1.5pt}
\multirow{3}{*}{\makecell[l]{\textbf{Scenario:} \\ Method}} & \multicolumn{6}{c|}{Toys4k} & \multicolumn{6}{c}{OmniObject3D}\\
\cline{2-13}
& \multicolumn{3}{c|}{Without Occlusion} & \multicolumn{3}{c|}{With Occlusion} & \multicolumn{3}{c|}{Without Occlusion} & \multicolumn{3}{c}{With Occlusion} \\ [1ex]

& F-score $\uparrow$ & vIoU $\uparrow$ & CD$\downarrow_{\times10}$ & F-score $\uparrow$ & vIoU $\uparrow$ & CD$\downarrow_{\times10}$ & F-score $\uparrow$ & vIoU $\uparrow$ & CD$\downarrow_{\times10}$ & F-score $\uparrow$ & vIoU $\uparrow$ & CD$\downarrow_{\times10}$ \\

 \Xhline{1.5pt}

\textbf{\textit{Single-View:}} & & & &&&&&&&&&\\ [1ex]

Amodal3R~\cite{wu2025amodal3r} &
0.7310 &
0.2129 &
0.5460  &
0.6877  &
0.1886  &
0.6144  &
0.4847  &
0.1459  &
1.1406  &
0.4724  &
0.1382  &
1.1174  \\ [-1ex]
& \pmrow{0.2365}{0.1139}{0.6305}{0.2381}{0.1063}{0.6495}{0.2646}{0.0897}{0.9197}{0.2587}{0.0845}{0.9336}\\ [1ex]

SAM 3D~\cite{chen2025sam3d} &
0.7160 &
0.2039  &
0.5672  &
0.6764  &
0.1836  &
0.6375  &
0.4808  &
0.1440  &
1.1984  &
0.4666  &
0.1407  &
1.2409  \\
& \pmrow{0.2491}{0.1049}{0.6464}{0.2504}{0.1018}{0.6871}{0.2715}{0.0932}{0.9989}{0.2709}{0.0932}{1.0098}\\ [1ex]

Hy3D-Omni~\cite{hunyuan3d2025hunyuan3domni} &
0.7811  &
0.2496  &
0.4491  &
0.6593  &
0.1802  &
0.8123  &
\cellcolor{secondyellow}0.7401  &
0.2007  &
0.4913  &
0.6028  &
0.1406  &
0.9018  \\
& \pmrow{0.2065}{0.1471}{0.4261}{0.2614}{0.1297}{0.8209}{0.1984}{0.1147}{0.4132}{0.2386}{0.1036}{0.7425}\\ [1ex]

ShapeR~\cite{siddiqui2026shaper} &
\cellcolor{secondyellow}0.7920 &
\cellcolor{secondyellow}0.2573  &
\cellcolor{secondyellow}0.3981  &
\cellcolor{secondyellow}0.7453  &
\cellcolor{secondyellow}0.2285  &
\cellcolor{secondyellow}0.4622  &
0.7135  &
\cellcolor{secondyellow}0.2160  &
\cellcolor{secondyellow}0.4820  &
\cellcolor{secondyellow}0.6816  &
\cellcolor{secondyellow}0.2012  &
\cellcolor{secondyellow}0.5338  \\
\scalebox{0.9}{(concurrent)} & \pmrow{0.2127}{0.1235}{0.4121}{0.2249}{0.1208}{0.4321}{0.2210}{0.1008}{0.3853}{0.2306}{0.0961}{0.4236}\\ [1ex]

\textbf{Ours} &
\cellcolor{bestgreen}0.9221 &
\cellcolor{bestgreen}0.3470  &
\cellcolor{bestgreen}0.2157  &
\cellcolor{bestgreen}0.9046  &
\cellcolor{bestgreen}0.3265  &
\cellcolor{bestgreen}0.2402  &
\cellcolor{bestgreen}0.8716  &
\cellcolor{bestgreen}0.2792  &
\cellcolor{bestgreen}0.2854  &
\cellcolor{bestgreen}0.8613  &
\cellcolor{bestgreen}0.2724  &
\cellcolor{bestgreen}0.2971  \\
& \pmrow{0.1044}{0.0988}{0.1820}{0.1140}{0.1037}{0.2079}{0.1187}{0.0856}{0.2269}{0.1191}{0.0848}{0.2228}\\ [1ex]

\Xhline{1.5pt}

\textbf{\textit{Multi-View (6):}} & & & &&&&&&&&&\\ [1ex]

Amodal3R~\cite{wu2025amodal3r} &
0.8134 &
0.2516  &
0.3935  &
0.7529  &
0.2166  &
0.4999  &
0.5250  &
0.1585  &
1.0733  &
0.5046  &
0.1484  &
1.1127  \\
& \pmrow{0.2012}{0.1236}{0.5012}{0.2198}{0.1154}{0.6021}{0.2741}{0.0968}{0.9534}{0.2667}{0.0905}{0.9520}\\ [1ex]

Hy3D-Omni~\cite{hunyuan3d2025hunyuan3domni} &
0.9211 &
0.3860  &
0.2111  &
0.8920  &
0.3355  &
0.2738  &
0.9549  &
0.3638  &
0.1677  &
0.9353  &
0.3314  &
0.2005  \\
& \pmrow{0.1138}{0.1587}{0.2108}{0.1482}{0.1581}{0.3667}{0.0797}{0.1397}{0.1580}{0.1043}{0.1377}{0.2317}\\ [1ex]

ShapeR~\cite{siddiqui2026shaper} &
\cellcolor{secondyellow}0.9433 &
\cellcolor{bestgreen}0.4476  &
\cellcolor{secondyellow}0.1649  &
\cellcolor{secondyellow}0.9190  &
\cellcolor{secondyellow}0.4002  &
\cellcolor{secondyellow}0.2040  &
\cellcolor{secondyellow}0.9629  &
\cellcolor{bestgreen}0.4412  &
\cellcolor{bestgreen}0.1329  &
\cellcolor{secondyellow}0.9532  &
\cellcolor{bestgreen}0.4123  &
\cellcolor{bestgreen}0.1443  \\
\scalebox{0.9}{(concurrent)} & \pmrow{0.1303}{0.1449}{0.3096}{0.1528}{0.1553}{0.3416}{0.0985}{0.1318}{0.2112}{0.0989}{0.1313}{0.1842}\\ [1ex]

\textbf{Ours} &
\cellcolor{bestgreen}0.9768 &
\cellcolor{secondyellow}0.4424  &
\cellcolor{bestgreen}0.1351  &
\cellcolor{bestgreen}0.9689  &
\cellcolor{bestgreen}0.4187  &
\cellcolor{bestgreen}0.1448  &
\cellcolor{bestgreen}0.9842  &
\cellcolor{secondyellow}0.4127  &
\cellcolor{secondyellow}0.1389  &
\cellcolor{bestgreen}0.9755  &
\cellcolor{secondyellow}0.3967  &
\cellcolor{secondyellow}0.1467  \\
& \pmrow{0.0445}{0.1077}{0.0543}{0.0534}{0.1121}{0.0685}{0.0425}{0.0903}{0.1240}{0.0473}{0.0889}{0.1030}\\ [1ex]

\Xhline{1.5pt}

\end{tabular}
}

\endgroup
\end{table*}

\begin{table*}[t]

\centering 

\definecolor{pmGray}{HTML}{6B6B6B}

\newcommand{\pmitem}[1]{%
    \textcolor{pmGray}{\scalebox{0.7}{$\pm$ #1}}%
}

\newcommand{\pmrow}[6]{%
\pmitem{#1} & 
\pmitem{#2} & 
\pmitem{#3} & 
\pmitem{#4} & 
\pmitem{#5} & 
\pmitem{#6} 
}

\newcommand{\skipcell}{\multicolumn{1}{c}{\multirow{2}{*}{N/A}}}

\begin{minipage}[t]{0.48\linewidth}
\centering
\captionof{table}{Quantitative results with single-view \textbf{predicted depth} on Toys4k.}
\label{tab:depth}
\resizebox{\linewidth}{!} {
\begin{tabular}{@{}l||lll|lll@{}}
\Xhline{1.0pt}
\multirow{2}{*}{Method} & \multicolumn{3}{c|}{w/o Camera Perturb} & \multicolumn{3}{c}{w/ Camera Perturb} \\

 & F-score $\uparrow$ & vIoU $\uparrow$ & CD$\downarrow_{\times10}$ & F-score $\uparrow$ & vIoU $\uparrow$ & CD$\downarrow_{\times10}$ \\

 \Xhline{1.0pt}

SAM3D &
\cellcolor{secondyellow}{0.7062} &
\cellcolor{secondyellow}{0.1993} &
\cellcolor{secondyellow}{0.5898} &
\skipcell &
\skipcell &
\skipcell \\
{} &
\pmitem{0.2538} &
\pmitem{0.1049} &
\pmitem{0.6643} \\

Hy3D-Omni &
{0.6837} &
{0.1729} &
{0.5926} &
\skipcell &
\skipcell &
\skipcell \\
{} &
\pmitem{0.2170} &
\pmitem{0.1249} &
\pmitem{0.4585} \\

ShapeR &
0.5626&
0.1407&
0.7767&
\cellcolor{secondyellow}0.5605&
\cellcolor{secondyellow}0.1409&
\cellcolor{secondyellow}0.7750\\
\scalebox{0.9}{(concurrent)} & \pmrow{0.2408}{0.0847}{0.5569}{0.2398}{0.0863}{0.5518}\\

\textbf{Ours} &
\cellcolor{bestgreen}0.7616&
\cellcolor{bestgreen}0.1903&
\cellcolor{bestgreen}0.4235&
\cellcolor{bestgreen}0.7620&
\cellcolor{bestgreen}0.1901&
\cellcolor{bestgreen}0.4230\\
& \pmrow{0.1731}{0.0908}{0.3135}{0.1712}{0.0908}{0.3138}\\

\Xhline{1.0pt}

\end{tabular}

}
\end{minipage}\hfill
\begin{minipage}[t]{0.48\linewidth}
\centering
\captionof{table}{Quantitative results with \textbf{2048} input points on Toys4k with occlusions.}
\label{tab:2kpts}
\resizebox{\linewidth}{!} {
\begin{tabular}{@{}l||lll|lll@{}}
\Xhline{1.0pt}
\multirow{2}{*}{Method} & \multicolumn{3}{c|}{Single-View} & \multicolumn{3}{c}{Multi-View} \\

 & F-score $\uparrow$ & vIoU $\uparrow$ & CD$\downarrow_{\times10}$ & F-score $\uparrow$ & vIoU $\uparrow$ & CD$\downarrow_{\times10}$ \\

 \Xhline{1.0pt}

Hy3D-Omni &
0.6594&
0.1802&
0.8119&
0.8920&
0.3356&
0.2738\\
& \pmrow{0.2614}{0.1297}{0.8208}{0.1482}{0.1581}{0.3667}\\

ShapeR &
\cellcolor{secondyellow}0.6956&
\cellcolor{secondyellow}0.2004&
\cellcolor{secondyellow}0.5377&
\cellcolor{secondyellow}0.9126&
\cellcolor{secondyellow}0.3734&
\cellcolor{secondyellow}0.2119\\
\scalebox{0.9}{(concurrent)} & \pmrow{0.2317}{0.1106}{0.4505}{0.1466}{0.1505}{0.3078}\\

\textbf{Ours} &
\cellcolor{bestgreen}0.8738&
\cellcolor{bestgreen}0.2766&
\cellcolor{bestgreen}0.2838&
\cellcolor{bestgreen}0.9653&
\cellcolor{bestgreen}0.3996&
\cellcolor{bestgreen}0.1521\\
& \pmrow{0.1291}{0.0958}{0.2303}{0.0568}{0.1103}{0.0696}\\

\Xhline{1.0pt}

\end{tabular}
}

\end{minipage}

\end{table*}

\subsection{Quantitative Results}\label{ssec:quant}
We present quantitative results in Tb.~\ref{tab:quant}. 
Under single-view conditions, our method significantly outperforms the baselines including concurrent work, ShapeR. In multi-view conditions, our method shows superior performance against state-of-the-art and comparable with ShapeR. In all scenarios, our method achieves lower standard deviation showing more consistent and stable performance. 

We evaluate robustness to realistic inputs using predicted monocular depth without occlusions in Tb.~\ref{tab:depth}. Our method achieves the best geometric fidelity across all metrics, outperforming the baselines. Compared to the ground-truth depth setting in Tb.~\ref{tab:quant}, the performance gap narrows under predicted depth, with our method, ShapeR, and Hy3D-Omni showing larger performance drops, reflecting the sensitivity of geometry-guided reconstruction to errors in monocular depth estimation. In contrast, SAM3D being trained directly on predicted depth exhibit smaller relative degradation, but remain less faithful to the input geometry thus achieving lower overall performance.
We also simulate camera perturbations in Tb.~\ref{tab:depth}, where our method and ShapeR are both resilient to moderate camera noise through use of Pl\"ucker embeddings. Other baselines are not evaluated as they do not condition on camera parameters.
We additionally report results using 2,048 input points for all methods in Tb.~\ref{tab:2kpts}. Our method shows stable performance under weaker geometric conditioning while ShapeR saw larger performance drop in single-view compared to Tb.~\ref{tab:quant}. 

\newcommand{\mvwidth}{0.23\linewidth}

\begin{figure*}[!t]
  \centering

  \begin{subfigure}{\linewidth}
    \centering

  \resizebox{0.95\linewidth}{!} {
    \begin{tabular}{@{}cc|ccccc|c@{}}

\scalebox{0.5}{Input View} &
\scalebox{0.5}{Input Points} & 
\scalebox{0.5}{Amodal3R} & 
\scalebox{0.5}{SAM3D} & 
\scalebox{0.5}{Hy3D-Omni} & 
\scalebox{0.5}{ShapeR} & 
\scalebox{0.5}{\textbf{Ours}} & 
\scalebox{0.5}{Ground Truth} \\ 

\includegraphics[width=\qcwidth, valign=m]{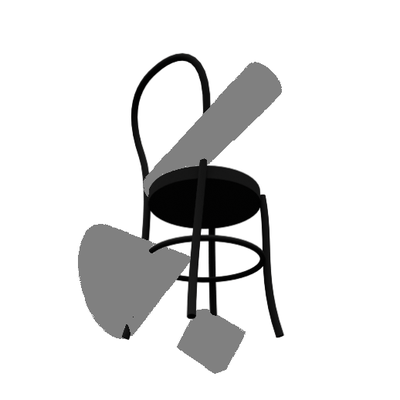} & 
\includegraphics[width=\qcwidth, valign=m]{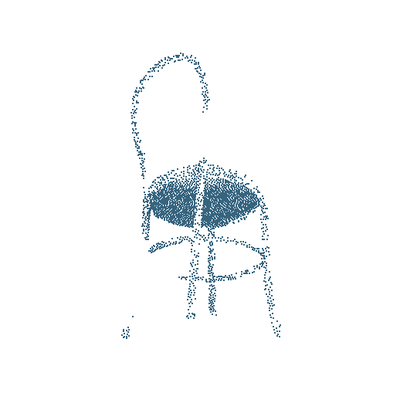} & 
\includegraphics[width=\qcwidth, valign=m]{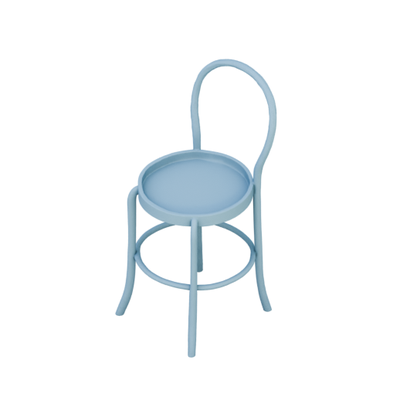} & 
\includegraphics[width=\qcwidth, valign=m]{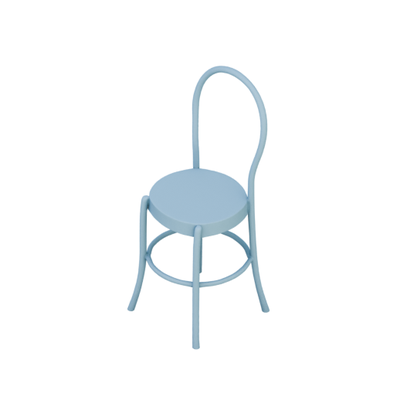} & 
\includegraphics[width=\qcwidth, valign=m]{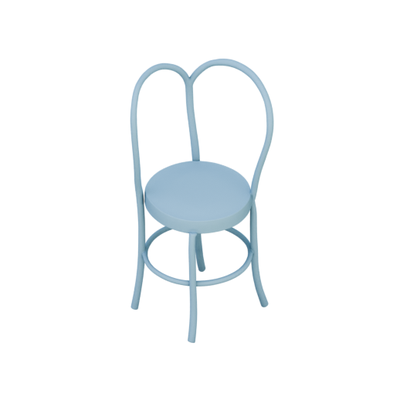} & 
\includegraphics[width=\qcwidth, valign=m]{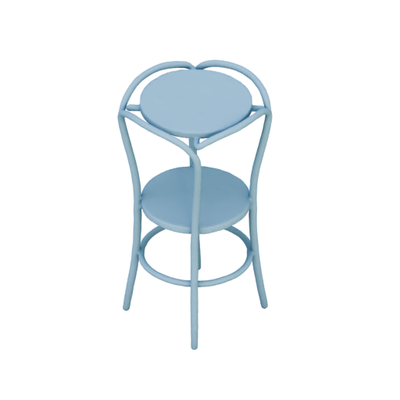} & 
\includegraphics[width=\qcwidth, valign=m]{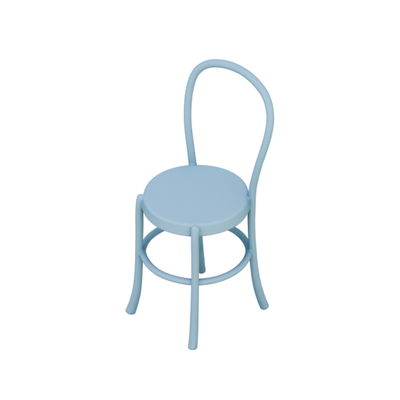} & 
\includegraphics[width=\qcwidth, valign=m]{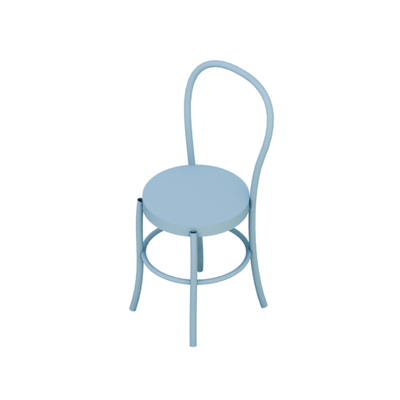} \\

\includegraphics[width=\qcwidth, valign=m]{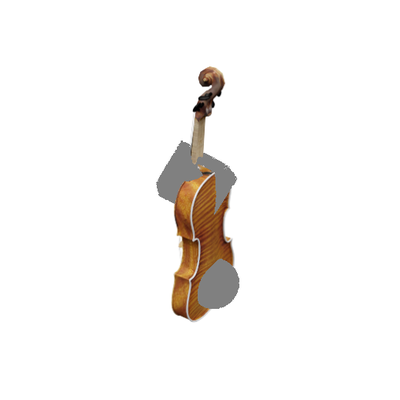} & 
\includegraphics[width=\qcwidth, valign=m]{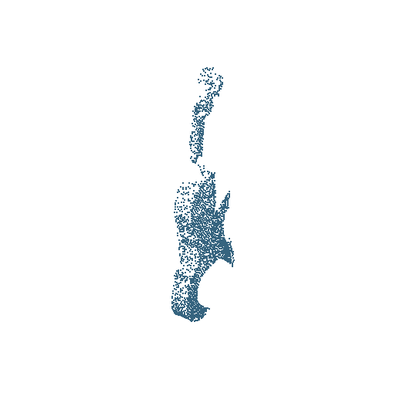} & 
\includegraphics[width=\qcwidth, valign=m]{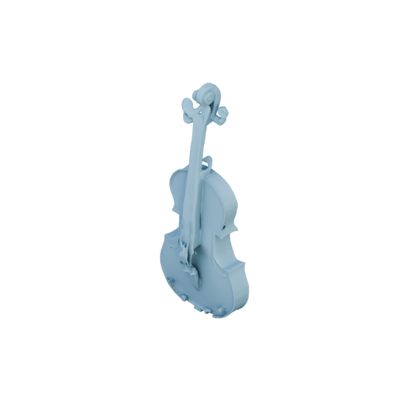} & 
\includegraphics[width=\qcwidth, valign=m]{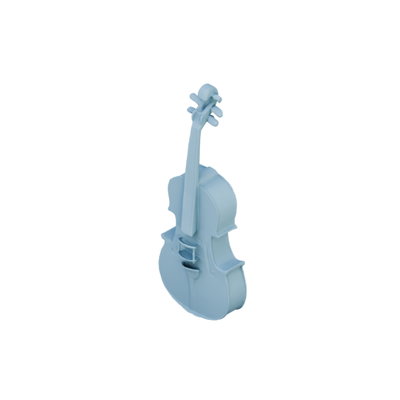} & 
\includegraphics[width=\qcwidth, valign=m]{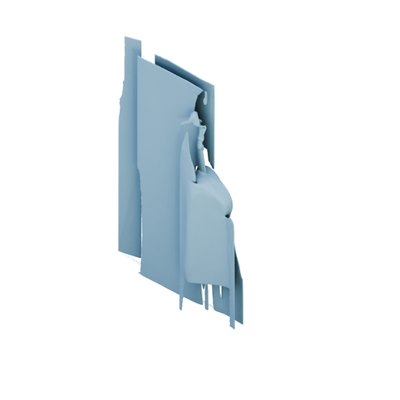} & 
\includegraphics[width=\qcwidth, valign=m]{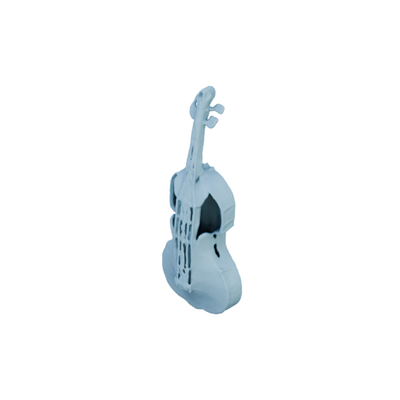} & 
\includegraphics[width=\qcwidth, valign=m]{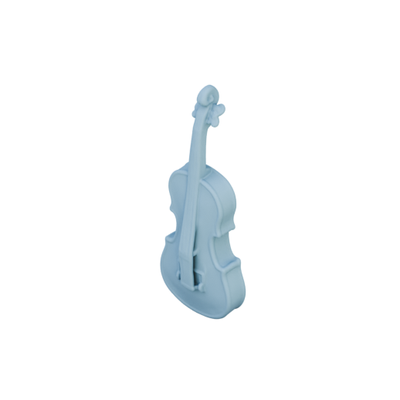} & 
\includegraphics[width=\qcwidth, valign=m]{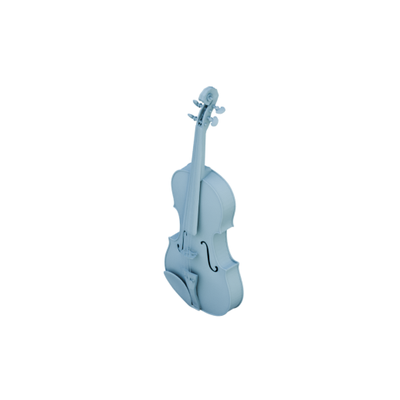} \\

\includegraphics[width=\qcwidth, valign=m]{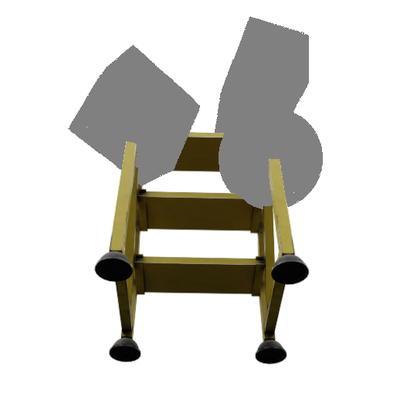} & 
\includegraphics[width=\qcwidth, valign=m]{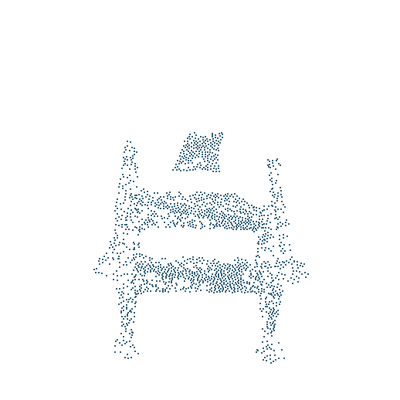} & 
\includegraphics[width=\qcwidth, valign=m]{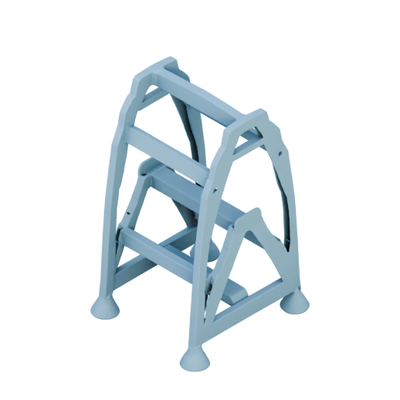} & 
\includegraphics[width=\qcwidth, valign=m]{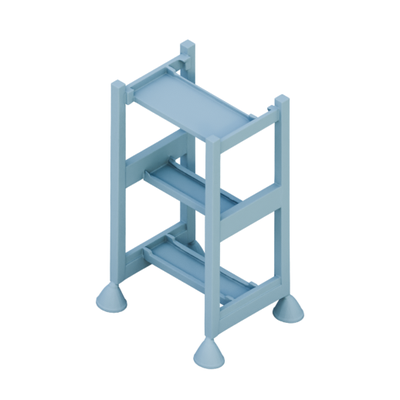} & 
\includegraphics[width=\qcwidth, valign=m]{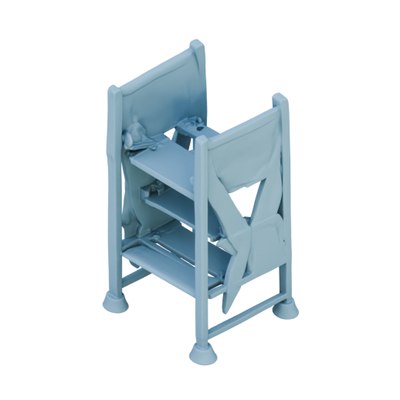} & 
\includegraphics[width=\qcwidth, valign=m]{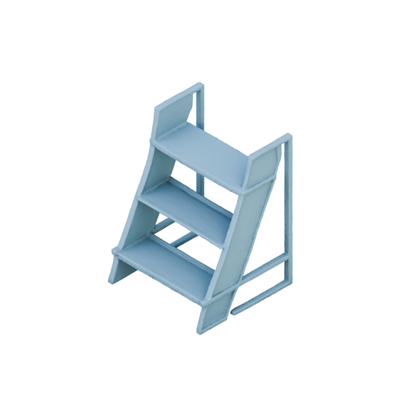} & 
\includegraphics[width=\qcwidth, valign=m]{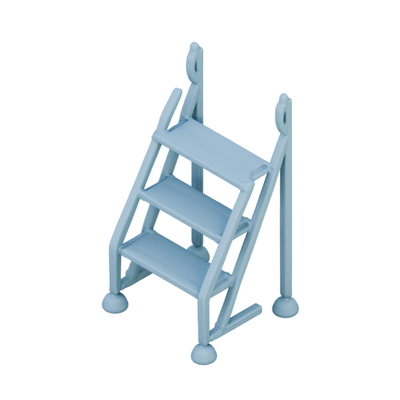} & 
\includegraphics[width=\qcwidth, valign=m]{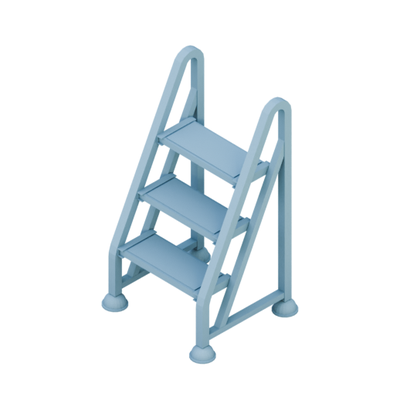} \\

\includegraphics[width=\qcwidth, valign=m]{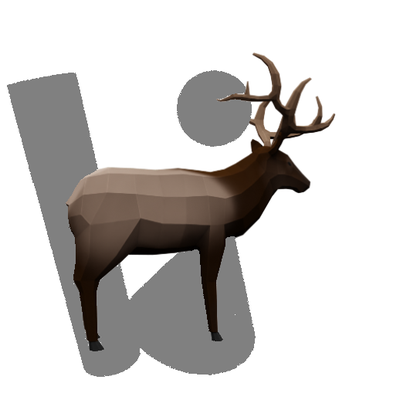} & 
\includegraphics[width=\qcwidth, valign=m]{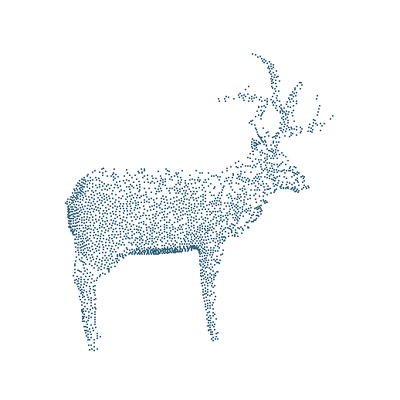} & 
\includegraphics[width=\qcwidth, valign=m]{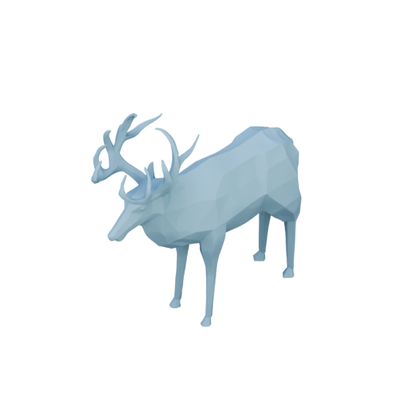} & 
\includegraphics[width=\qcwidth, valign=m]{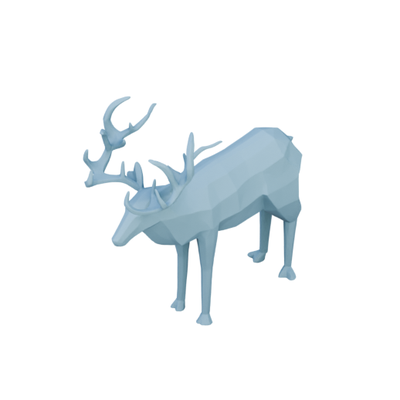} & 
\includegraphics[width=\qcwidth, valign=m]{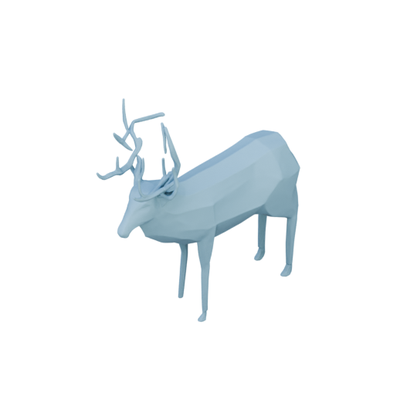} & 
\includegraphics[width=\qcwidth, valign=m]{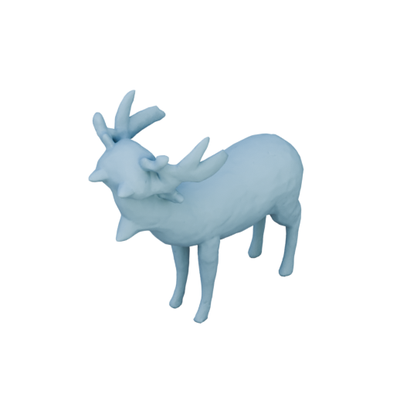} & 
\includegraphics[width=\qcwidth, valign=m]{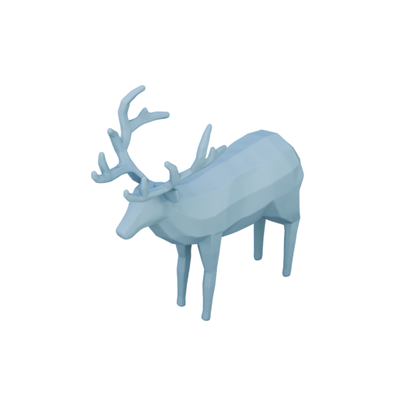} & 
\includegraphics[width=\qcwidth, valign=m]{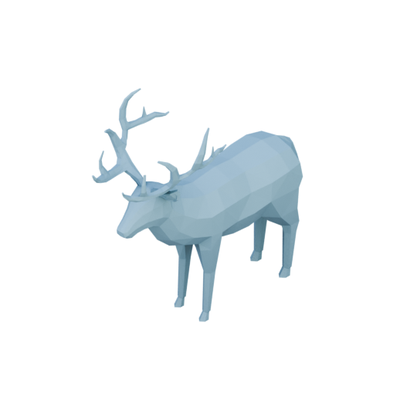} \\

\includegraphics[width=\qcwidth, valign=m]{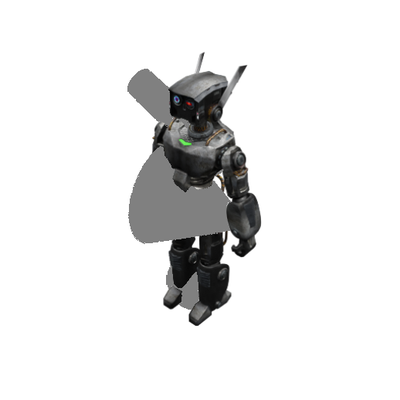} & 
\includegraphics[width=\qcwidth, valign=m]{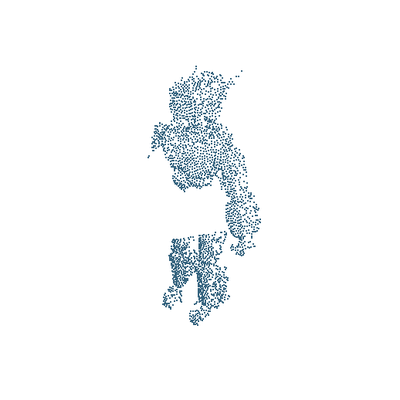} & 
\includegraphics[width=\qcwidth, valign=m]{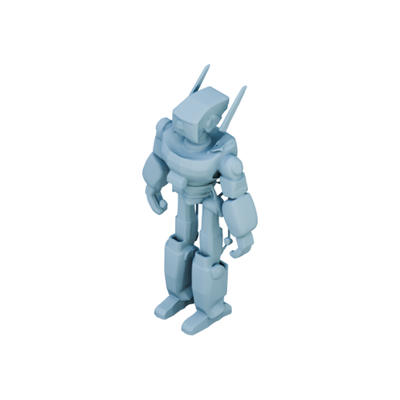} & 
\includegraphics[width=\qcwidth, valign=m]{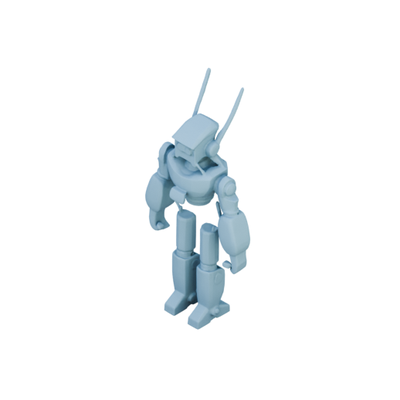} & 
\includegraphics[width=\qcwidth, valign=m]{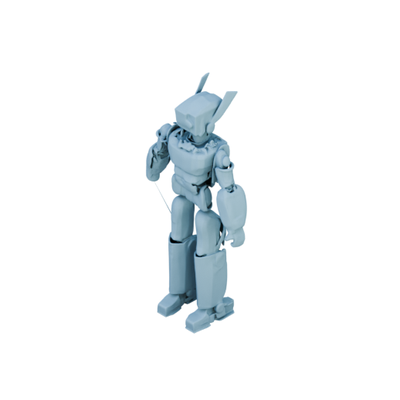} & 
\includegraphics[width=\qcwidth, valign=m]{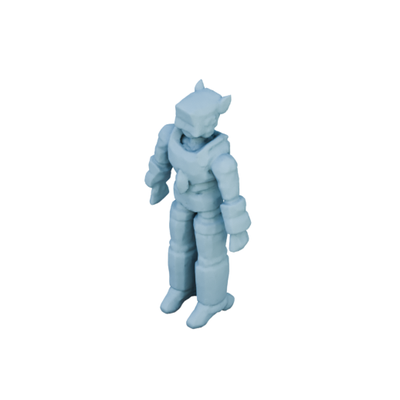} & 
\includegraphics[width=\qcwidth, valign=m]{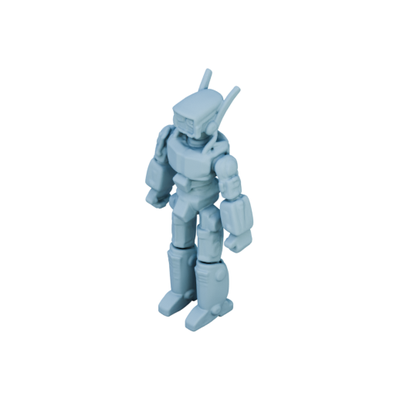} & 
\includegraphics[width=\qcwidth, valign=m]{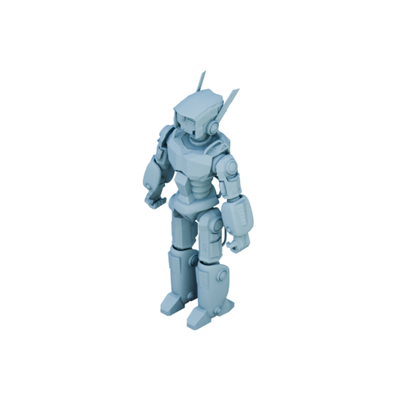} \\

\includegraphics[width=\qcwidth, valign=m]{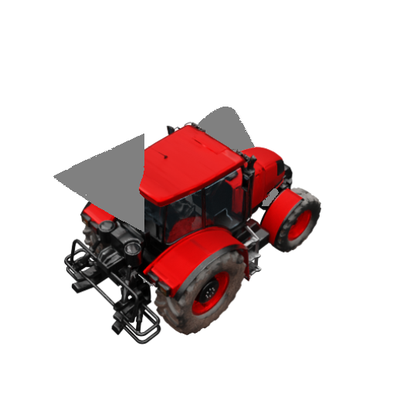} & 
\includegraphics[width=\qcwidth, valign=m]{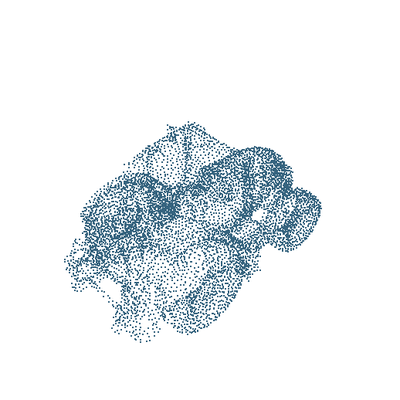} & 
\includegraphics[width=\qcwidth, valign=m]{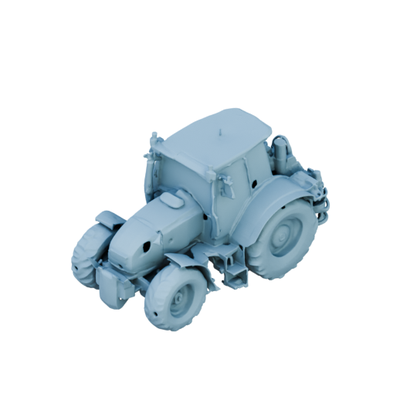} & 
\includegraphics[width=\qcwidth, valign=m]{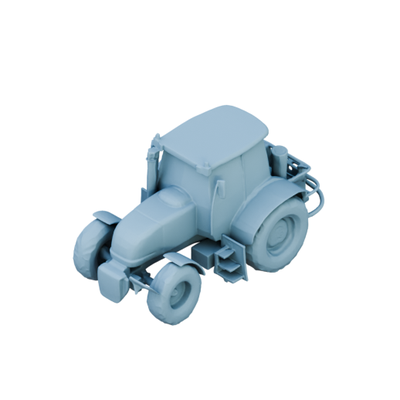} & 
\includegraphics[width=\qcwidth, valign=m]{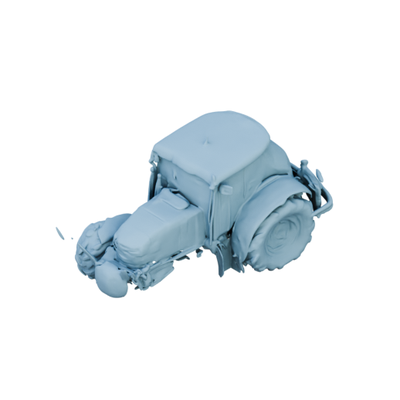} & 
\includegraphics[width=\qcwidth, valign=m]{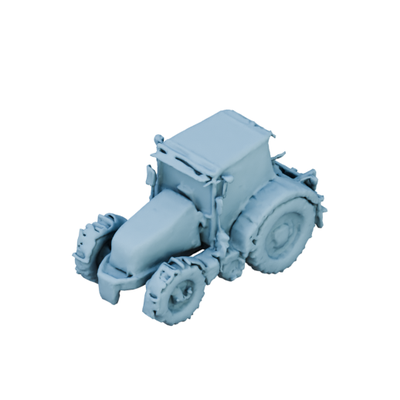} & 
\includegraphics[width=\qcwidth, valign=m]{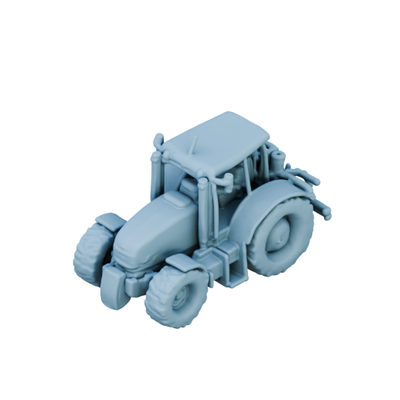} & 
\includegraphics[width=\qcwidth, valign=m]{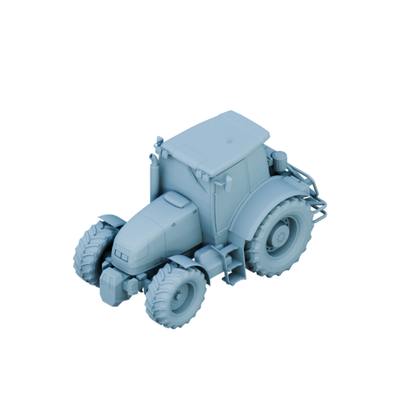} \\

  \end{tabular}
}
  
    \caption{Single-view input results}
    \label{fig:qual_compare_sv}
  \end{subfigure}

  \vspace{2mm}

  \begin{subfigure}{\linewidth}

  \centering
\resizebox{\linewidth}{!} {
\begin{tabular}{@{}cc|cccc|c@{}}

\scalebox{0.5}{Input View} &
\scalebox{0.5}{Input Points} & 
\scalebox{0.5}{Amodal3R} & 
\scalebox{0.5}{Hy3D-Omni} & 
\scalebox{0.5}{ShapeR} & 
\scalebox{0.5}{\textbf{Ours}} & 
\scalebox{0.5}{Ground Truth} \\ 

\includegraphics[width=\mvwidth, valign=m]{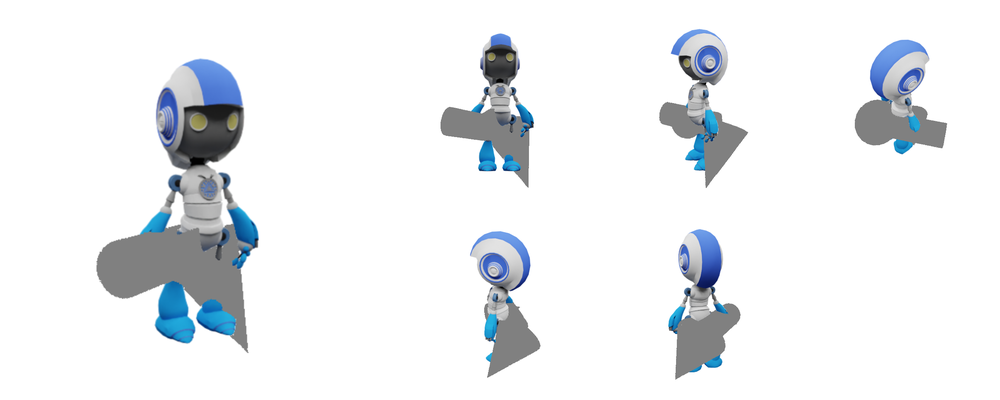} & 
\includegraphics[width=\qcwidth, valign=m]{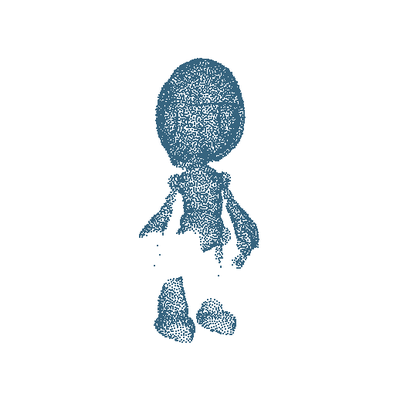} & 
\includegraphics[width=\qcwidth, valign=m]{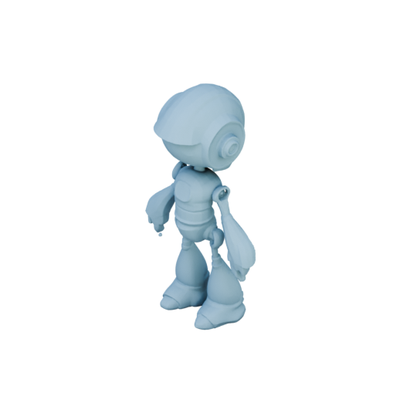} & 
\includegraphics[width=\qcwidth, valign=m]{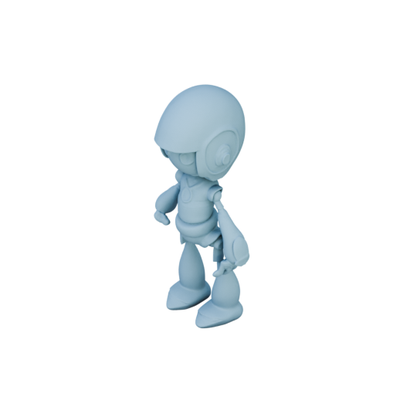} & 
\includegraphics[width=\qcwidth, valign=m]{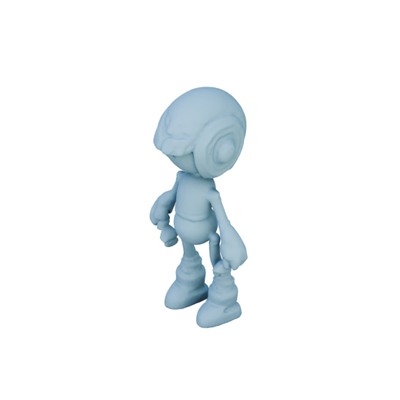} & 
\includegraphics[width=\qcwidth, valign=m]{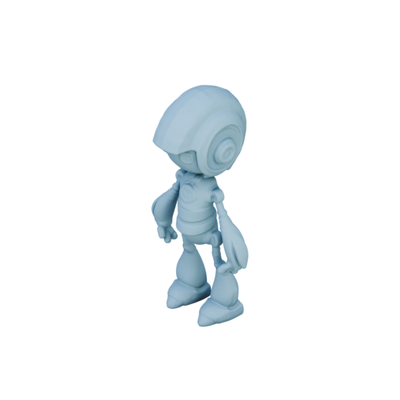} & 
\includegraphics[width=\qcwidth, valign=m]{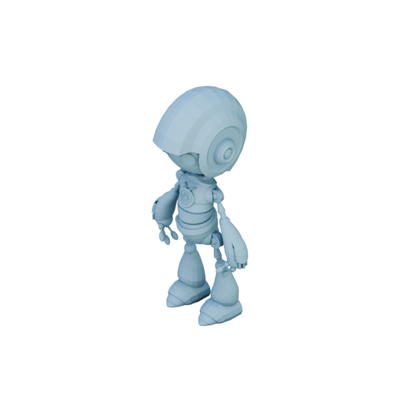} \\

\includegraphics[width=\mvwidth, valign=m]{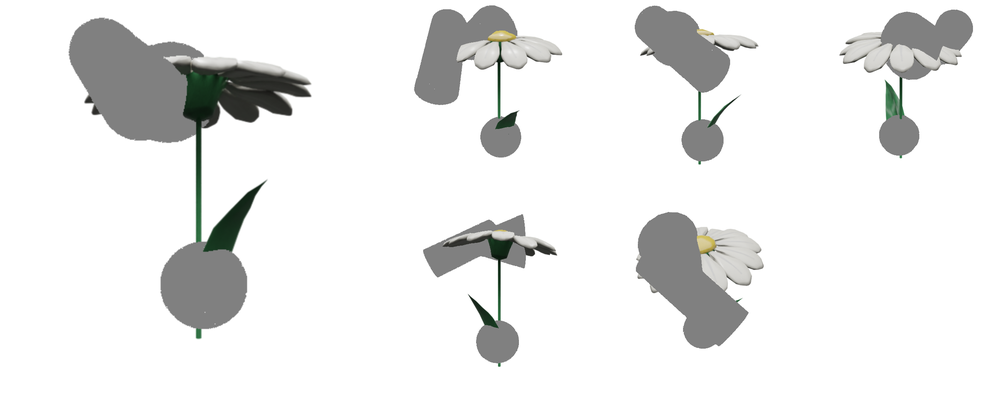} &
\includegraphics[width=\qcwidth, valign=m]{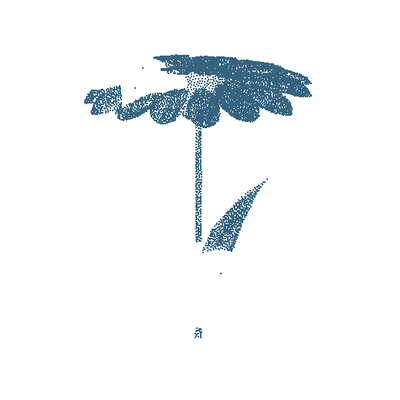} & 
\includegraphics[width=\qcwidth, valign=m]{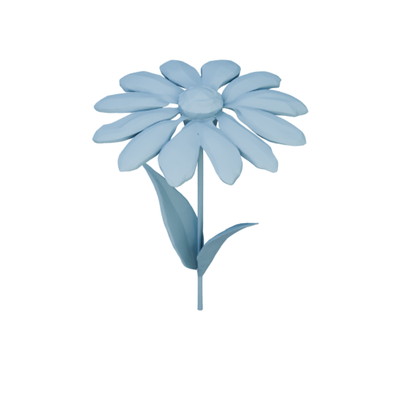} & 
\includegraphics[width=\qcwidth, valign=m]{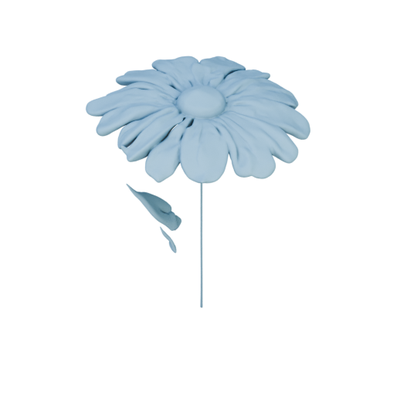} & 
\includegraphics[width=\qcwidth, valign=m]{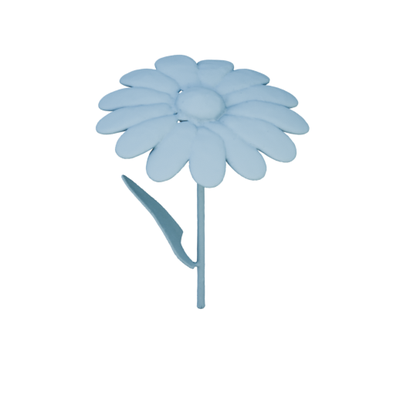} & 
\includegraphics[width=\qcwidth, valign=m]{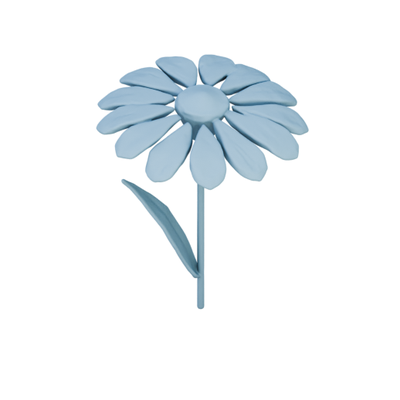} & 
\includegraphics[width=\qcwidth, valign=m]{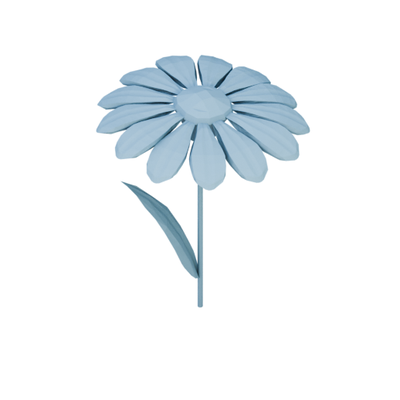} \\

\includegraphics[width=\mvwidth, valign=m]{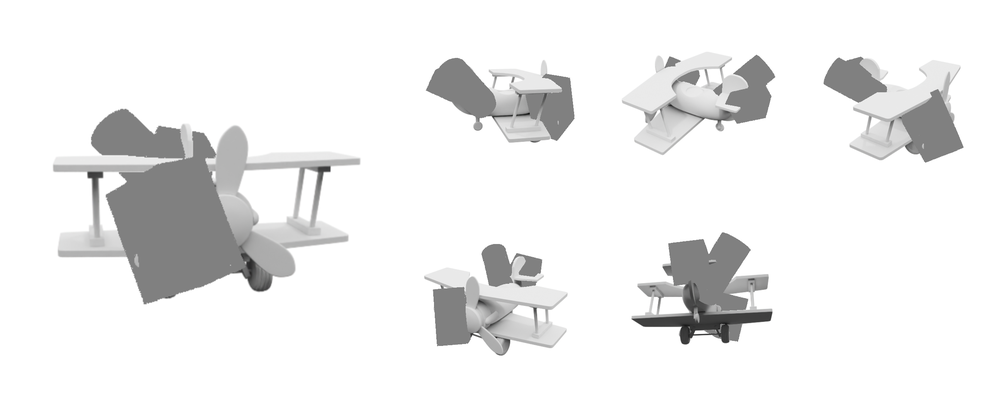} & 
\includegraphics[width=\qcwidth, valign=m]{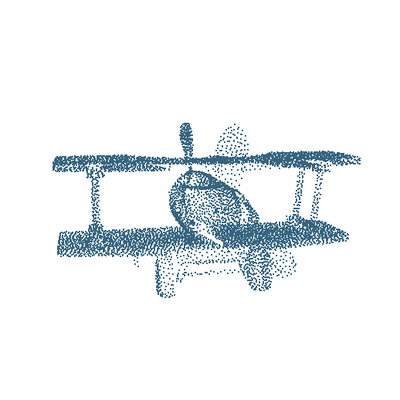} & 
\includegraphics[width=\qcwidth, valign=m]{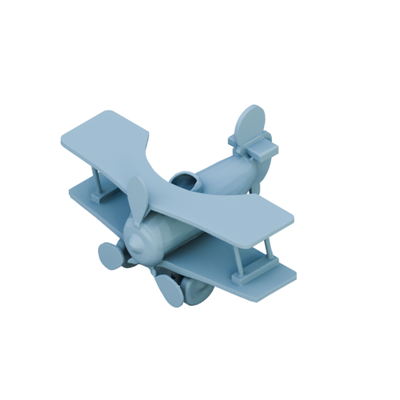} & 
\includegraphics[width=\qcwidth, valign=m]{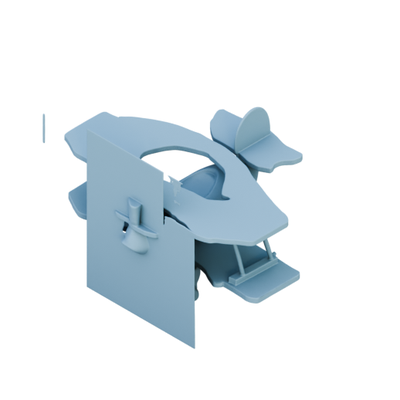} & 
\includegraphics[width=\qcwidth, valign=m]{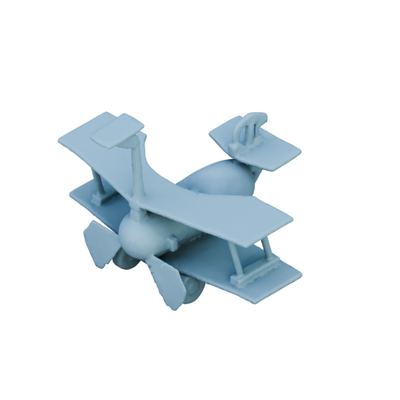} & 
\includegraphics[width=\qcwidth, valign=m]{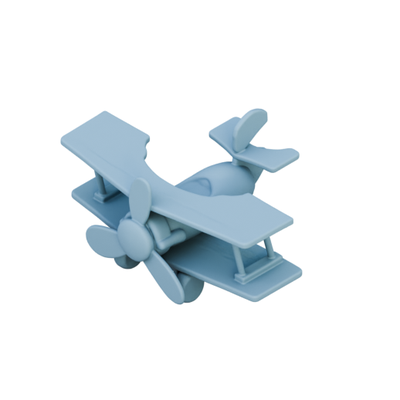} & 
\includegraphics[width=\qcwidth, valign=m]{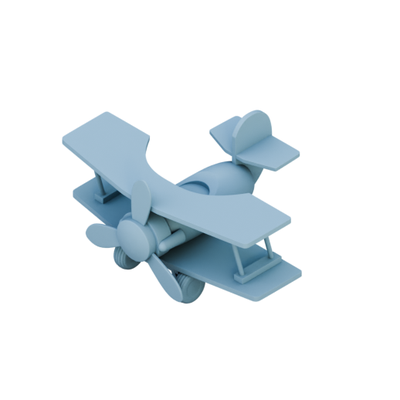} \\

\includegraphics[width=\mvwidth, valign=m]{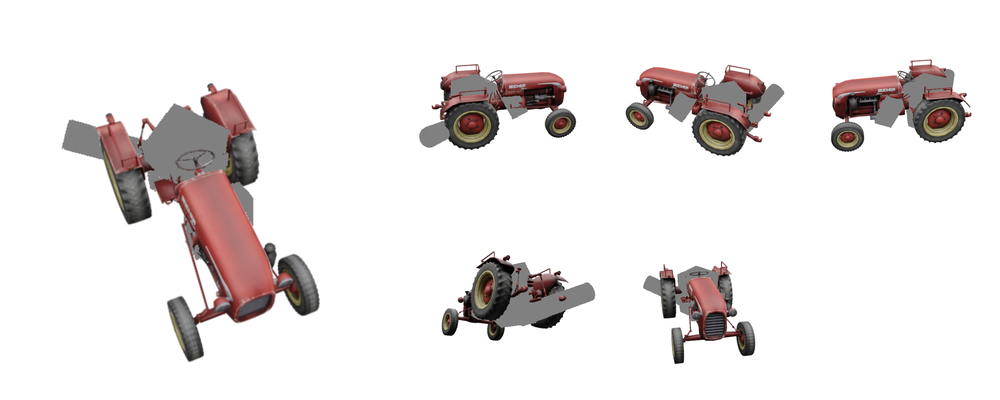} & 
\includegraphics[width=\qcwidth, valign=m]{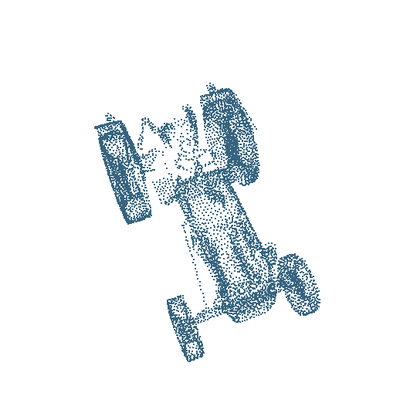} & 
\includegraphics[width=\qcwidth, valign=m]{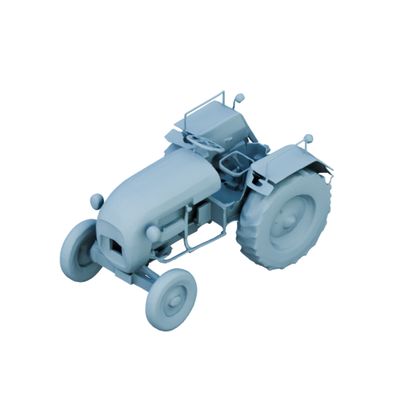} & 
\includegraphics[width=\qcwidth, valign=m]{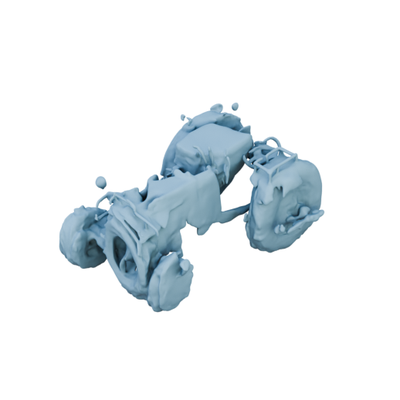} & 
\includegraphics[width=\qcwidth, valign=m]{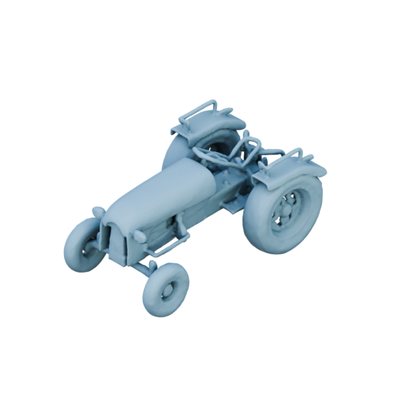} & 
\includegraphics[width=\qcwidth, valign=m]{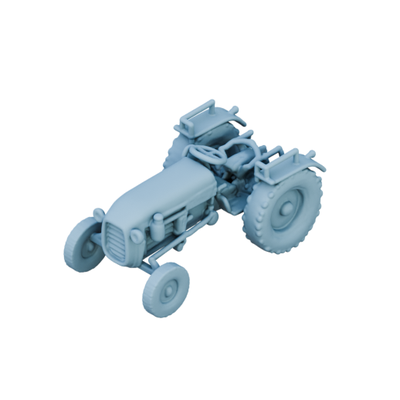} & 
\includegraphics[width=\qcwidth, valign=m]{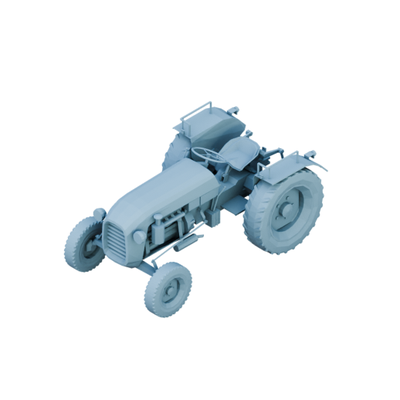} \\

\includegraphics[width=\mvwidth, valign=m]{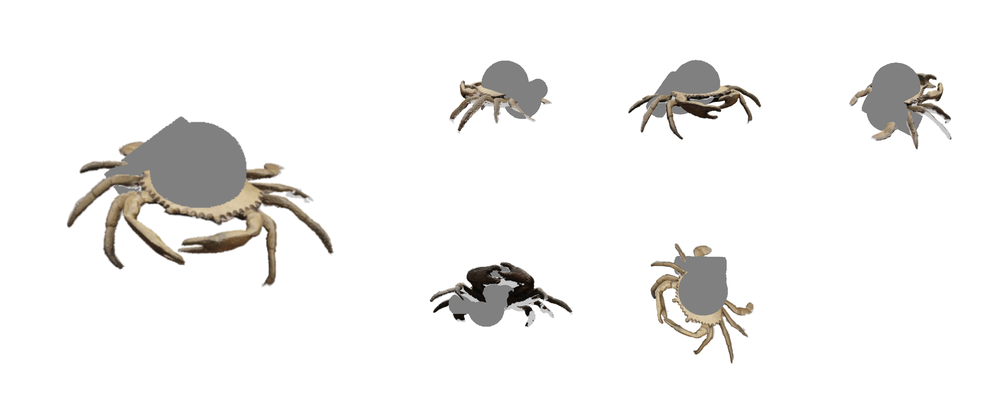} & 
\includegraphics[width=\qcwidth, valign=m]{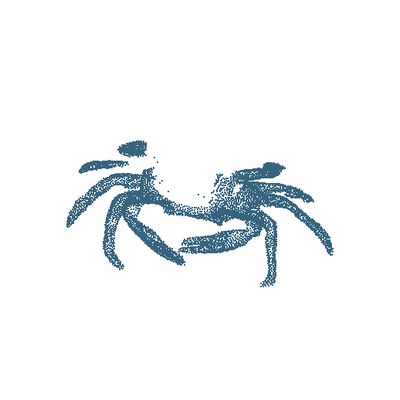} & 
\includegraphics[width=\qcwidth, valign=m]{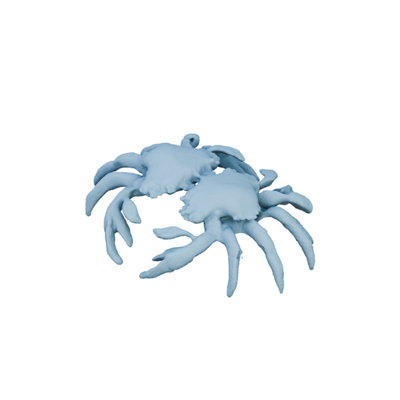} & 
\includegraphics[width=\qcwidth, valign=m]{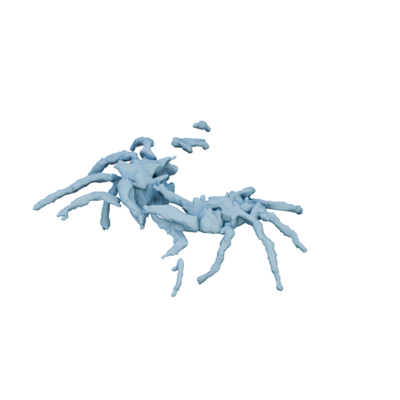} & 
\includegraphics[width=\qcwidth, valign=m]{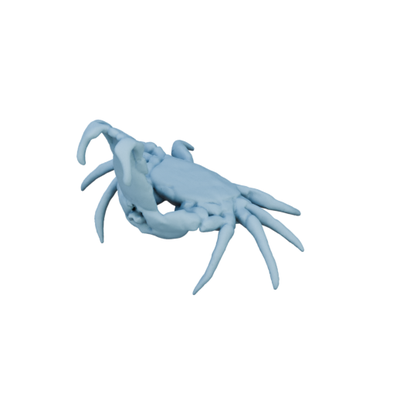} & 
\includegraphics[width=\qcwidth, valign=m]{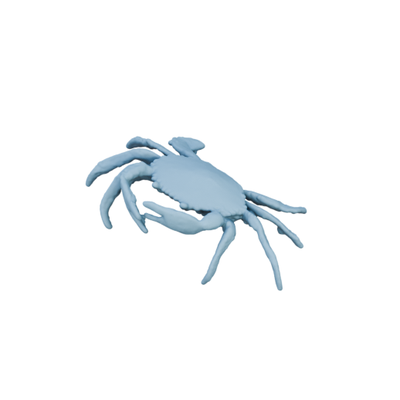} & 
\includegraphics[width=\qcwidth, valign=m]{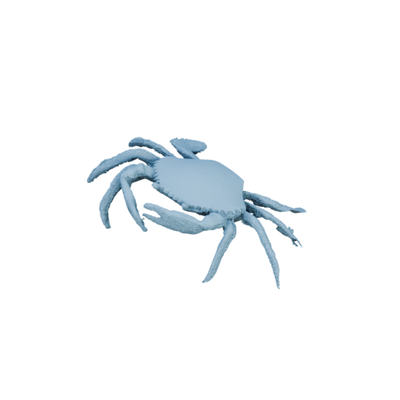} \\

\includegraphics[width=\mvwidth, valign=m]{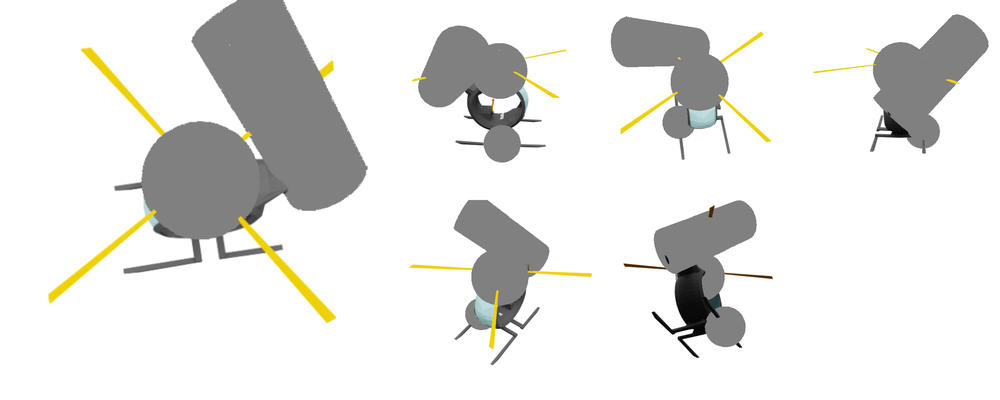} & 
\includegraphics[width=\qcwidth, valign=m]{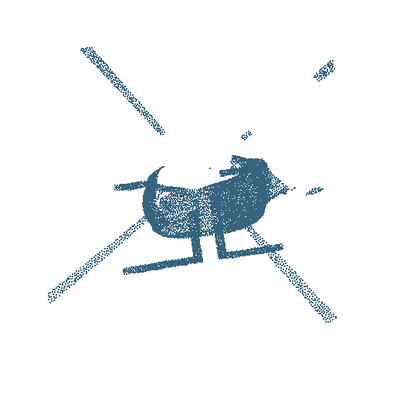} & 
\includegraphics[width=\qcwidth, valign=m]{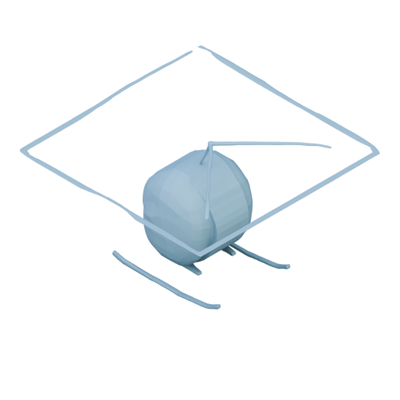} & 
\includegraphics[width=\qcwidth, valign=m]{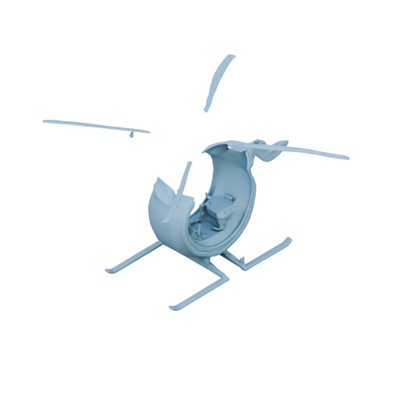} & 
\includegraphics[width=\qcwidth, valign=m]{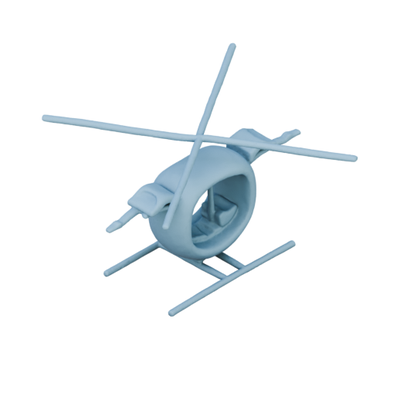} & 
\includegraphics[width=\qcwidth, valign=m]{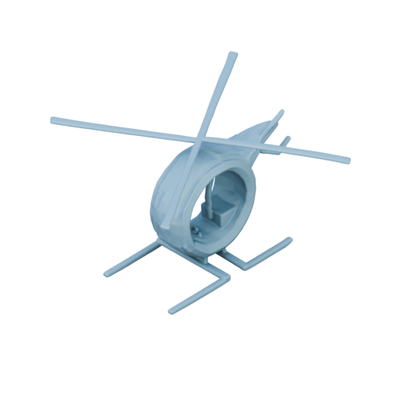} & 
\includegraphics[width=\qcwidth, valign=m]{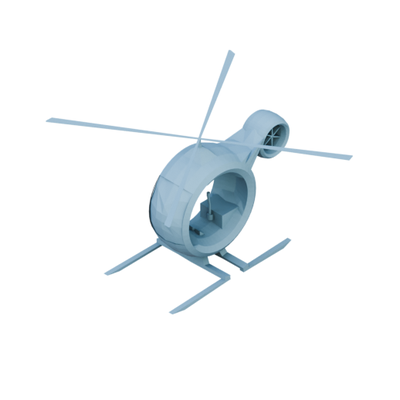} \\

  \end{tabular}
}

    \caption{Multi-view input results}
    \label{fig:qual_compare_mv}
  \end{subfigure}

  \caption{\textbf{Qualitative Comparisons}: Single- and multi-view completion results against baselines. Occluders are shown in grey in the input view. Zoom in for details.}
  \label{fig:qual_combined}
\end{figure*}

\subsection{Qualitative Results}\label{ssec:qual}

\begingroup
\setlength{\tabcolsep}{1pt}
    
\renewcommand{\arraystretch}{0}

\begin{figure}[tb]
	\centering

\resizebox{\linewidth}{!} {
\begin{tabular}{@{}cc|cc|cc|cc|cc@{}}

\scalebox{0.5}{Input View} &
\scalebox{0.5}{Input Points} & 
\multicolumn{2}{c|}{\scalebox{0.5}{SAM3D}} & 
\multicolumn{2}{c|}{\scalebox{0.5}{Hy3D-Omni}} & 
\multicolumn{2}{c|}{\scalebox{0.5}{ShapeR}} & 
\multicolumn{2}{c}{\scalebox{0.5}{Ours}}  \\

\includegraphics[width=\qcwidth, valign=m]{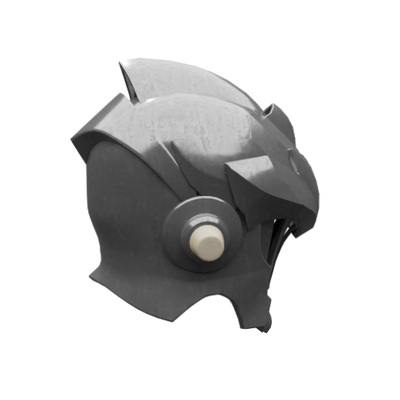} & 
\includegraphics[width=\qcwidth, valign=m]{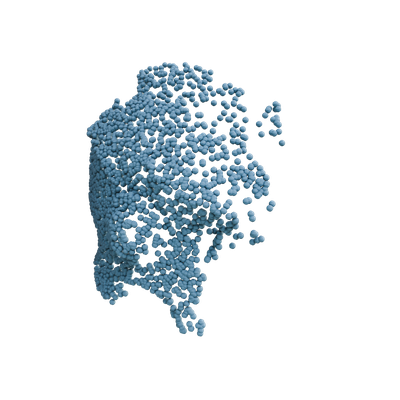} & 
\includegraphics[width=\qcwidth, valign=m]{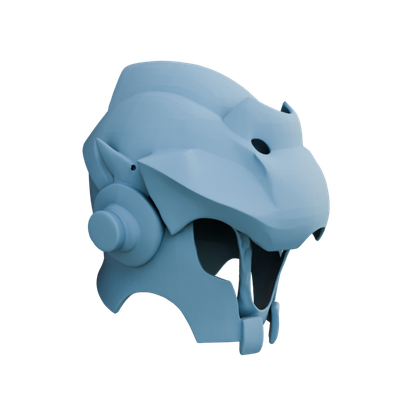} & 
\includegraphics[width=\qcwidth, valign=m]{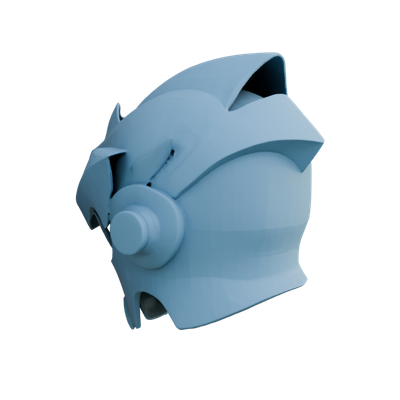} & 
\includegraphics[width=\qcwidth, valign=m]{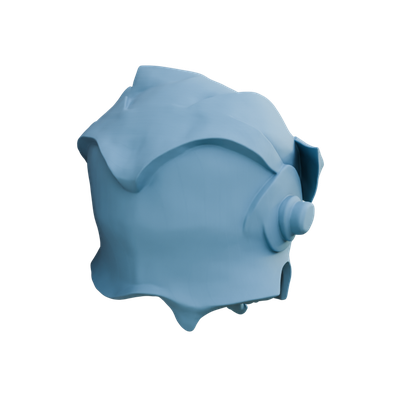} & 
\includegraphics[width=\qcwidth, valign=m]{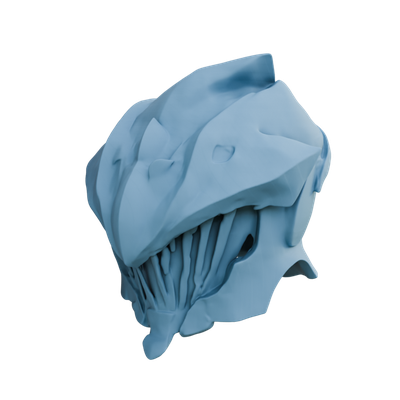} & 
\includegraphics[width=\qcwidth, valign=m]{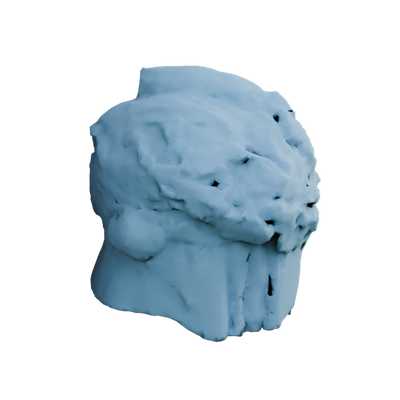} & 
\includegraphics[width=\qcwidth, valign=m]{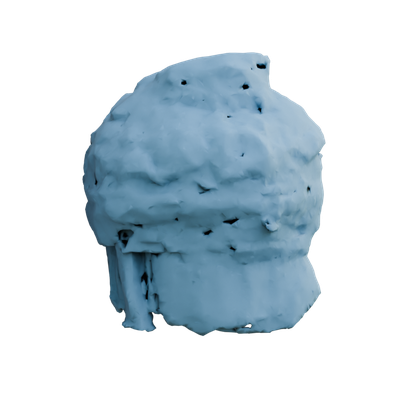} & 
\includegraphics[width=\qcwidth, valign=m]{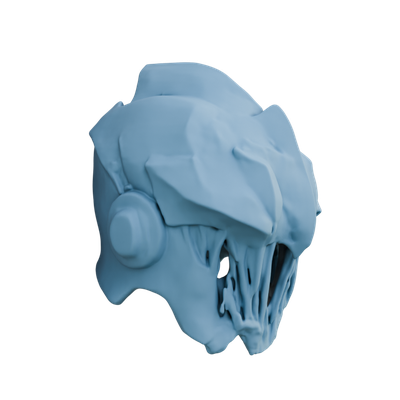} & 
\includegraphics[width=\qcwidth, valign=m]{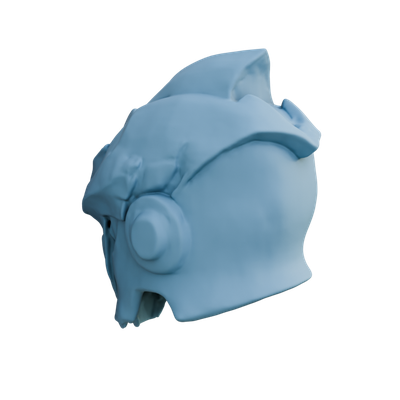} \\

\includegraphics[width=\qcwidth, valign=m]{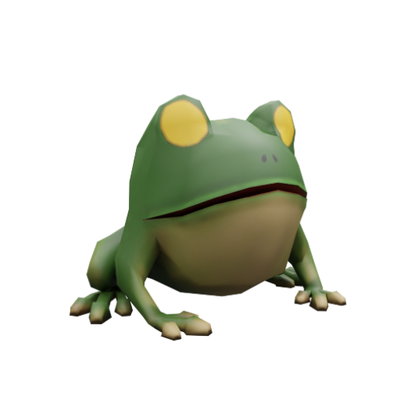} & 
\includegraphics[width=\qcwidth, valign=m]{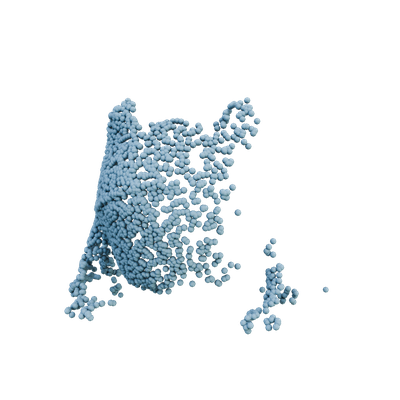} & 
\includegraphics[width=\qcwidth, valign=m]{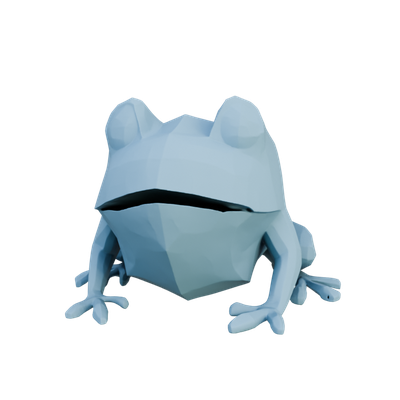} & 
\includegraphics[width=\qcwidth, valign=m]{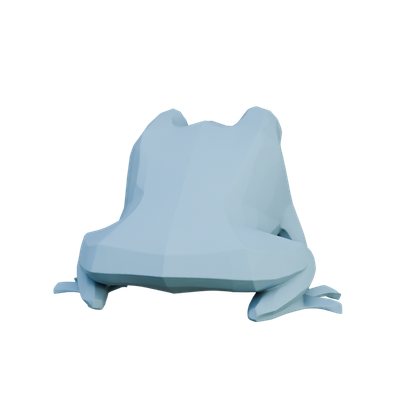} & 
\includegraphics[width=\qcwidth, valign=m]{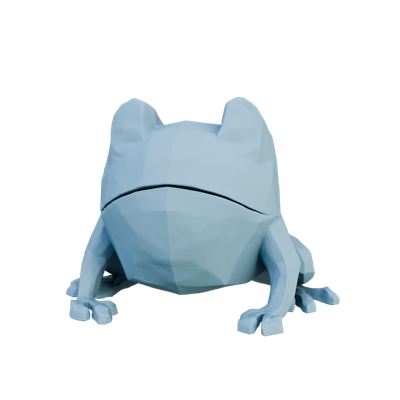} & 
\includegraphics[width=\qcwidth, valign=m]{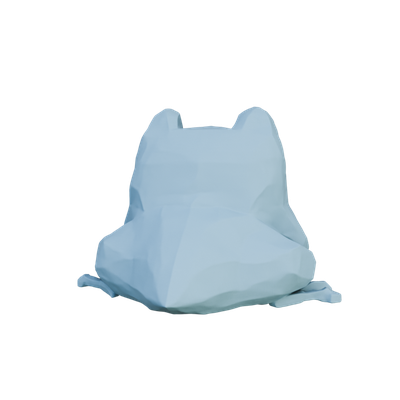} & 
\includegraphics[width=\qcwidth, valign=m]{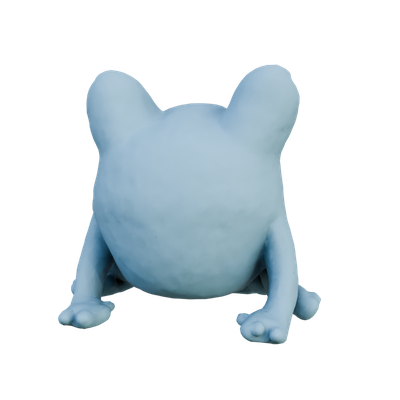} & 
\includegraphics[width=\qcwidth, valign=m]{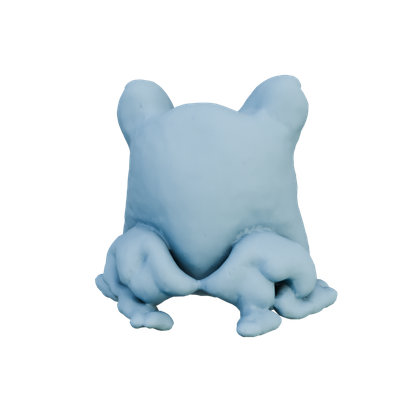} & 
\includegraphics[width=\qcwidth, valign=m]{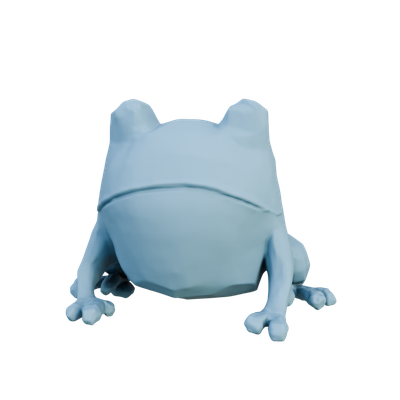} & 
\includegraphics[width=\qcwidth, valign=m]{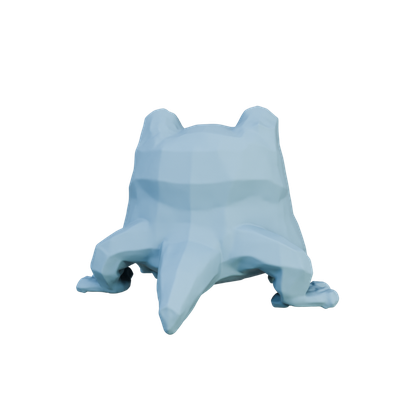} \\

\includegraphics[width=\qcwidth, valign=m]{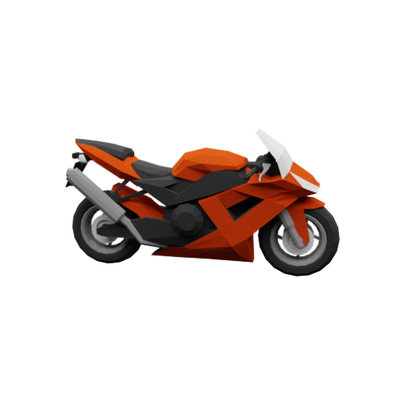} & 
\includegraphics[width=\qcwidth, valign=m]{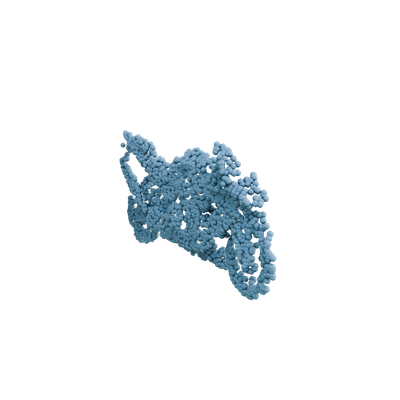} & 
\includegraphics[width=\qcwidth, valign=m]{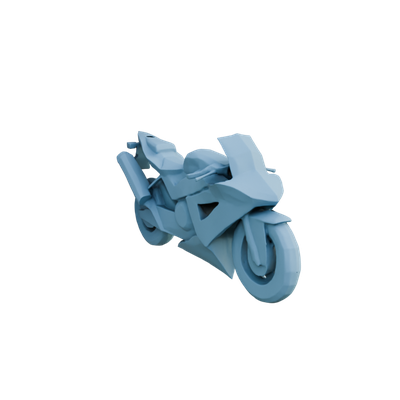} & 
\includegraphics[width=\qcwidth, valign=m]{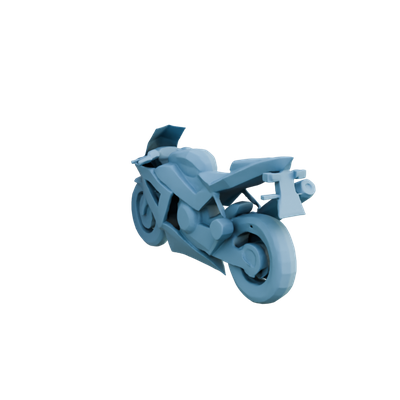} & 
\includegraphics[width=\qcwidth, valign=m]{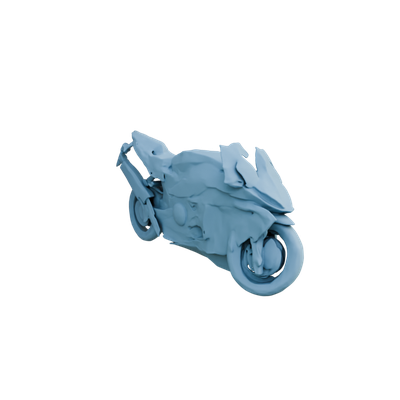} & 
\includegraphics[width=\qcwidth, valign=m]{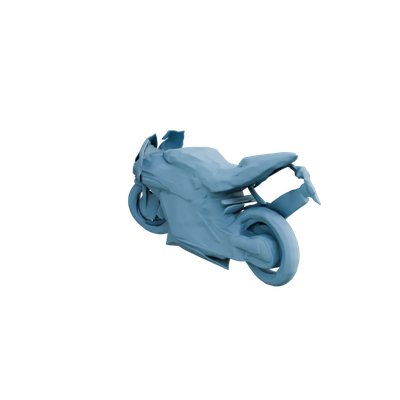} & 
\includegraphics[width=\qcwidth, valign=m]{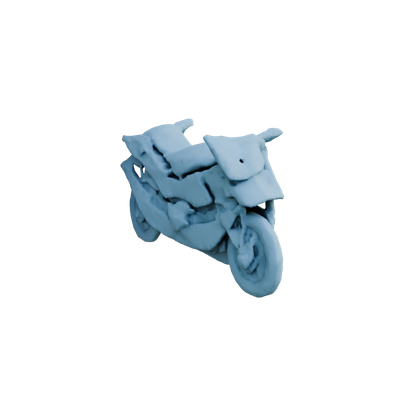} & 
\includegraphics[width=\qcwidth, valign=m]{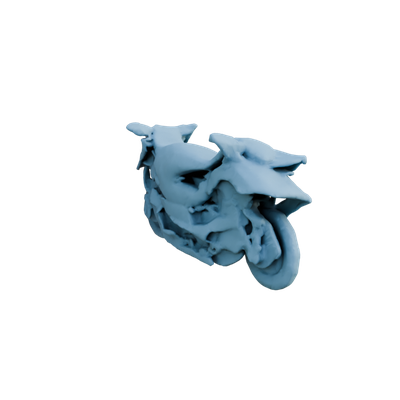} & 
\includegraphics[width=\qcwidth, valign=m]{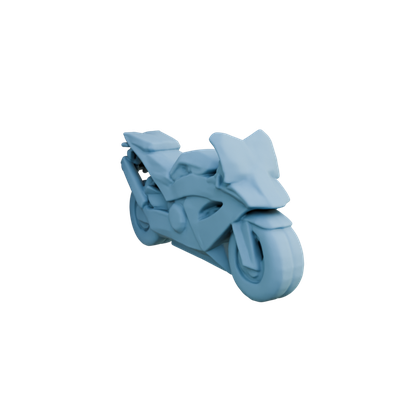} & 
\includegraphics[width=\qcwidth, valign=m]{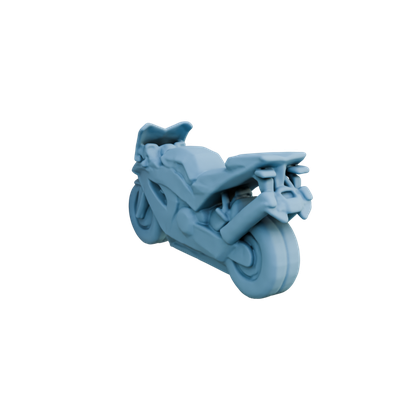} \\

\includegraphics[width=\qcwidth, valign=m]{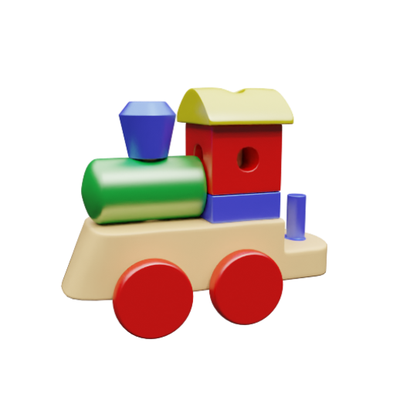} & 
\includegraphics[width=\qcwidth, valign=m]{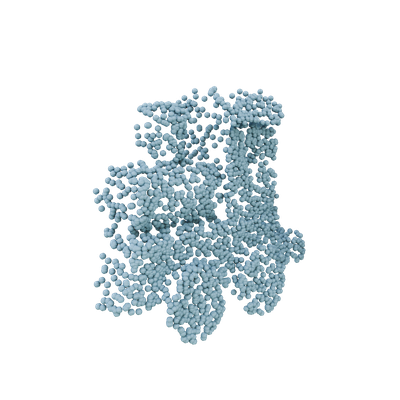} & 
\includegraphics[width=\qcwidth, valign=m]{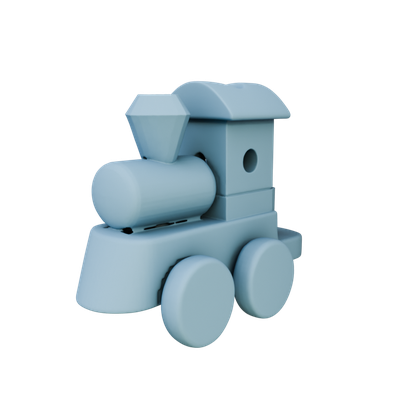} & 
\includegraphics[width=\qcwidth, valign=m]{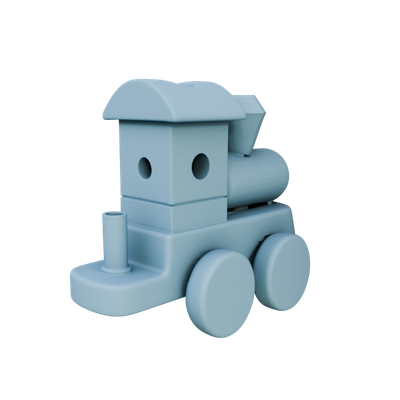} & 
\includegraphics[width=\qcwidth, valign=m]{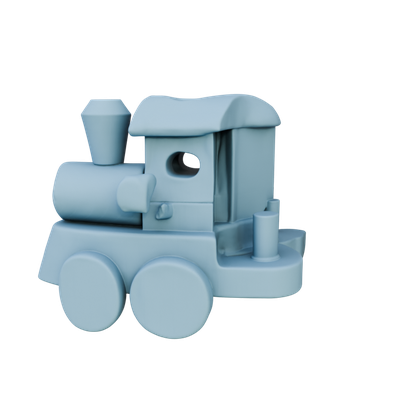} & 
\includegraphics[width=\qcwidth, valign=m]{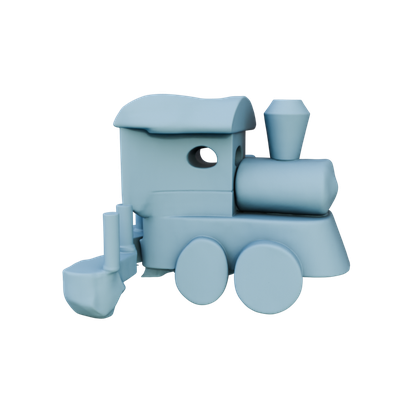} & 
\includegraphics[width=\qcwidth, valign=m]{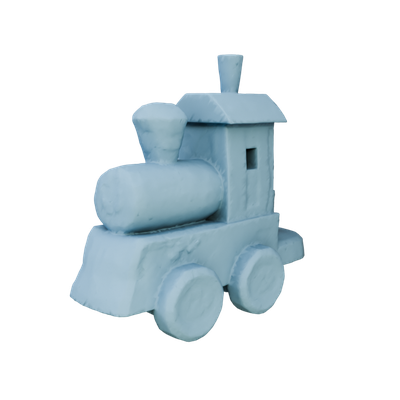} & 
\includegraphics[width=\qcwidth, valign=m]{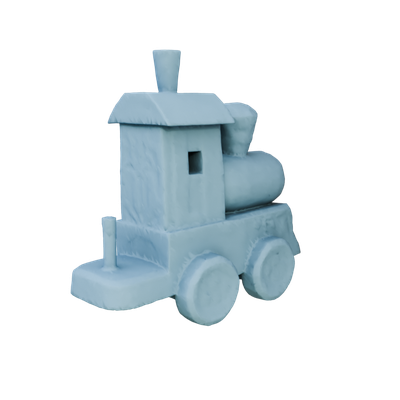} & 
\includegraphics[width=\qcwidth, valign=m]{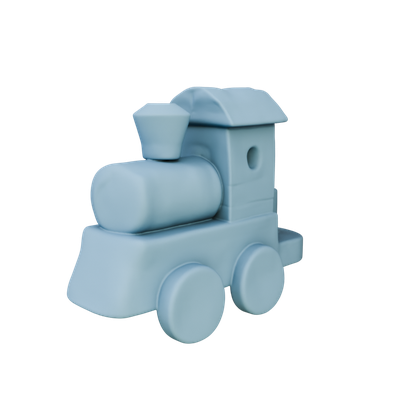} & 
\includegraphics[width=\qcwidth, valign=m]{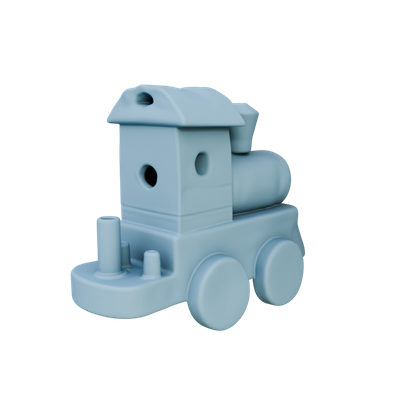} \\

\includegraphics[width=\qcwidth, valign=m]{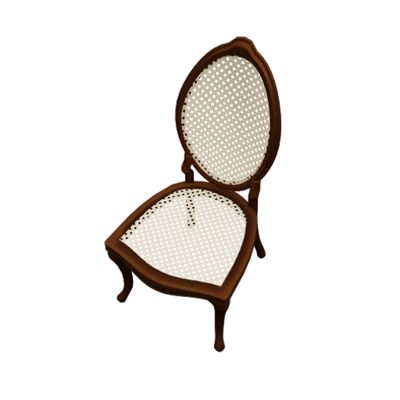} & 
\includegraphics[width=\qcwidth, valign=m]{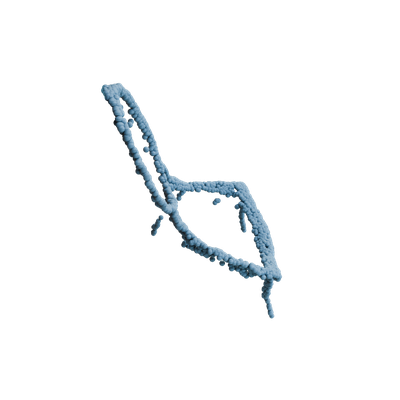} & 
\includegraphics[width=\qcwidth, valign=m]{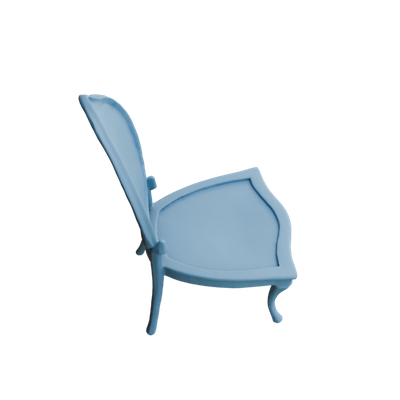} & 
\includegraphics[width=\qcwidth, valign=m]{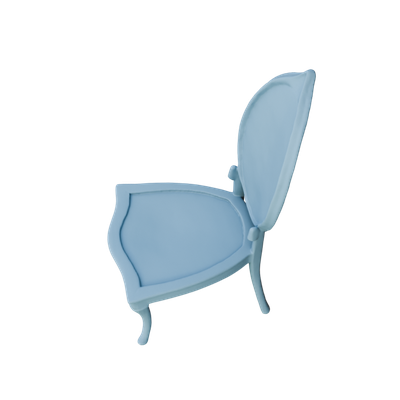} & 
\includegraphics[width=\qcwidth, valign=m]{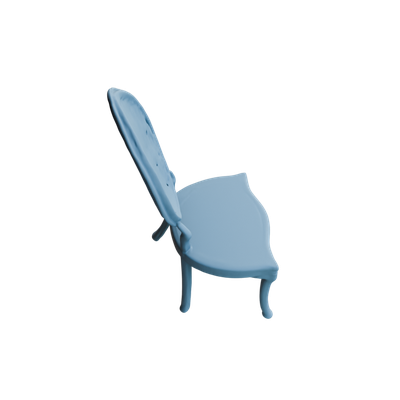} & 
\includegraphics[width=\qcwidth, valign=m]{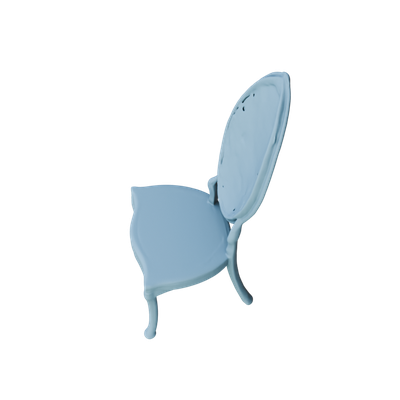} & 
\includegraphics[width=\qcwidth, valign=m]{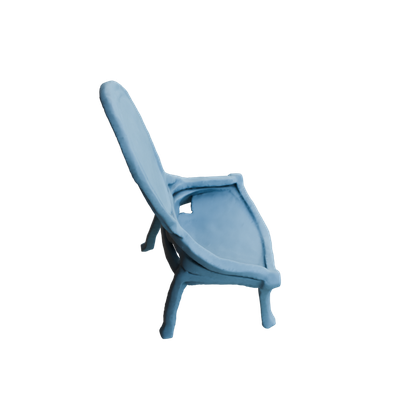} & 
\includegraphics[width=\qcwidth, valign=m]{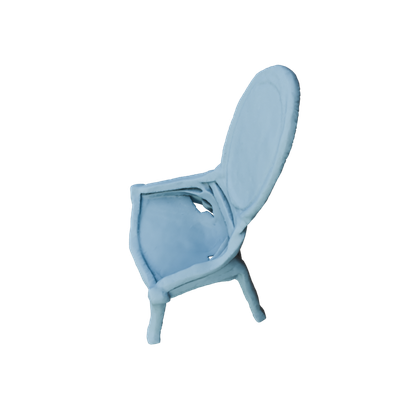} & 
\includegraphics[width=\qcwidth, valign=m]{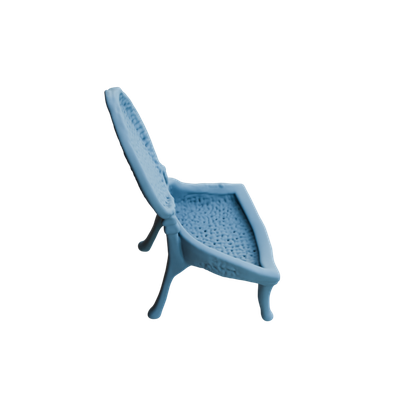} & 
\includegraphics[width=\qcwidth, valign=m]{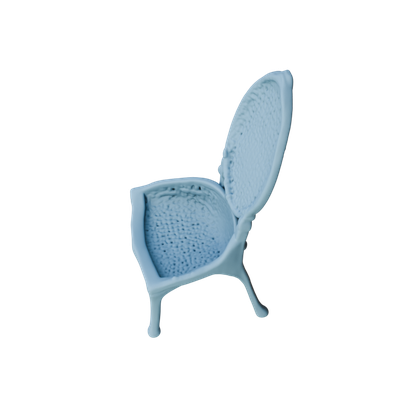} \\

\includegraphics[width=\qcwidth, valign=m]{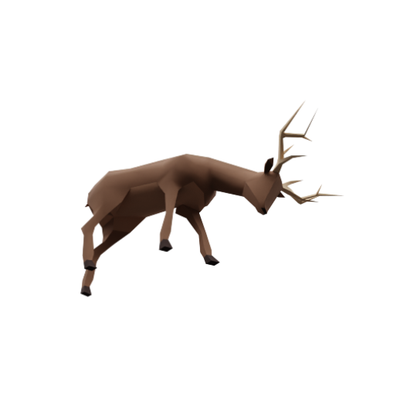} & 
\includegraphics[width=\qcwidth, valign=m]{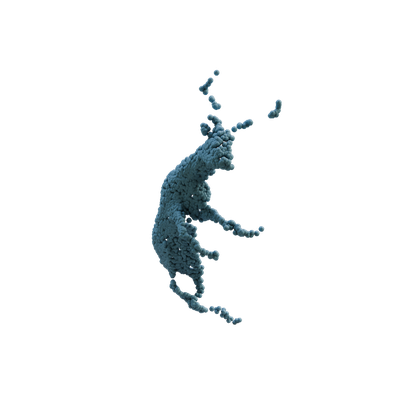} & 
\includegraphics[width=\qcwidth, valign=m]{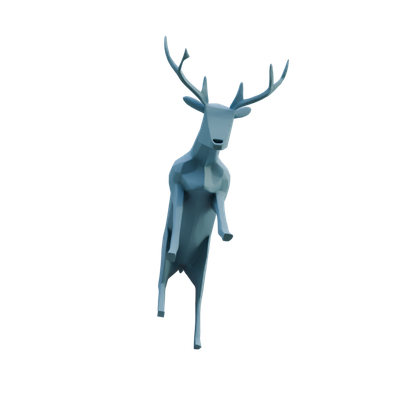} & 
\includegraphics[width=\qcwidth, valign=m]{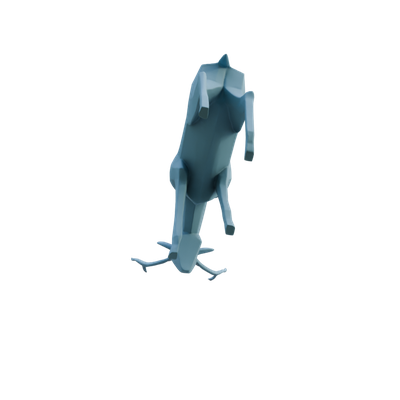} & 
\includegraphics[width=\qcwidth, valign=m]{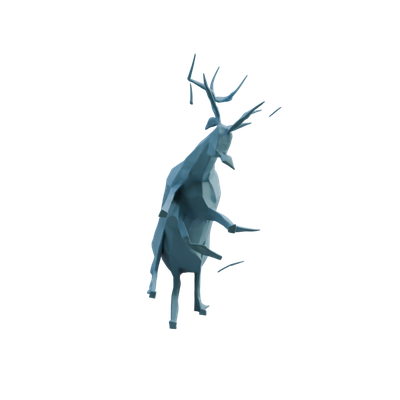} & 
\includegraphics[width=\qcwidth, valign=m]{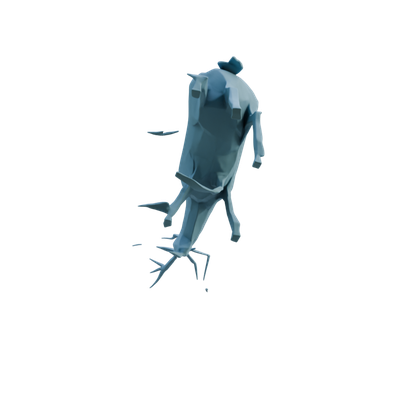} & 
\includegraphics[width=\qcwidth, valign=m]{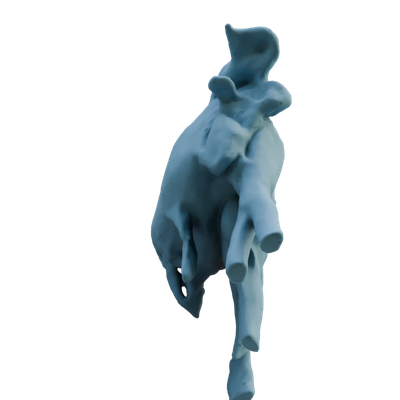} & 
\includegraphics[width=\qcwidth, valign=m]{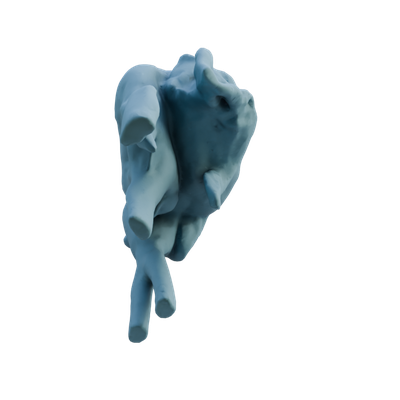} & 
\includegraphics[width=\qcwidth, valign=m]{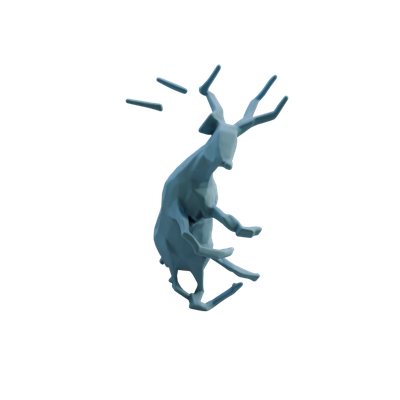} & 
\includegraphics[width=\qcwidth, valign=m]{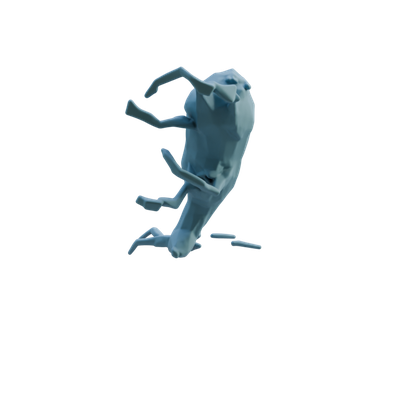} \\

  \end{tabular}
}
	\caption{\textbf{Predicted Depth Comparisons}: Unlike SAM3D, our method prioritizes adherence to input points, yielding more faithful completion, but degrades under severe depth prediction errors---a performance trade off. Zoom in for details.}
	\label{fig:qual_compare_depth}
\end{figure}

\endgroup

\begingroup
\setlength{\tabcolsep}{1pt}
    
\renewcommand{\arraystretch}{0}

\begin{figure*}[t]
	\centering

\resizebox{\linewidth}{!} {
\begin{tabular}{@{}ccc|ccc|ccc|c@{}}

\scalebox{0.5}{Input View} &
\scalebox{0.5}{Input Points} & 
\scalebox{0.5}{Output} & 
\scalebox{0.5}{Input View} &
\scalebox{0.5}{Input Points} & 
\scalebox{0.5}{Output} & 
\scalebox{0.5}{Input View} &
\scalebox{0.5}{Input Points} & 
\scalebox{0.5}{Output} & 
\scalebox{0.5}{Ground Truth} \\ 

\includegraphics[width=\qcwidth, valign=m]{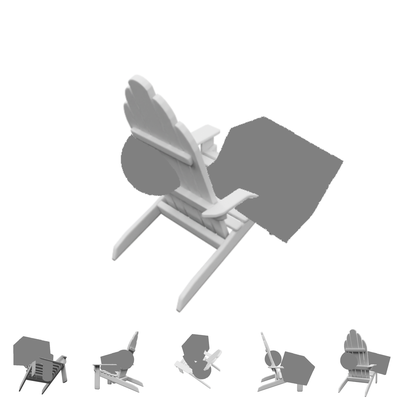} & 
\includegraphics[width=\qcwidth, valign=m]{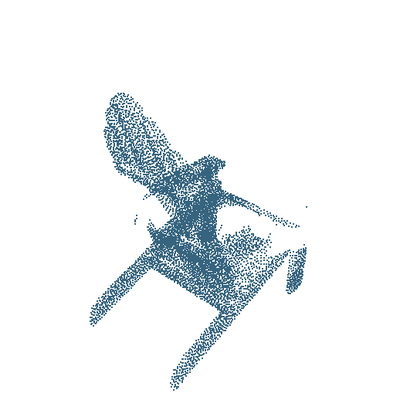} & 
\includegraphics[width=\qcwidth, valign=m]{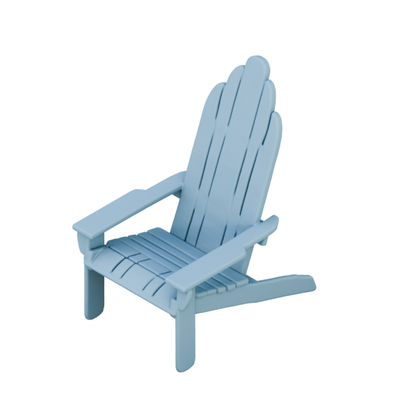} & 
\includegraphics[width=\qcwidth, valign=m]{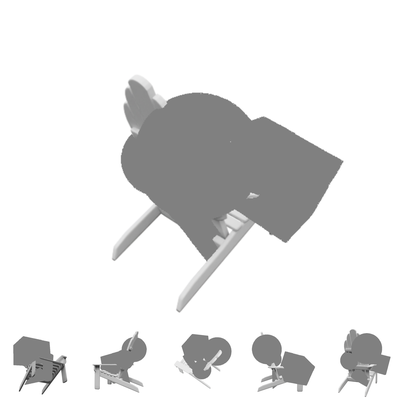} & 
\includegraphics[width=\qcwidth, valign=m]{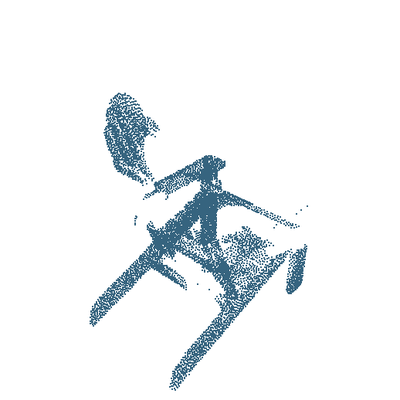} & 
\includegraphics[width=\qcwidth, valign=m]{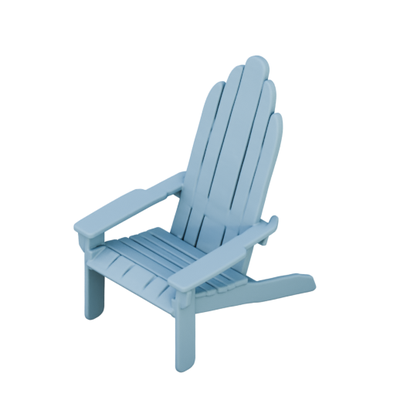} & 
\includegraphics[width=\qcwidth, valign=m]{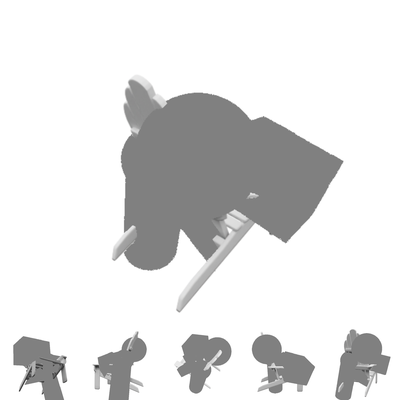} & 
\includegraphics[width=\qcwidth, valign=m]{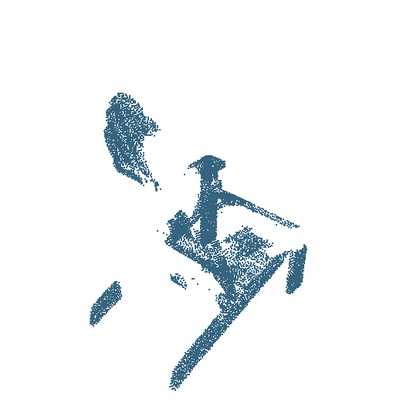} & 
\includegraphics[width=\qcwidth, valign=m]{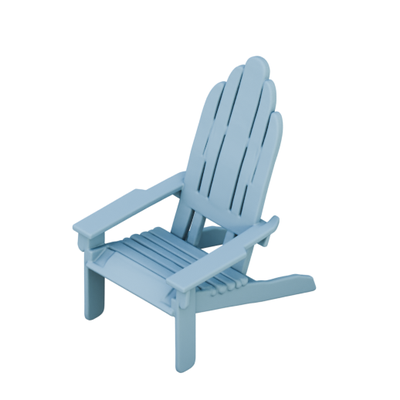} &  
\includegraphics[width=\qcwidth, valign=m]{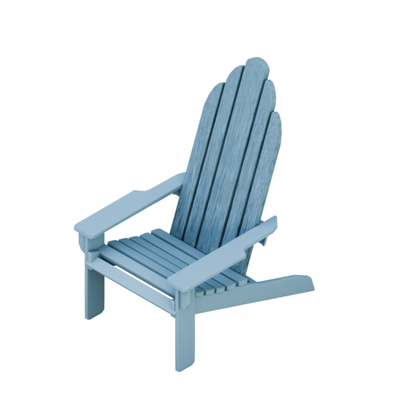} \\

\includegraphics[width=\qcwidth, valign=m]{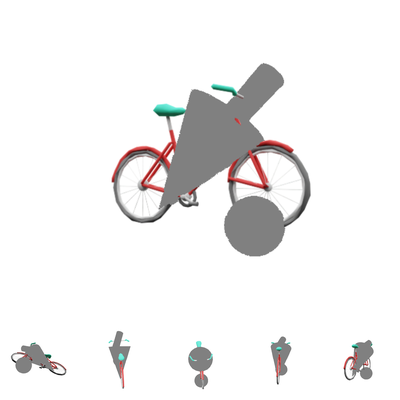} & 
\includegraphics[width=\qcwidth, valign=m]{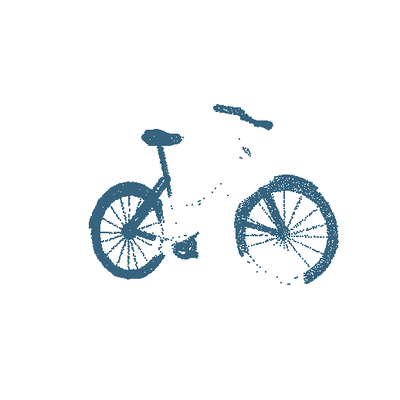} & 
\includegraphics[width=\qcwidth, valign=m]{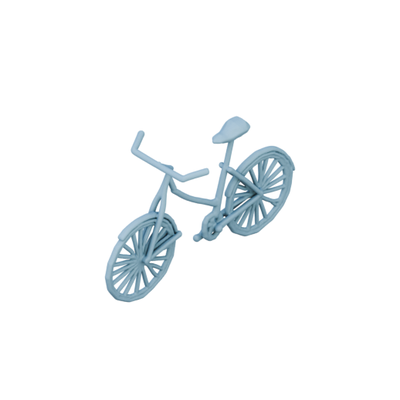} & 
\includegraphics[width=\qcwidth, valign=m]{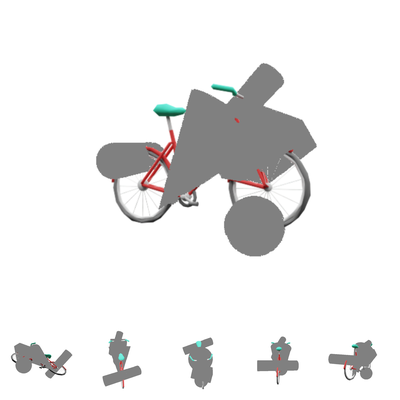} & 
\includegraphics[width=\qcwidth, valign=m]{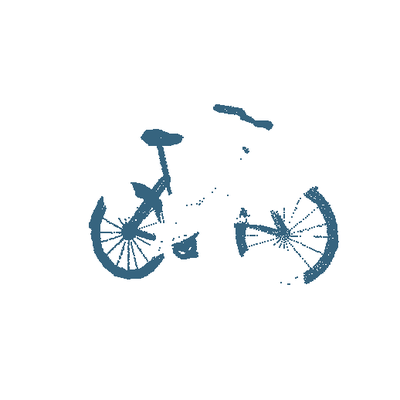} & 
\includegraphics[width=\qcwidth, valign=m]{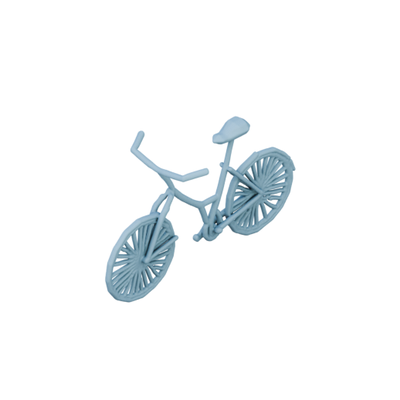} & 
\includegraphics[width=\qcwidth, valign=m]{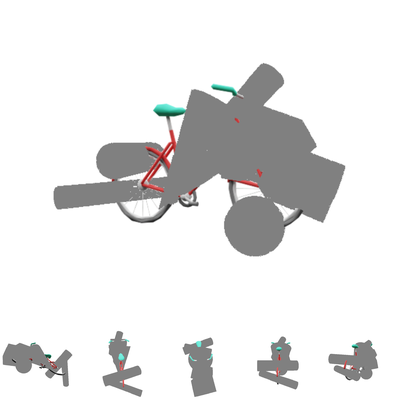} & 
\includegraphics[width=\qcwidth, valign=m]{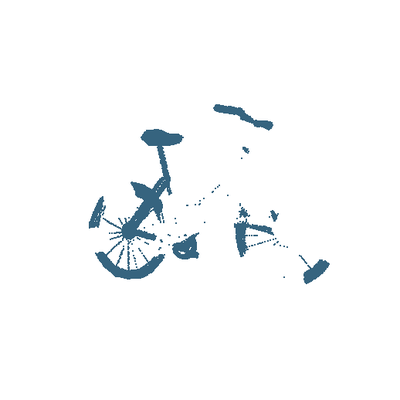} & 
\includegraphics[width=\qcwidth, valign=m]{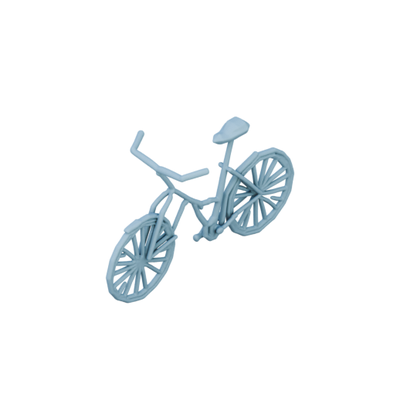} &  
\includegraphics[width=\qcwidth, valign=m]{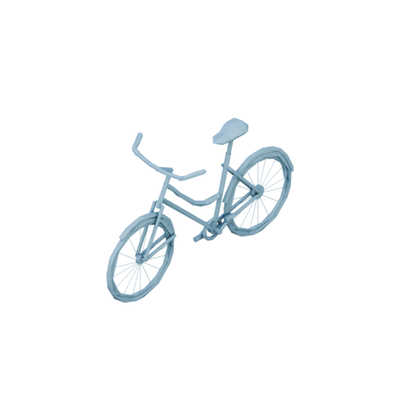} \\

\includegraphics[width=\qcwidth, valign=m]{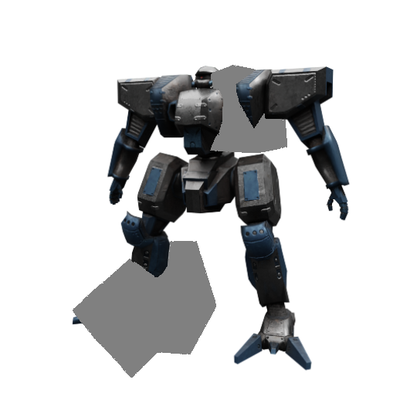} & 
\includegraphics[width=\qcwidth, valign=m]{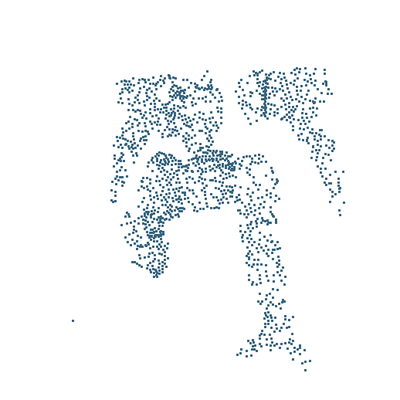} & 
\includegraphics[width=\qcwidth, valign=m]{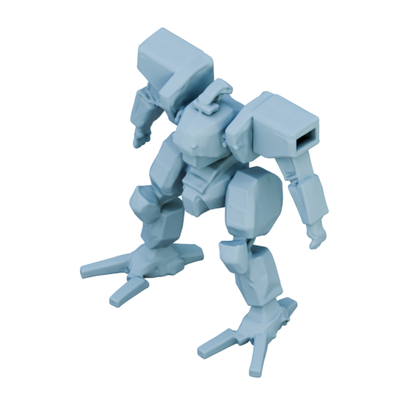} & 
\includegraphics[width=\qcwidth, valign=m]{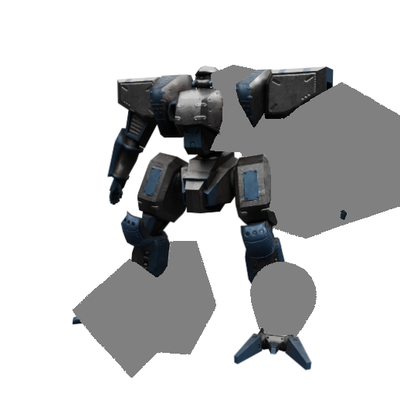} & 
\includegraphics[width=\qcwidth, valign=m]{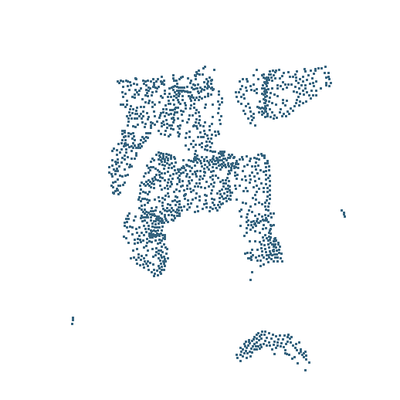} & 
\includegraphics[width=\qcwidth, valign=m]{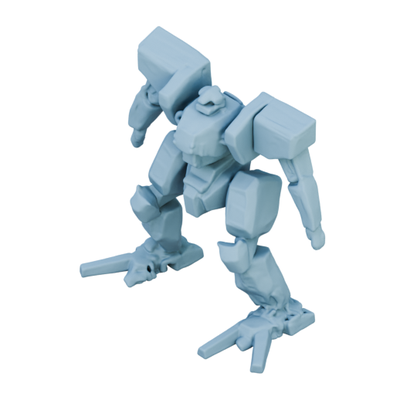} & 
\includegraphics[width=\qcwidth, valign=m]{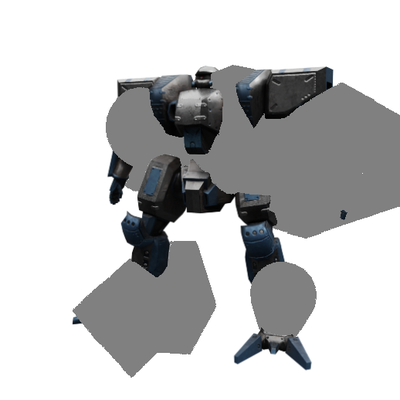} & 
\includegraphics[width=\qcwidth, valign=m]{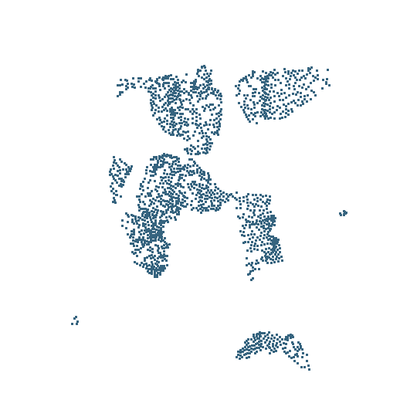} & 
\includegraphics[width=\qcwidth, valign=m]{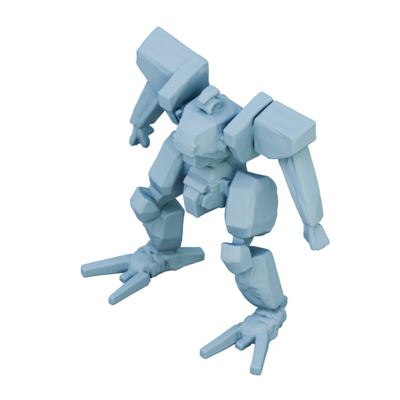} &  
\includegraphics[width=\qcwidth, valign=m]{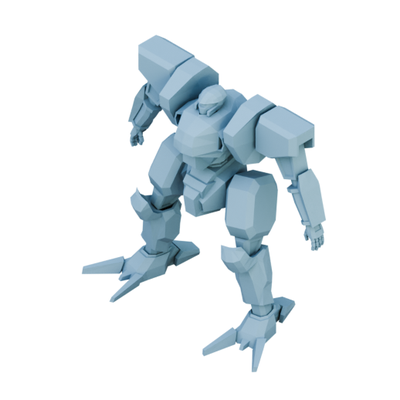} \\

\includegraphics[width=\qcwidth, valign=m]{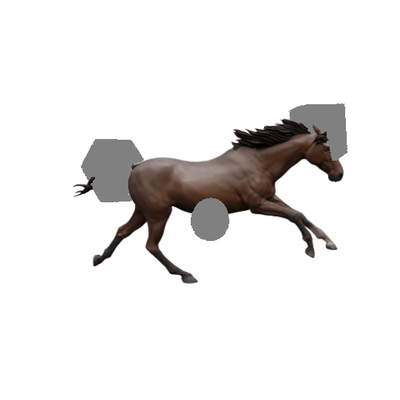} & 
\includegraphics[width=\qcwidth, valign=m]{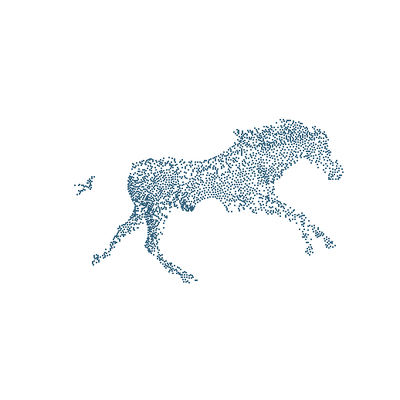} & 
\includegraphics[width=\qcwidth, valign=m]{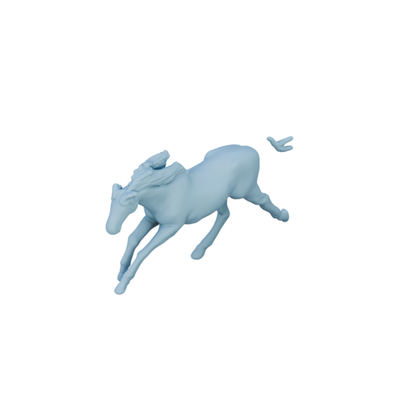} & 
\includegraphics[width=\qcwidth, valign=m]{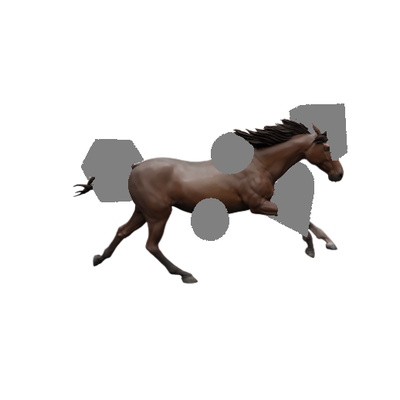} & 
\includegraphics[width=\qcwidth, valign=m]{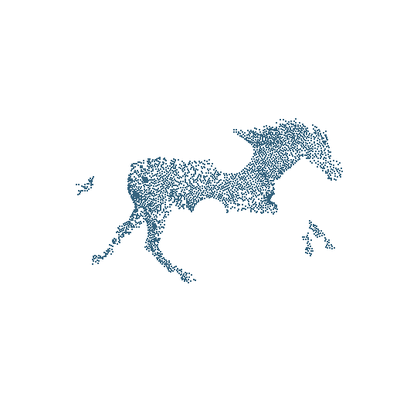} & 
\includegraphics[width=\qcwidth, valign=m]{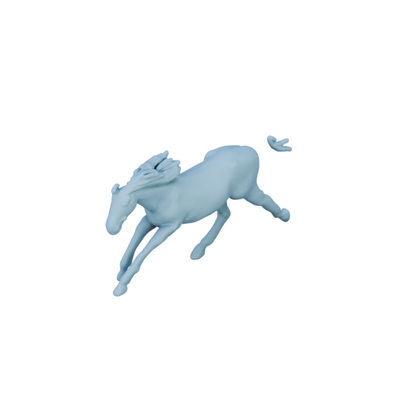} & 
\includegraphics[width=\qcwidth, valign=m]{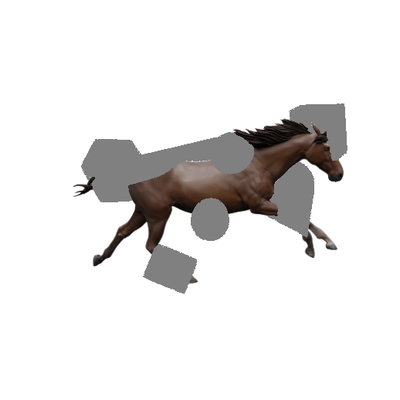} & 
\includegraphics[width=\qcwidth, valign=m]{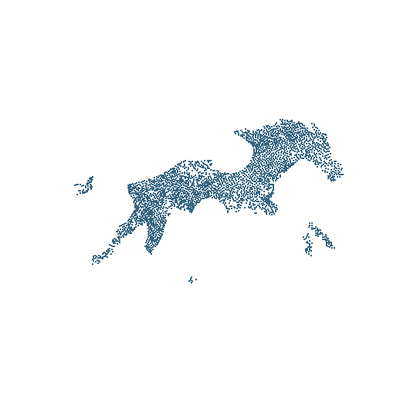} & 
\includegraphics[width=\qcwidth, valign=m]{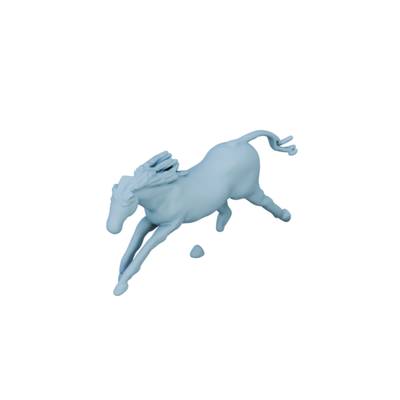} &  
\includegraphics[width=\qcwidth, valign=m]{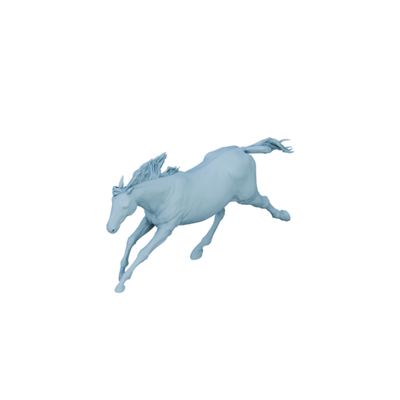} \\

  \end{tabular}
}
	\caption{\textbf{Completion Stress Test}: Model performance on challenging single- and multi-view examples with increasing amounts of occluders (3, 5, 7). Zoom in for details.}
	\label{fig:occlusions}
\end{figure*}

\endgroup

Fig.~\ref{fig:qual_compare_sv} and \ref{fig:qual_compare_mv} show qualitative results on challenging occlusion senarios (see Supplement for results without occlusion).
{\ourmodel} leverages conditioning points to produce faithful reconstructions, retaining details in known regions that other methods may simplify. 
With our data augmentation, our method learns to exploit the natural symmetry of objects, handling difficult regions---such as the deer in Fig.~\ref{fig:qual_compare_sv}---that others struggle to reconstruct.
In certain examples, baselines produce unwanted structures, whereas our method preserves structural consistency with the provided guidance.
Even in failure cases, when our method fails to complete all missing areas, our approach maintains high visual quality in the regions constrained by the conditioning points.

Fig.~\ref{fig:qual_compare_depth} evaluates robustness under single-view predicted depth. While SAM3D produces high-quality reconstructions, the resulting shapes often deviate from the input point geometry. Hy3D-Omni shows occasional inconsistencies in object orientation, likely due to the absence of explicit camera conditioning. ShapeR is the most sensitive to noise in the predicted depth points, leading to noticeable degradation in mesh quality. 
In contrast, our method produces reconstructions that are more consistent with the input observations, while remaining robust under moderate depth noise, though performance degrades in cases of severe depth estimation errors.

To stress test our method, we show examples under increasing amounts of occlusion in Fig.~\ref{fig:occlusions}. 
Our approach effectively handles severe occlusions by using information from similar unoccluded regions, as shown with the bike wheels. We also include challenging cases such as the horse, where thin structures sometimes fail to fully connect, highlighting areas for improvement. 

\subsection{Ablation Study}\label{ssec:ablation}

\begin{figure*}[!t]

\captionof{table}{\textbf{Ablation Study.} 
\begingroup
\setlength{\fboxsep}{0pt}
\colorbox{bestgreen}{\strut Best} and
\colorbox{secondyellow}{\strut second-best}
\endgroup
results are highlighted accordingly.
}

\centering 
\label{tab:ablation}

\begingroup
\setlength{\tabcolsep}{3pt} 

\definecolor{pmGray}{HTML}{6B6B6B}

\newcommand{\pmitem}[1]{%
    \textcolor{pmGray}{\scalebox{0.7}{$\pm$ #1}}%
}

\newcommand{\pmrow}[6]{%
\pmitem{#1} & 
\pmitem{#2} & 
\pmitem{#3} & 
\pmitem{#4} & 
\pmitem{#5} & 
\pmitem{#6} 
}

\resizebox{\linewidth}{!} {
\begin{tabular}{@{}l||lll|lll@{}}
\Xhline{1.0pt}
\multirow{2}{*}{Variant} & \multicolumn{3}{c|}{Single-View} & \multicolumn{3}{c}{Multi-View} \\

 & F-score $\uparrow$ & vIoU $\uparrow$ & CD$\downarrow_{\times10}$ & F-score $\uparrow$ & vIoU $\uparrow$ & CD$\downarrow_{\times10}$ \\

 \Xhline{1.0pt}

w/o Points &
0.8074 \pmitem{0.1738} &
0.2253 \pmitem{0.0969} &
0.3589 \pmitem{0.2993} &
0.9097 \pmitem{0.0989} &
0.2881 \pmitem{0.0971} &
0.2231 \pmitem{0.1344} \\

w/o Mask &
0.8968 \pmitem{0.1311} &
0.3225 \pmitem{0.1068} &
0.2517 \pmitem{0.2568} &
0.9655 \pmitem{0.0597} &
0.4151 \pmitem{0.1192} &
0.1463 \pmitem{0.0751} \\

w/o Cameras &
\cellcolor{secondyellow}0.9002 \pmitem{0.1224} &
\cellcolor{secondyellow}0.3228 \pmitem{0.1048} &
\cellcolor{secondyellow}0.2509 \pmitem{0.2414} &
\cellcolor{bestgreen}0.9689 \pmitem{0.0582} &
\cellcolor{secondyellow}0.4153 \pmitem{0.1149} &
\cellcolor{secondyellow}0.1462 \pmitem{0.0878} \\

\textbf{Full Model} &
\cellcolor{bestgreen}0.9036 \pmitem{0.1164} &
\cellcolor{bestgreen}0.3262 \pmitem{0.1039} &
\cellcolor{bestgreen}0.2423 \pmitem{0.2167} &
\cellcolor{secondyellow}0.9688 \pmitem{0.0535} &
\cellcolor{bestgreen}0.4185 \pmitem{0.1123} &
\cellcolor{bestgreen}0.1450 \pmitem{0.0685} \\

\Xhline{1.0pt}

\end{tabular}
}

\endgroup

\bigskip 

\begingroup
\setlength{\tabcolsep}{1pt}
    
\renewcommand{\arraystretch}{0}

\resizebox{\linewidth}{!} {
\begin{tabular}{@{}cc|cccc|c@{}}

\scalebox{0.5}{Input View} &
\scalebox{0.5}{Input Points} & 
\scalebox{0.5}{Without Points} & 
\scalebox{0.5}{Without Mask} & 
\scalebox{0.5}{Without Camera} & 
\scalebox{0.5}{\textbf{Full Model}} & 
\scalebox{0.5}{Ground Truth} \\

\includegraphics[width=\qcwidth, valign=m]{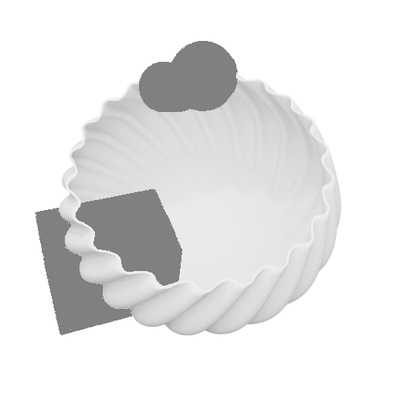} & 
\includegraphics[width=\qcwidth, valign=m]{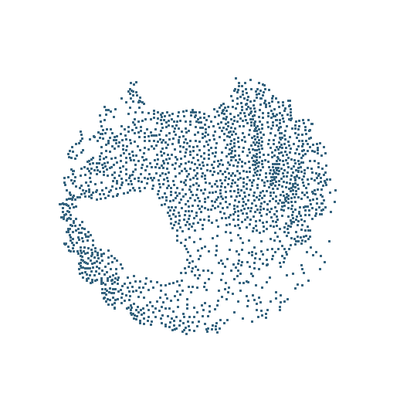} & 
\includegraphics[width=\qcwidth, valign=m]{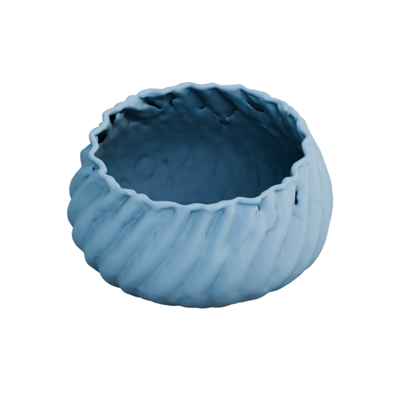} & 
\includegraphics[width=\qcwidth, valign=m]{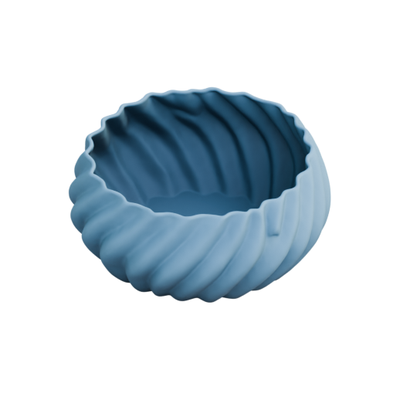} & 
\includegraphics[width=\qcwidth, valign=m]{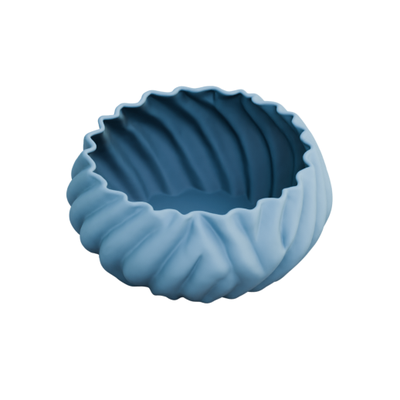} & 
\includegraphics[width=\qcwidth, valign=m]{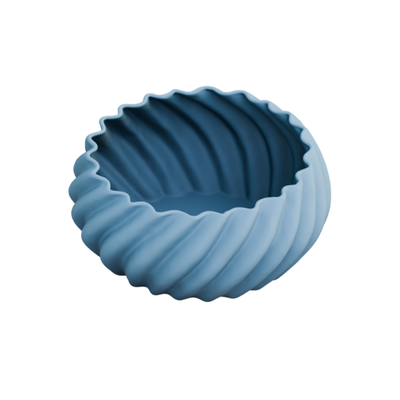} & 
\includegraphics[width=\qcwidth, valign=m]{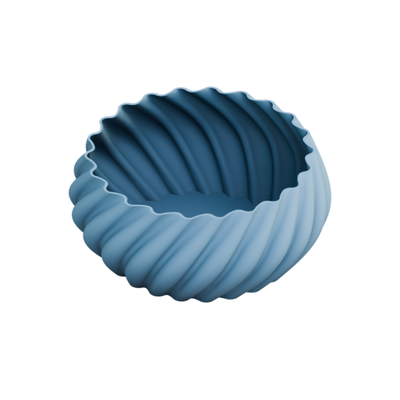} \\

\includegraphics[width=\qcwidth, valign=m]{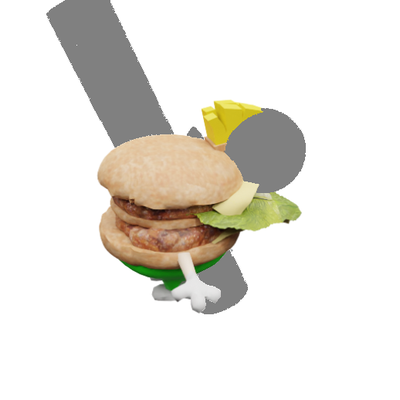} & 
\includegraphics[width=\qcwidth, valign=m]{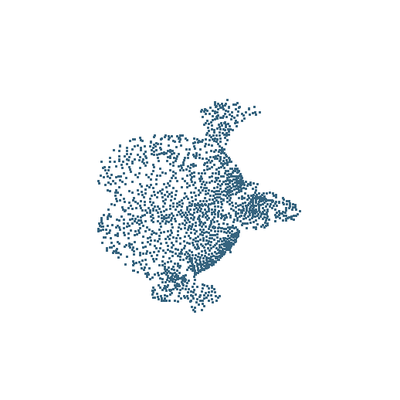} & 
\includegraphics[width=\qcwidth, valign=m]{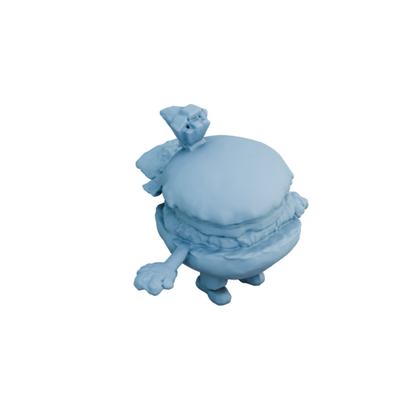} & 
\includegraphics[width=\qcwidth, valign=m]{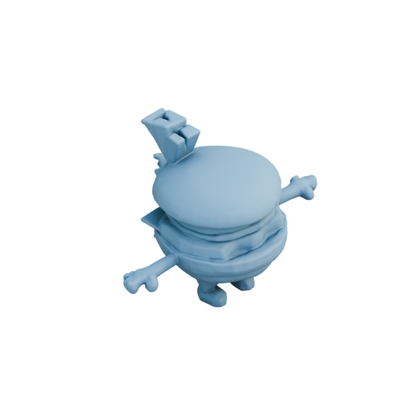} & 
\includegraphics[width=\qcwidth, valign=m]{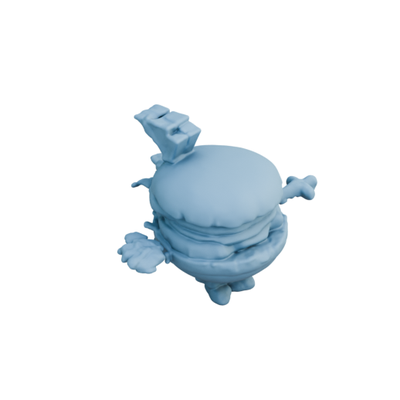} & 
\includegraphics[width=\qcwidth, valign=m]{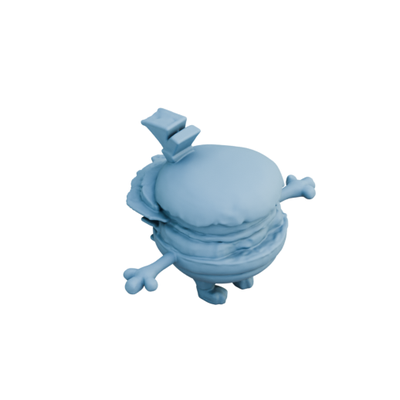} & 
\includegraphics[width=\qcwidth, valign=m]{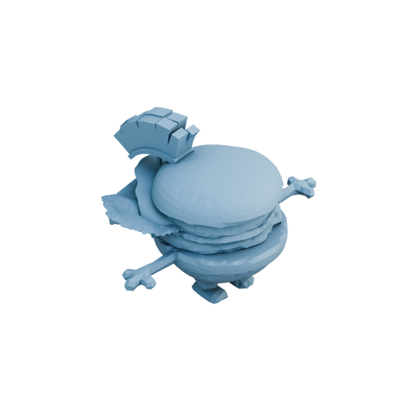} \\

\includegraphics[width=\qcwidth, valign=m]{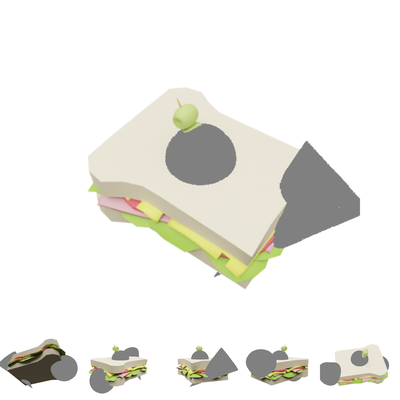} & 
\includegraphics[width=\qcwidth, valign=m]{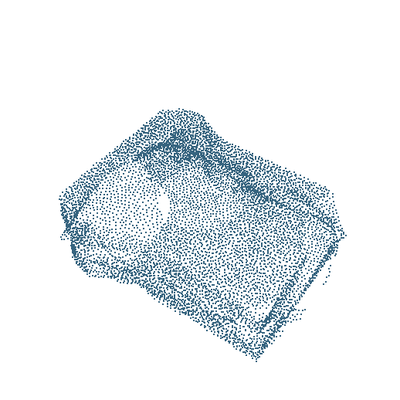} & 
\includegraphics[width=\qcwidth, valign=m]{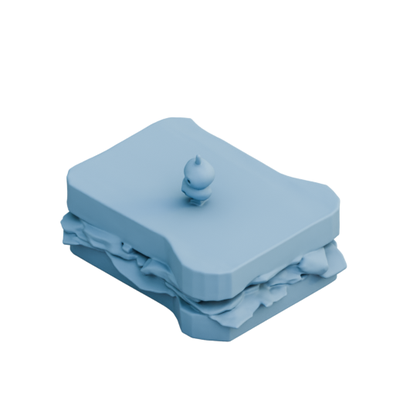} & 
\includegraphics[width=\qcwidth, valign=m]{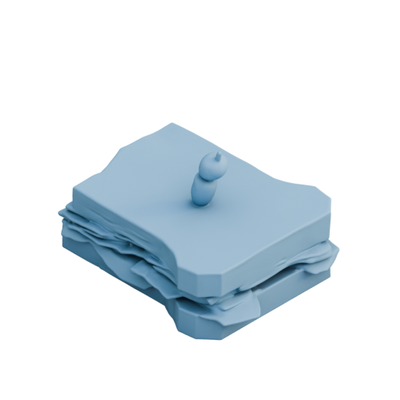} & 
\includegraphics[width=\qcwidth, valign=m]{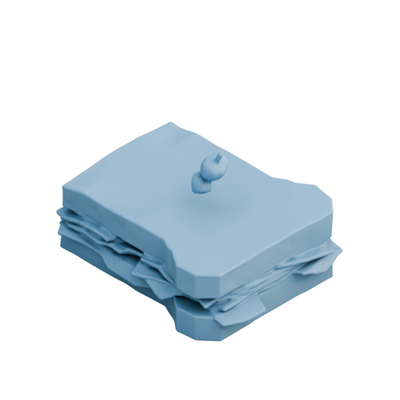} & 
\includegraphics[width=\qcwidth, valign=m]{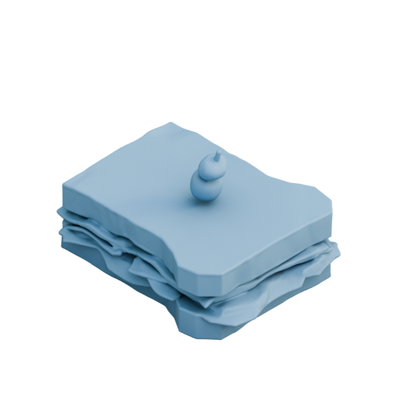} & 
\includegraphics[width=\qcwidth, valign=m]{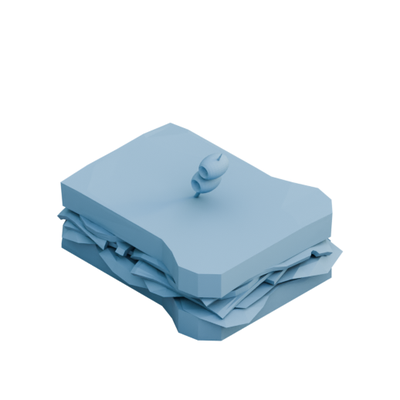} \\

  \end{tabular}
}
\captionof{figure}{\textbf{Ablation Examples}: Comparing completion results for the full model and ablated variants. Occluders are shown in grey in the input view. Zoom in for details.}
\label{fig:ablation_compare}

\endgroup

\end{figure*}

We conducted ablations to analyze the contribution of the various components of {\ourmodel} and report the results
in Tb.~\ref{tab:ablation} and Fig.~\ref{fig:ablation_compare}. For these experiments, we evaluate using Toys4k dataset under single- and multi-view occlusion scenarios. 

\subsubsection{Effect of Point Condition:} To analysis the importance of geometric conditioning, we remove the point condition along with multi-modal fusion layers, only passing in the output from visual encoding to the cross-attention layers (see \S\ref{ssec:model}). Removing the point condition leads to the largest drop in quantitative performance and visibly poorer completed objects, indicating that explicit geometric conditioning is crucial for maintaining accuracy in known regions and producing overall more consistent and complete 3D geometry.

\subsubsection{Effect of Masked Attention:} To assess the contribution of visibility-aware attention, we remove the attention bias from image masks applied to cross-attention layers. 
Removing the masking causes a small quantitative drop, while visual differences remain minimal, mostly appearing as subtle changes to global geometry such as slight thickness variations. Fine surface details and appearance remain largely unchanged, as these are captured by the unmasked image tokens. This suggests that masking primarily serves as a safeguard against potential interference from occluded regions, influencing global structural reasoning rather than perceptually noticeable features.

\subsubsection{Effect of Camera Condition:} To examine the impact of explicit camera parameterization, we replace camera Pl\"ucker embeddings with per-view learnable embeddings to differentiate between the patch tokens from each view. 
In multi-view settings, metrics remain close to the full model, suggesting the network can largely infer relative camera positions from visual cues when multiple views are available. In single-view scenarios, performance degrades more noticeably, as the model struggles to align 3D points with image features without explicit camera information. This highlights the importance of camera conditioning for reliable object completion under sparse observations.

\subsection{Applications}\label{ssec:applications}

\begin{figure*}[!tb]
  \centering

  \begin{subfigure}{\linewidth}
    \centering

\resizebox{\linewidth}{!} {
\begin{tabular}{@{}cccc|cccc@{}}

\scalebox{0.5}{Original View} &
\scalebox{0.5}{Edited View} & 
\scalebox{0.5}{Edited Points} & 
\scalebox{0.5}{Output} &
\scalebox{0.5}{Original View} &
\scalebox{0.5}{Edited View} & 
\scalebox{0.5}{Edited Points} & 
\scalebox{0.5}{Output} \\

\includegraphics[width=\qcwidth, valign=m]{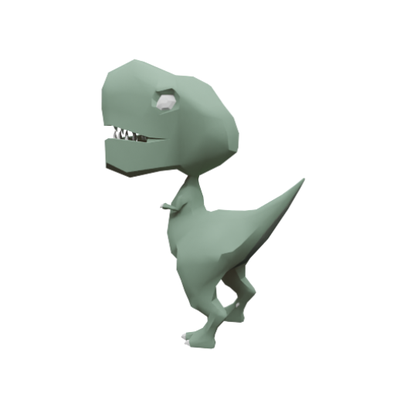} & 
\includegraphics[width=\qcwidth, valign=m]{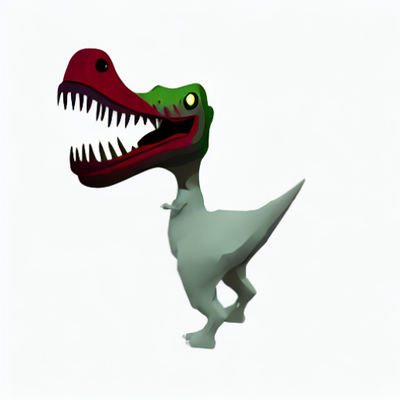} & 
\includegraphics[width=\qcwidth, valign=m]{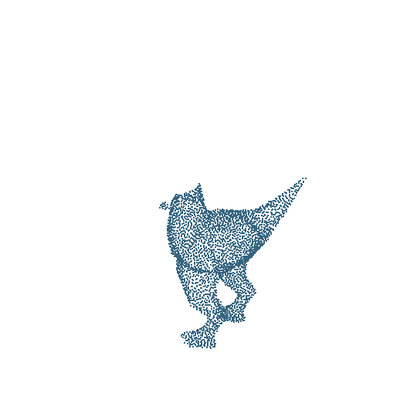} & 
\includegraphics[width=\qcwidth, valign=m]{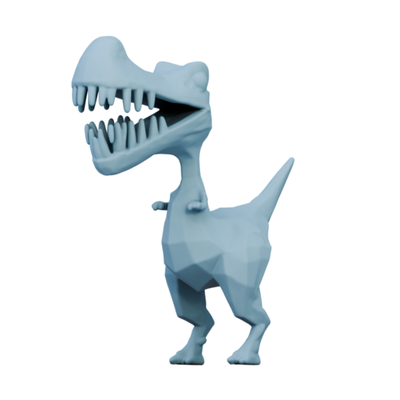} &

\includegraphics[width=\qcwidth, valign=m]{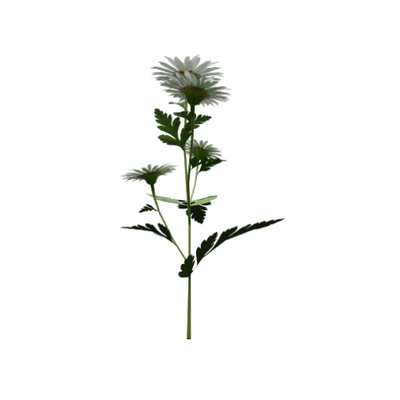} & 
\includegraphics[width=\qcwidth, valign=m]{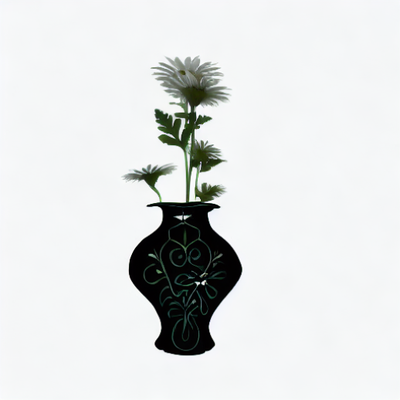} & 
\includegraphics[width=\qcwidth, valign=m]{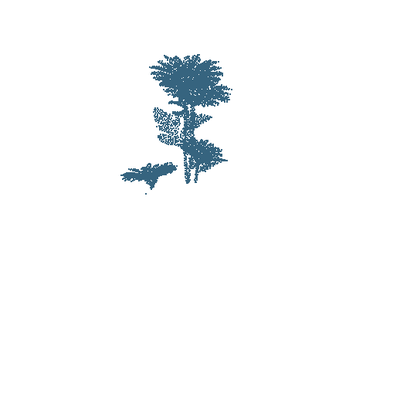} & 
\includegraphics[width=\qcwidth, valign=m]{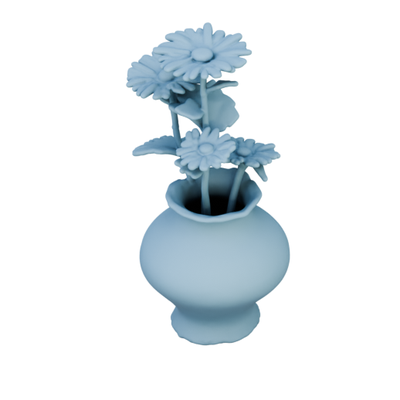} \\

\includegraphics[width=\qcwidth, valign=m]{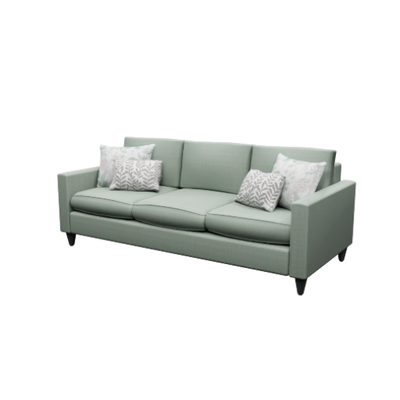} & 
\includegraphics[width=\qcwidth, valign=m]{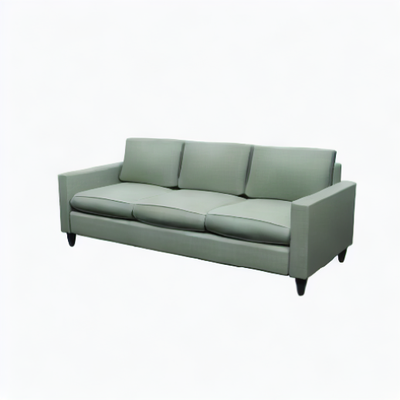} & 
\includegraphics[width=\qcwidth, valign=m]{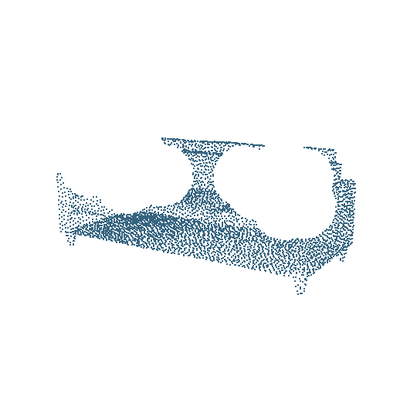} & 
\includegraphics[width=\qcwidth, valign=m]{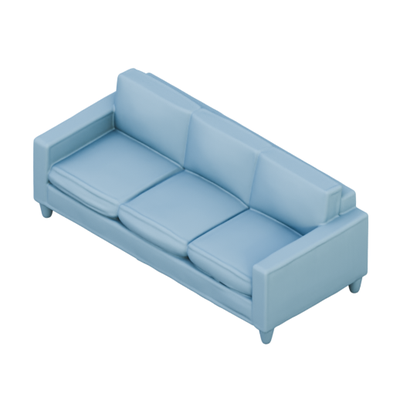} &

\includegraphics[width=\qcwidth, valign=m]{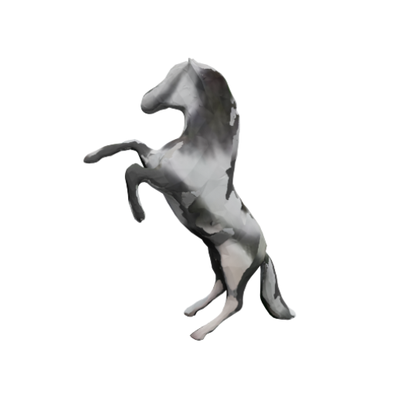} & 
\includegraphics[width=\qcwidth, valign=m]{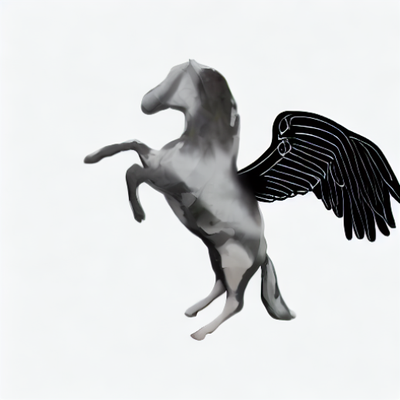} & 
\includegraphics[width=\qcwidth, valign=m]{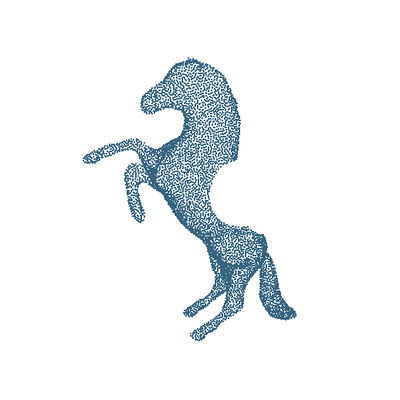} & 
\includegraphics[width=\qcwidth, valign=m]{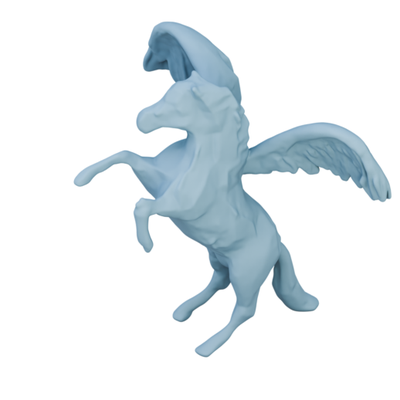} \\

\includegraphics[width=\qcwidth, valign=m]{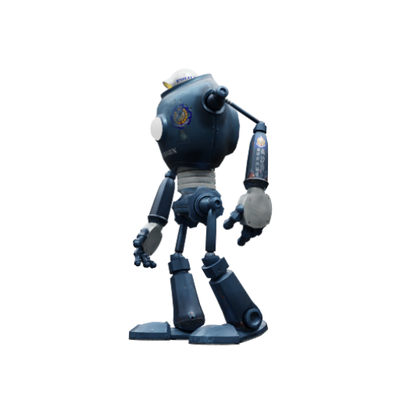} & 
\includegraphics[width=\qcwidth, valign=m]{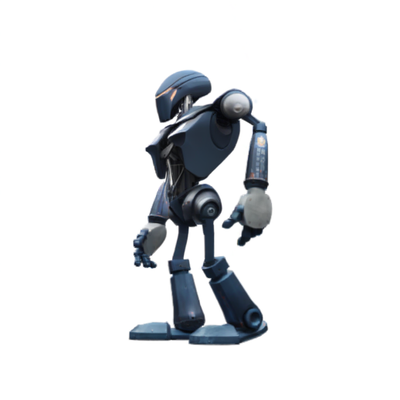} & 
\includegraphics[width=\qcwidth, valign=m]{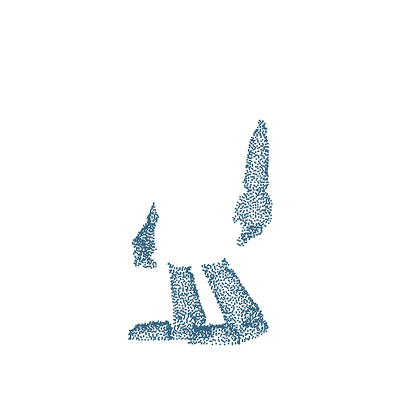} & 
\includegraphics[width=\qcwidth, valign=m]{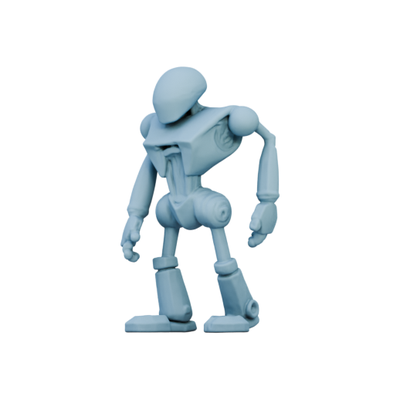} &

\includegraphics[width=\qcwidth, valign=m]{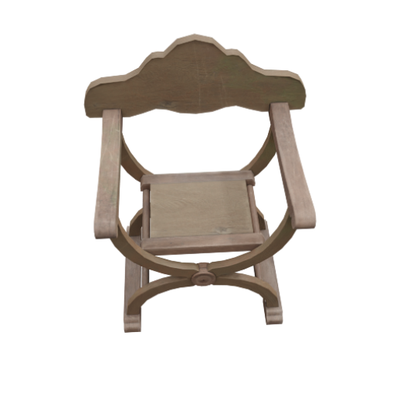} & 
\includegraphics[width=\qcwidth, valign=m]{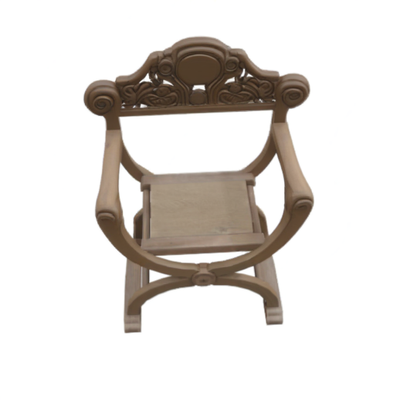} & 
\includegraphics[width=\qcwidth, valign=m]{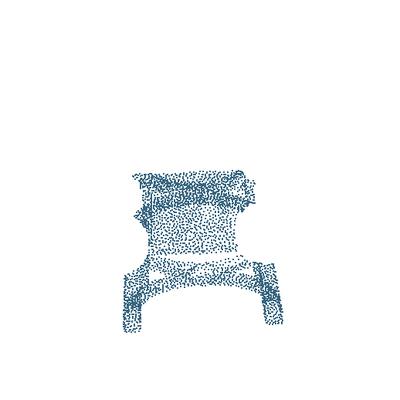} & 
\includegraphics[width=\qcwidth, valign=m]{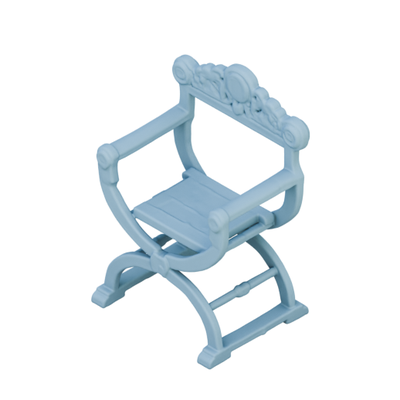} \\

  \end{tabular}
}
  
    \caption{\textbf{Shape editing} with user edited single-view image through image inpainting.}
    \label{fig:edit}
  \end{subfigure}

  \vspace{2mm}

  \begin{subfigure}{\linewidth}

  \centering

\resizebox{\linewidth}{!} {
\begin{tabular}{@{}ccc|ccc|ccc@{}}

\scalebox{0.5}{Input View} &
\scalebox{0.5}{Input Points} & 
\scalebox{0.5}{Output} & 
\scalebox{0.5}{Input View} &
\scalebox{0.5}{Input Points} & 
\scalebox{0.5}{Output} & 
\scalebox{0.5}{Input View} &
\scalebox{0.5}{Input Points} & 
\scalebox{0.5}{Output} \\ 

\includegraphics[width=\qcwidth, valign=m]{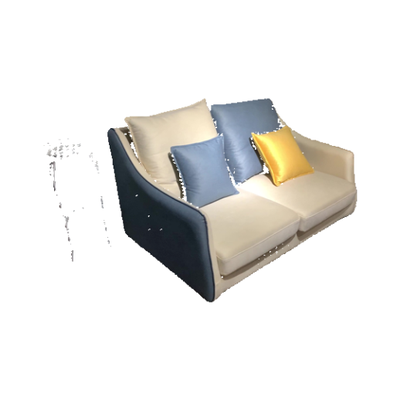} & 
\includegraphics[width=\qcwidth, valign=m]{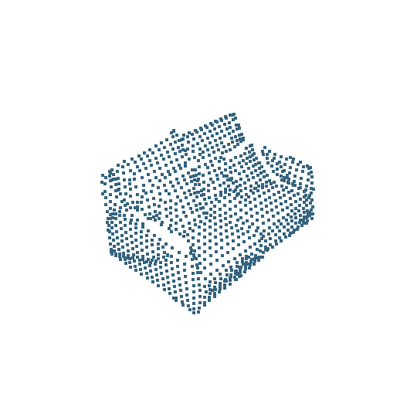} & 
\includegraphics[width=\qcwidth, valign=m]{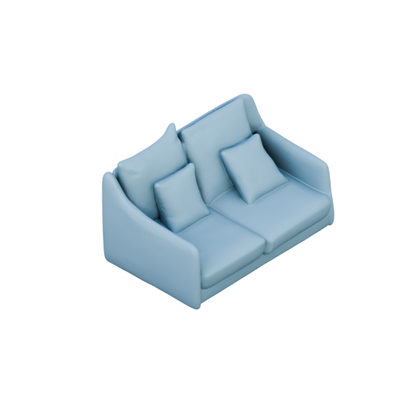} & 
\includegraphics[width=\qcwidth, valign=m]{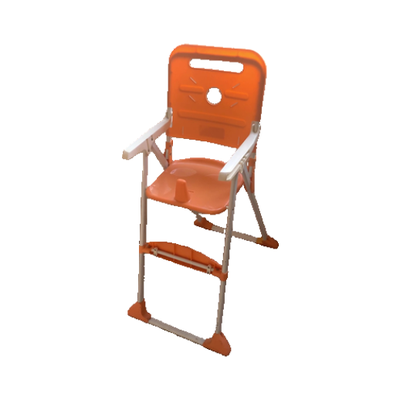} & 
\includegraphics[width=\qcwidth, valign=m]{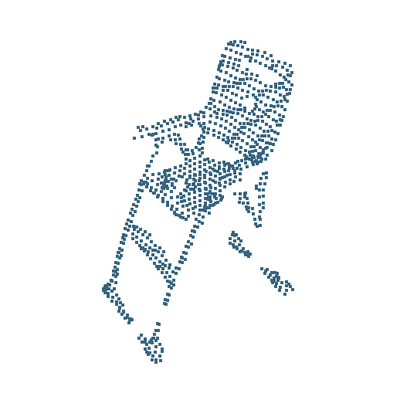} & 
\includegraphics[width=\qcwidth, valign=m]{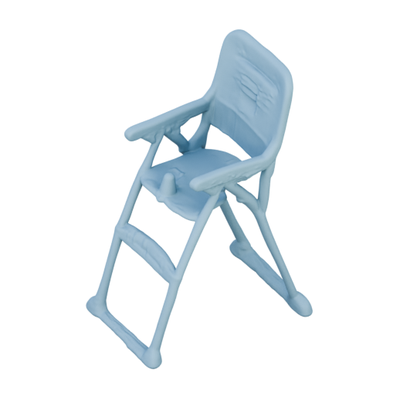} & 
\includegraphics[width=\qcwidth, valign=m]{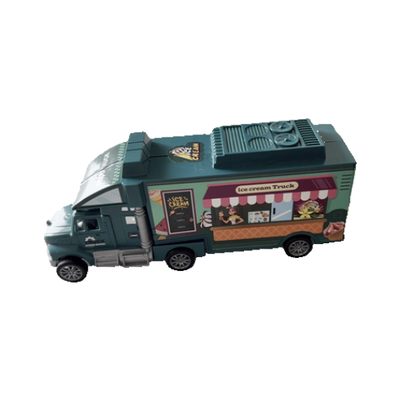} & 
\includegraphics[width=\qcwidth, valign=m]{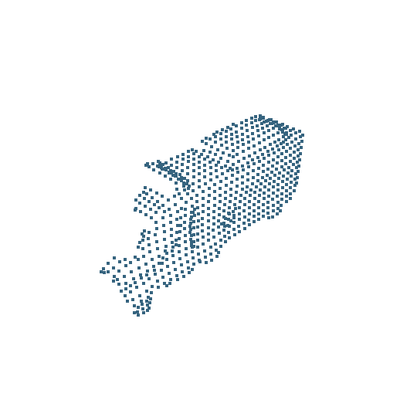} & 
\includegraphics[width=\qcwidth, valign=m]{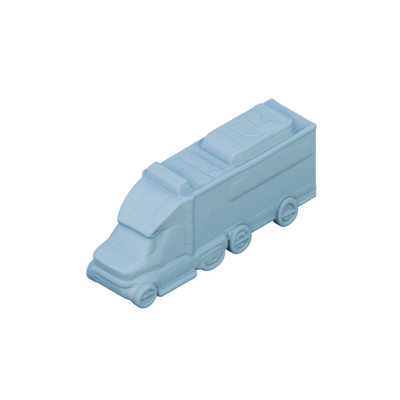} \\

\includegraphics[width=\qcwidth, valign=m]{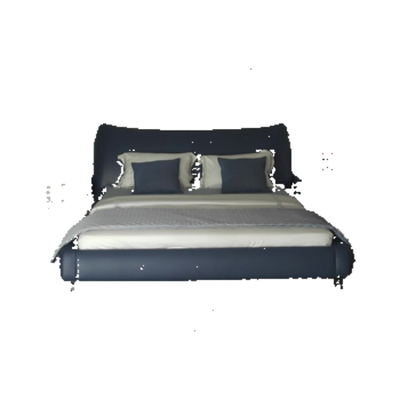} & 
\includegraphics[width=\qcwidth, valign=m]{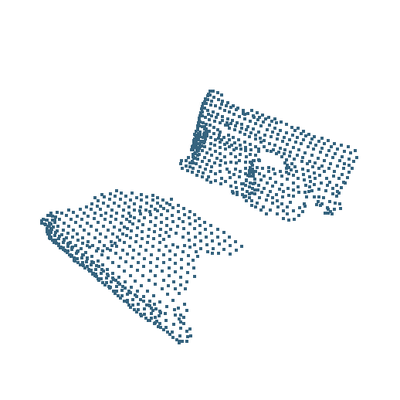} & 
\includegraphics[width=\qcwidth, valign=m]{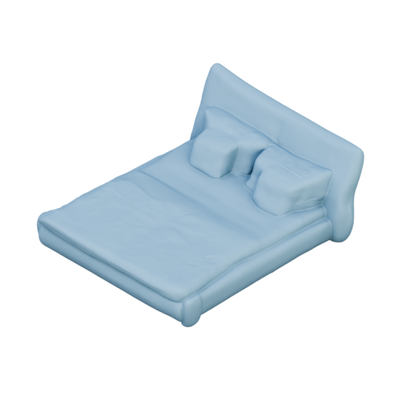} & 
\includegraphics[width=\qcwidth, valign=m]{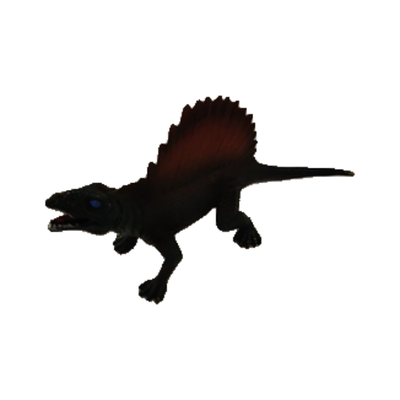} & 
\includegraphics[width=\qcwidth, valign=m]{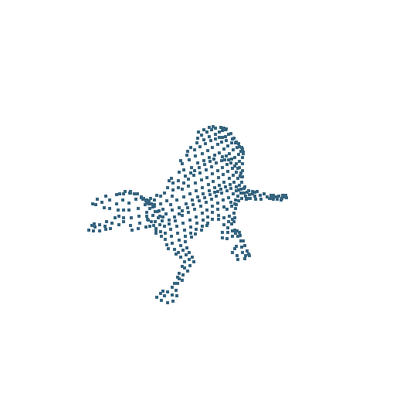} & 
\includegraphics[width=\qcwidth, valign=m]{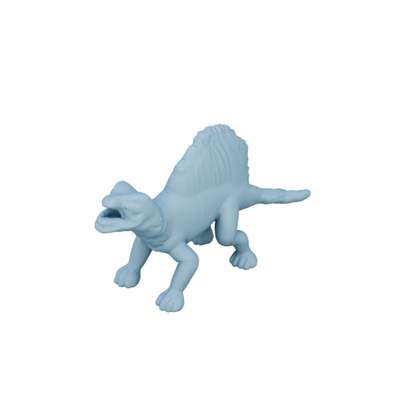} & 
\includegraphics[width=\qcwidth, valign=m]{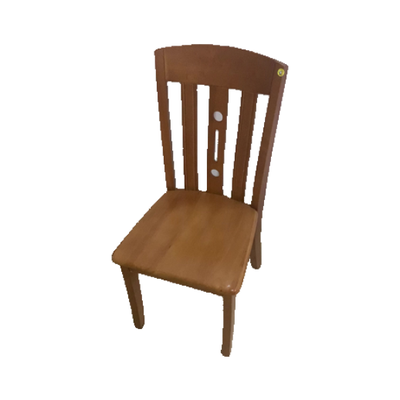} & 
\includegraphics[width=\qcwidth, valign=m]{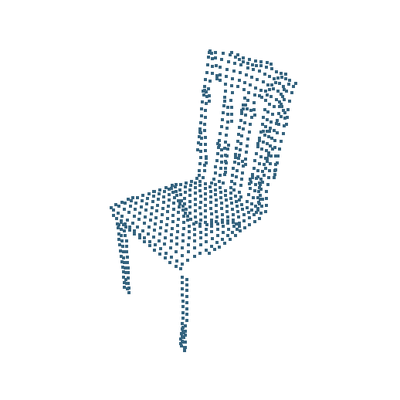} & 
\includegraphics[width=\qcwidth, valign=m]{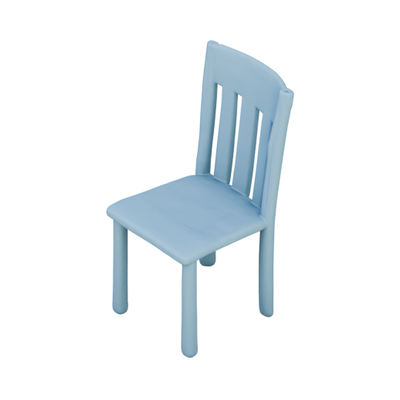} \\

  \end{tabular}
}

    \caption{\textbf{Image to 3D} with SAM 2 \cite{ravi2025sam2} segmentation and MapAnything \cite{keetha2026mapanything} outputs.}
    \label{fig:mapanything}
  \end{subfigure}

  \caption{\textbf{Applications:} {\ourmodel} enables precise object edits and accurate single-view generation with completion of missing regions. Zoom in for details.}
  \label{fig:applications}
\end{figure*}

\subsubsection{Shape Editing:} 
Our method can also be effectively applied for 3D shape editing (Fig.\ref{fig:edit}). The user selects a view, masks and edits it, e.g.\ using an image inpainting model like \cite{rombach2022stablediffusion}. This forms the single-view condition to our model. To provide conditioning points, we sample the surface of the original shape and discard points that project into the masked region of the edited view. Fig.\ref{fig:edit} shows effective 3D edits that remain faithful to the unedited parts of the shape.

\subsubsection{Image to 3D:}
While our method is trained on points sampled from meshes, it can be combined with approaches like \cite{keetha2026mapanything} to extend its usability to scenarios where points are not available. For example, to produce 3D shapes from a single image (Fig.\ref{fig:mapanything}), 
we segment the image using SAM 2 \cite{ravi2025sam2} and estimate object points and camera parameters using MapAnything \cite{keetha2026mapanything}. The object points are aligned and normalized, before being passed to our method (see Supplement), resulting in faithful 3D generations from a single view (Fig.\ref{fig:mapanything}).

\subsubsection{Simulation:}
Accurate and faithful completion is also important for physical applications, such as physics simulation of segmented captured scenes, as shown in \cite{tsang2026artisangs,dagli2025vomp}. See Supplement for examples of this application.

\section{Conclusion}

We presented {\ourmodel}, a unified framework for multi-modal, occlusion-aware 3D generation that completes and edits objects from partial observations. By jointly leveraging images, visibility masks, camera parameters, and partial point clouds, our model enforces geometric consistency while integrating visual and 3D evidence. While previous methods focus on specific tasks in isolation, {\ourmodel} provides a single framework capable of handling diverse reconstruction and editing scenarios. Experiments show state-of-the-art performance in sparse-view conditions with and without occlusions. These results highlight its potential for real-world 3D reconstruction and controllable geometry-aware editing. While {\ourmodel} is trained to faithfully follow input points, future work could explore balancing this fidelity with robustness to noisy point clouds, as well as supporting more diverse camera positions and textures.

\bibliographystyle{splncs04}
\bibliography{main}

\begin{thebibliography}{10}
\providecommand{\url}[1]{\texttt{#1}}
\providecommand{\urlprefix}{URL }
\providecommand{\doi}[1]{https://doi.org/#1}

\bibitem{barda2025instant3dit}
Barda, A., Gadelha, M., Kim, V.G., Aigerman, N., Bermano, A.H., Groueix, T.:
  Instant3dit: Multiview inpainting for fast editing of 3d objects. In:
  Proceedings of the IEEE/CVF Conference on Computer Vision and Pattern
  Recognition (CVPR). pp. 16273--16282 (June 2025)

\bibitem{besl1992method}
Besl, P.J., McKay, N.D.: Method for registration of 3-d shapes. In: Sensor
  fusion IV: control paradigms and data structures. vol.~1611, pp. 586--606.
  Spie (1992)

\bibitem{chang2015shapenet}
Chang, A.X., Funkhouser, T., Guibas, L., Hanrahan, P., Huang, Q., Li, Z.,
  Savarese, S., Savva, M., Song, S., Su, H., et~al.: Shapenet: An
  information-rich 3d model repository. arXiv preprint arXiv:1512.03012  (2015)

\bibitem{chang2026reconviagen}
Chang, J., Ye, C., Wu, Y., Chen, Y., Zhang, Y., Luo, Z., Li, C., Zhi, Y., Han,
  X.: Reconviagen: Towards accurate multi-view 3d object reconstruction via
  generation. In: The Fourteenth International Conference on Learning
  Representations (2026), \url{https://openreview.net/forum?id=z0QLeooEEf}

\bibitem{chen2025sam3d}
Chen, X., Chu, F.J., Gleize, P., Liang, K.J., Sax, A., Tang, H., Wang, W., Guo,
  M., Hardin, T., Li, X., et~al.: Sam 3d: 3dfy anything in images. arXiv
  preprint arXiv:2511.16624  (2025)

\bibitem{collins2022abo}
Collins, J., Goel, S., Deng, K., Luthra, A., Xu, L., Gundogdu, E., Zhang, X.,
  Vicente, T.F.Y., Dideriksen, T., Arora, H., Guillaumin, M., Malik, J.: Abo:
  Dataset and benchmarks for real-world 3d object understanding. In:
  Proceedings of the IEEE/CVF Conference on Computer Vision and Pattern
  Recognition (CVPR). pp. 21126--21136 (June 2022)

\bibitem{dagli2025vomp}
Dagli, R., Xiang, D., Modi, V., Loop, C., Tsang, C.F., Chen, A.H., Hu, A.,
  State, G., Levin, D.I., Shugrina, M.: Vomp: Predicting volumetric mechanical
  property fields. arXiv preprint arXiv:2510.22975  (2025)

\bibitem{deitke2023objaverseXL}
Deitke, M., Liu, R., Wallingford, M., Ngo, H., Michel, O., Kusupati, A., Fan,
  A., Laforte, C., Voleti, V., Gadre, S.Y., et~al.: Objaverse-xl: A universe of
  10m+ 3d objects. Advances in Neural Information Processing Systems
  \textbf{36},  35799--35813 (2023)

\bibitem{engel2018slam}
Engel, J., Koltun, V., Cremers, D.: Direct sparse odometry. IEEE Transactions
  on Pattern Analysis and Machine Intelligence  \textbf{40}(3),  611--625
  (2018). \doi{10.1109/TPAMI.2017.2658577}

\bibitem{erkoc2025preditor3d}
Erko\c{c}, Z., G\"umeli, C., Wang, C., Nie{\ss}ner, M., Dai, A., Wonka, P.,
  Lee, H.Y., Zhuang, P.: Preditor3d: Fast and precise 3d shape editing. In:
  Proceedings of the Computer Vision and Pattern Recognition Conference (CVPR).
  pp. 640--649 (June 2025)

\bibitem{feng2025objfiller3d}
Feng, H., Liu, J., Tang, J., Wu, G., Chen, B., Lai, J., Wang, G.: Objfiller-3d:
  Consistent multi-view 3d inpainting via video diffusion models. arXiv
  preprint arXiv:2508.18271  (2025)

\bibitem{fu20213dfuture}
Fu, H., Jia, R., Gao, L., Gong, M., Zhao, B., Maybank, S., Tao, D.: 3d-future:
  3d furniture shape with texture. International Journal of Computer Vision
  \textbf{129}(12),  3313--3337 (2021). \doi{10.1007/s11263-021-01534-z}

\bibitem{tsang2026artisangs}
{Fuji Tsang}, C., Hu, A., Perel, O., Kolve, C., Shugrina, M.: Artisangs:
  Interactive tools for gaussian splat selection with ai and human in the loop.
  arXiv preprint arXiv:2602.10173  (2026)

\bibitem{grattafiori2024llama3}
Grattafiori, A., Dubey, A., Jauhri, A., Pandey, A., Kadian, A., Al-Dahle, A.,
  Letman, A., Mathur, A., Schelten, A., Vaughan, A., et~al.: The llama 3 herd
  of models. arXiv preprint arXiv:2407.21783  (2024)

\bibitem{he2015delving}
He, K., Zhang, X., Ren, S., Sun, J.: Delving deep into rectifiers: Surpassing
  human-level performance on imagenet classification. In: Proceedings of the
  IEEE International Conference on Computer Vision (ICCV). pp. 1026--1034
  (2015). \doi{10.1109/ICCV.2015.123}

\bibitem{ho2020ddpm}
Ho, J., Jain, A., Abbeel, P.: Denoising diffusion probabilistic models. In:
  Advances in Neural Information Processing Systems. vol.~33, pp. 6840--6851
  (2020),
  \url{https://proceedings.neurips.cc/paper_files/paper/2020/file/4c5bcfec8584af0d967f1ab10179ca4b-Paper.pdf}

\bibitem{ho2021cfg}
Ho, J., Salimans, T.: Classifier-free diffusion guidance. In: NeurIPS 2021
  Workshop on Deep Generative Models and Downstream Applications (2021),
  \url{https://openreview.net/forum?id=qw8AKxfYbI}

\bibitem{hsiao2026vecsetedit}
Hsiao, T.F., Ruan, B.K., Liu, Y.L., Shuai, H.H.: Vecset-edit: Unleashing
  pre-trained lrm for mesh editing from single image. arXiv preprint
  arXiv:2602.04349  (2026)

\bibitem{huang2025midi}
Huang, Z., Guo, Y.C., An, X., Yang, Y., Li, Y., Zou, Z.X., Liang, D., Liu, X.,
  Cao, Y.P., Sheng, L.: Midi: Multi-instance diffusion for single image to 3d
  scene generation. In: Proceedings of the Computer Vision and Pattern
  Recognition Conference (CVPR). pp. 23646--23657 (June 2025)

\bibitem{hunyuan3d2025hunyuan3d}
Hunyuan3D, T., Yang, S., Yang, M., Feng, Y., Huang, X., Zhang, S., He, Z., Luo,
  D., Liu, H., Zhao, Y., et~al.: Hunyuan3d 2.1: From images to high-fidelity 3d
  assets with production-ready pbr material. arXiv preprint arXiv:2506.15442
  (2025)

\bibitem{hunyuan3d2025hunyuan3domni}
Hunyuan3D, T., Zhang, B., Guo, C., Liu, H., Yan, H., Shi, H., Huang, J., Yu,
  J., Li, K., Wang, P., et~al.: Hunyuan3d-omni: A unified framework for
  controllable generation of 3d assets. arXiv preprint arXiv:2509.21245  (2025)

\bibitem{keetha2026mapanything}
Keetha, N., M\"uller, N., Sch\"onberger, J., Porzi, L., Zhang, Y., Fischer, T.,
  Knapitsch, A., Zauss, D., Weber, E., Antunes, N., Luiten, J.,
  Lopez-Antequera, M., Bul\`o, S.R., Richardt, C., Ramanan, D., Scherer, S.,
  Kontschieder, P.: {MapAnything}: Universal feed-forward metric {3D}
  reconstruction. In: International Conference on 3D Vision (3DV). IEEE (2026)

\bibitem{kerbl2023gsplat}
Kerbl, B., Kopanas, G., Leimkuehler, T., Drettakis, G.: 3d gaussian splatting
  for real-time radiance field rendering. ACM Transactions On Graphics (TOG)
  \textbf{42}(4) (2023). \doi{10.1145/3592433},
  \url{https://doi.org/10.1145/3592433}

\bibitem{kerbl3Dgaussians}
Kerbl, B., Kopanas, G., Leimk{\"u}hler, T., Drettakis, G.: 3d gaussian
  splatting for real-time radiance field rendering. ACM Transactions On
  Graphics (TOG)  \textbf{42}(4) (July 2023),
  \url{https://repo-sam.inria.fr/fungraph/3d-gaussian-splatting/}

\bibitem{khanna2024hssd}
Khanna, M., Mao, Y., Jiang, H., Haresh, S., Shacklett, B., Batra, D., Clegg,
  A., Undersander, E., Chang, A.X., Savva, M.: Habitat synthetic scenes dataset
  (hssd-200): An analysis of 3d scene scale and realism tradeoffs for
  objectgoal navigation. In: Proceedings of the IEEE/CVF Conference on Computer
  Vision and Pattern Recognition (CVPR). pp. 16384--16393 (June 2024)

\bibitem{kong2024eschernet}
Kong, X., Liu, S., Lyu, X., Taher, M., Qi, X., Davison, A.J.: Eschernet: A
  generative model for scalable view synthesis. In: Proceedings of the IEEE/CVF
  Conference on Computer Vision and Pattern Recognition (CVPR). pp. 9503--9513
  (June 2024)

\bibitem{lin2026depthanything}
Lin, H., Chen, S., Liew, J.H., Chen, D.Y., Li, Z., Zhao, Y., Peng, S., Guo, H.,
  Zhou, X., Shi, G., Feng, J., Kang, B.: Depth anything 3: Recovering the
  visual space from any views. In: The Fourteenth International Conference on
  Learning Representations (2026),
  \url{https://openreview.net/forum?id=yirunib8l8}

\bibitem{liu2025one2345pp}
Liu, M., Shi, R., Chen, L., Zhang, Z., Xu, C., Wei, X., Chen, H., Zeng, C., Gu,
  J., Su, H.: One-2-3-45++: Fast single image to 3d objects with consistent
  multi-view generation and 3d diffusion. In: Proceedings of the IEEE/CVF
  Conference on Computer Vision and Pattern Recognition (CVPR). pp.
  10072--10083 (June 2024)

\bibitem{loshchilov2018adamw}
Loshchilov, I., Hutter, F.: Decoupled weight decay regularization. In:
  International Conference on Learning Representations (ICLR) (2019),
  \url{https://openreview.net/forum?id=Bkg6RiCqY7}

\bibitem{modi2025lab}
Modi, V., Fuji~Tsang, C., Shugrina, M., Daviet, G.: Tackling gaussian splats,
  physics simulation and visualization with nvidia kaolin and warp libraries.
  In: Proceedings of the Special Interest Group on Computer Graphics and
  Interactive Techniques Conference Labs. SIGGRAPH Labs '25, Association for
  Computing Machinery, New York, NY, USA (2025). \doi{10.1145/3721251.3736529},
  \url{https://doi.org/10.1145/3721251.3736529}

\bibitem{modi2024simplicits}
Modi, V., Sharp, N., Perel, O., Sueda, S., Levin, D.I.: Simplicits: Mesh-free,
  geometry-agnostic elastic simulation. ACM Transactions on Graphics (TOG)
  \textbf{43}(4),  1--11 (2024)

\bibitem{moenne20243d}
Moenne-Loccoz, N., Mirzaei, A., Perel, O., de~Lutio, R., Martinez~Esturo, J.,
  State, G., Fidler, S., Sharp, N., Gojcic, Z.: 3d gaussian ray tracing: Fast
  tracing of particle scenes. ACM Transactions on Graphics (TOG)
  \textbf{43}(6),  1--19 (2024)

\bibitem{oquab2023dinov2}
Oquab, M., Darcet, T., Moutakanni, T., Vo, H., Szafraniec, M., Khalidov, V.,
  Fernandez, P., Haziza, D., Massa, F., El-Nouby, A., et~al.: Dinov2: Learning
  robust visual features without supervision. arXiv preprint arXiv:2304.07193
  (2023)

\bibitem{ozguroglu2024pix2gest}
Ozguroglu, E., Liu, R., Sur{\'\i}s, D., Chen, D., Dave, A., Tokmakov, P.,
  Vondrick, C.: pix2gestalt: Amodal segmentation by synthesizing wholes. In:
  Proceedings of the IEEE/CVF Conference on Computer Vision and Pattern
  Recognition (CVPR). pp. 3931--3940 (June 2024)

\bibitem{qi2017fps}
Qi, C.R., Yi, L., Su, H., Guibas, L.J.: Pointnet++: Deep hierarchical feature
  learning on point sets in a metric space. In: Advances in Neural Information
  Processing Systems. vol.~30 (2017),
  \url{https://proceedings.neurips.cc/paper_files/paper/2017/file/d8bf84be3800d12f74d8b05e9b89836f-Paper.pdf}

\bibitem{qu2025deocc}
Qu, Y., Dai, S., Li, X., Wang, Y., Shen, Y., Cao, L., Ji, R.: Deocc-1-to-3: 3d
  de-occlusion from a single image via self-supervised multi-view diffusion.
  arXiv preprint arXiv:2506.21544  (2025)

\bibitem{ravi2025sam2}
Ravi, N., Gabeur, V., Hu, Y.T., Hu, R., Ryali, C., Ma, T., Khedr, H.,
  R{\"a}dle, R., Rolland, C., Gustafson, L., Mintun, E., Pan, J., Alwala, K.V.,
  Carion, N., Wu, C.Y., Girshick, R., Dollar, P., Feichtenhofer, C.: {SAM} 2:
  Segment anything in images and videos. In: The Thirteenth International
  Conference on Learning Representations (2025),
  \url{https://openreview.net/forum?id=Ha6RTeWMd0}

\bibitem{rombach2022stablediffusion}
Rombach, R., Blattmann, A., Lorenz, D., Esser, P., Ommer, B.: High-resolution
  image synthesis with latent diffusion models. In: Proceedings of the IEEE/CVF
  Conference on Computer Vision and Pattern Recognition (CVPR). pp.
  10684--10695 (June 2022)

\bibitem{schonberger2016sfm}
Schonberger, J.L., Frahm, J.M.: Structure-from-motion revisited. In:
  Proceedings of the IEEE Conference on Computer Vision and Pattern Recognition
  (CVPR) (June 2016)

\bibitem{siddiqui2026shaper}
Siddiqui, Y., Frost, D., Aroudj, S., Avetisyan, A., Howard-Jenkins, H., DeTone,
  D., Moulon, P., Wu, Q., Li, Z., Straub, J., Newcombe, R., Engel, J.: Shaper:
  Robust conditional 3d shape generation from casual captures. arXiv preprint
  arXiv:2601.11514  (2026)

\bibitem{sitzmann2021plucker}
Sitzmann, V., Rezchikov, S., Freeman, B., Tenenbaum, J., Durand, F.: Light
  field networks: Neural scene representations with single-evaluation
  rendering. In: Advances in Neural Information Processing Systems. vol.~34,
  pp. 19313--19325 (2021),
  \url{https://proceedings.neurips.cc/paper_files/paper/2021/file/a11ce019e96a4c60832eadd755a17a58-Paper.pdf}

\bibitem{stojanov2021toys4k}
Stojanov, S., Thai, A., Rehg, J.M.: Using shape to categorize: Low-shot
  learning with an explicit shape bias. In: Proceedings of the IEEE/CVF
  Conference on Computer Vision and Pattern Recognition (CVPR). pp. 1798--1808
  (June 2021)

\bibitem{suvorov2022lama}
Suvorov, R., Logacheva, E., Mashikhin, A., Remizova, A., Ashukha, A.,
  Silvestrov, A., Kong, N., Goka, H., Park, K., Lempitsky, V.:
  Resolution-robust large mask inpainting with fourier convolutions. In:
  Proceedings of the IEEE/CVF Winter Conference on Applications of Computer
  Vision (WACV). pp. 2149--2159 (January 2022)

\bibitem{wang2025vggt}
Wang, J., Chen, M., Karaev, N., Vedaldi, A., Rupprecht, C., Novotny, D.: Vggt:
  Visual geometry grounded transformer. In: Proceedings of the IEEE/CVF
  Conference on Computer Vision and Pattern Recognition (CVPR). pp. 5294--5306
  (June 2025)

\bibitem{wang2023neus}
Wang, P., Liu, L., Liu, Y., Theobalt, C., Komura, T., Wang, W.: Neus: Learning
  neural implicit surfaces by volume rendering for multi-view reconstruction.
  arXiv preprint arXiv:2106.10689  (2023)

\bibitem{wang2026pi3}
Wang, Y., Zhou, J., Zhu, H., Chang, W., Zhou, Y., Li, Z., Chen, J., Pang, J.,
  Shen, C., He, T.: \${\textbackslash}pi{\textasciicircum}3\$:
  Permutation-equivariant visual geometry learning. In: The Fourteenth
  International Conference on Learning Representations (2026),
  \url{https://openreview.net/forum?id=DTQIjngDta}

\bibitem{wang2026orient}
Wang, Z., Zhang, Z., Xu, J., Wang, J., Pang, T., Du, C., Zhao, H., Zhao, Z.:
  Orient anything v2: Unifying orientation and rotation understanding. In: The
  Thirty-ninth Annual Conference on Neural Information Processing Systems
  (2026), \url{https://openreview.net/forum?id=n3armuTFit}

\bibitem{wu20253dgut}
Wu, Q., Martinez~Esturo, J., Mirzaei, A., Moenne-Loccoz, N., Gojcic, Z.: 3dgut:
  Enabling distorted cameras and secondary rays in gaussian splatting.
  Conference on Computer Vision and Pattern Recognition (CVPR)  (2025)

\bibitem{wu2025amodal3r}
Wu, T., Zheng, C., Guan, F., Vedaldi, A., Cham, T.J.: Amodal3r: Amodal 3d
  reconstruction from occluded 2d images. In: Proceedings of the IEEE/CVF
  International Conference on Computer Vision. pp. 9181--9193 (2025)

\bibitem{wu2023omniobject3d}
Wu, T., Zhang, J., Fu, X., Wang, Y., Ren, J., Pan, L., Wu, W., Yang, L., Wang,
  J., Qian, C., Lin, D., Liu, Z.: Omniobject3d: Large-vocabulary 3d object
  dataset for realistic perception, reconstruction and generation. In:
  Proceedings of the IEEE/CVF Conference on Computer Vision and Pattern
  Recognition (CVPR). pp. 803--814 (June 2023)

\bibitem{xiang2025trellis2}
Xiang, J., Chen, X., Xu, S., Wang, R., Lv, Z., Deng, Y., Zhu, H., Dong, Y.,
  Zhao, H., Yuan, N.J., Yang, J.: Native and compact structured latents for 3d
  generation. Tech report  (2025)

\bibitem{xiang2025structured}
Xiang, J., Lv, Z., Xu, S., Deng, Y., Wang, R., Zhang, B., Chen, D., Tong, X.,
  Yang, J.: Structured 3d latents for scalable and versatile 3d generation. In:
  Proceedings of the IEEE/CVF conference on computer vision and pattern
  recognition. pp. 21469--21480 (2025)

\bibitem{xu2024sparp}
Xu, C., Li, A., Chen, L., Liu, Y., Shi, R., Su, H., Liu, M.: Sparp: Fast 3d
  object reconstruction and pose estimation from sparse views. In: Proceedings
  of the European Conference on Computer Vision (ECCV). p. 143–163 (2024)

\bibitem{yan2025posemaster}
Yan, H., Luo, K., Li, W., Liang, Y., Li, S., Huang, J., Guo, C., Tan, P.:
  Posemaster: Generating 3d characters in arbitrary poses from a single image.
  arXiv preprint arXiv:2506.21076  (2025)

\bibitem{yao2025cast}
Yao, K., Zhang, L., Yan, X., Zeng, Y., Zhang, Q., Xu, L., Yang, W., Gu, J., Yu,
  J.: Cast: Component-aligned 3d scene reconstruction from an rgb image. ACM
  Transactions On Graphics (TOG)  \textbf{44}(4) (2025). \doi{10.1145/3730841},
  \url{https://doi.org/10.1145/3730841}

\bibitem{yu2023mvimgnet}
Yu, X., Xu, M., Zhang, Y., Liu, H., Ye, C., Wu, Y., Yan, Z., Zhu, C., Xiong,
  Z., Liang, T., Chen, G., Cui, S., Han, X.: Mvimgnet: A large-scale dataset of
  multi-view images. In: Proceedings of the IEEE/CVF Conference on Computer
  Vision and Pattern Recognition (CVPR). pp. 9150--9161 (June 2023)

\bibitem{github2025vecsetx}
Zhang, B.: Vecsetx. \url{https://github.com/1zb/VecSetX} (2025)

\bibitem{zhang20233dshape2vecset}
Zhang, B., Tang, J., Niessner, M., Wonka, P.: 3dshape2vecset: A 3d shape
  representation for neural fields and generative diffusion models. ACM
  Transactions On Graphics (TOG)  \textbf{42}(4),  1--16 (2023)

\bibitem{zhang2024clay}
Zhang, L., Wang, Z., Zhang, Q., Qiu, Q., Pang, A., Jiang, H., Yang, W., Xu, L.,
  Yu, J.: Clay: A controllable large-scale generative model for creating
  high-quality 3d assets. ACM Transactions On Graphics (TOG)  \textbf{43}(4)
  (2024). \doi{10.1145/3658146}

\bibitem{zhang2025eschernetpp}
Zhang, X., Irshad, M.Z., Yezzi, A., Tsai, Y.C., Kira, Z.: Eschernet++:
  Simultaneous amodal completion and scalable view synthesis through masked
  fine-tuning and enhanced feed-forward 3d reconstruction. arXiv preprint
  arXiv:2507.07410  (2025)

\bibitem{zhou2025amodalgen3d}
Zhou, J., Tai, Y.W.: Amodalgen3d: Generative amodal 3d object reconstruction
  from sparse unposed views. arXiv preprint arXiv:2511.21945  (2025)

\bibitem{zuo2024gobjaverse}
Zuo, Q., Gu, X., Dong, Y., Zhao, Z., Yuan, W., Qiu, L., Bo, L., Dong, Z.:
  High-fidelity 3d textured shapes generation by sparse encoding and
  adversarial decoding. In: Proceedings of the European Conference on Computer
  Vision (ECCV). pp. 52--69 (2024). \doi{10.1007/978-3-031-72684-2_4}

\end{thebibliography}

\clearpage
\appendix

\begin{center}
{\Large\bfseries Supplementary Material}
\end{center}

\addtocontents{toc}{\protect\setcounter{tocdepth}{2}}
\setcounter{tocdepth}{2}
\renewcommand\contentsname{}
\begingroup
\let\clearpage\relax
{\raggedright\large\bfseries Contents\par\vspace{-2em}}
\tableofcontents
\endgroup

\section{Model Architecture}\label{supp:model}
\subsection{Gated Feed-Forward Network}\label{supp:model_ffn}
We adopt the feed-forward block used in LLaMA-3 \cite{grattafiori2024llama3} and SAM 3D \cite{chen2025sam3d}.
Given an input $x \in \mathbb{R}^d$, the feed-forward layer is defined as:

\begin{equation}
\text{FFN}(x) = W_2 \left( \text{SiLU}(W_1 x) \odot (W_3 x) \right),
\end{equation}

where $W_1, W_3 \in \mathbb{R}^{d \times h}$,
$W_2 \in \mathbb{R}^{h \times d_{\text{out}}}$. Here, $h$ is the hidden dimension,
$d$ and $d_{\text{out}}$ are the input and output dimensions,
$\odot$ denotes element-wise multiplication,
and $\text{SiLU}(z) = z \cdot \sigma(z)$.

\section{Training Data Processing}\label{supp:train_data}

\subsection{Surface Point Sampling}\label{supp:surf_pts}
When preprocessing meshes for training, we generate surface point cloud $P^k_\text{surface}$ to use with our data augmentation pipeline to generate conditioning points. To mimic the points a real-world depth sensor would capture, internal components that are fully or mostly enclosed within larger parts of the mesh are removed to avoid sampling from hidden geometry. Points are then uniformly sampled from the remaining external surfaces, along with their normals. To identify internal components, the mesh is first split into connected components. Each component is voxelized and filled to represent its full volume, then aligned in a global voxel grid for comparison. 
For each component, we compare against all components larger than itself to determine if it should be removed. 
Let $V_i$ and $V_j$ denote the sets of voxels occupied by components $i$ and $j$, respectively. 
For a component $i$ being checked against a larger component $j$, we define the \emph{overlap fraction} as
\begin{equation}
f_{i,j} = \frac{|V_i \cap V_j|}{|V_i|}.
\end{equation}
Component $i$ is removed if there exists a larger component $j$ such that $f_{i,j} \geq 0.95$.

\subsection{Data Augmentation Functions}\label{supp:data_aug}

Here we provide details on algorithms used during data augmentation. We assume a setting in which per-view depth maps are not provided, as is often the case in rendering pipelines for 3D generation models, such as TRELLIS \cite{xiang2025structured}.

Let the mesh surface points $P_\text{surface}$ consist of points $\{p_i\}_{i=1}^S$ and corresponding normals $\{n_i\}_{i=1}^S$ where $S$ is the number of points. Alg.~\ref{alg:pts_in_view} filters the points to produce a binary point visibility mask 
$V \in \{0,1\}^S$ for a given camera view with intrinsics $K$ and extrinsics $E$. 
We first perform back-face culling using the point normals and the camera backward vector. 
Next, for each point $p_i$, the function 
$\textsc{Project}(p_i, E, K)$ returns the corresponding normalized image coordinates ($u_i$, $v_i$) and depth $d_i$.
Due to the sparsity of points, a coarse z-buffer $Z$ is constructed by projecting the points onto a low-resolution image grid and storing the minimum depth in each cell. 
This provides an approximation of point visibility across the image. 
To incorporate the occlusion mask $M_{occ}$ (a binary mask where 1 indicates an occluded pixel), the z-buffer is upsampled to $Z_{up}$, matching the resolution of $M_{occ}$.
Finally, points whose depth is within a threshold $\epsilon=0.05$ of the z-buffer and not occluded in $M_{occ}$ are marked visible. 
The algorithm is applied for all $N$ conditioning views and the resulting masks $\{V_j\}_{j=1}^N$ are combined with a point-wise logical OR to obtain the final visibility mask to obtain the conditioning points $P$.

\newcommand{\Output}[1]{\item[\textbf{Output:}] #1}

\begin{algorithm}
\caption{Select Visible Points in a Camera View}\label{alg:pts_in_view}
\begin{algorithmic}[1]
\Require 3D points $\mathbf{P_\text{surface}} = \{\mathbf{p}_i\}_{i=1}^S$, normals $\{\mathbf{n}_i\}_{i=1}^S$
\Require Camera extrinsics $\mathbf{E}$, intrinsics $\mathbf{K}$
\Require Optional occlusion mask $\mathbf{M_{occ}} \in \mathbb{R}^{H \times W}$
\Output Point visibility mask $\mathbf{V} \in \{0,1\}^{S}$

\State $\mathbf{b} \gets -\mathbf{E}_{3,1:3}$ \Comment{Camera backward vector}
\For{$i = 1$ to $S$}  \Comment{Back-face culling (normal filtering)}
    \State $f_i \gets (\mathbf{n}_i^\top \mathbf{b} > 0)$
\EndFor

\For{$i = 1$ to $S$} \Comment{Project points to image coordinates and compute depth}
    \State $(u_i, v_i, d_i) \gets \textsc{Project}(\mathbf{p}_i, \mathbf{E}, \mathbf{K})$
\EndFor

\State $h \gets H/8$, \quad $w \gets W/8$
\State Initialize z-buffer $\mathbf{Z} \in \mathbb{R}^{h \times w}$ with $+\infty$

\For{$i = 1$ to $S$} \Comment{Coarse z-buffer construction}
    \State $x_i, y_i \gets \text{clip}(\lfloor u_i \cdot (w-1) \rfloor, 0, w-1), \text{clip}(\lfloor v_i \cdot (h-1) \rfloor, 0, h-1)$
    \State $\mathbf{Z}[y_i, x_i] \gets \min(\mathbf{Z}[y_i, x_i], d_i)$
\EndFor

\State $\mathbf{Z}_\text{up} \gets \textsc{BilinearUpsample}(\mathbf{Z}, H, W)$ \Comment{Upsample z-buffer to full resolution}

\If{$\mathbf{M_{occ}}$ is not provided} \Comment{Set occlusion mask to all visible if not provided}
    \State $\mathbf{M_{occ}} \gets \mathbf{0}^{H \times W}$
\EndIf
\State Initialize $\mathbf{V} \gets \mathbf{0}^{S}$ \Comment{All pixels initially visible}
\For{$i = 1$ to $S$} \Comment{Set point visible if within depth threshold and not occluded}
    \State $x_i, y_i \gets \text{clip}(\lfloor u_i \cdot (W-1) \rfloor, 0, W-1), \quad \text{clip}(\lfloor v_i \cdot (H-1) \rfloor, 0, H-1)$
    \If{$|d_i - \mathbf{Z}_\text{up}[y_i, x_i]| < \epsilon \;\wedge\; \mathbf{M_{occ}}[y_i, x_i] < 1$}
        \State $V[i] \gets 1$
    \EndIf
\EndFor

\State \Return $\mathbf{V}$

\end{algorithmic}
\end{algorithm}

Alg.~\ref{alg:occlusion_mask} is used during the edit scenario to generate visibility masks $\{M_i\}_{i=1}^N$ from 3D bounding boxes. 
For each bounding box $B \in \mathbb{R}^{8 \times 3}$, the corners are projected onto the image and the convex hull of the projected points is used to create the visibility mask.
While this efficiently approximates occluded regions, it may over-occlude areas where the bounding box corners are behind or within the object.

\begin{algorithm}
\caption{Create Visibility Mask from 3D Bounding Box}\label{alg:occlusion_mask}
\begin{algorithmic}[1]
\Require Bounding box points $\mathbf{B} \in \mathbb{R}^{8 \times 3}$
\Require Camera extrinsics $\mathbf{E}$, intrinsics $\mathbf{K}$
\Output Binary visibility mask $\mathbf{M} \in \{0,1\}^{H \times W}$

\State Initialize $\mathbf{M} \gets \mathbf{1}^{H \times W}$ \Comment{All pixels initially visible}

\For{$j = 1$ to $8$} \Comment{Project bounding box points to image coordinates}
    \State $(u_j, v_j) \gets \textsc{Project}(\mathbf{B}_j, \mathbf{E}, \mathbf{K})$
    \State $x_j, y_j \gets \text{clip}(\lfloor u_j \cdot (W-1) \rfloor, 0, W-1), \quad \text{clip}(\lfloor v_j \cdot (H-1) \rfloor, 0, H-1)$
\EndFor
\State $p_\text{corners} \gets \text{stack}(x_j, y_j)$ \Comment{Pixel coordinates of box corners}

\State $hull \gets \text{ConvexHull}(p_\text{corners})$ \Comment{Ordered list of points along the convex hull}
\State $\text{FillContours}(\mathbf{M}, hull, 0)$ \Comment{Set all mask pixels inside the convex hull to 0}

\State \Return $\mathbf{M}$

\end{algorithmic}
\end{algorithm}

\subsection{Training Data Filtering}\label{supp:data_filter}
We train our model on 407k shapes from TRELLIS-500K dataset \cite{xiang2025structured} which consists of aesthetics filtered objects from ObjaverseXL \cite{deitke2023objaverseXL}, ABO \cite{collins2022abo}, 3D-FUTURE \cite{fu20213dfuture}, and HSSD \cite{khanna2024hssd}.
While the ObjaverseXL subset in TRELLIS-500K has already been filtered for aesthetics, it still contains many entries that correspond to scene-level assets rather than single objects, which are the focus of our work. To address this, we leverage the G-Objaverse \cite{zuo2024gobjaverse} dataset, which provides human-provided category labels for the original Objaverse dataset. It defines 10 general categories, including Human-Shape (41,557), Animals (28,882), Daily-Used (220,222), Furnitures (19,284), Buildings \& Outdoor (116,545), Transportations (20,075), Plants (7,195), Food (5,314), Electronics (13,252), and Poor-quality (107,001).
To focus on single-object assets, we remove entries in the Buildings \& Outdoor category. We accomplish this by training a classifier head on top of pretrained DINOv3 embeddings. Given a rendered view of an object, the classifier predicts its general category, and we aggregate predictions across multiple views by averaging to obtain a robust per-object classification. This approach allows us to automatically filter out scene-like objects for training, removing 66k objects from our downloaded subset of TRELLIS-500K. Examples of objects removed are shown in Fig.\ref{fig:supp_data_filter}.

\providecommand{\qcwidth}{0.09\linewidth}

\begingroup
\setlength{\tabcolsep}{1pt}
    
\renewcommand{\arraystretch}{0}

\begin{figure*}[tb]
	\centering

\resizebox{\linewidth}{!} {
\begin{tabular}{@{}ccccc@{}}

\includegraphics[width=\qcwidth, valign=m]{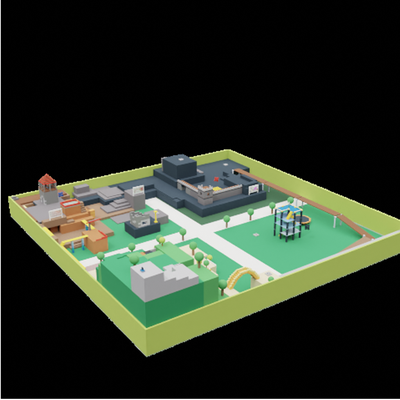} & 
\includegraphics[width=\qcwidth, valign=m]{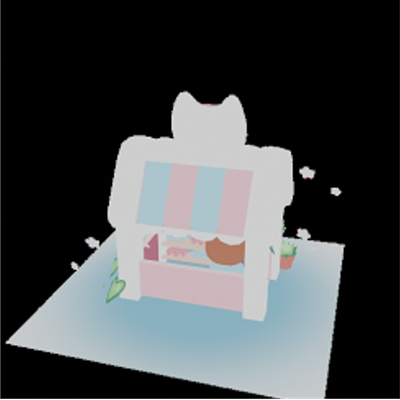} & 
\includegraphics[width=\qcwidth, valign=m]{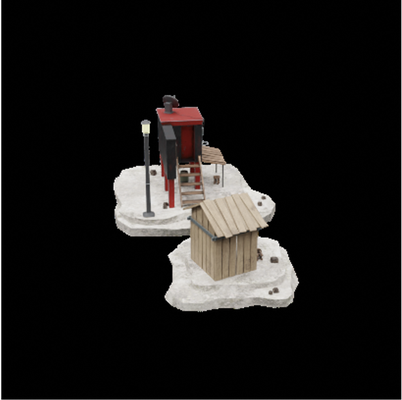} & 
\includegraphics[width=\qcwidth, valign=m]{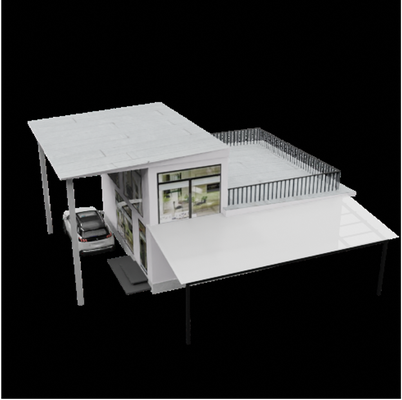} & 
\includegraphics[width=\qcwidth, valign=m]{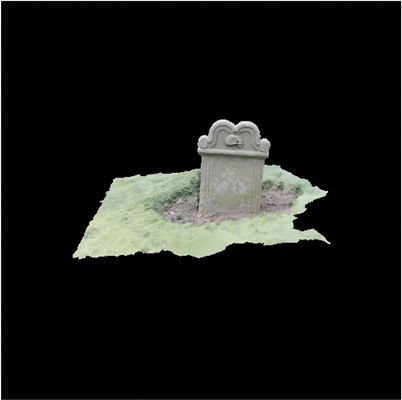}  \\

  \end{tabular}
}
	\caption{\textbf{Training Data Filtering}: Example of objects that are removed.}
	\label{fig:supp_data_filter}
\end{figure*}

\endgroup

\section{Experiment Details}\label{supp:expr}

\subsection{Occlusion Mask Generation}\label{supp:expr_eval}

\begingroup
\setlength{\tabcolsep}{1pt}
    
\renewcommand{\arraystretch}{0}

\begin{figure*}[tb]
	\centering

\resizebox{\linewidth}{!} {
\begin{tabular}{@{}ccccc@{}}

\includegraphics[width=\qcwidth, valign=m]{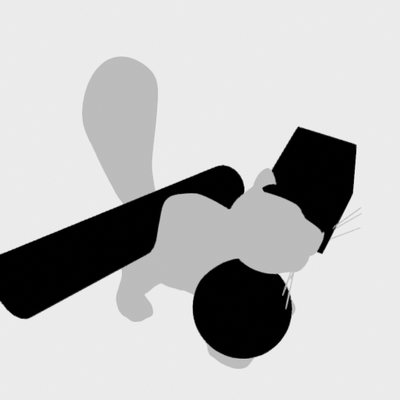} & 
\includegraphics[width=\qcwidth, valign=m]{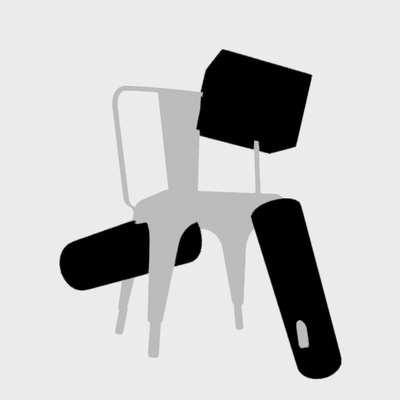} & 
\includegraphics[width=\qcwidth, valign=m]{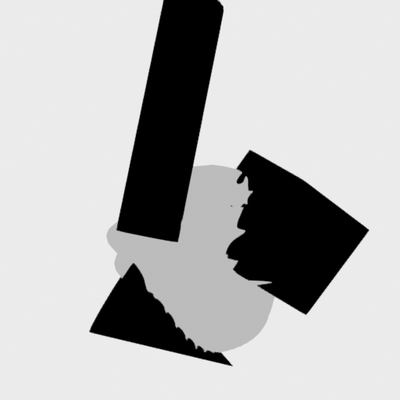} & 
\includegraphics[width=\qcwidth, valign=m]{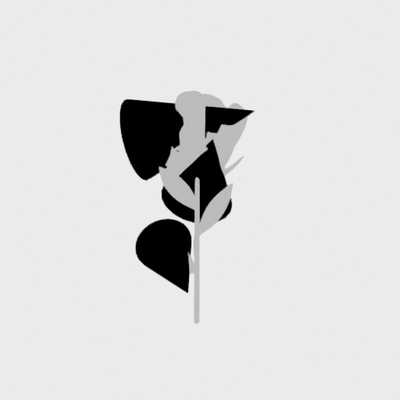} & 
\includegraphics[width=\qcwidth, valign=m]{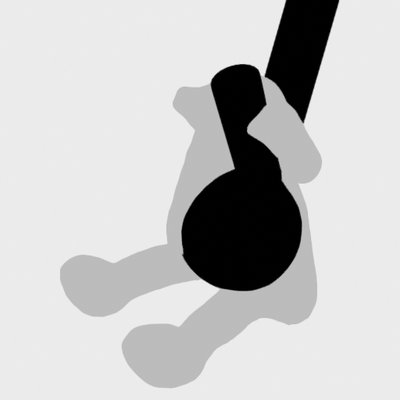}  \\

  \end{tabular}
}
	\caption{\textbf{3D-Consistent Masks}: Example of masks with generated occluders.}
	\label{fig:supp_occluders}
\end{figure*}

\endgroup

For quantitative and qualitative evaluation, we simulate realistic 3D-consistent occlusions by adding random occluders from a set of basic shapes (box, cylinder, sphere, and cone). 
In Amodal3R \cite{wu2025amodal3r}, a 3D-consistent mask is generated with a random-walk strategy
to iteratively select neighboring faces of a mesh. 
Instead of sampling directly from the mesh surface---which can reveal the geometry of the occluded region---we use occluder shapes to prevent the model from inferring the hidden structure from the mask.
For each mesh, we sample 3 occluders from the set of basic shapes with equal probability. 

For each mesh, we sample three occluders from the set of basic shapes with equal probability. 
To create an occluder, we first compute the object axis-aligned bounding box and its volume $V_{obj}$. 
We then sample a linear scale $s \sim U(0.25, 5)$ and define the target occluder volume as
\begin{equation}
V_{occ} = s^3 \cdot V_{obj}.
\end{equation}

To determine the occluder position, we uniformly sample 100k points from the mesh surface and apply voxel downsampling (with voxel size $0.1$) to obtain a spatially distributed set of candidate locations. 
One point is randomly selected as the occluder center, ensuring the occluder is placed near the object geometry. 

Given the target volume $V_{\text{occ}}$, the dimensions of each occluder type are sampled as follows:

\begin{itemize}
\item \textbf{Box.} Independent aspect ratios $r_x, r_y, r_z \sim U(0.6, 1.4)$ are sampled, and the side lengths are scaled to match the target volume $V_{\text{occ}}$, yielding
\[
l_i = r_i \left(\frac{V_{\text{occ}}}{r_x r_y r_z}\right)^{1/3}, \quad i \in \{x, y, z\}.
\]
This ensures that $l_x \cdot l_y \cdot l_z = V_{\text{occ}}$.

\item \textbf{Sphere.} The radius is computed directly from the target volume:
\[
r = \left(\frac{3V_{\text{occ}}}{4\pi}\right)^{1/3}.
\]

\item \textbf{Cylinder.} The aspect ratio between height and diameter is sampled as $a \sim U(2.0, 6.0)$. The radius is then
\[
r = \left(\frac{V_{\text{occ}}}{2\pi a}\right)^{1/3},
\]
and the height is $h = 2ar$.

\item \textbf{Cone.} The height-to-radius ratio is sampled as $a \sim U(1.5, 4.0)$. The radius is
\[
r = \left(\frac{3V_{\text{occ}}}{\pi a}\right)^{1/3},
\]
and the height is $h = ar$.
\end{itemize}

Each occluder is centered at the sampled location and used to render the final occlusion mask in each camera view. Example masks are shown in Fig. \ref{fig:supp_occluders}.

\subsection{View Selection for Hunyuan3D-Omni}\label{supp:expr_hy3d}
During evaluation, we used Hunyuan3D-Omni (Hy3D-Omni) \cite{hunyuan3d2025hunyuan3domni} as a baseline for both single- and multi-view completion. Although Hy3D-Omni is designed to take only a single-view image as input, it can be conditioned on point clouds from multiple views. To leverage this capability, we select the single input image that provides the most informative view while conditioning on multi-view points. Below, we detail our view selection strategy.

For each view, with occlusion mask $M_{occ}$ and foreground object mask $M_{obj}$, we compute two metrics:
\begin{equation}
F_m = \frac{\sum M_{occ}}{\sum M_{obj}}, \quad 
F_c = \frac{\sum M_{obj}}{\text{R}^2},
\end{equation}
where \(F_m\) denotes the fraction of foreground pixels that are occluded, \(F_c\) is the fraction of image pixels occupied by the object, and \(\text{R}\) represents the image resolution. For unoccluded scenarios, the view with the largest object coverage, \emph{i.e.}, the view that maximizes \(F_c\), is selected. In occluded scenarios, the view is selected to maximize both object coverage and visibility by choosing the view that maximizes \(F_c \cdot (1 - F_m)\). This procedure ensures that the selected image provides the most visible and informative view of the object for Hy3D-Omni.

\subsection{Image-to-3D with MapAnything}\label{supp:expr_appl}
\begin{figure}[tb]
  \centering
  \includegraphics[width=0.95\linewidth]{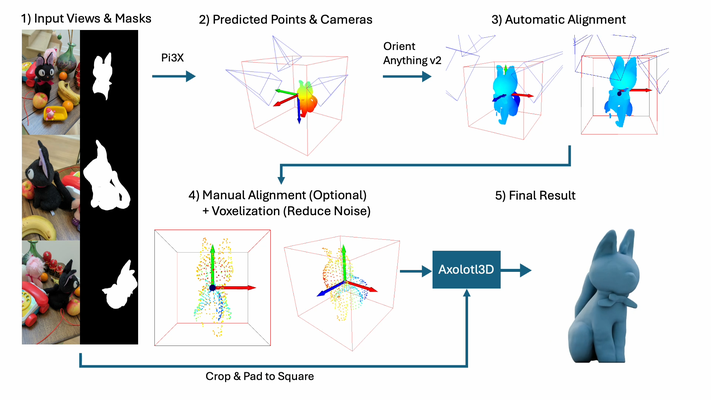}
  \caption{\textbf{Image-to-3D Pipeline:} Fully automated with SAM 2 \cite{ravi2025sam2} segmentation, MapAnything \cite{keetha2026mapanything} points and cameras, and Orient Anything v2 \cite{wang2026orient} point alignment.}
  \label{fig:supp_imgto3d}
\end{figure}
In our paper, we showed results on 3D shape generation directly from images using MapAnything \cite{keetha2026mapanything} outputs. Here we provide more details on the pipeline, specifically regarding object orientation. 
The input image is first segmented using SAM 2 \cite{ravi2025sam2} to obtain the object segment and ground floor segment. The input image is then passed to MapAnything \cite{keetha2026mapanything} to produce points and camera parameters. The object and ground points are obtained using the segmentation masks. 
To match the orientation of objects in the training data (+Y up, +Z forward), the ground points are first used to estimate a ground plane, and the points are rotated so that its normal aligns with the +Y axis. The object points are then projected onto the X–Z plane, and their dominant horizontal axis (first principal component) is aligned with +Z using singular value decomposition (SVD). Alternatively, when ground points are unavailable, Orient Anything v2 \cite{wang2026orient} can be applied to a view to predict the orientation as shown in Fig.~\ref{fig:supp_imgto3d} but may require manual correction. Finally, the points are centered and scaled to $[-1, 1]$.
Depending on the surface quality of the predicted points, we apply voxel downsampling and outlier point removal to reduce surface artifacts in our predict output. See \S\ref{supp:limit} for further discussion regarding limitations. 

\subsection{3D Capture \& Simulation}\label{supp:expr_simulation}
Accurate geometry is important for applications such as robot simulation in digital twins or interaction in captured environments.
Here we show extraction of objects from a real-world captured scene and simulation in the trained gaussian splat scene. 

\subsubsection{Generating Meshes from Captures:} For each object, we manually select a set of representative views and segment the object using SAM~2~\cite{ravi2025sam2}. The segmented regions are then cropped and padded to form the input images.
Since multi-view depth prediction methods often suffer from view alignment inconsistencies, we instead estimate depth from a single image using Depth Anything~3~\cite{lin2026depthanything}. To enable comparison with the gaussian splatting \cite{kerbl2023gsplat} scene representation, we scale and shift the predicted depth to align with the depth rendered from the gaussian splat scene.
The aligned depth map is then masked using object segmentation masks and back-projected using the corresponding camera parameters from the gaussian splat scene to obtain aligned conditioning points. These points are subsequently manually oriented to $+Y$ up, $+Z$ forward and rescaled to the range $[-1, 1]$ to match the input convention of our method. The inverse transformation will be applied to the predicted output to revert to the original scene coordinate frame.
Depending on the quality of the predicted depth, we apply voxel downsampling and outlier point removal to reduce noise.
To incorporate cropped multi-view image conditions, we apply our method without camera conditioning, as described in the ablation section of the main paper.
\subsubsection{Simulation:} We follow the Kaolin hand-on lab \cite{modi2025lab} that utilizes representation-agnostic physics simulation \cite{modi2024simplicits} and mixed splat-mesh rendering \cite{wu20253dgut,moenne20243d} to show our results; splat objects are segmented following an algorithm similar to \cite{tsang2026artisangs}.
The resulting simulation is shown in the supplementary video.
Unlike gaussian splat object segments which contain large missing areas, our geometries are complete, and can be used for more accurate physics simulation of the captured environment.

\section{Additional Results}\label{supp:expr_qual}
\subsection{Quantitative Results without ICP}\label{supp:expr_icp}
\begin{table}[!tb]

\caption{Quantitative evaluation without ICP alignment on Toys4k.}
\centering 
\label{tab:icp}

\begingroup
\setlength{\tabcolsep}{6pt} 

\newcommand{\pmitem}[1]{\scalebox{0.7}{$\pm$ #1}}

\newcommand{\pmrow}[6]{%
\pmitem{#1} & 
\pmitem{#2} & 
\pmitem{#3} & 
\pmitem{#4} & 
\pmitem{#5} & 
\pmitem{#6} 
}

\resizebox{\linewidth}{!} {
\begin{tabular}{@{}l||lll|lll@{}}
\Xhline{1.0pt}
\multirow{2}{*}{\makecell[l]{\textbf{Scenario:} \\ Method}} & \multicolumn{3}{c|}{Single-View} & \multicolumn{3}{c}{Multi-View} \\

 & F-score $\uparrow$ & vIoU $\uparrow$ & CD$\downarrow_{\times10}$ & F-score $\uparrow$ & vIoU $\uparrow$ & CD$\downarrow_{\times10}$ \\

 \Xhline{1.0pt}

\textbf{\textit{w/ occlusion:}} &&&&&& \\

Hy3D-Omni &
0.6166&
0.1800&
0.8635&
0.8646&
0.3346&
0.3092\\
& \pmrow{0.2762}{0.1291}{0.8270}{0.1809}{0.1574}{0.3803}\\

ShapeR &
\cellcolor{secondyellow}0.6851&
\cellcolor{secondyellow}0.2282&
\cellcolor{secondyellow}0.5237&
\cellcolor{secondyellow}0.8901&
\cellcolor{secondyellow}0.3998&
\cellcolor{secondyellow}0.2448\\
\scalebox{0.9}{(concurrent)} & \pmrow{0.2545}{0.1203}{0.4498}{0.1896}{0.1554}{0.3854}\\

\textbf{Ours} &
\cellcolor{bestgreen}0.8898&
\cellcolor{bestgreen}0.3277&
\cellcolor{bestgreen}0.2559&
\cellcolor{bestgreen}0.9615&
\cellcolor{bestgreen}0.4196&
\cellcolor{bestgreen}0.1540\\
& \pmrow{0.1328}{0.1035}{0.2207}{0.0762}{0.1126}{0.0865}\\

\Xhline{1.0pt}

\textbf{\textit{w/o occlusion:}} &&&&&& \\

Hy3D-Omni &
\cellcolor{secondyellow}0.7512&
0.2486&
0.4858&
0.9071&
0.3849&
0.2333\\
& \pmrow{0.2257}{0.1462}{0.4408}{0.1351}{0.1591}{0.2254}\\

ShapeR &
0.7386&
\cellcolor{secondyellow}0.2566&
\cellcolor{secondyellow}0.4549&
\cellcolor{secondyellow}0.9294&
\cellcolor{bestgreen}0.4470&
\cellcolor{secondyellow}0.1897\\
\scalebox{0.9}{(concurrent)} & \pmrow{0.2445}{0.1230}{0.4318}{0.1546}{0.1450}{0.3343}\\

\textbf{Ours} &
\cellcolor{bestgreen}0.9118&
\cellcolor{bestgreen}0.3481&
\cellcolor{bestgreen}0.2278&
\cellcolor{bestgreen}0.9767&
\cellcolor{secondyellow}0.4438&
\cellcolor{bestgreen}0.1372\\
& \pmrow{0.1171}{0.0986}{0.1942}{0.0468}{0.1078}{0.0571}\\

\Xhline{1.0pt}

\end{tabular}
}

\endgroup
\end{table}

\begingroup
\setlength{\tabcolsep}{1pt}
    
\renewcommand{\arraystretch}{0}

\begin{figure}[tb]
	\centering

\resizebox{0.9\linewidth}{!} {
\begin{tabular}{@{}cc|ccc@{}}

\scalebox{0.5}{Edited View} & 
\scalebox{0.5}{Edited Points} & 
\scalebox{0.5}{Hy3D-Omni} & 
\scalebox{0.5}{ShapeR} & 
\scalebox{0.5}{Ours} \\ 

\includegraphics[width=\qcwidth, valign=m]{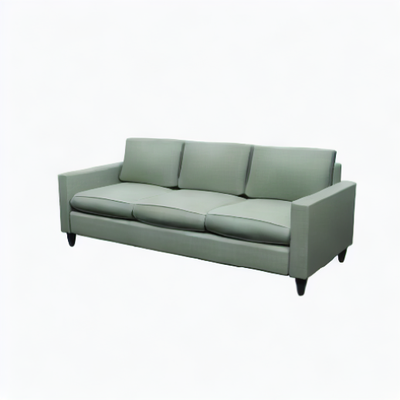} & 
\includegraphics[width=\qcwidth, valign=m]{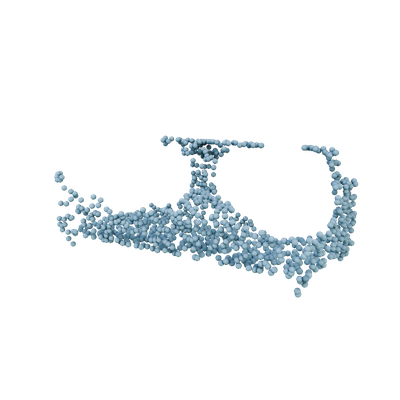} & 
\includegraphics[width=\qcwidth, valign=m]{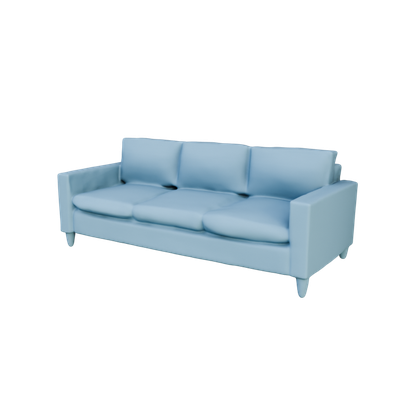} & 
\includegraphics[width=\qcwidth, valign=m]{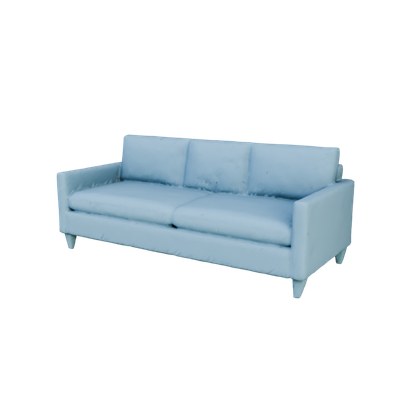} & 
\includegraphics[width=\qcwidth, valign=m]{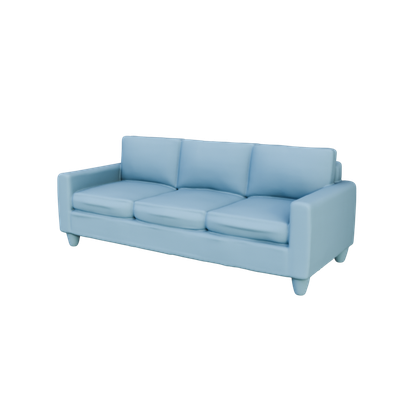} \\

\includegraphics[width=\qcwidth, valign=m]{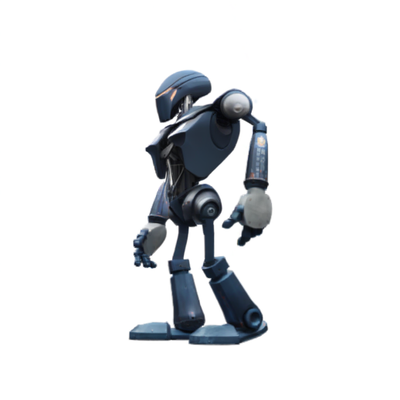} & 
\includegraphics[width=\qcwidth, valign=m]{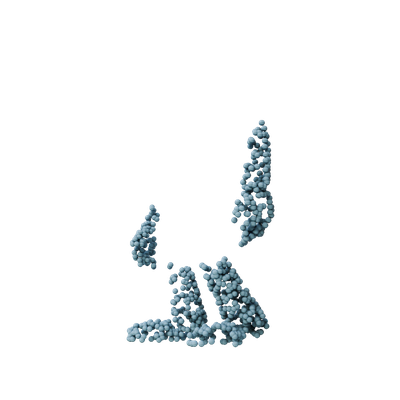} & 
\includegraphics[width=\qcwidth, valign=m]{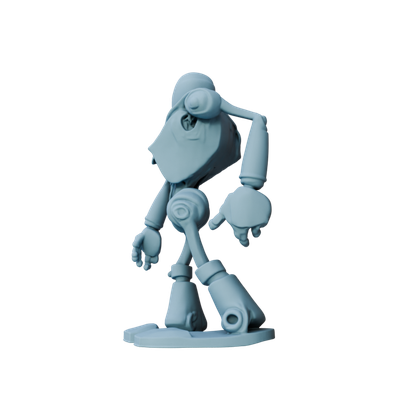} & 
\includegraphics[width=\qcwidth, valign=m]{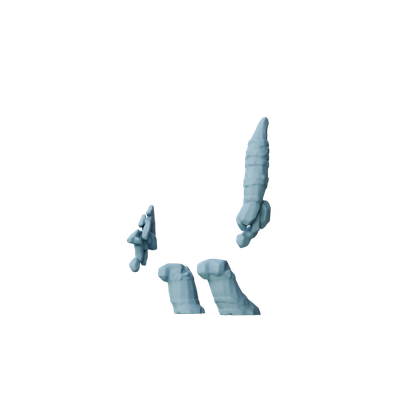} & 
\includegraphics[width=\qcwidth, valign=m]{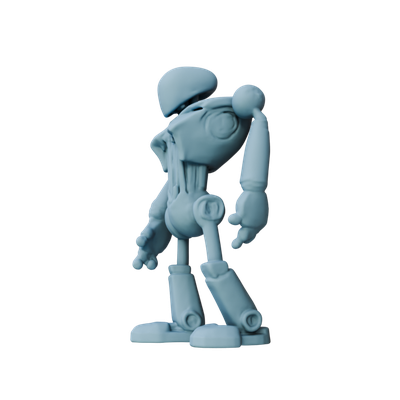} \\

\includegraphics[width=\qcwidth, valign=m]{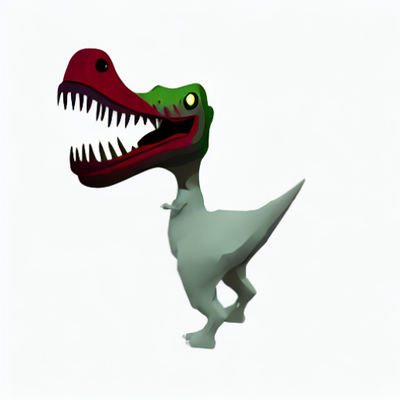} & 
\includegraphics[width=\qcwidth, valign=m]{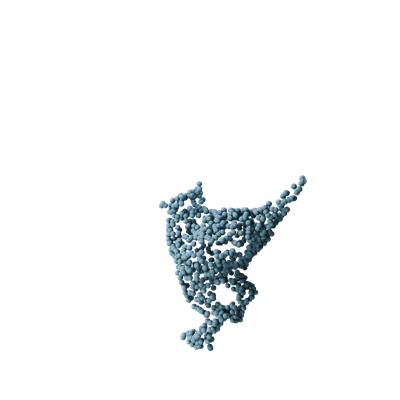} & 
\includegraphics[width=\qcwidth, valign=m]{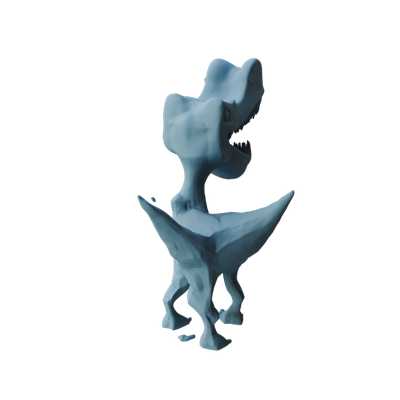} & 
\includegraphics[width=\qcwidth, valign=m]{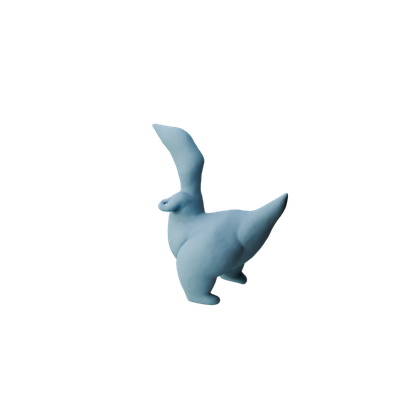} & 
\includegraphics[width=\qcwidth, valign=m]{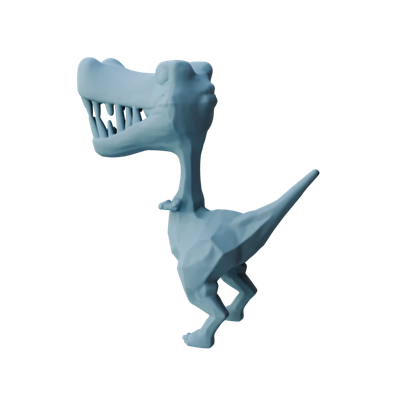} \\

  \end{tabular}
}
	\caption{\textbf{Object edit comparison} against baselines with 2048 input points.}
	\label{fig:supp_compare_edit}

\end{figure}

\endgroup

We apply ICP to all methods for fairness as Amodal3R does not use point conditioning and SAM3D uses camera-space pointmaps. Metrics without alignment would largely reflect coordinate-frame mismatch rather than reconstruction accuracy.
Tb.~\ref{tab:icp} shows results without ICP. Scores decrease for all methods and relative rankings mostly unchanged.

\subsection{Object Edit Comparisons}\label{supp:expr_edit}
In Fig.~\ref{fig:supp_compare_edit}, we show comparisons against point conditioned baselines in object edit task. The baselines were not explicitly trained for object editing, and therefore struggle to handle more challenging edits. Hunyuan3D-Omni (Hy3D-Omni) lacks camera conditioning, which results in ambiguity in object orientation. In contrast, our method respects the conditioning points to preserve unedited regions, while jointly leveraging camera parameters and the input image to produce faithful edits.

\subsection{Qualitative Comparisons}
\begin{figure*}[!t]
  \centering

  \begin{subfigure}{\linewidth}
    \centering

  \resizebox{\linewidth}{!} {
    \begin{tabular}{@{}cc|ccccc|c@{}}

\scalebox{0.5}{Input View} &
\scalebox{0.5}{Input Points} & 
\scalebox{0.5}{Amodal3R} & 
\scalebox{0.5}{SAM3D} & 
\scalebox{0.5}{Hy3D-Omni} & 
\scalebox{0.5}{ShapeR} & 
\scalebox{0.5}{\textbf{Ours}} & 
\scalebox{0.5}{Ground Truth} \\ 

\includegraphics[width=\qcwidth, valign=m]{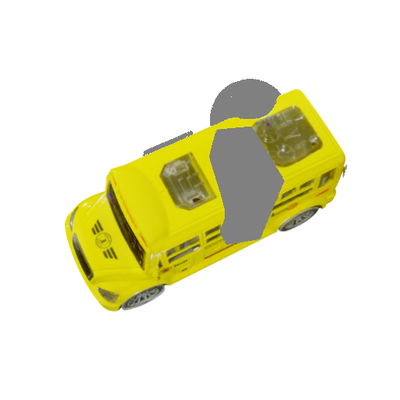} & 
\includegraphics[width=\qcwidth, valign=m]{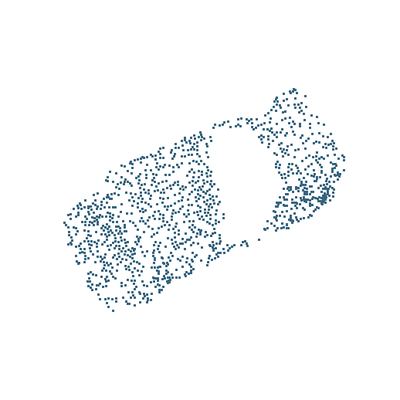} & 
\includegraphics[width=\qcwidth, valign=m]{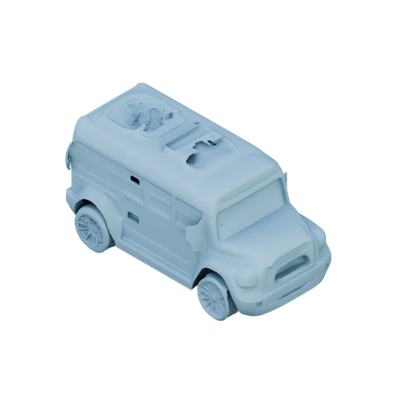} & 
\includegraphics[width=\qcwidth, valign=m]{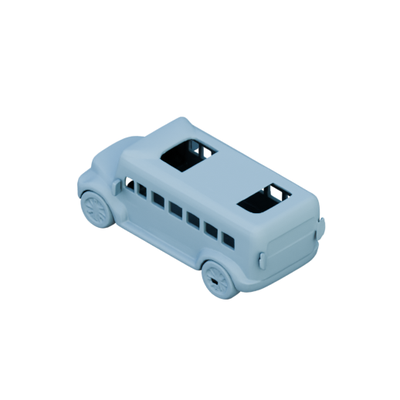} & 
\includegraphics[width=\qcwidth, valign=m]{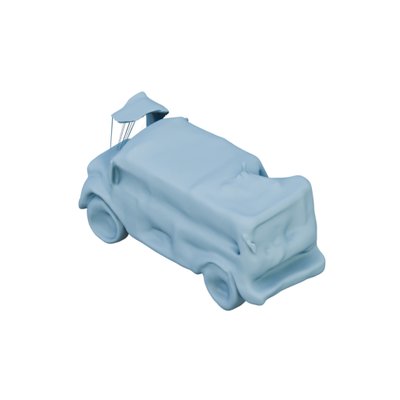} & 
\includegraphics[width=\qcwidth, valign=m]{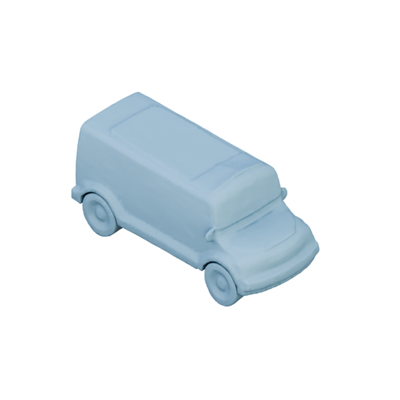} & 
\includegraphics[width=\qcwidth, valign=m]{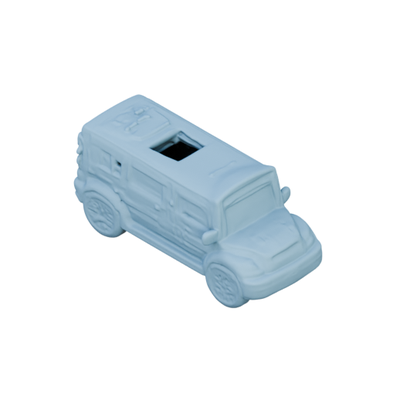} & 
\includegraphics[width=\qcwidth, valign=m]{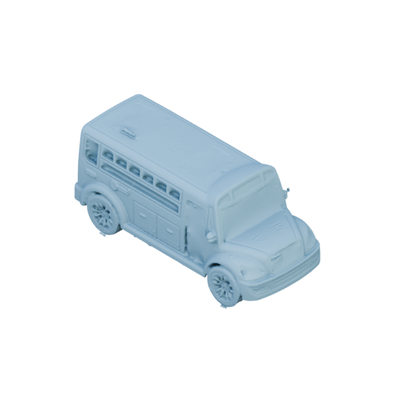} \\

\includegraphics[width=\qcwidth, valign=m]{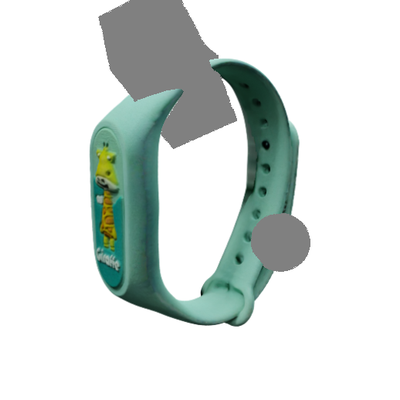} & 
\includegraphics[width=\qcwidth, valign=m]{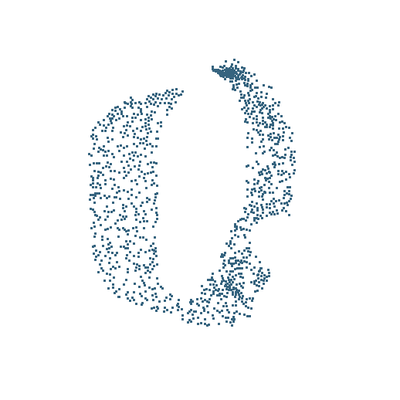} & 
\includegraphics[width=\qcwidth, valign=m]{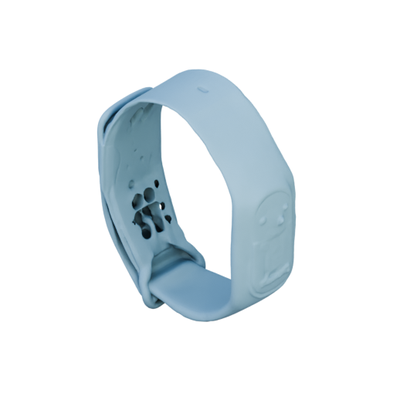} & 
\includegraphics[width=\qcwidth, valign=m]{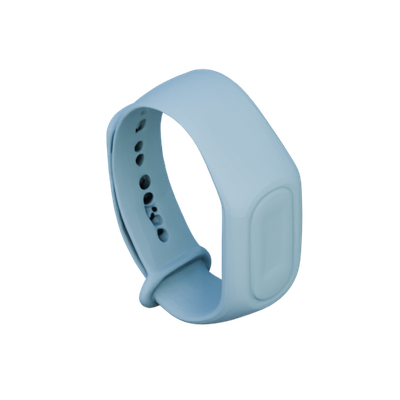} & 
\includegraphics[width=\qcwidth, valign=m]{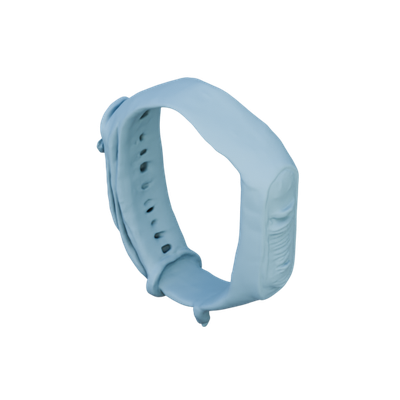} & 
\includegraphics[width=\qcwidth, valign=m]{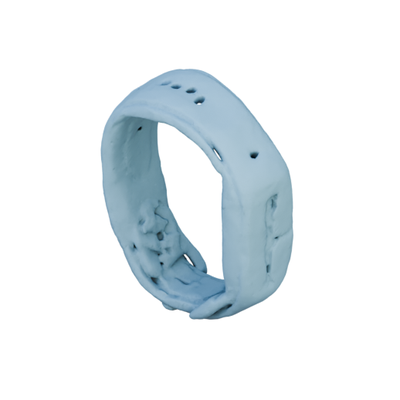} & 
\includegraphics[width=\qcwidth, valign=m]{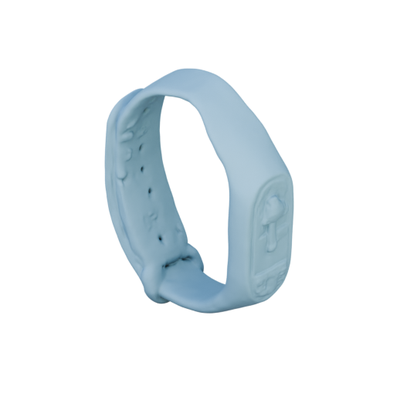} & 
\includegraphics[width=\qcwidth, valign=m]{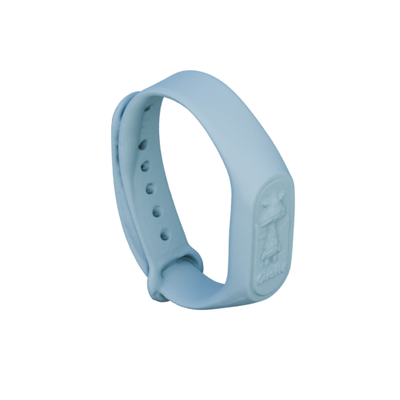} \\

\includegraphics[width=\qcwidth, valign=m]{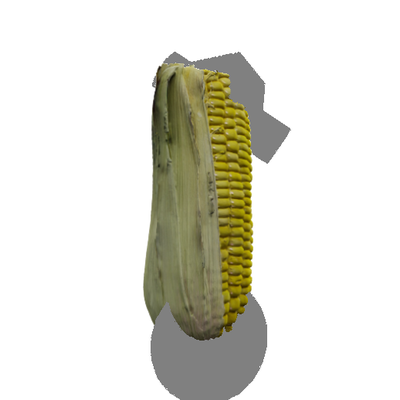} & 
\includegraphics[width=\qcwidth, valign=m]{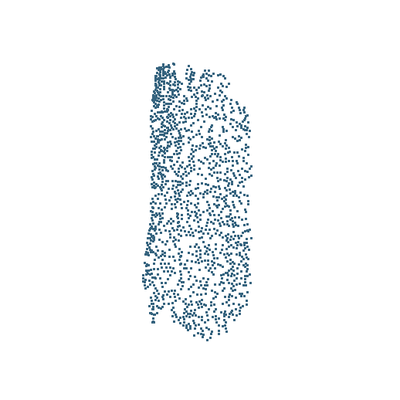} & 
\includegraphics[width=\qcwidth, valign=m]{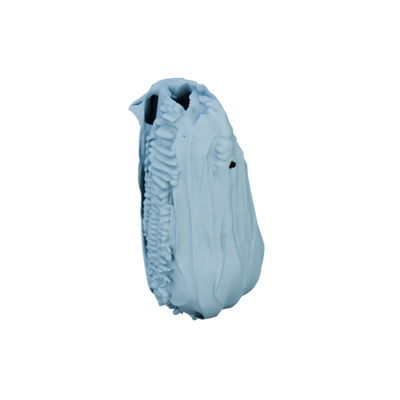} & 
\includegraphics[width=\qcwidth, valign=m]{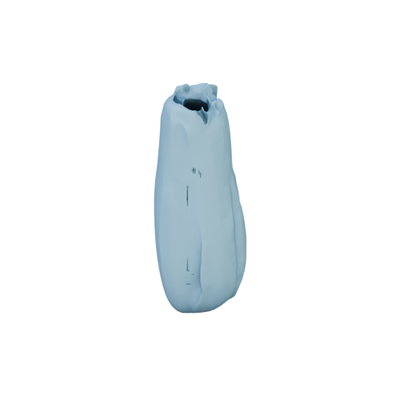} & 
\includegraphics[width=\qcwidth, valign=m]{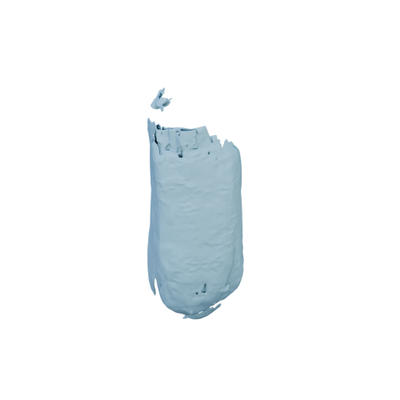} & 
\includegraphics[width=\qcwidth, valign=m]{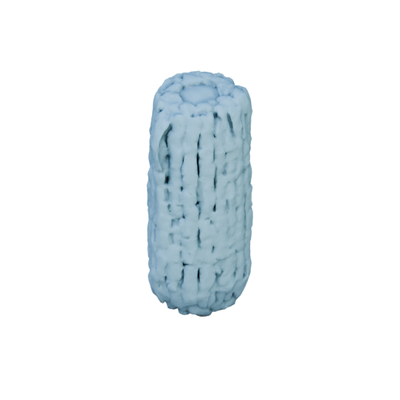} & 
\includegraphics[width=\qcwidth, valign=m]{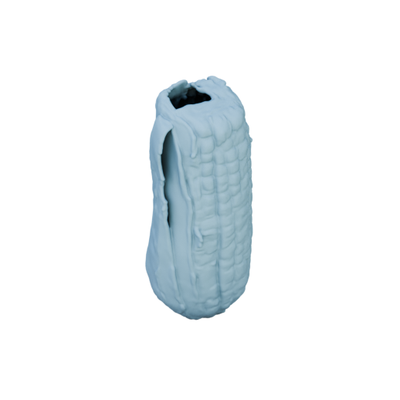} & 
\includegraphics[width=\qcwidth, valign=m]{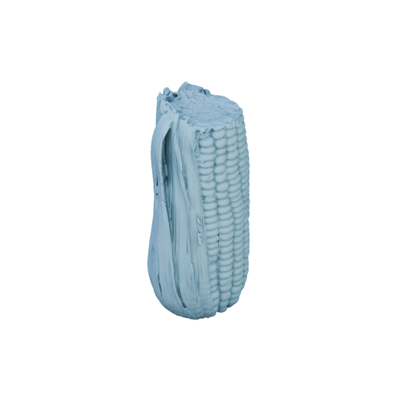} \\

\includegraphics[width=\qcwidth, valign=m]{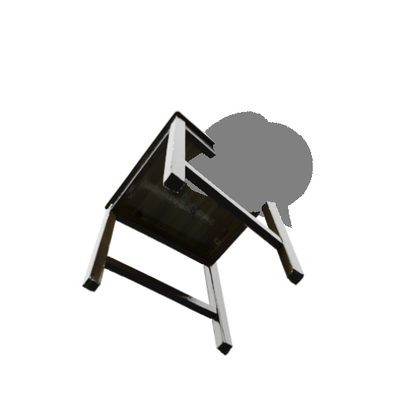} & 
\includegraphics[width=\qcwidth, valign=m]{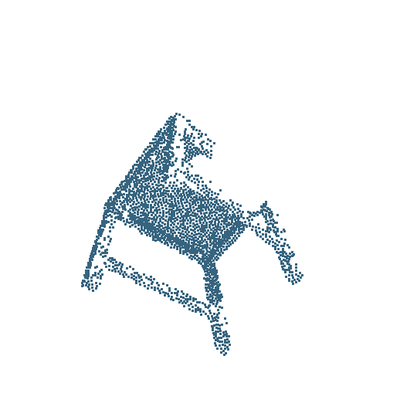} & 
\includegraphics[width=\qcwidth, valign=m]{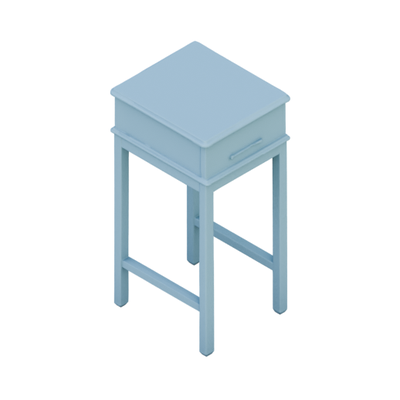} & 
\includegraphics[width=\qcwidth, valign=m]{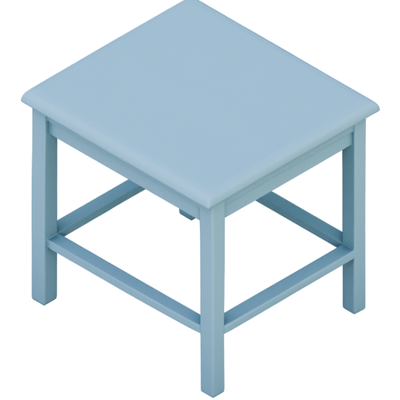} & 
\includegraphics[width=\qcwidth, valign=m]{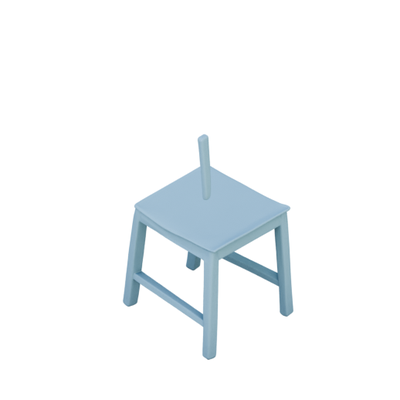} & 
\includegraphics[width=\qcwidth, valign=m]{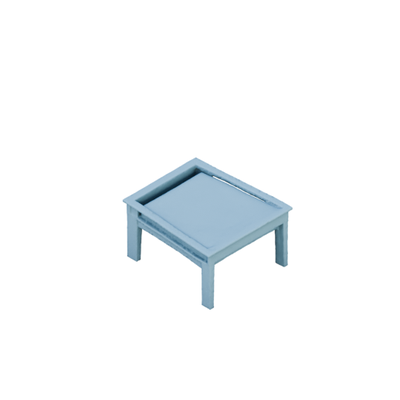} & 
\includegraphics[width=\qcwidth, valign=m]{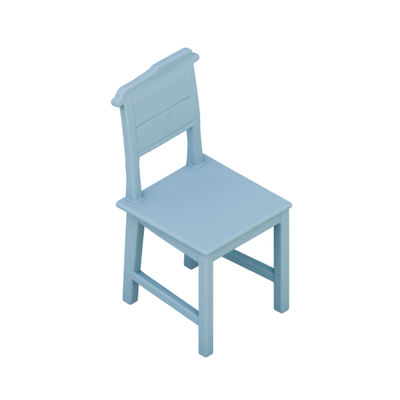} & 
\includegraphics[width=\qcwidth, valign=m]{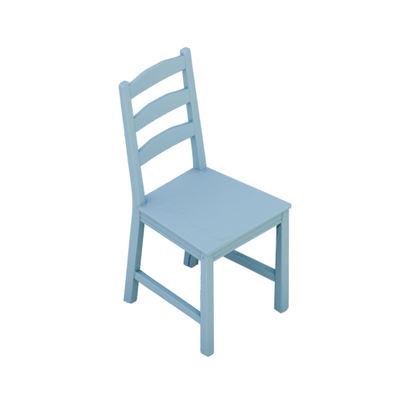} \\

\includegraphics[width=\qcwidth, valign=m]{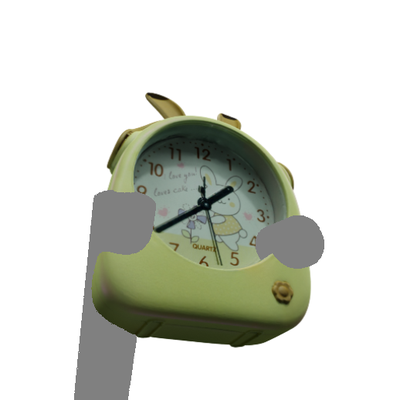} & 
\includegraphics[width=\qcwidth, valign=m]{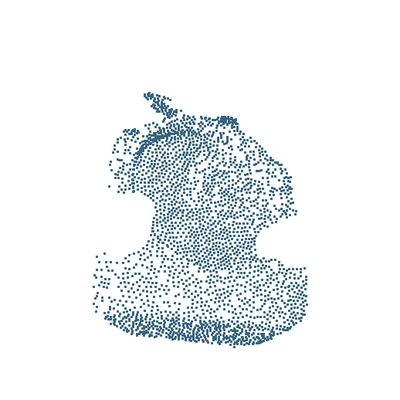} & 
\includegraphics[width=\qcwidth, valign=m]{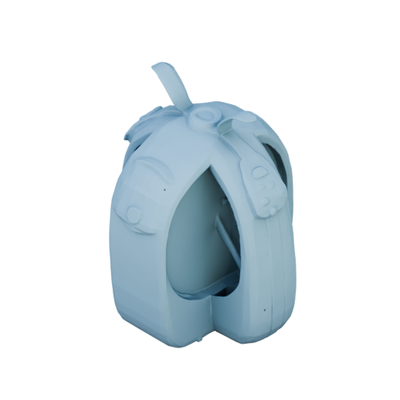} & 
\includegraphics[width=\qcwidth, valign=m]{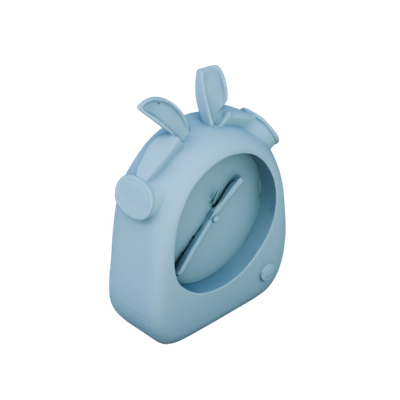} & 
\includegraphics[width=\qcwidth, valign=m]{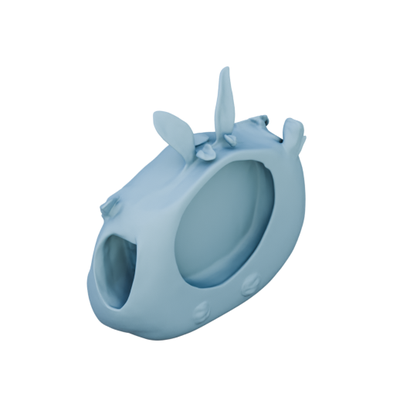} & 
\includegraphics[width=\qcwidth, valign=m]{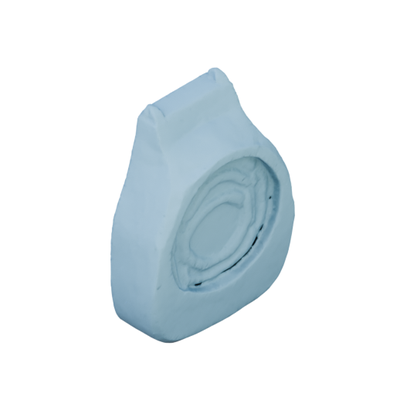} & 
\includegraphics[width=\qcwidth, valign=m]{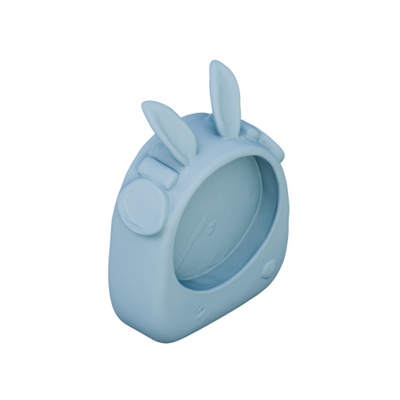} & 
\includegraphics[width=\qcwidth, valign=m]{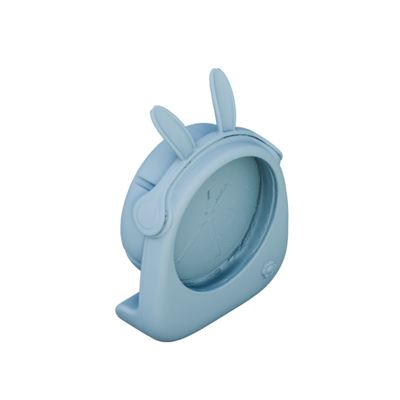} \\

  \end{tabular}
}
  
    \caption{Single-view input results}
    \label{fig:supp_omni_sv_with_occ}
  \end{subfigure}

  \vspace{2mm}

  \begin{subfigure}{\linewidth}

  \centering
\resizebox{\linewidth}{!} {
\begin{tabular}{@{}cc|cccc|c@{}}

\scalebox{0.5}{Input View} &
\scalebox{0.5}{Input Points} & 
\scalebox{0.5}{Amodal3R} & 
\scalebox{0.5}{Hy3D-Omni} & 
\scalebox{0.5}{ShapeR} & 
\scalebox{0.5}{\textbf{Ours}} & 
\scalebox{0.5}{Ground Truth} \\ 

\includegraphics[width=\qcwidth, valign=m]{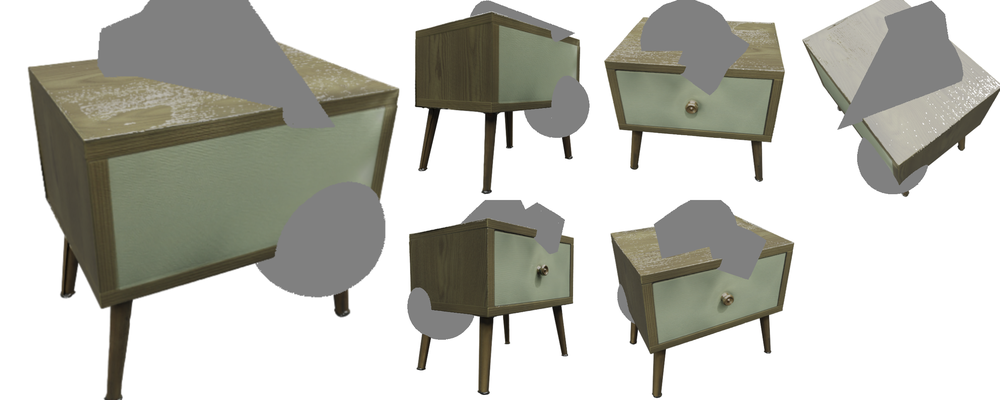} & 
\includegraphics[width=\qcwidth, valign=m]{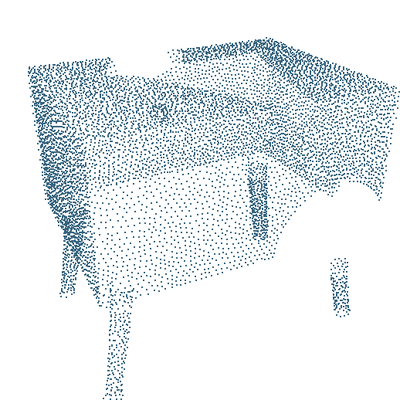} & 
\includegraphics[width=\qcwidth, valign=m]{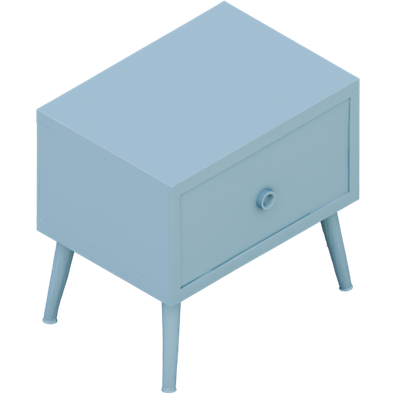} & 
\includegraphics[width=\qcwidth, valign=m]{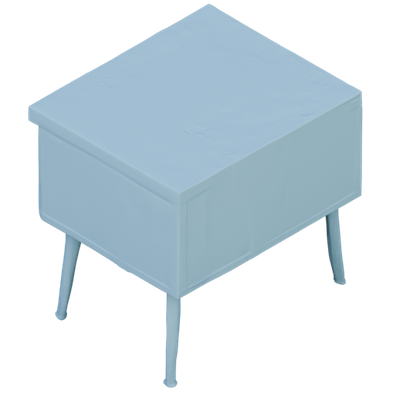} & 
\includegraphics[width=\qcwidth, valign=m]{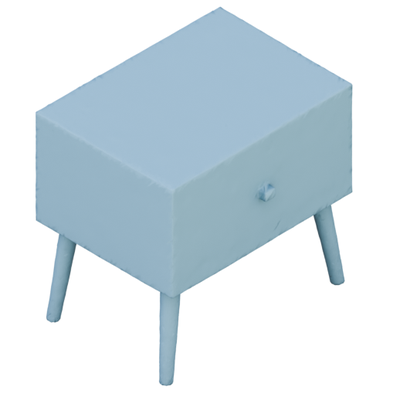} & 
\includegraphics[width=\qcwidth, valign=m]{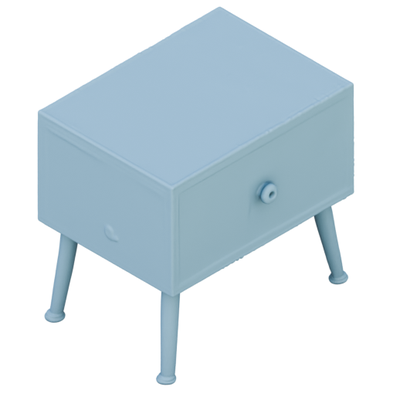} & 
\includegraphics[width=\qcwidth, valign=m]{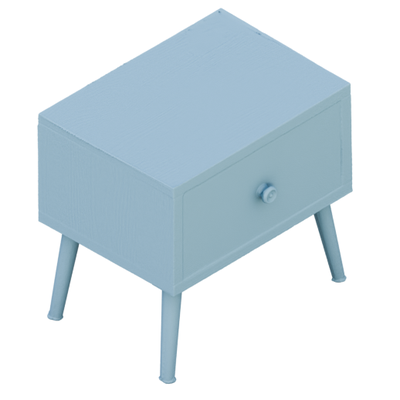} \\

\includegraphics[width=\qcwidth, valign=m]{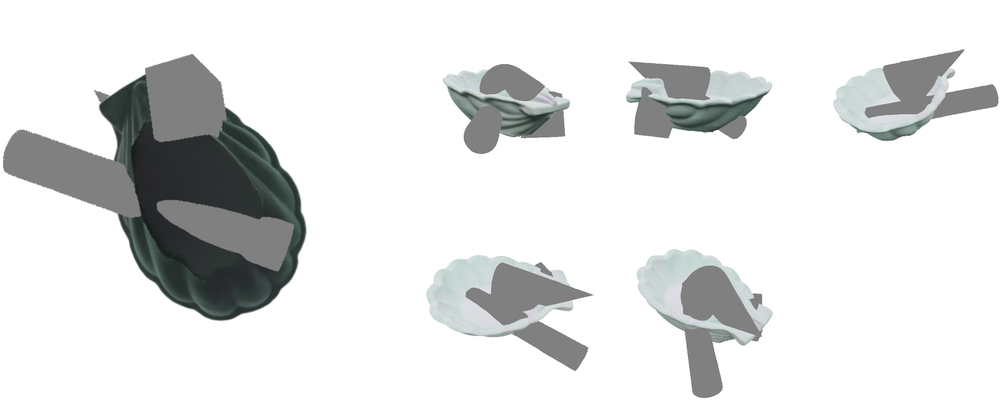} & 
\includegraphics[width=\qcwidth, valign=m]{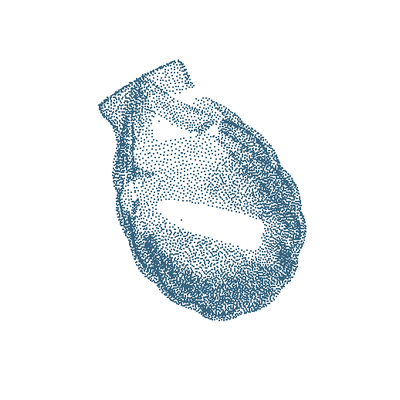} & 
\includegraphics[width=\qcwidth, valign=m]{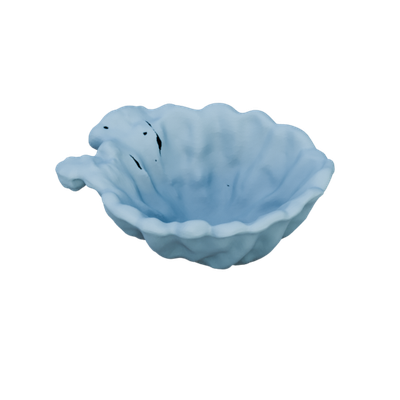} & 
\includegraphics[width=\qcwidth, valign=m]{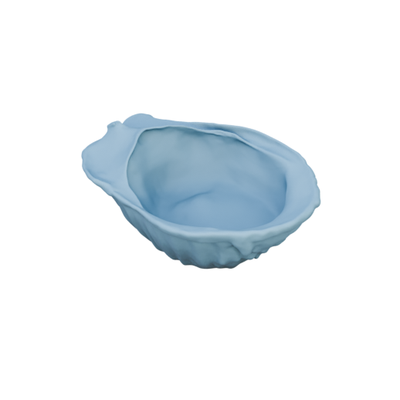} & 
\includegraphics[width=\qcwidth, valign=m]{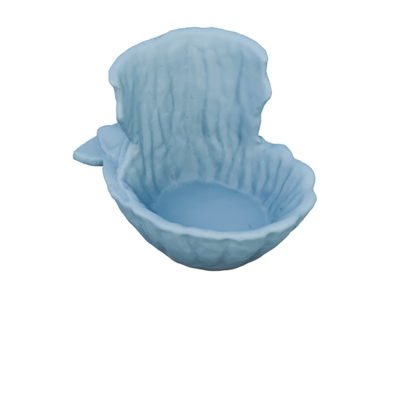} & 
\includegraphics[width=\qcwidth, valign=m]{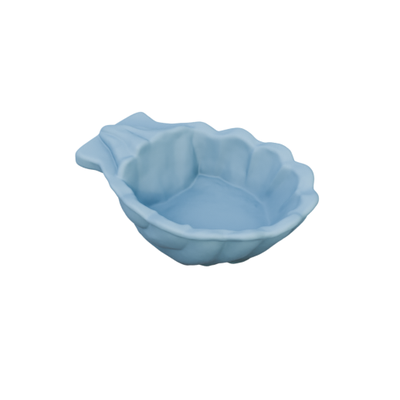} & 
\includegraphics[width=\qcwidth, valign=m]{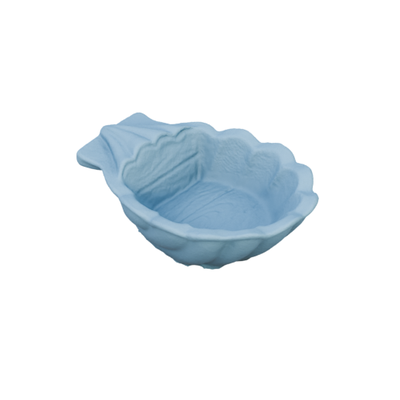} \\

\includegraphics[width=\qcwidth, valign=m]{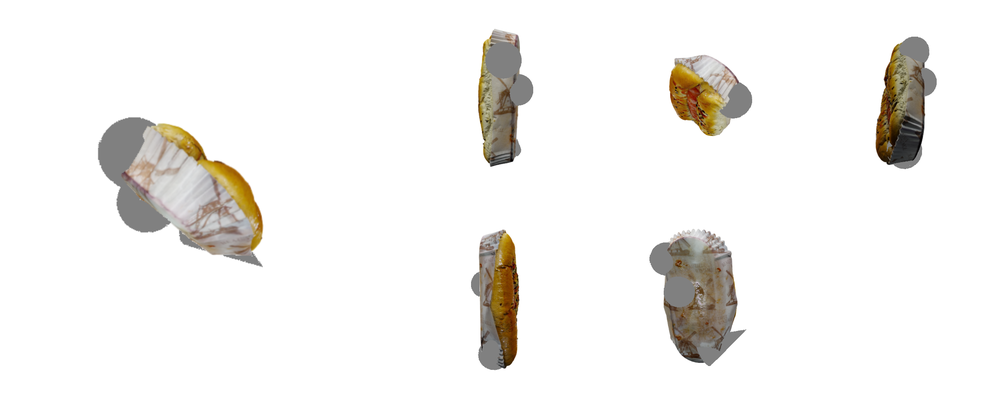} & 
\includegraphics[width=\qcwidth, valign=m]{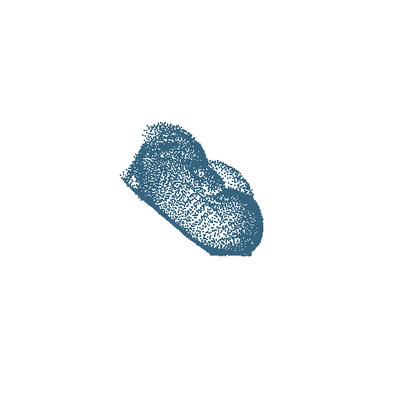} & 
\includegraphics[width=\qcwidth, valign=m]{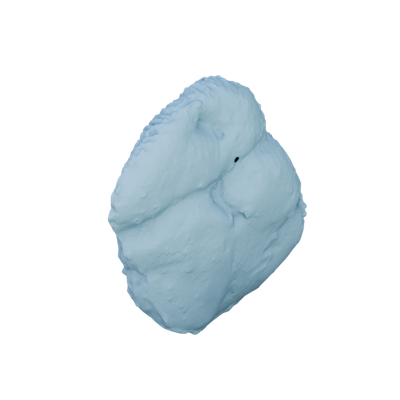} & 
\includegraphics[width=\qcwidth, valign=m]{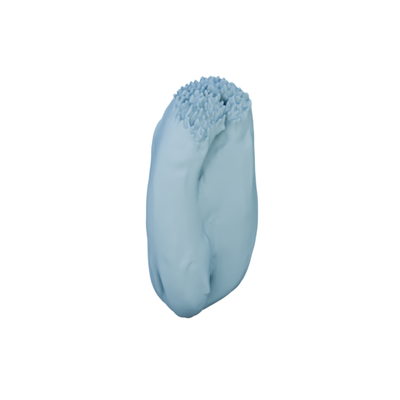} & 
\includegraphics[width=\qcwidth, valign=m]{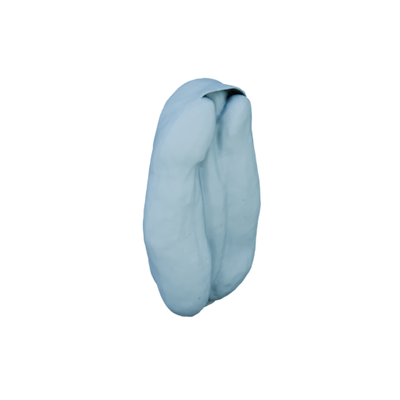} & 
\includegraphics[width=\qcwidth, valign=m]{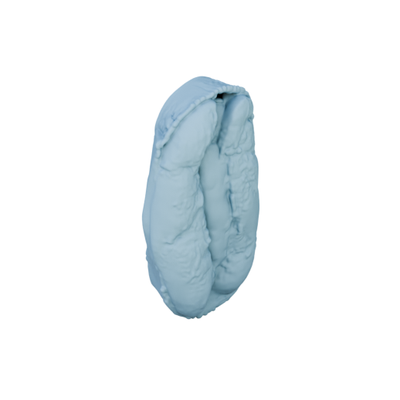} & 
\includegraphics[width=\qcwidth, valign=m]{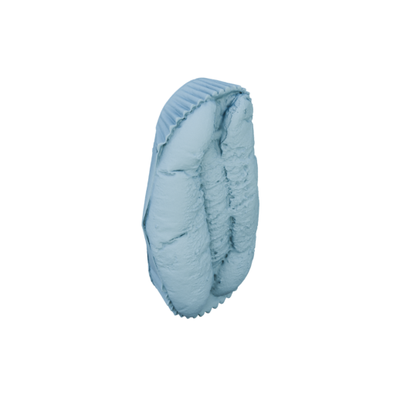} \\

\includegraphics[width=\qcwidth, valign=m]{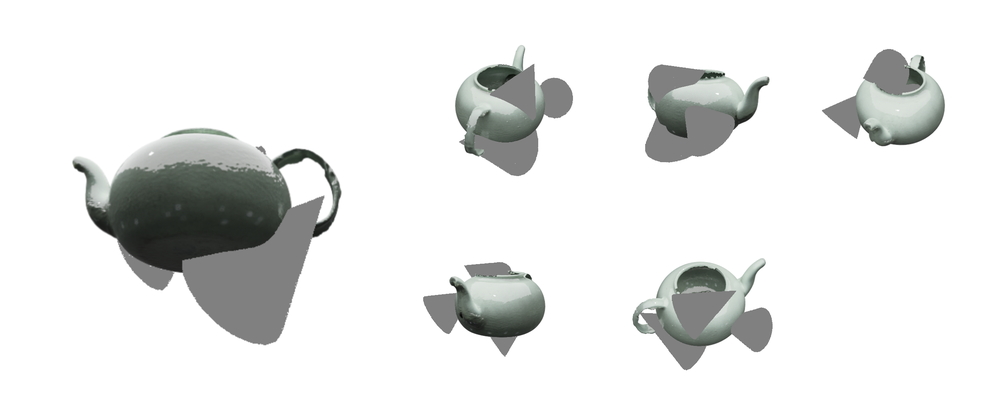} & 
\includegraphics[width=\qcwidth, valign=m]{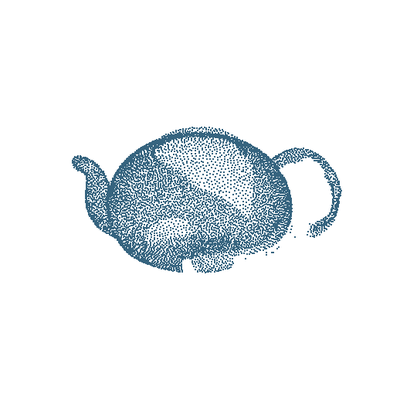} & 
\includegraphics[width=\qcwidth, valign=m]{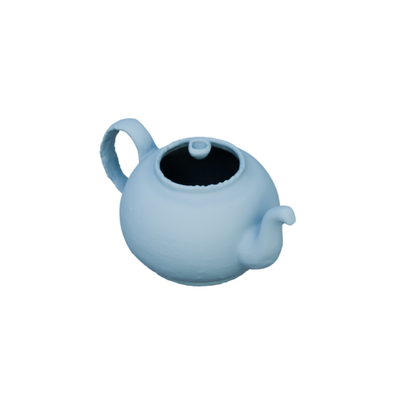} & 
\includegraphics[width=\qcwidth, valign=m]{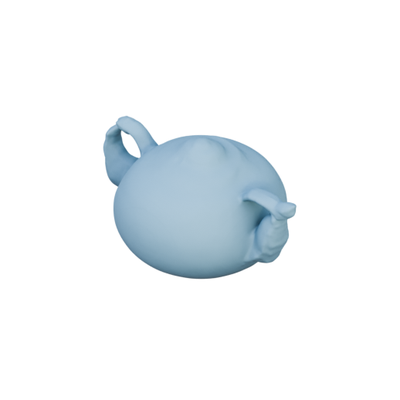} & 
\includegraphics[width=\qcwidth, valign=m]{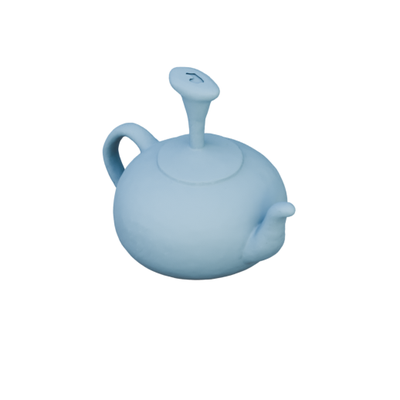} & 
\includegraphics[width=\qcwidth, valign=m]{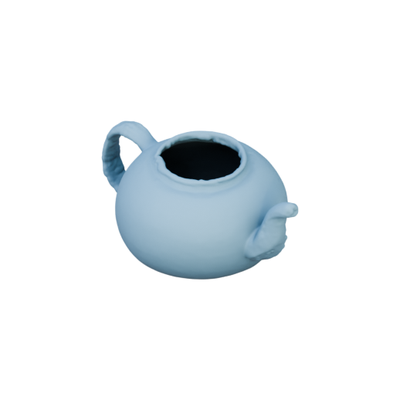} & 
\includegraphics[width=\qcwidth, valign=m]{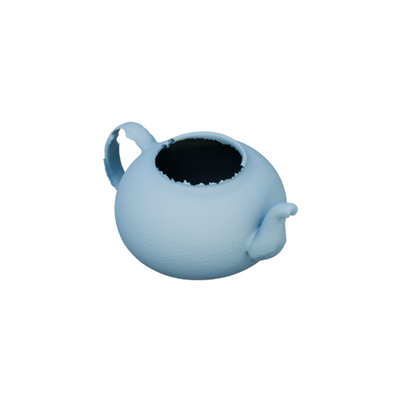} \\

\includegraphics[width=\qcwidth, valign=m]{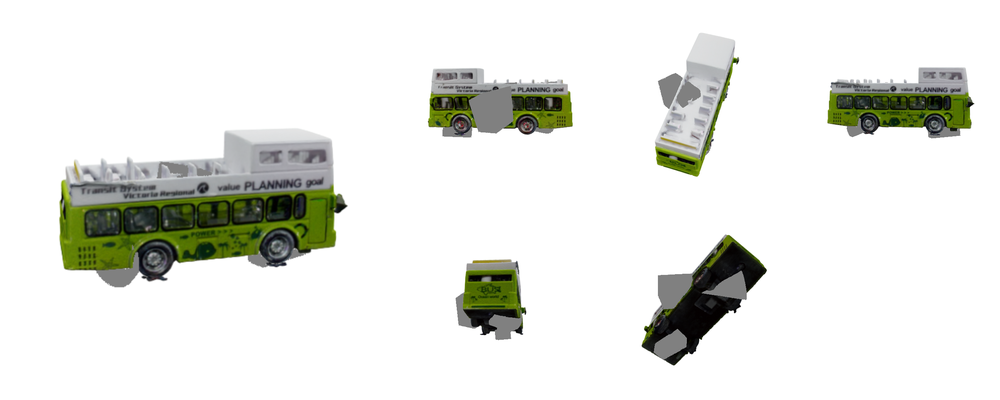} & 
\includegraphics[width=\qcwidth, valign=m]{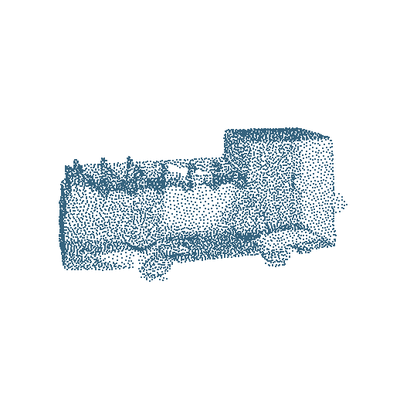} & 
\includegraphics[width=\qcwidth, valign=m]{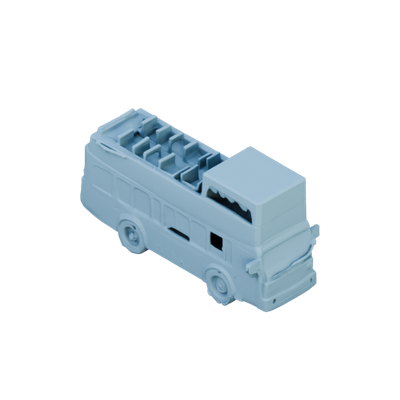} & 
\includegraphics[width=\qcwidth, valign=m]{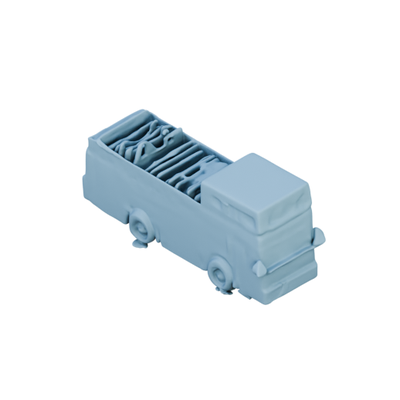} & 
\includegraphics[width=\qcwidth, valign=m]{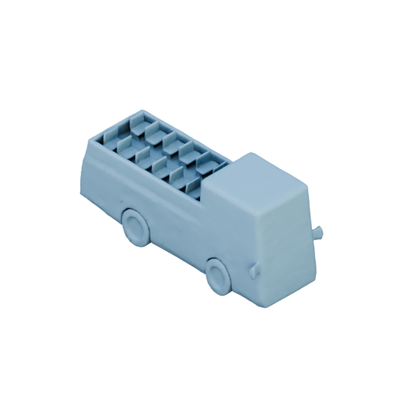} & 
\includegraphics[width=\qcwidth, valign=m]{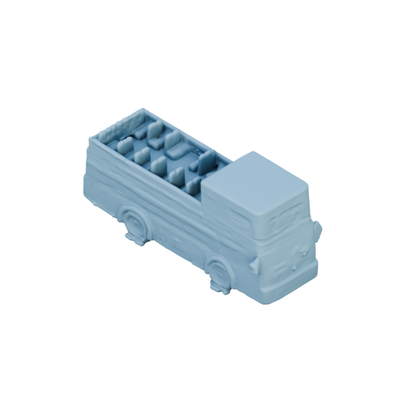} & 
\includegraphics[width=\qcwidth, valign=m]{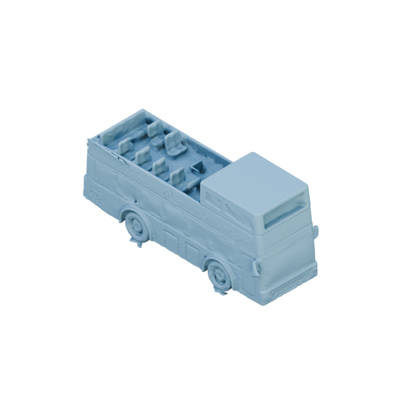} \\

\includegraphics[width=\qcwidth, valign=m]{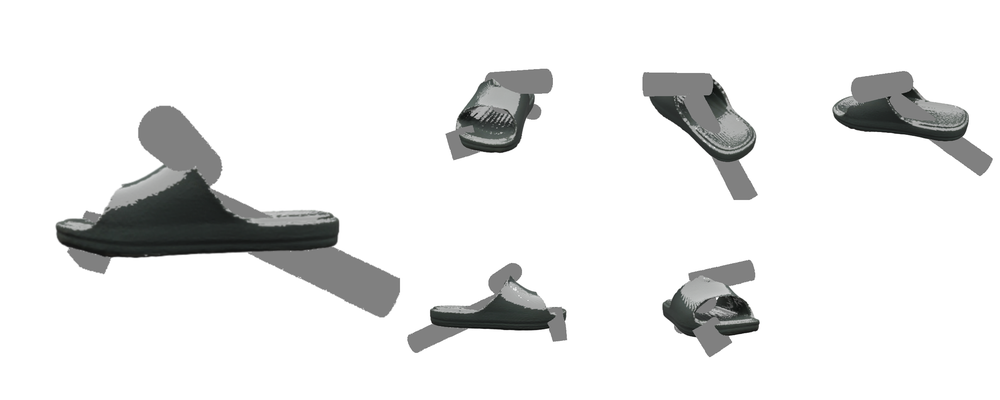} & 
\includegraphics[width=\qcwidth, valign=m]{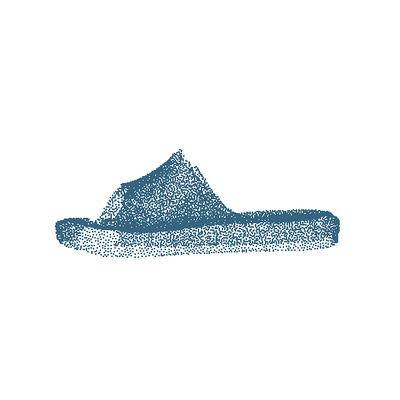} & 
\includegraphics[width=\qcwidth, valign=m]{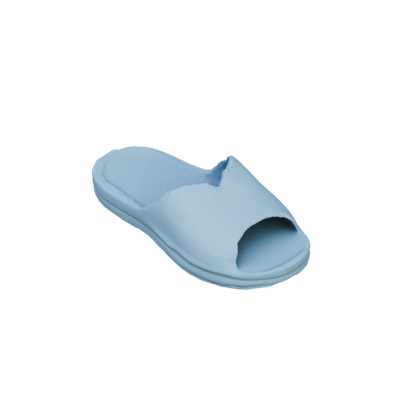} & 
\includegraphics[width=\qcwidth, valign=m]{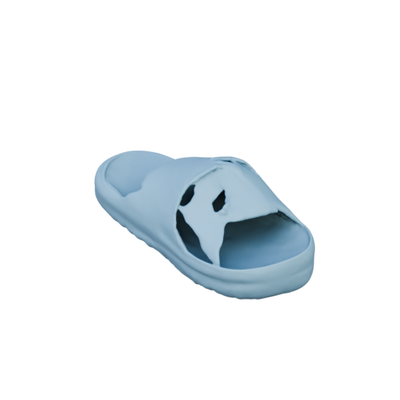} & 
\includegraphics[width=\qcwidth, valign=m]{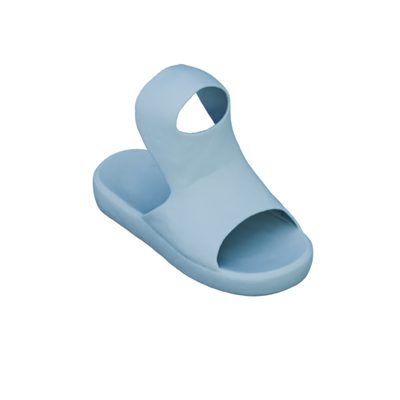} & 
\includegraphics[width=\qcwidth, valign=m]{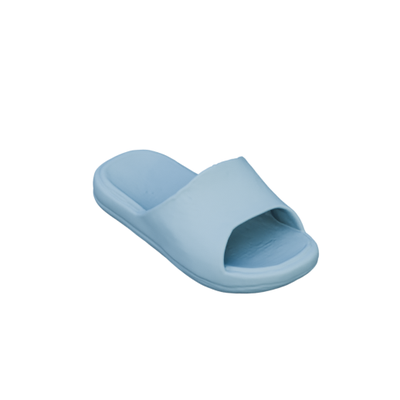} & 
\includegraphics[width=\qcwidth, valign=m]{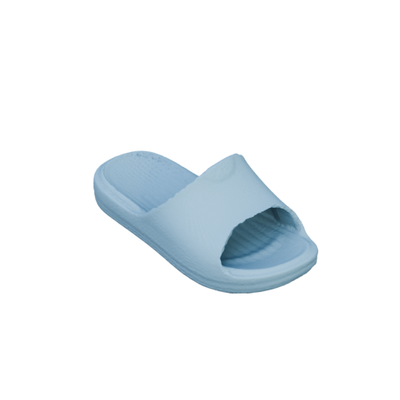} \\

  \end{tabular}
}

    \caption{Multi-view input results}
    \label{fig:supp_omni_mv_with_occ}
  \end{subfigure}

  \caption{\textbf{Qualitative Comparisons on OmniObject3D}: Single- and multi-view completion results against baselines with occlusions.}
  \label{fig:supp_omni_with_occ}
\end{figure*}

\begin{figure*}[!t]
  \centering

  \begin{subfigure}{\linewidth}
    \centering

  \resizebox{\linewidth}{!} {
    \begin{tabular}{@{}cc|ccccc|c@{}}

\scalebox{0.5}{Input View} &
\scalebox{0.5}{Input Points} & 
\scalebox{0.5}{Amodal3R} & 
\scalebox{0.5}{SAM3D} & 
\scalebox{0.5}{Hy3D-Omni} & 
\scalebox{0.5}{ShapeR} & 
\scalebox{0.5}{\textbf{Ours}} & 
\scalebox{0.5}{Ground Truth} \\ 

\includegraphics[width=\qcwidth, valign=m]{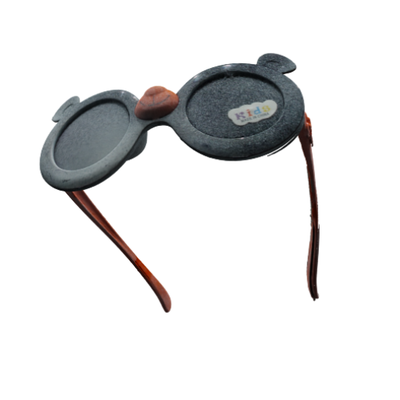} & 
\includegraphics[width=\qcwidth, valign=m]{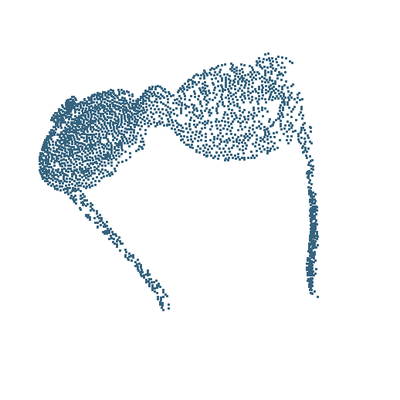} & 
\includegraphics[width=\qcwidth, valign=m]{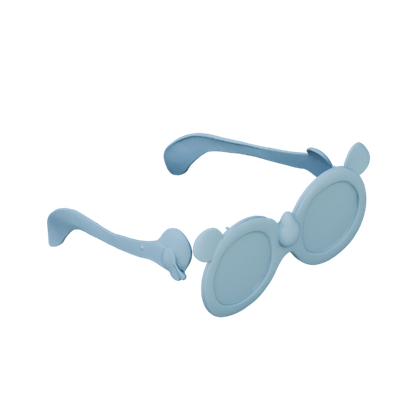} & 
\includegraphics[width=\qcwidth, valign=m]{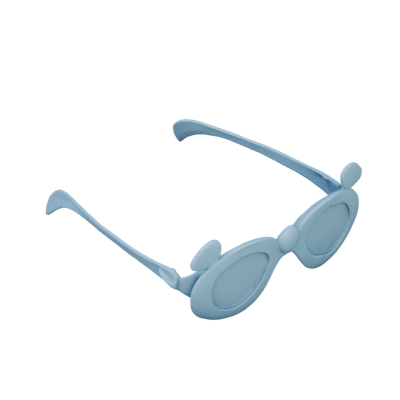} & 
\includegraphics[width=\qcwidth, valign=m]{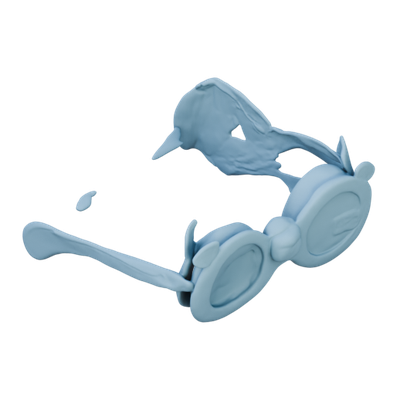} & 
\includegraphics[width=\qcwidth, valign=m]{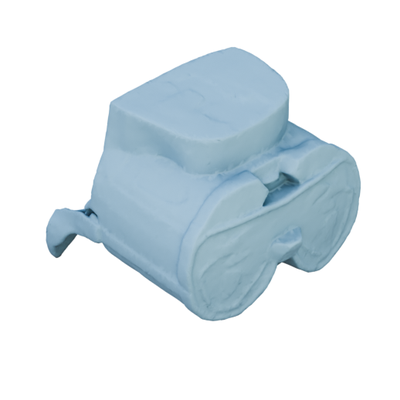} & 
\includegraphics[width=\qcwidth, valign=m]{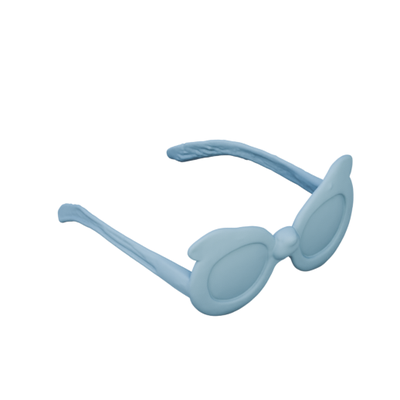} & 
\includegraphics[width=\qcwidth, valign=m]{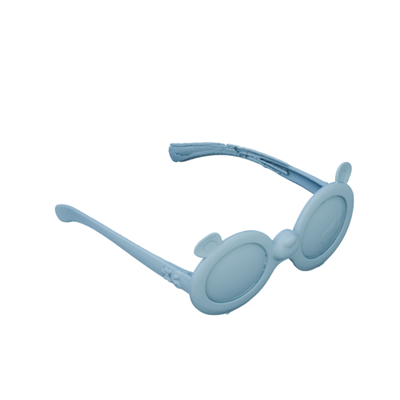} \\

\includegraphics[width=\qcwidth, valign=m]{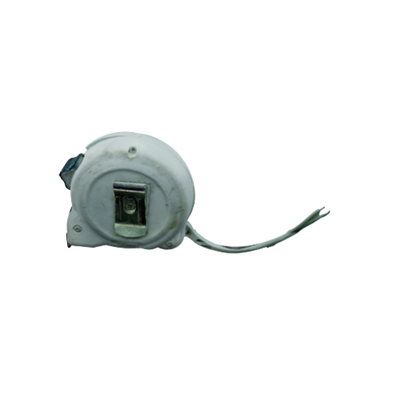} & 
\includegraphics[width=\qcwidth, valign=m]{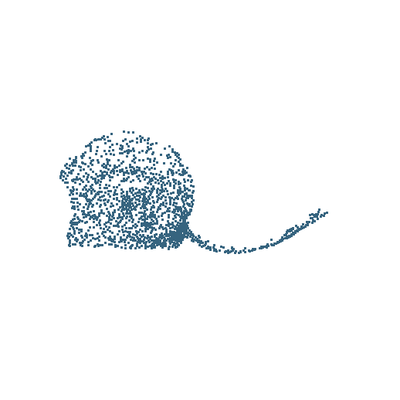} & 
\includegraphics[width=\qcwidth, valign=m]{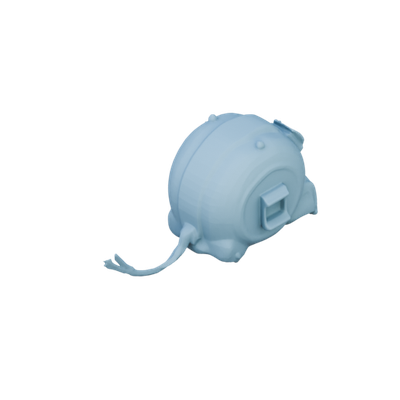} & 
\includegraphics[width=\qcwidth, valign=m]{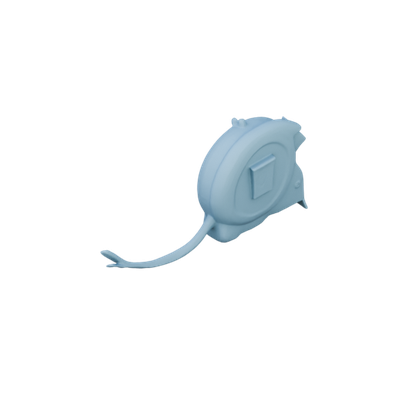} & 
\includegraphics[width=\qcwidth, valign=m]{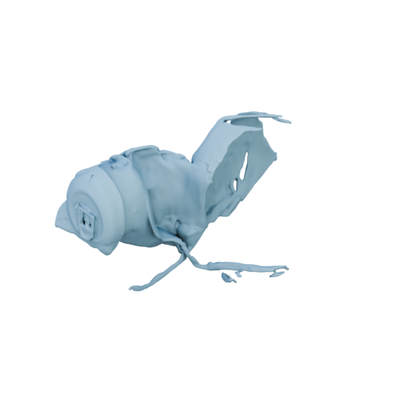} & 
\includegraphics[width=\qcwidth, valign=m]{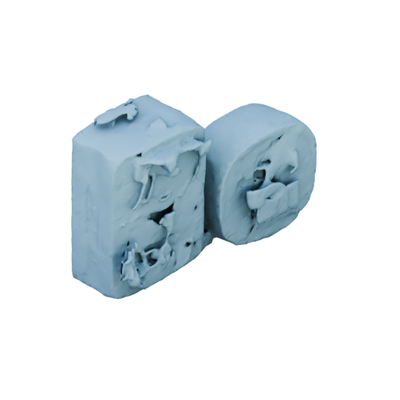} & 
\includegraphics[width=\qcwidth, valign=m]{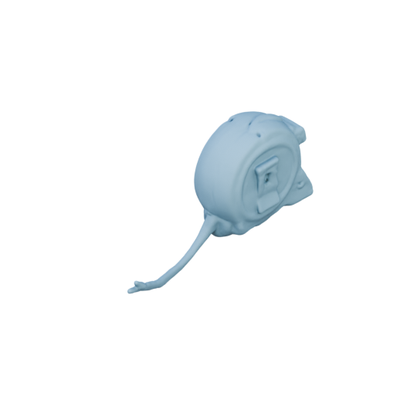} & 
\includegraphics[width=\qcwidth, valign=m]{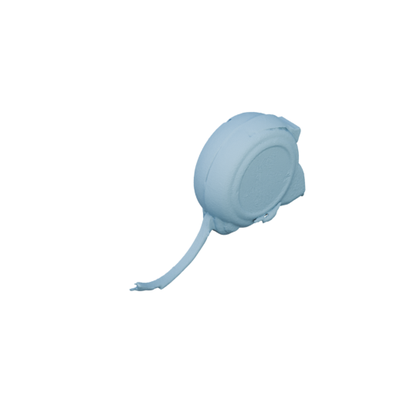} \\

\includegraphics[width=\qcwidth, valign=m]{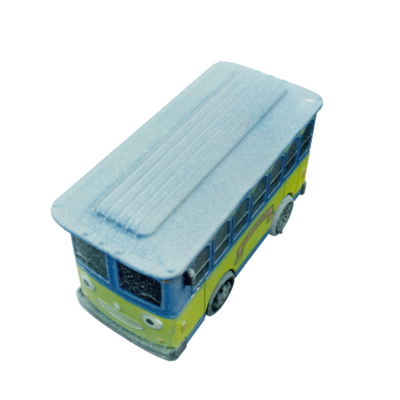} & 
\includegraphics[width=\qcwidth, valign=m]{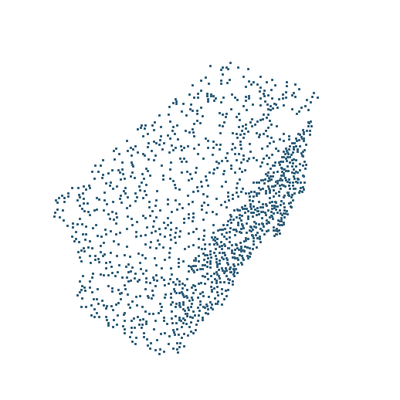} & 
\includegraphics[width=\qcwidth, valign=m]{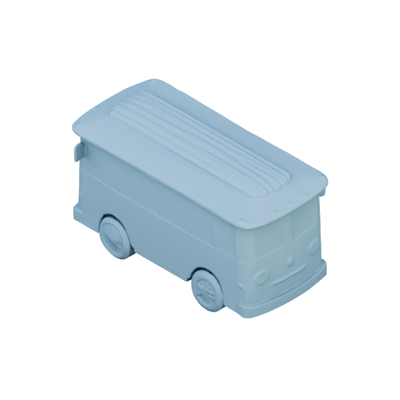} & 
\includegraphics[width=\qcwidth, valign=m]{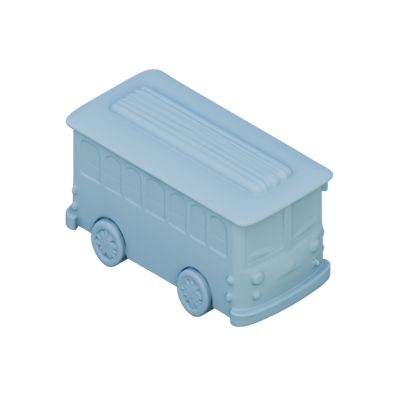} & 
\includegraphics[width=\qcwidth, valign=m]{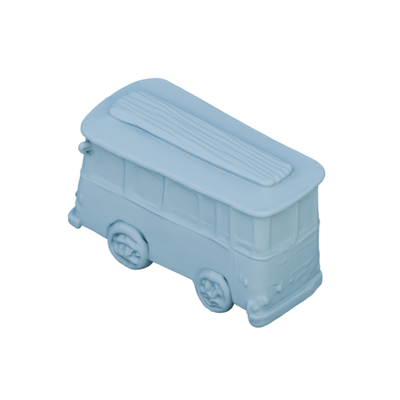} & 
\includegraphics[width=\qcwidth, valign=m]{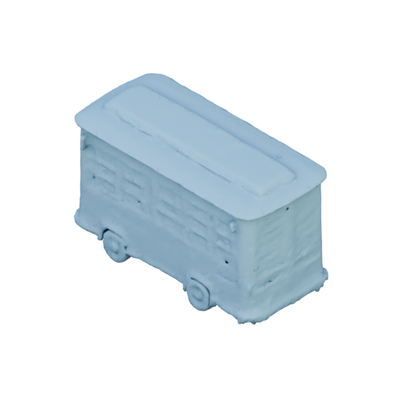} & 
\includegraphics[width=\qcwidth, valign=m]{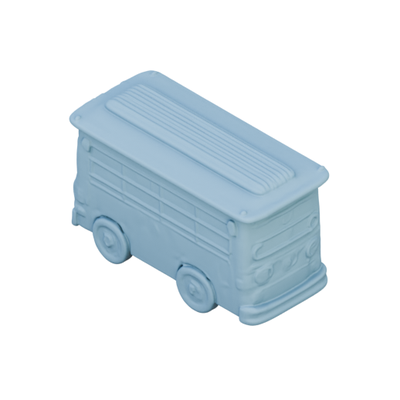} & 
\includegraphics[width=\qcwidth, valign=m]{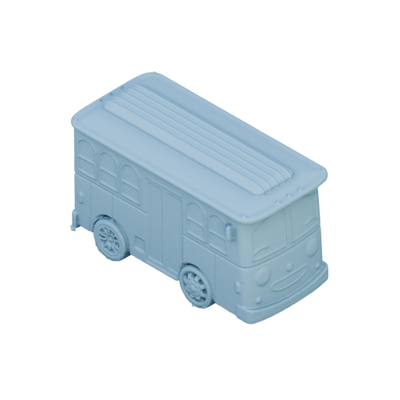} \\

\includegraphics[width=\qcwidth, valign=m]{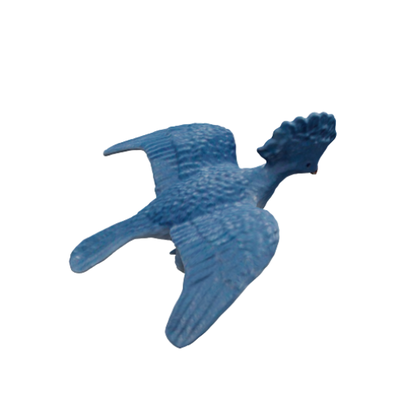} & 
\includegraphics[width=\qcwidth, valign=m]{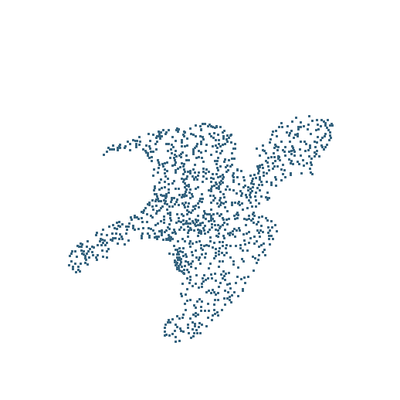} & 
\includegraphics[width=\qcwidth, valign=m]{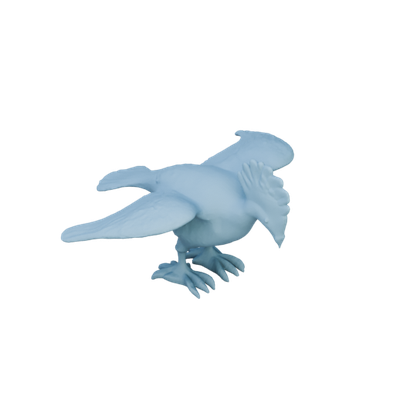} & 
\includegraphics[width=\qcwidth, valign=m]{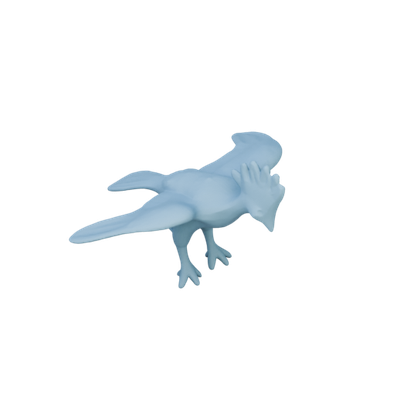} & 
\includegraphics[width=\qcwidth, valign=m]{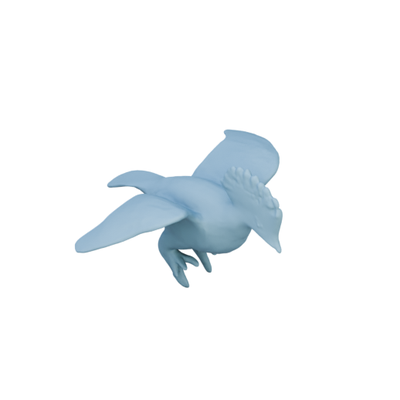} & 
\includegraphics[width=\qcwidth, valign=m]{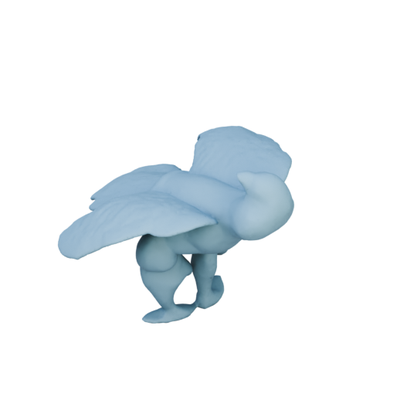} & 
\includegraphics[width=\qcwidth, valign=m]{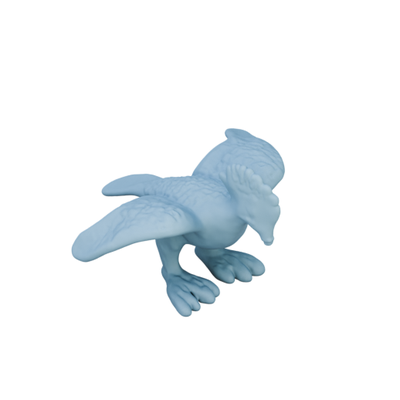} & 
\includegraphics[width=\qcwidth, valign=m]{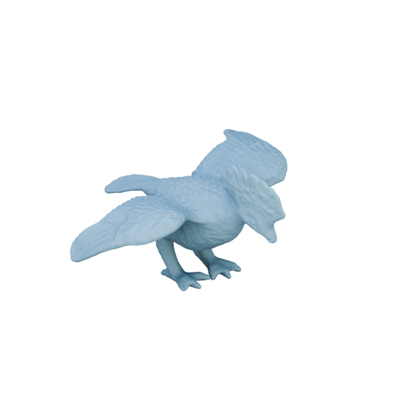} \\

\includegraphics[width=\qcwidth, valign=m]{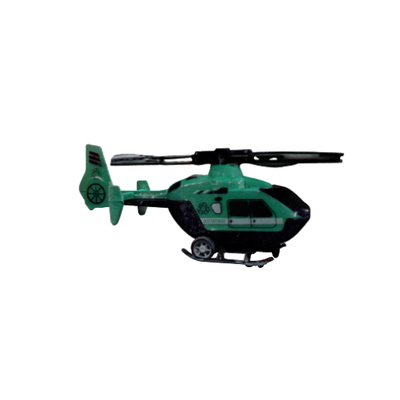} & 
\includegraphics[width=\qcwidth, valign=m]{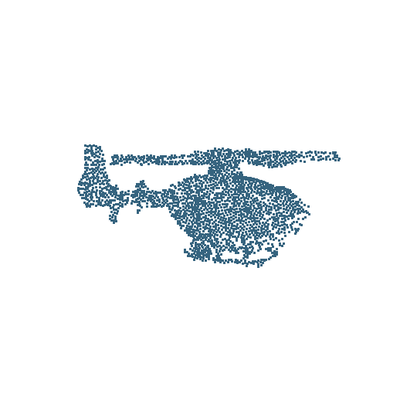} & 
\includegraphics[width=\qcwidth, valign=m]{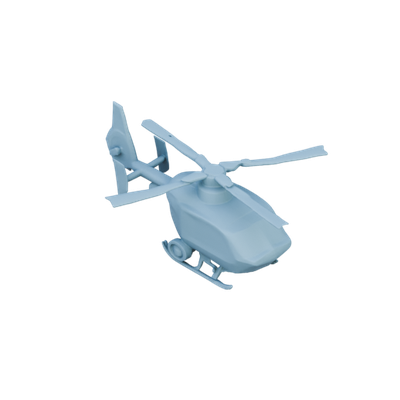} & 
\includegraphics[width=\qcwidth, valign=m]{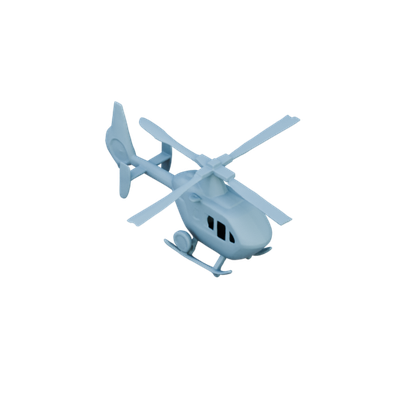} & 
\includegraphics[width=\qcwidth, valign=m]{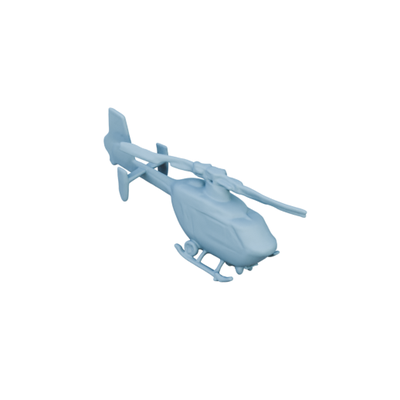} & 
\includegraphics[width=\qcwidth, valign=m]{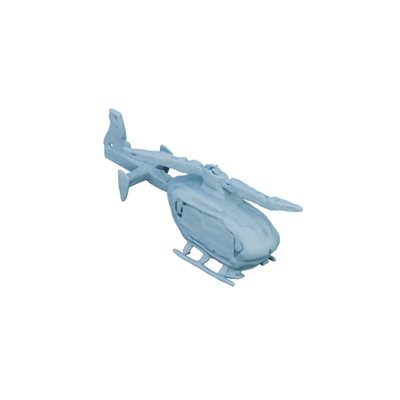} & 
\includegraphics[width=\qcwidth, valign=m]{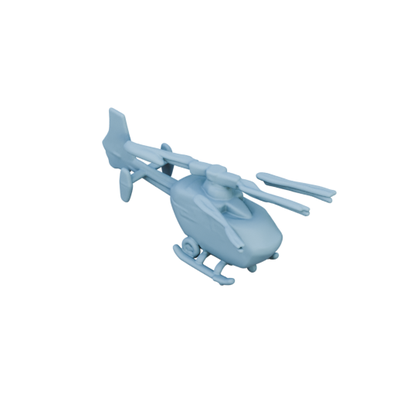} & 
\includegraphics[width=\qcwidth, valign=m]{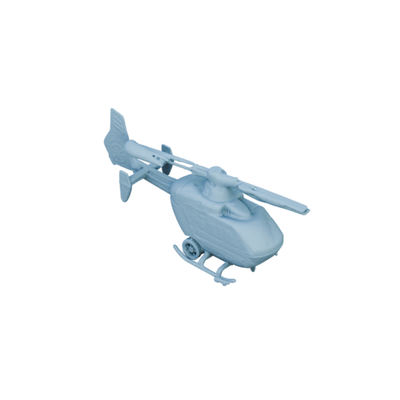} \\

  \end{tabular}
}
  
    \caption{Single-view input results}
    \label{fig:supp_omni_sv_no_occ}
  \end{subfigure}

  \vspace{2mm}

  \begin{subfigure}{\linewidth}

  \centering
\resizebox{\linewidth}{!} {
\begin{tabular}{@{}cc|cccc|c@{}}

\scalebox{0.5}{Input View} &
\scalebox{0.5}{Input Points} & 
\scalebox{0.5}{Amodal3R} & 
\scalebox{0.5}{Hy3D-Omni} & 
\scalebox{0.5}{ShapeR} & 
\scalebox{0.5}{\textbf{Ours}} & 
\scalebox{0.5}{Ground Truth} \\ 

\includegraphics[width=\qcwidth, valign=m]{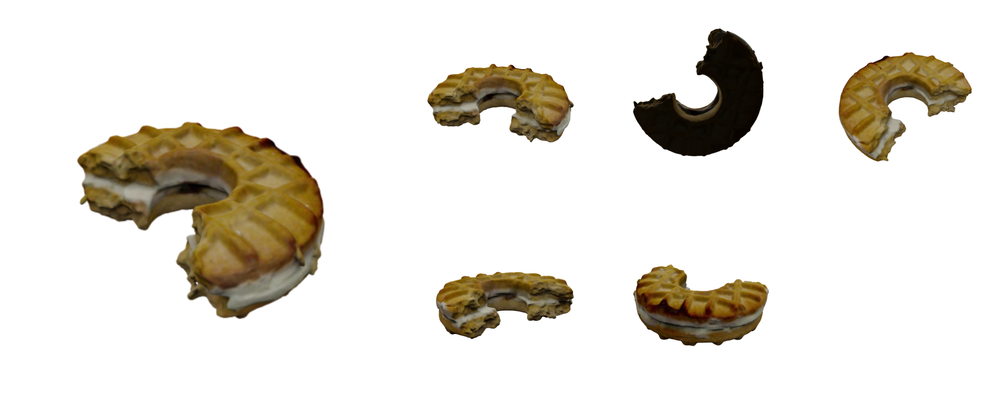} & 
\includegraphics[width=\qcwidth, valign=m]{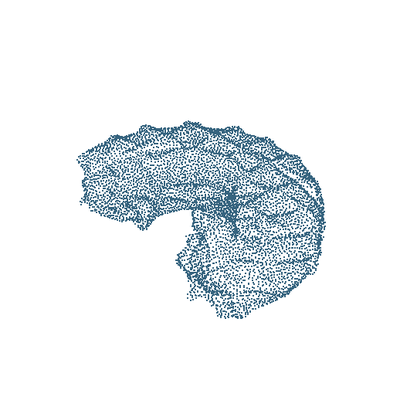} & 
\includegraphics[width=\qcwidth, valign=m]{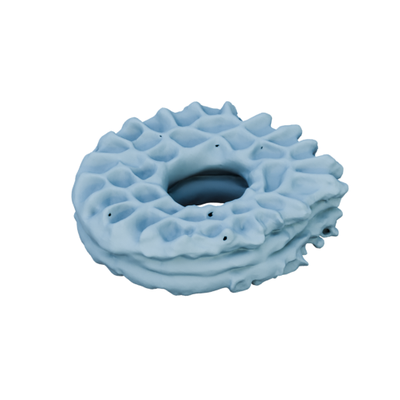} & 
\includegraphics[width=\qcwidth, valign=m]{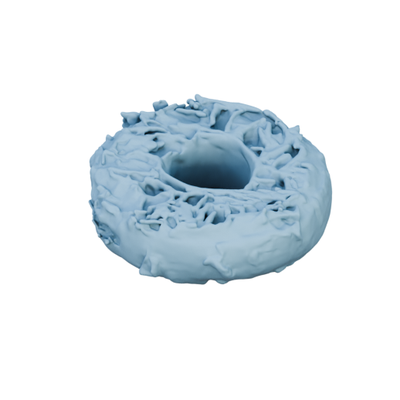} & 
\includegraphics[width=\qcwidth, valign=m]{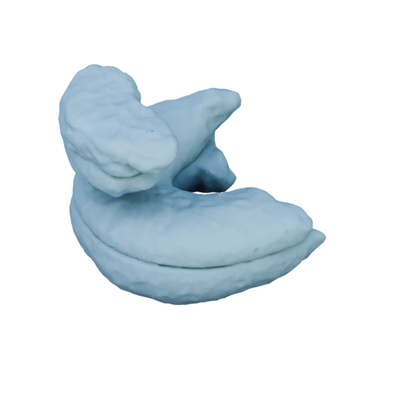} & 
\includegraphics[width=\qcwidth, valign=m]{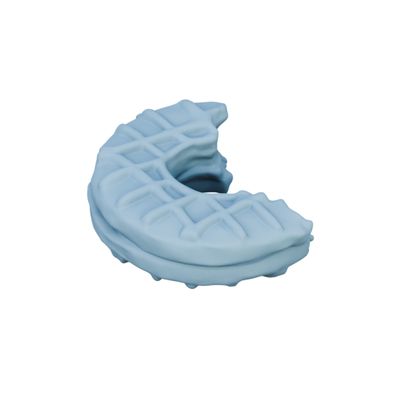} & 
\includegraphics[width=\qcwidth, valign=m]{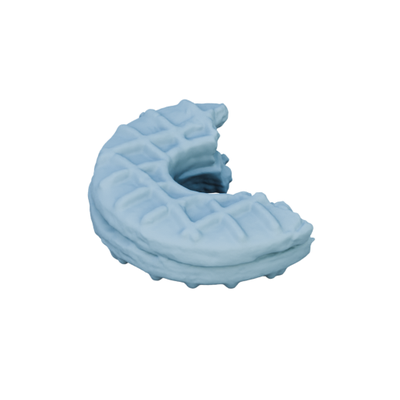} \\

\includegraphics[width=\qcwidth, valign=m]{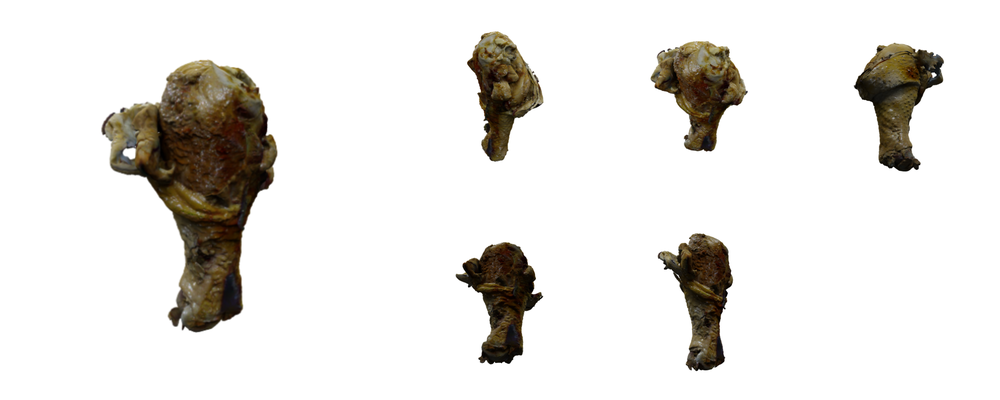} & 
\includegraphics[width=\qcwidth, valign=m]{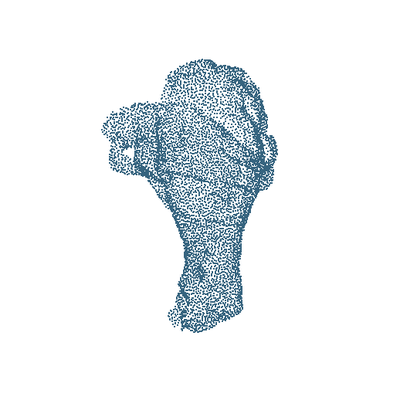} & 
\includegraphics[width=\qcwidth, valign=m]{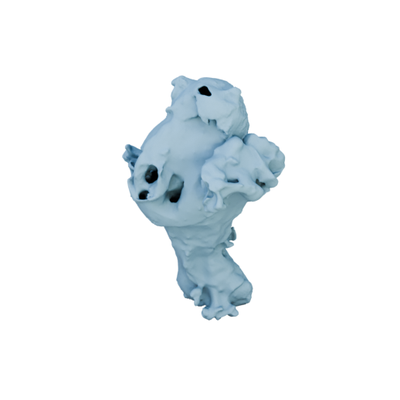} & 
\includegraphics[width=\qcwidth, valign=m]{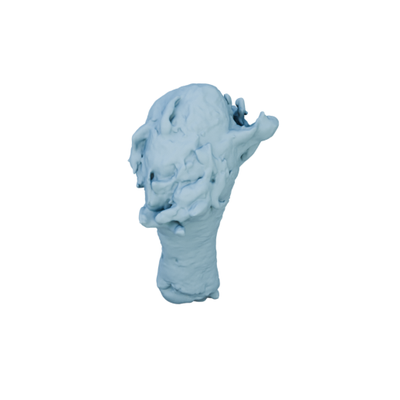} & 
\includegraphics[width=\qcwidth, valign=m]{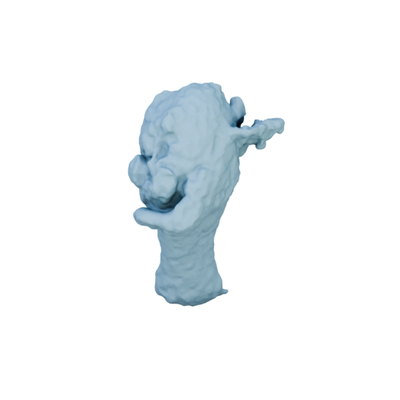} & 
\includegraphics[width=\qcwidth, valign=m]{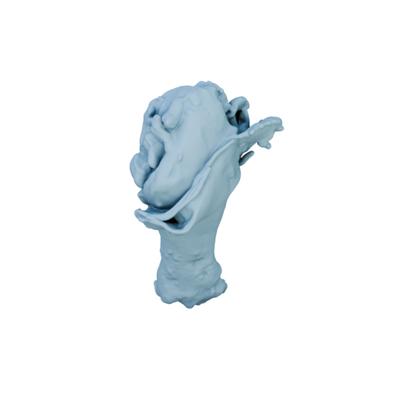} & 
\includegraphics[width=\qcwidth, valign=m]{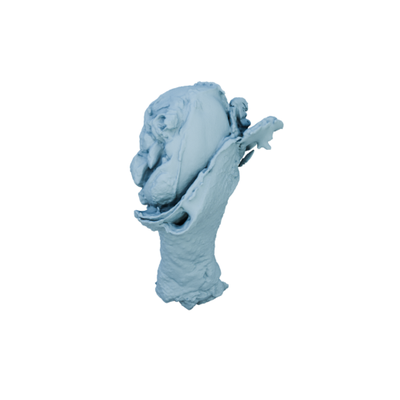} \\

\includegraphics[width=\qcwidth, valign=m]{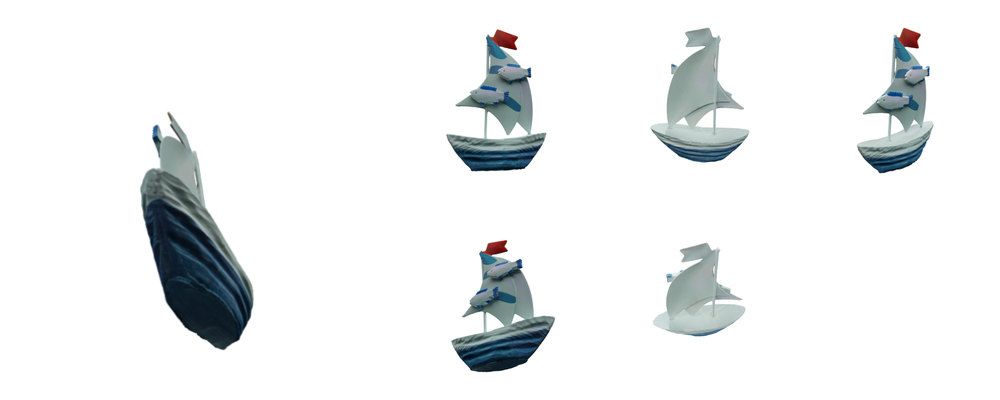} & 
\includegraphics[width=\qcwidth, valign=m]{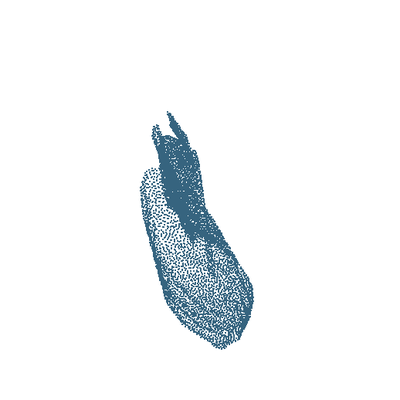} & 
\includegraphics[width=\qcwidth, valign=m]{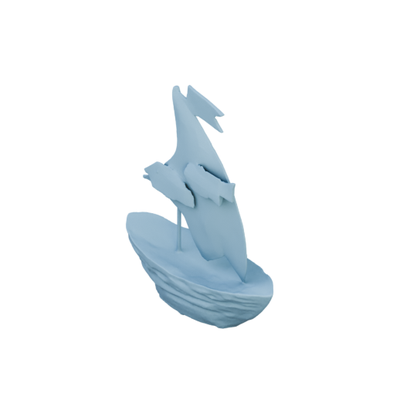} & 
\includegraphics[width=\qcwidth, valign=m]{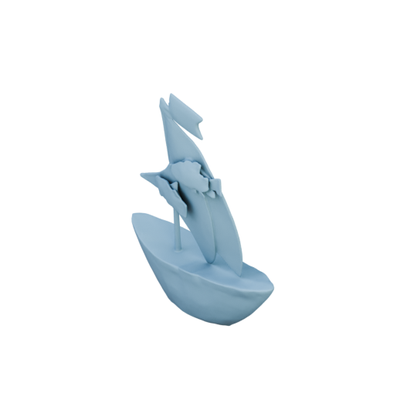} & 
\includegraphics[width=\qcwidth, valign=m]{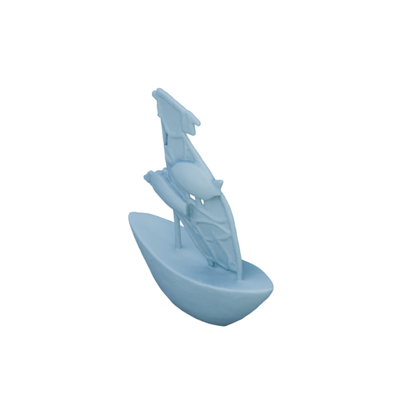} & 
\includegraphics[width=\qcwidth, valign=m]{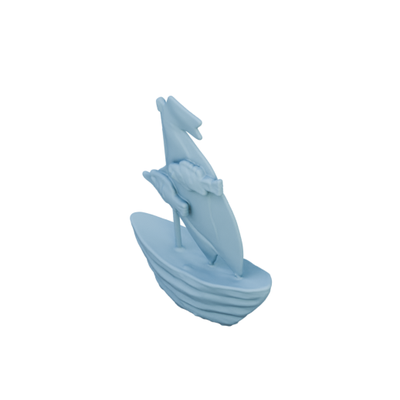} & 
\includegraphics[width=\qcwidth, valign=m]{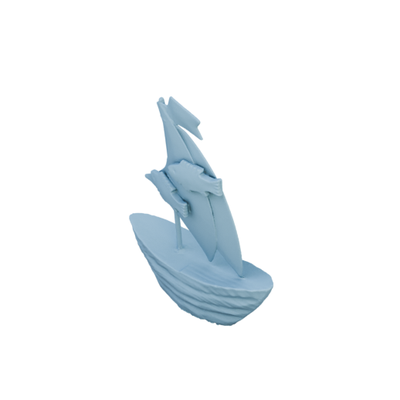} \\

\includegraphics[width=\qcwidth, valign=m]{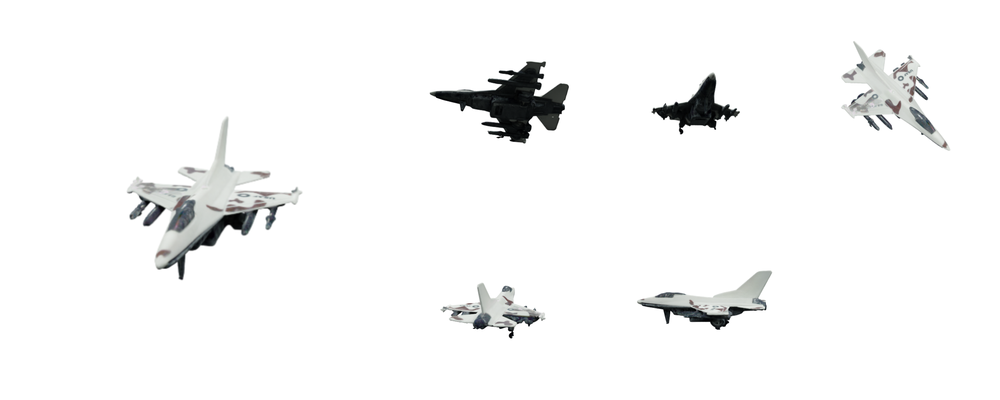} & 
\includegraphics[width=\qcwidth, valign=m]{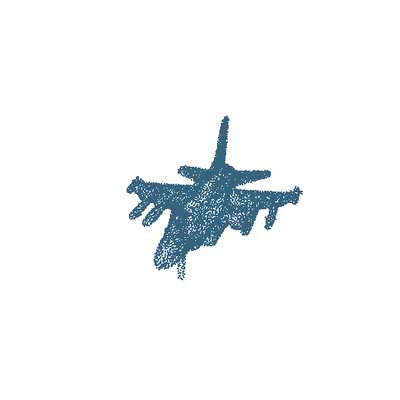} & 
\includegraphics[width=\qcwidth, valign=m]{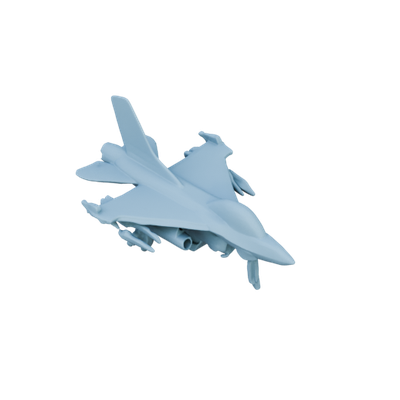} & 
\includegraphics[width=\qcwidth, valign=m]{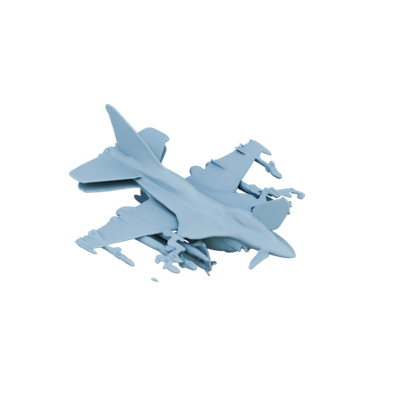} & 
\includegraphics[width=\qcwidth, valign=m]{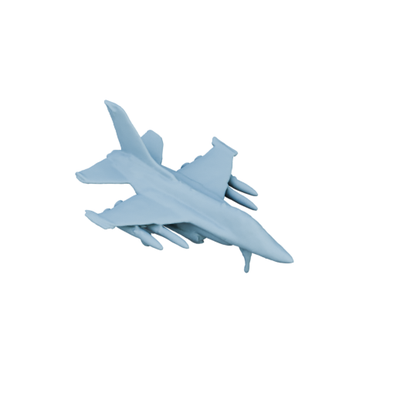} & 
\includegraphics[width=\qcwidth, valign=m]{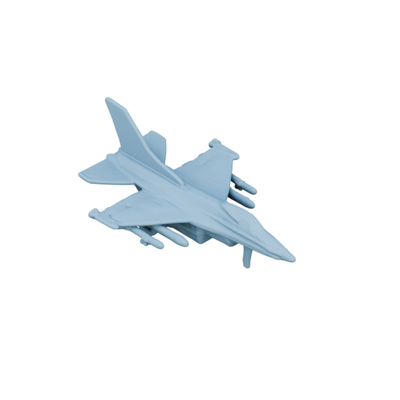} & 
\includegraphics[width=\qcwidth, valign=m]{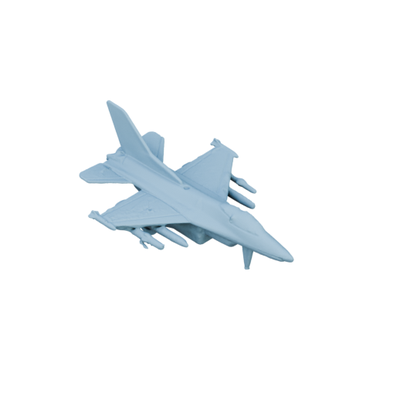} \\

\includegraphics[width=\qcwidth, valign=m]{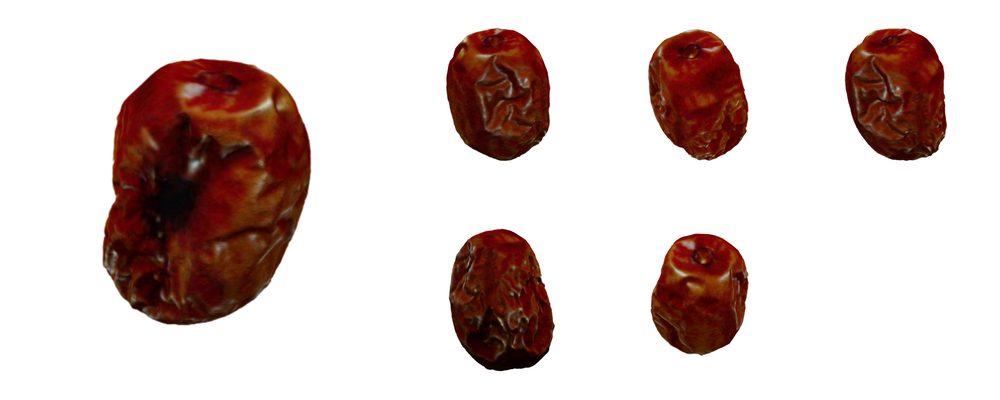} & 
\includegraphics[width=\qcwidth, valign=m]{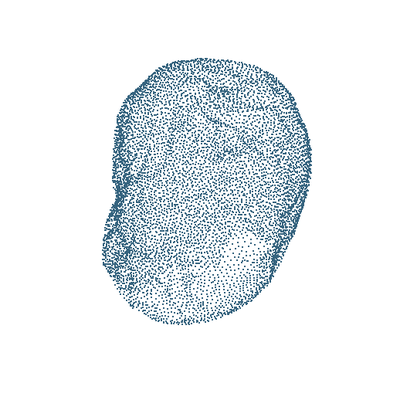} & 
\includegraphics[width=\qcwidth, valign=m]{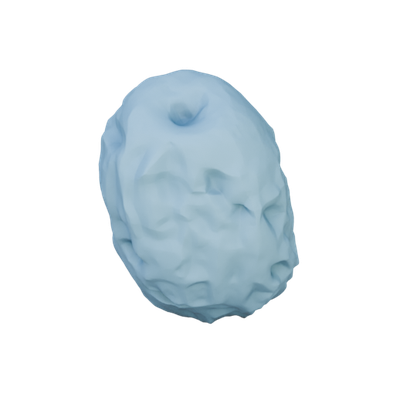} & 
\includegraphics[width=\qcwidth, valign=m]{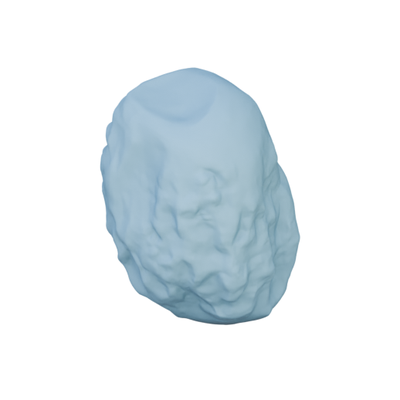} & 
\includegraphics[width=\qcwidth, valign=m]{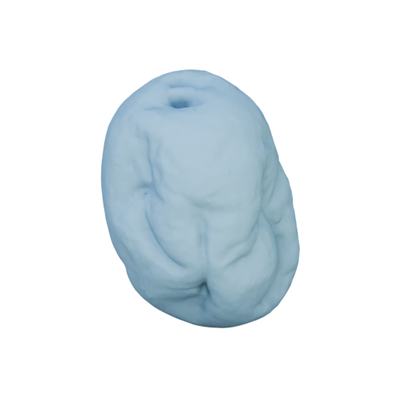} & 
\includegraphics[width=\qcwidth, valign=m]{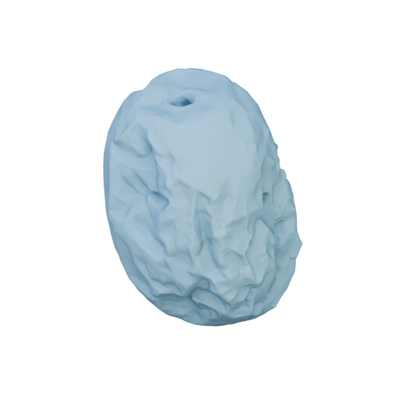} & 
\includegraphics[width=\qcwidth, valign=m]{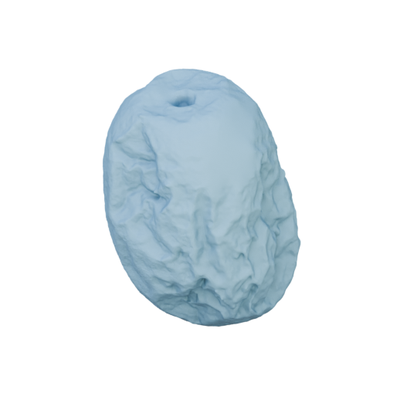} \\

\includegraphics[width=\qcwidth, valign=m]{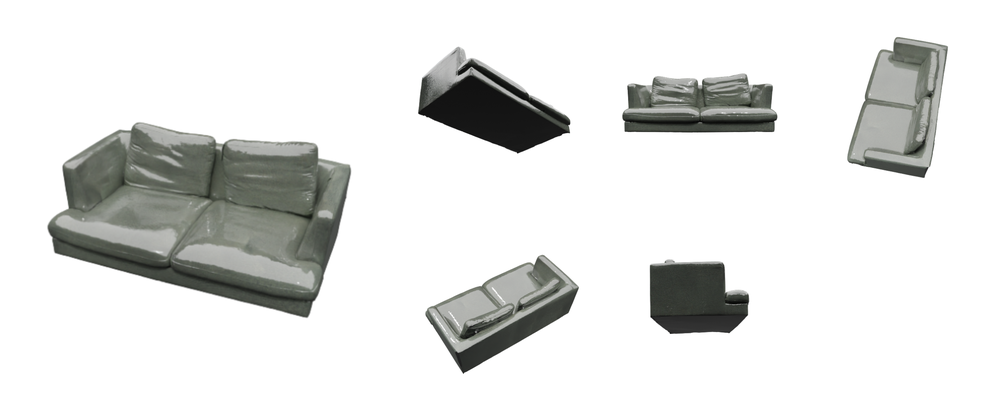} & 
\includegraphics[width=\qcwidth, valign=m]{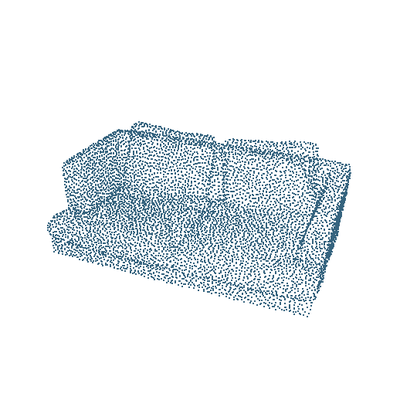} & 
\includegraphics[width=\qcwidth, valign=m]{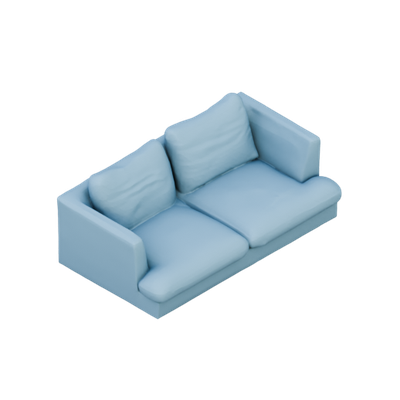} & 
\includegraphics[width=\qcwidth, valign=m]{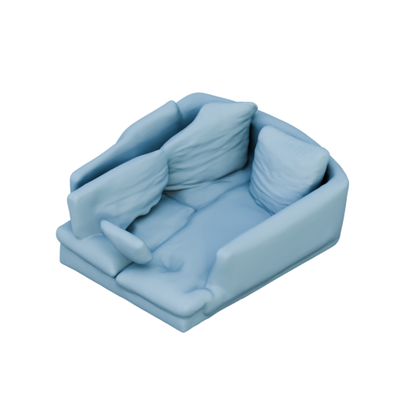} & 
\includegraphics[width=\qcwidth, valign=m]{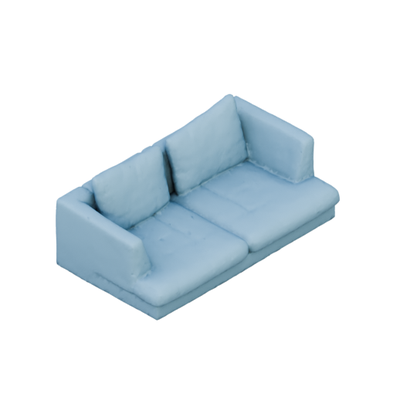} & 
\includegraphics[width=\qcwidth, valign=m]{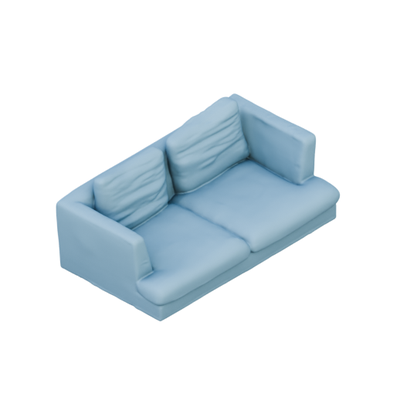} & 
\includegraphics[width=\qcwidth, valign=m]{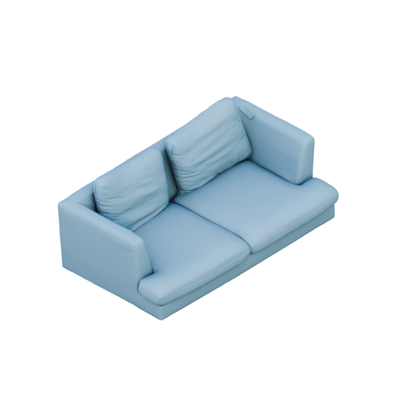} \\

  \end{tabular}
}

    \caption{Multi-view input results}
    \label{fig:supp_omni_mv_no_occ}
  \end{subfigure}

  \caption{\textbf{Qualitative Comparisons on OmniObject3D}: Single- and multi-view completion results against baselines without occlusions.}
  \label{fig:supp_omni_no_occ}
\end{figure*}

In this section, we provide additional qualitative comparisons against baseline methods to further demonstrate the performance of our method.
Fig.~\ref{fig:supp_omni_with_occ} and \ref{fig:supp_omni_no_occ} show results on the OmniObject3D dataset under occluded and unoccluded conditions. Fig.~\ref{fig:supp_toys4k} illustrates unoccluded scenarios on Toys4k.

\providecommand{\mvwidth}{0.23\linewidth}

\begin{figure*}[!t]
  \centering

  \begin{subfigure}{\linewidth}
    \centering

  \resizebox{\linewidth}{!} {
    \begin{tabular}{@{}cc|ccccc|c@{}}

\scalebox{0.5}{Input View} &
\scalebox{0.5}{Input Points} & 
\scalebox{0.5}{Amodal3R} & 
\scalebox{0.5}{SAM3D} & 
\scalebox{0.5}{Hy3D-Omni} & 
\scalebox{0.5}{ShapeR} & 
\scalebox{0.5}{\textbf{Ours}} & 
\scalebox{0.5}{Ground Truth} \\ 

\includegraphics[width=\qcwidth, valign=m]{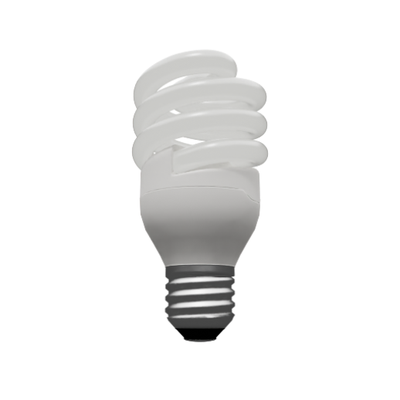} & 
\includegraphics[width=\qcwidth, valign=m]{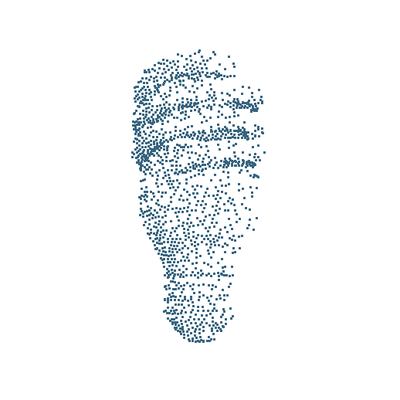} & 
\includegraphics[width=\qcwidth, valign=m]{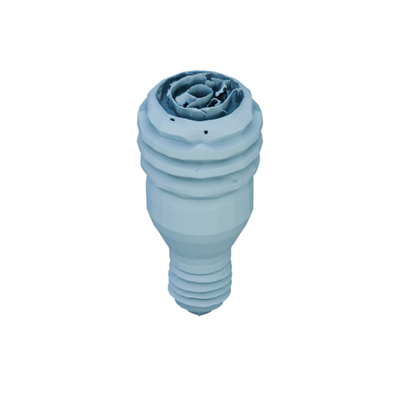} & 
\includegraphics[width=\qcwidth, valign=m]{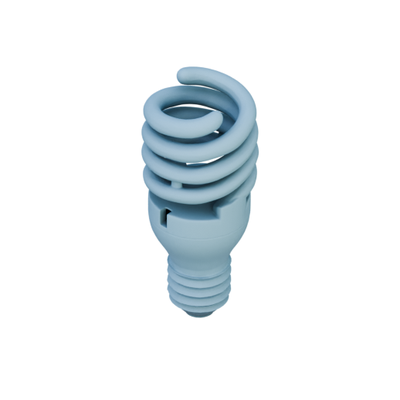} & 
\includegraphics[width=\qcwidth, valign=m]{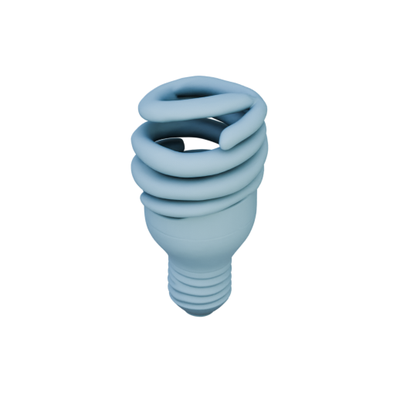} & 
\includegraphics[width=\qcwidth, valign=m]{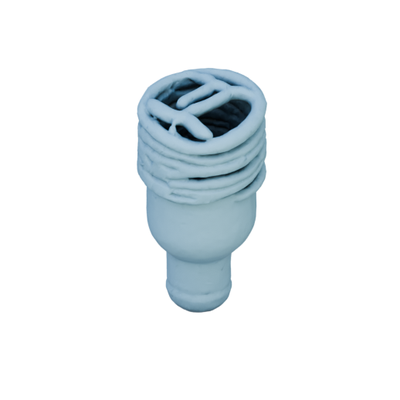} & 
\includegraphics[width=\qcwidth, valign=m]{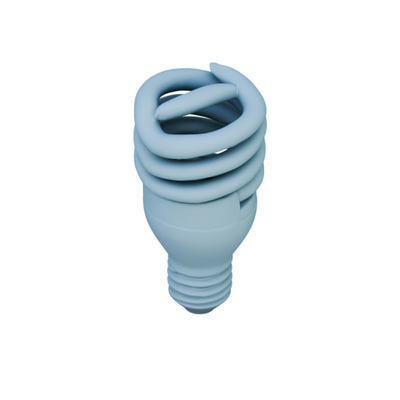} & 
\includegraphics[width=\qcwidth, valign=m]{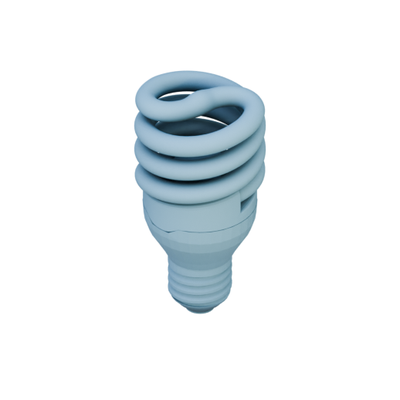} \\

\includegraphics[width=\qcwidth, valign=m]{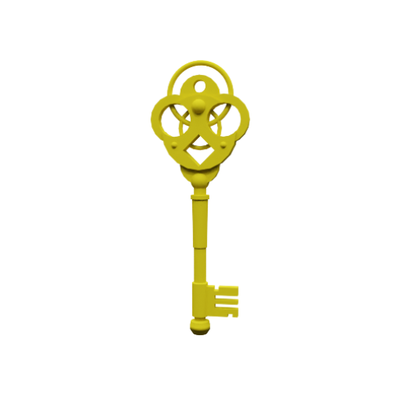} & 
\includegraphics[width=\qcwidth, valign=m]{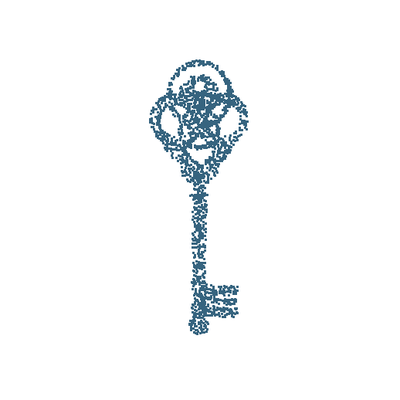} & 
\includegraphics[width=\qcwidth, valign=m]{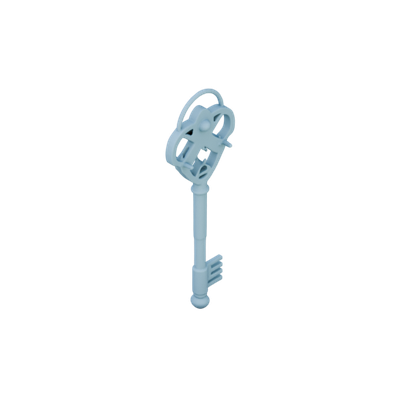} & 
\includegraphics[width=\qcwidth, valign=m]{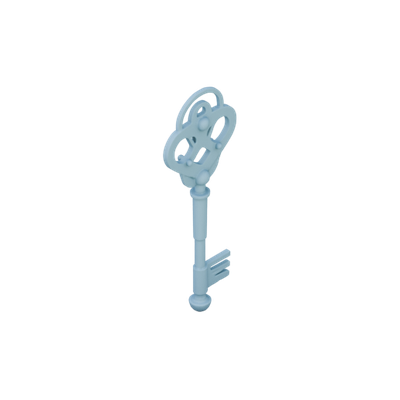} & 
\includegraphics[width=\qcwidth, valign=m]{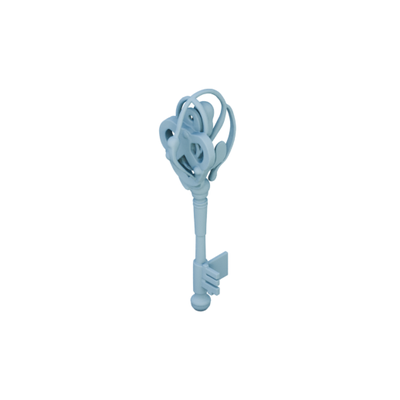} & 
\includegraphics[width=\qcwidth, valign=m]{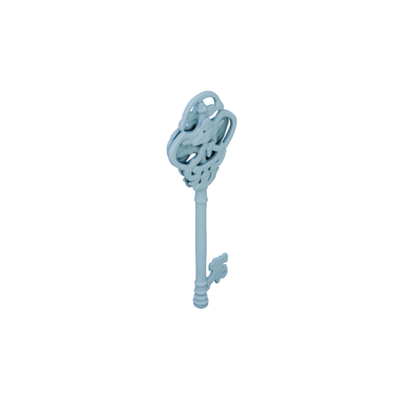} & 
\includegraphics[width=\qcwidth, valign=m]{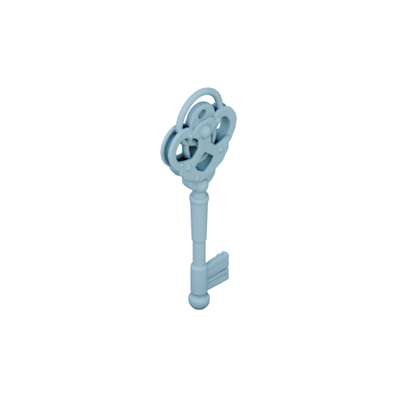} & 
\includegraphics[width=\qcwidth, valign=m]{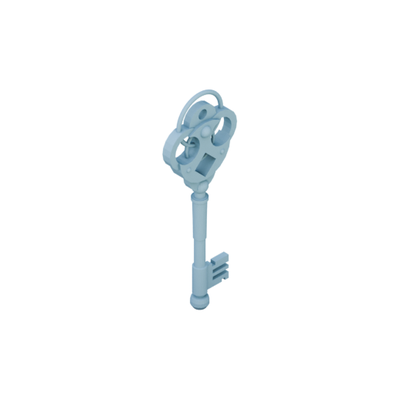} \\

\includegraphics[width=\qcwidth, valign=m]{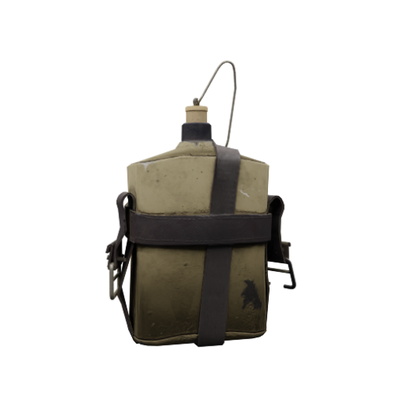} & 
\includegraphics[width=\qcwidth, valign=m]{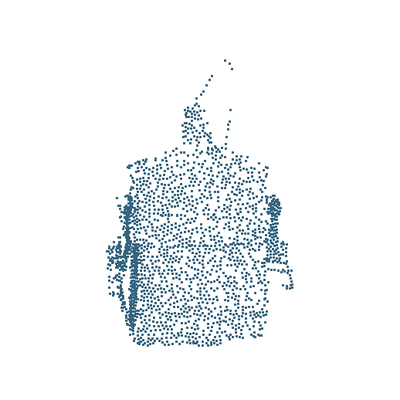} & 
\includegraphics[width=\qcwidth, valign=m]{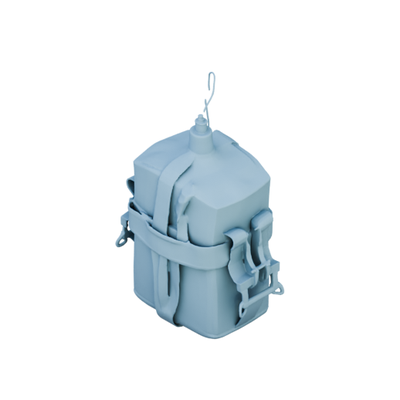} & 
\includegraphics[width=\qcwidth, valign=m]{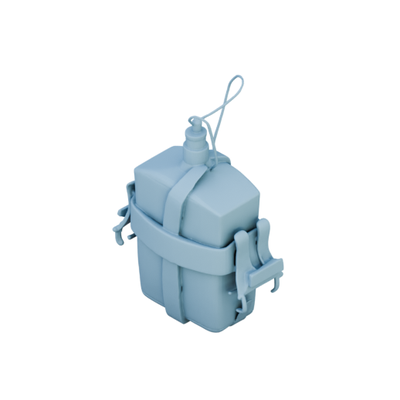} & 
\includegraphics[width=\qcwidth, valign=m]{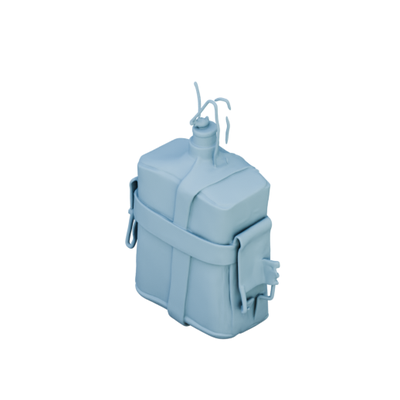} & 
\includegraphics[width=\qcwidth, valign=m]{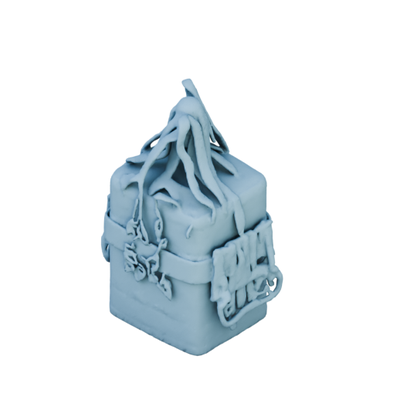} & 
\includegraphics[width=\qcwidth, valign=m]{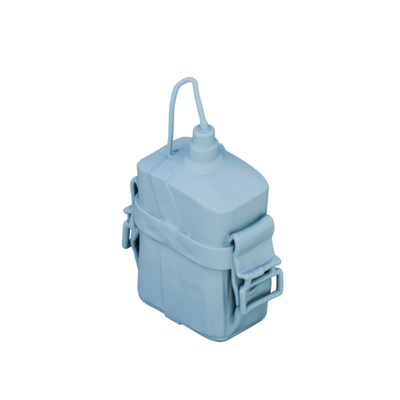} & 
\includegraphics[width=\qcwidth, valign=m]{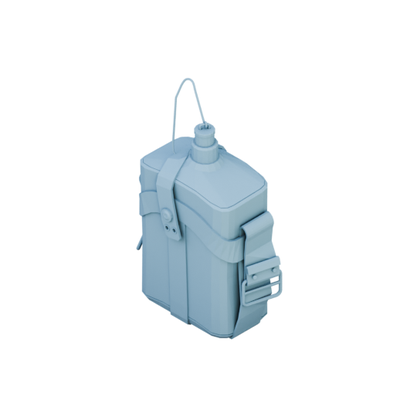} \\

\includegraphics[width=\qcwidth, valign=m]{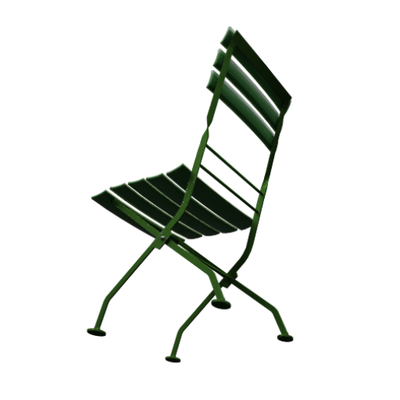} & 
\includegraphics[width=\qcwidth, valign=m]{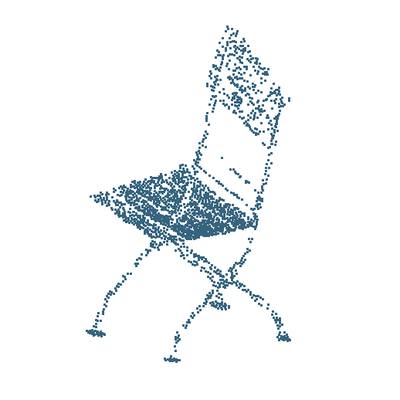} & 
\includegraphics[width=\qcwidth, valign=m]{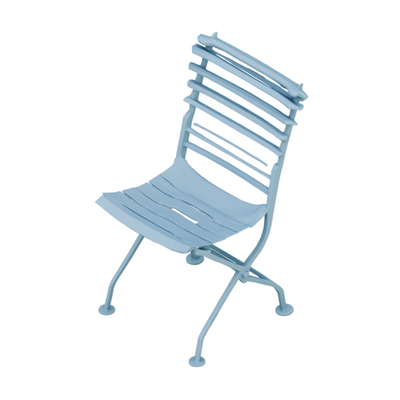} & 
\includegraphics[width=\qcwidth, valign=m]{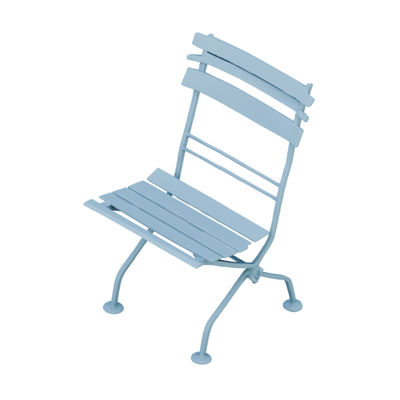} & 
\includegraphics[width=\qcwidth, valign=m]{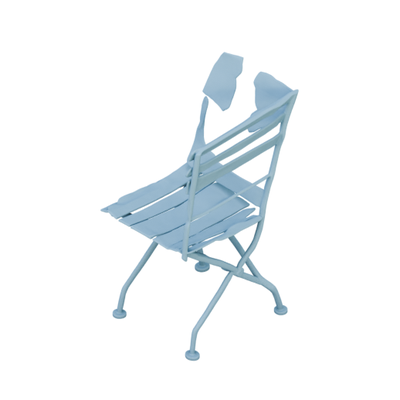} & 
\includegraphics[width=\qcwidth, valign=m]{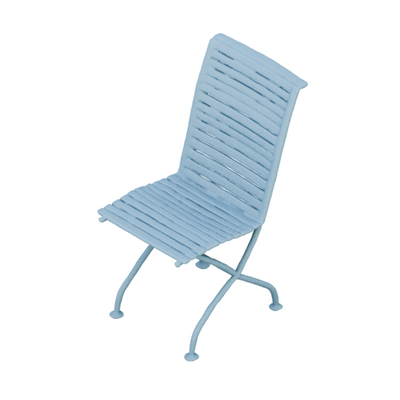} & 
\includegraphics[width=\qcwidth, valign=m]{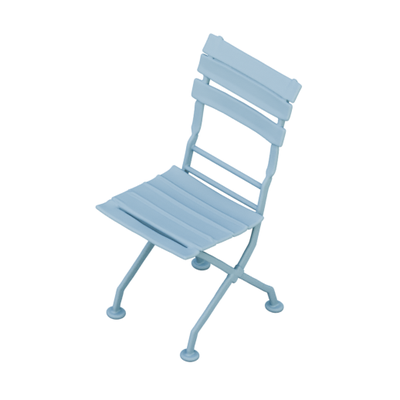} & 
\includegraphics[width=\qcwidth, valign=m]{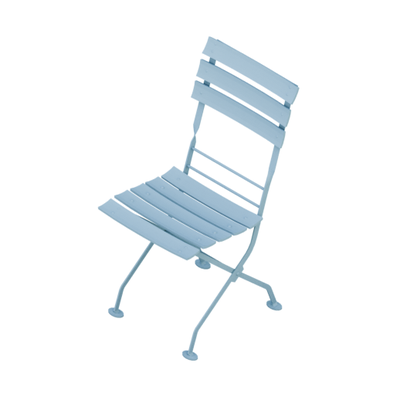} \\

\includegraphics[width=\qcwidth, valign=m]{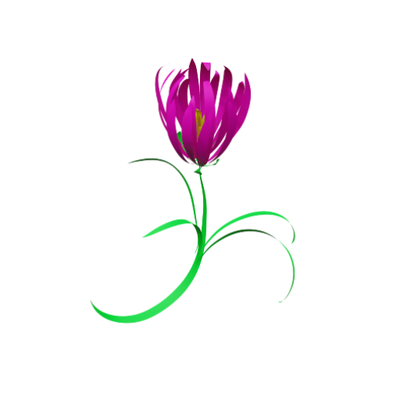} & 
\includegraphics[width=\qcwidth, valign=m]{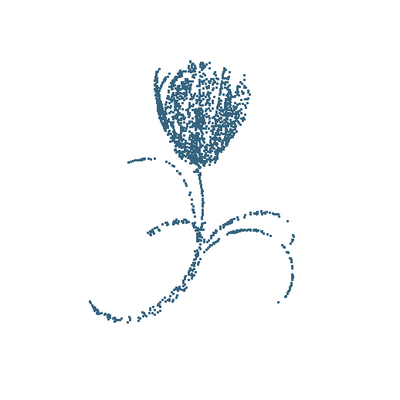} & 
\includegraphics[width=\qcwidth, valign=m]{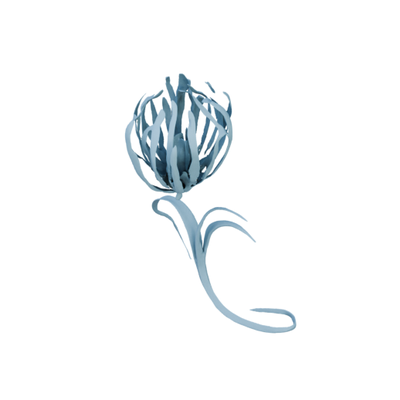} & 
\includegraphics[width=\qcwidth, valign=m]{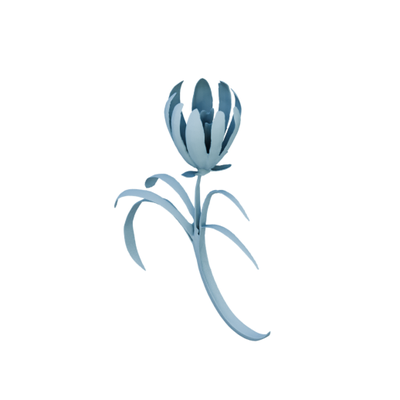} & 
\includegraphics[width=\qcwidth, valign=m]{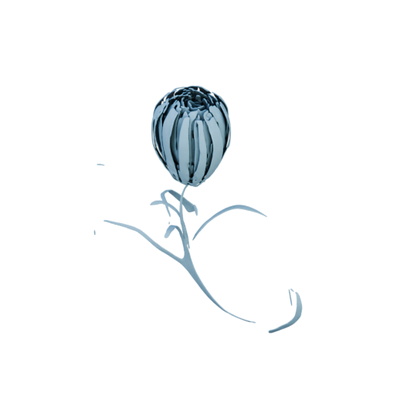} & 
\includegraphics[width=\qcwidth, valign=m]{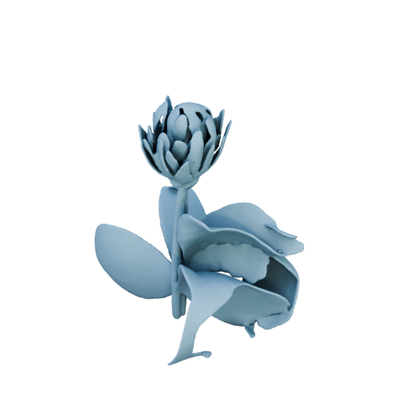} & 
\includegraphics[width=\qcwidth, valign=m]{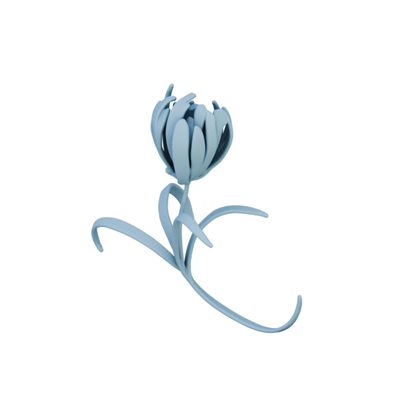} & 
\includegraphics[width=\qcwidth, valign=m]{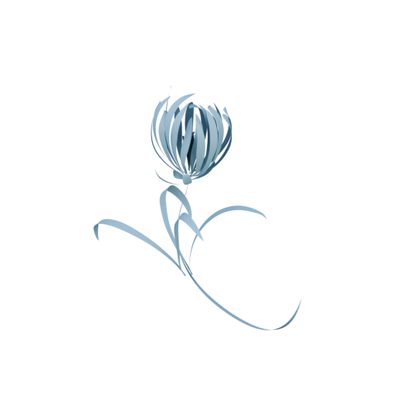} \\

\includegraphics[width=\qcwidth, valign=m]{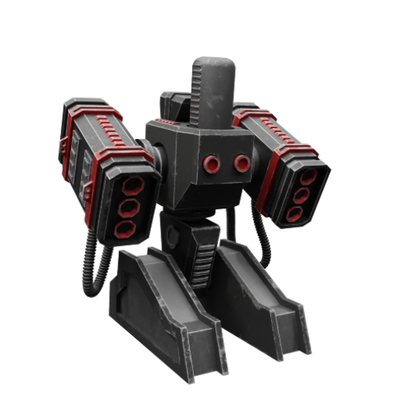} & 
\includegraphics[width=\qcwidth, valign=m]{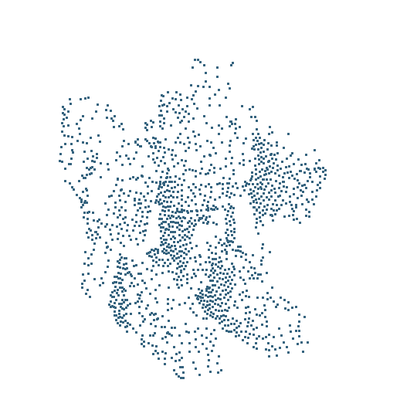} & 
\includegraphics[width=\qcwidth, valign=m]{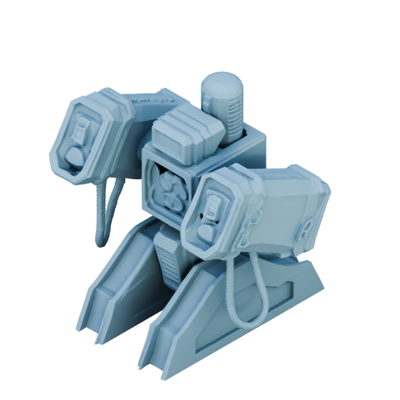} & 
\includegraphics[width=\qcwidth, valign=m]{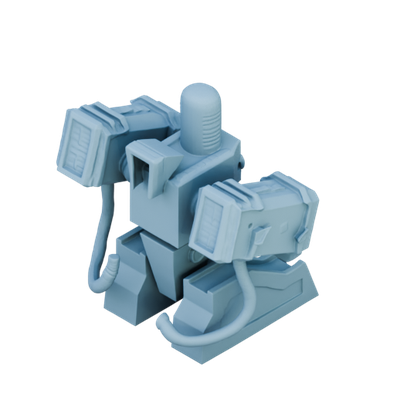} & 
\includegraphics[width=\qcwidth, valign=m]{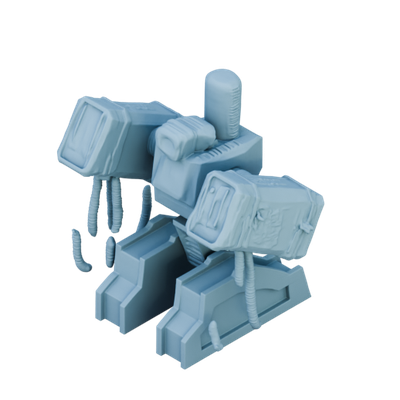} & 
\includegraphics[width=\qcwidth, valign=m]{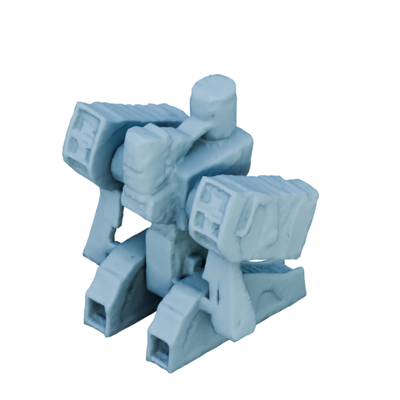} & 
\includegraphics[width=\qcwidth, valign=m]{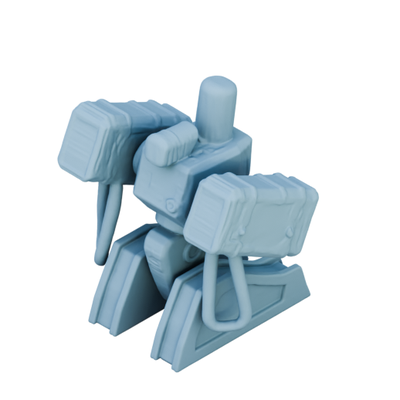} & 
\includegraphics[width=\qcwidth, valign=m]{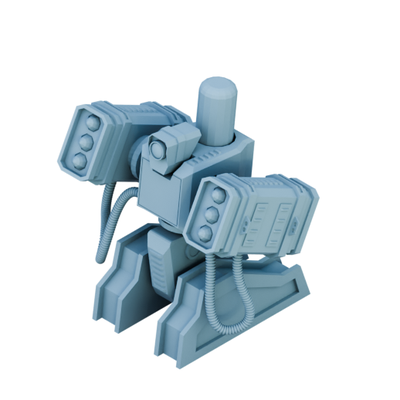} \\

  \end{tabular}
}
  
    \caption{Single-view input results}
    \label{fig:supp_toys4k_sv}
  \end{subfigure}

  \vspace{2mm}

  \begin{subfigure}{\linewidth}

  \centering
\resizebox{\linewidth}{!} {
\begin{tabular}{@{}cc|cccc|c@{}}

\scalebox{0.5}{Input View} &
\scalebox{0.5}{Input Points} & 
\scalebox{0.5}{Amodal3R} & 
\scalebox{0.5}{Hy3D-Omni} & 
\scalebox{0.5}{ShapeR} & 
\scalebox{0.5}{\textbf{Ours}} & 
\scalebox{0.5}{Ground Truth} \\ 

\includegraphics[width=\mvwidth, valign=m]{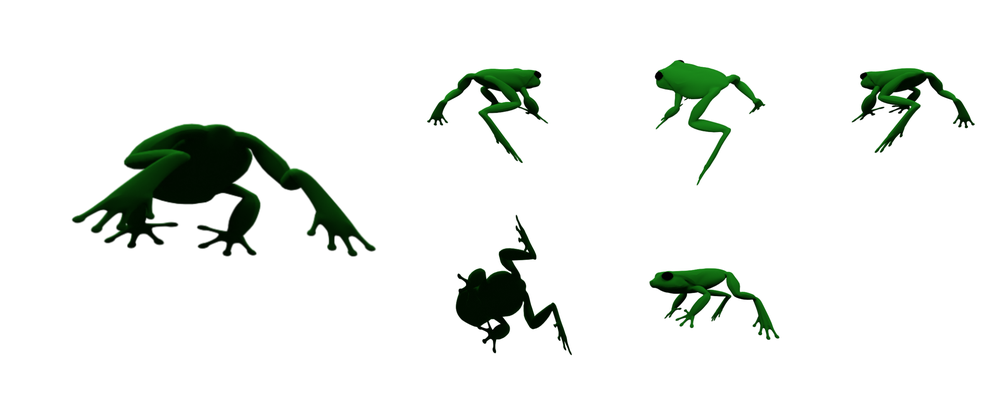} & 
\includegraphics[width=\qcwidth, valign=m]{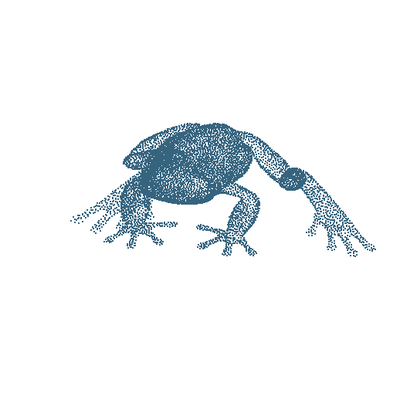} & 
\includegraphics[width=\qcwidth, valign=m]{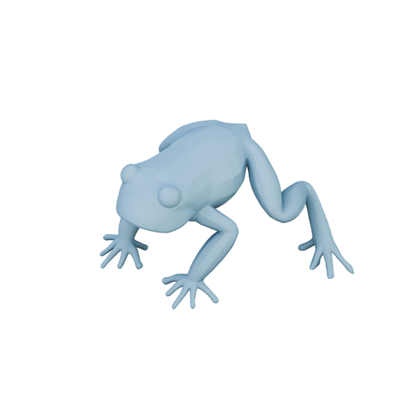} & 
\includegraphics[width=\qcwidth, valign=m]{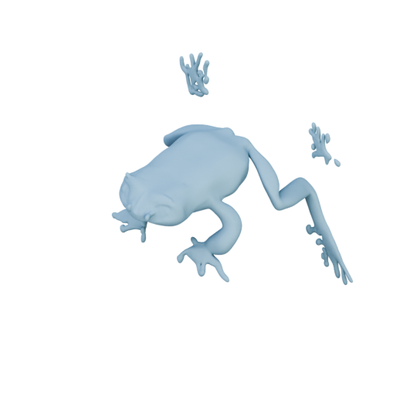} & 
\includegraphics[width=\qcwidth, valign=m]{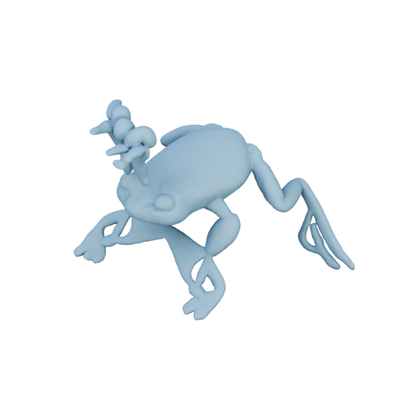} & 
\includegraphics[width=\qcwidth, valign=m]{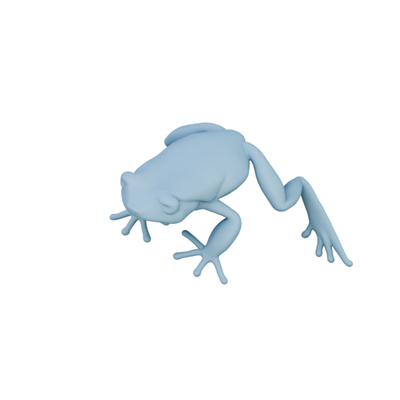} & 
\includegraphics[width=\qcwidth, valign=m]{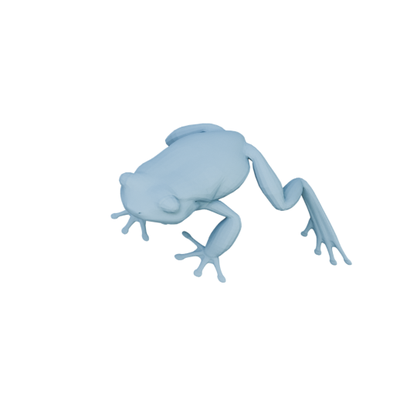} \\

\includegraphics[width=\mvwidth, valign=m]{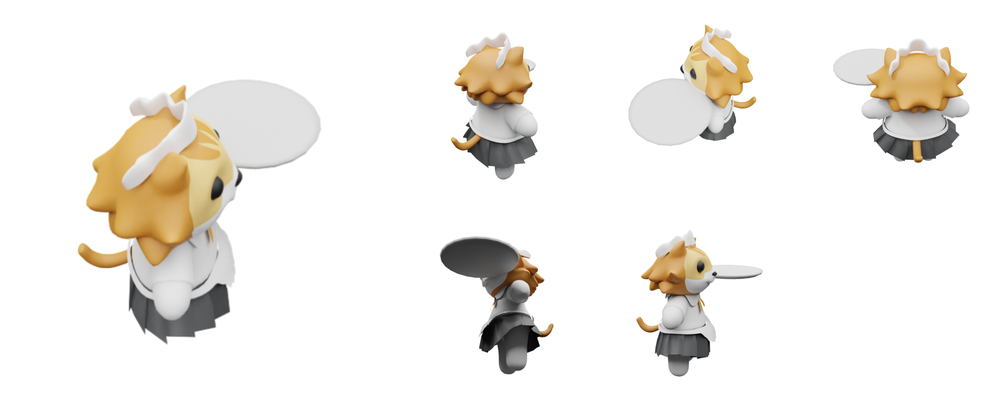} & 
\includegraphics[width=\qcwidth, valign=m]{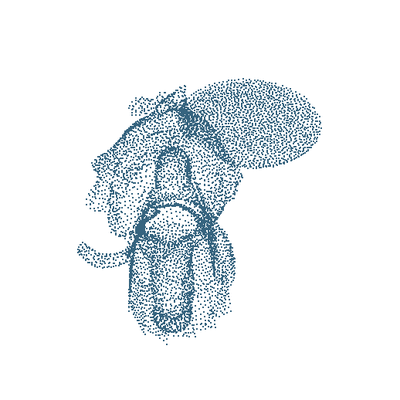} & 
\includegraphics[width=\qcwidth, valign=m]{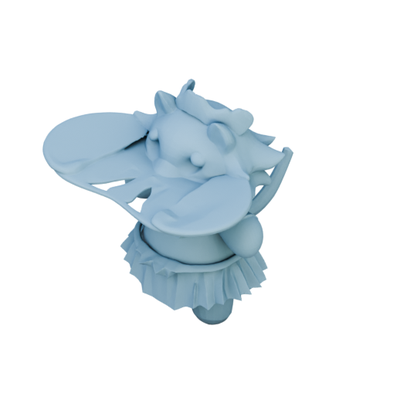} & 
\includegraphics[width=\qcwidth, valign=m]{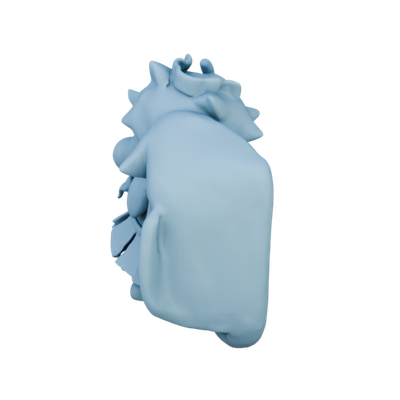} & 
\includegraphics[width=\qcwidth, valign=m]{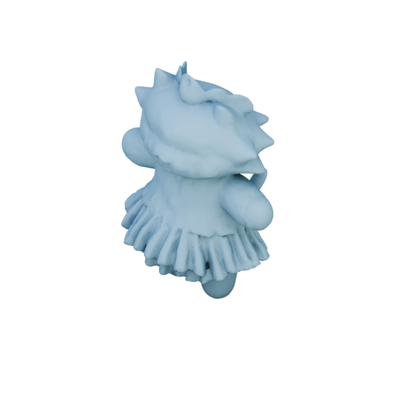} & 
\includegraphics[width=\qcwidth, valign=m]{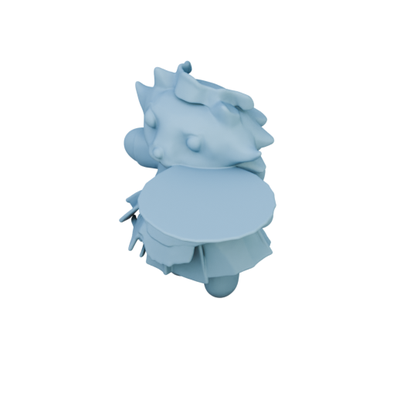} & 
\includegraphics[width=\qcwidth, valign=m]{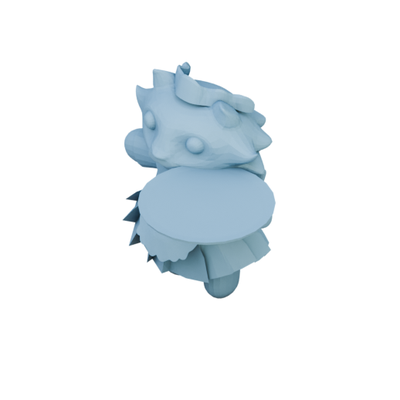} \\

\includegraphics[width=\mvwidth, valign=m]{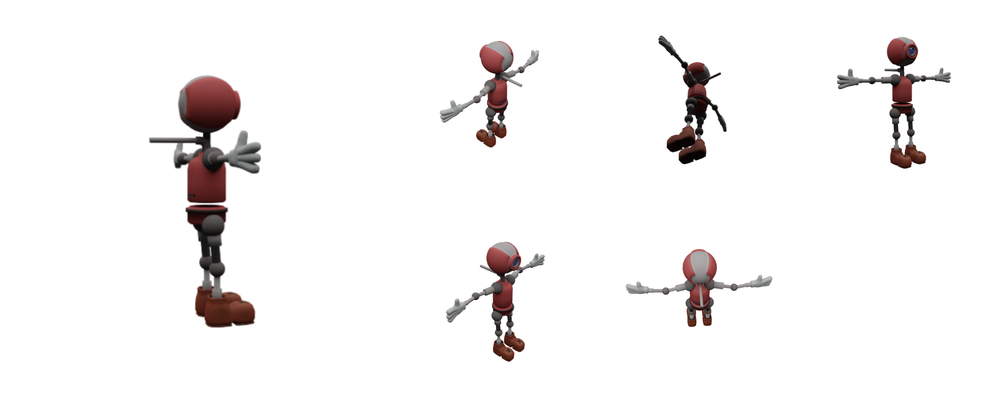} & 
\includegraphics[width=\qcwidth, valign=m]{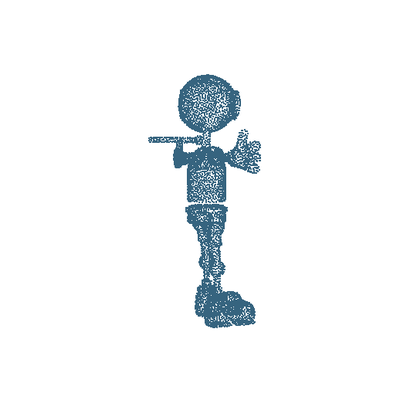} & 
\includegraphics[width=\qcwidth, valign=m]{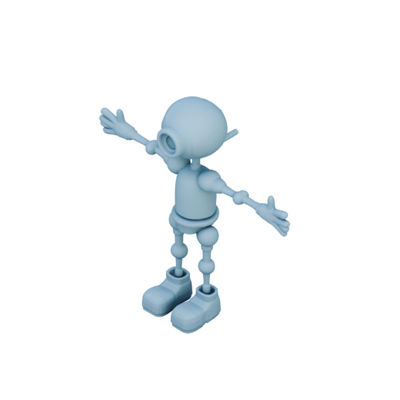} & 
\includegraphics[width=\qcwidth, valign=m]{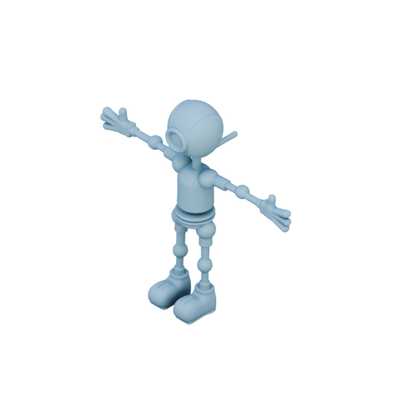} & 
\includegraphics[width=\qcwidth, valign=m]{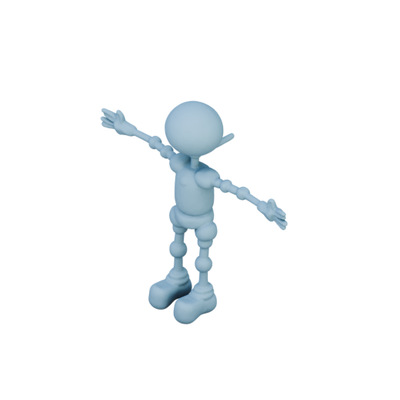} & 
\includegraphics[width=\qcwidth, valign=m]{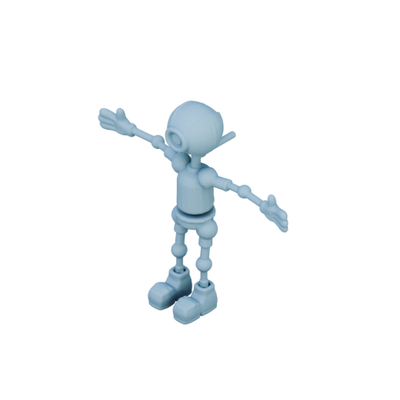} & 
\includegraphics[width=\qcwidth, valign=m]{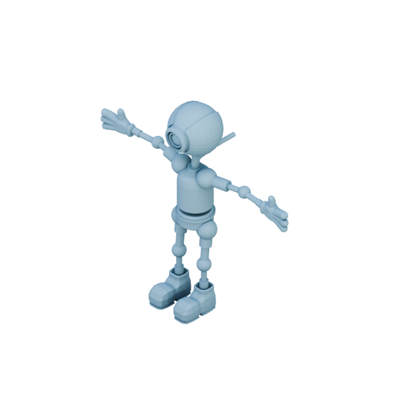} \\

\includegraphics[width=\mvwidth, valign=m]{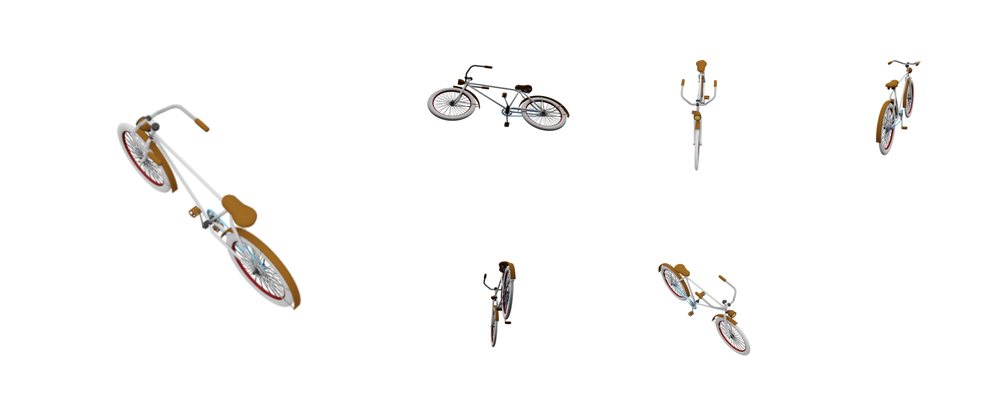} &
\includegraphics[width=\qcwidth, valign=m]{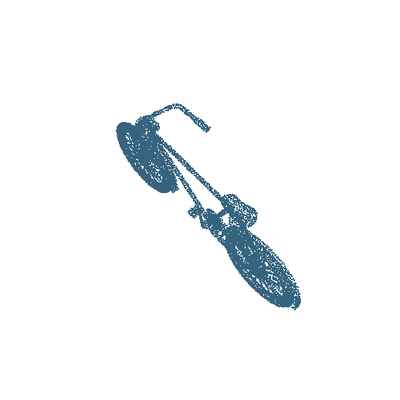} & 
\includegraphics[width=\qcwidth, valign=m]{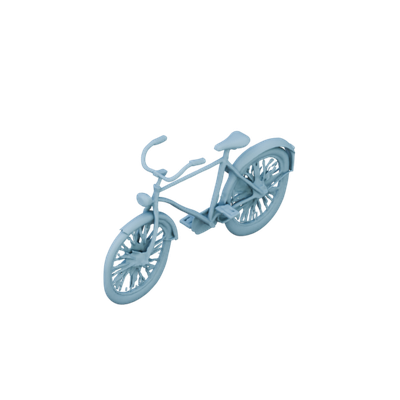} & 
\includegraphics[width=\qcwidth, valign=m]{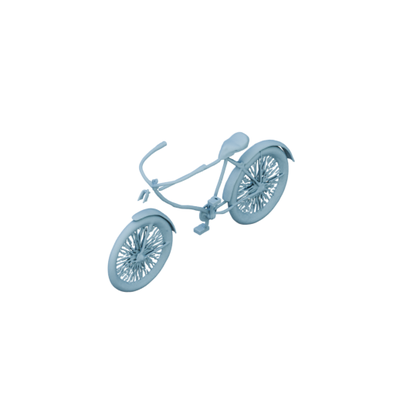} & 
\includegraphics[width=\qcwidth, valign=m]{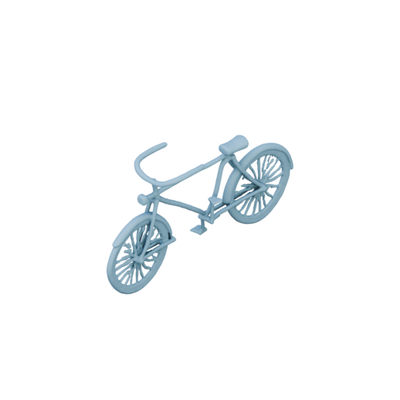} & 
\includegraphics[width=\qcwidth, valign=m]{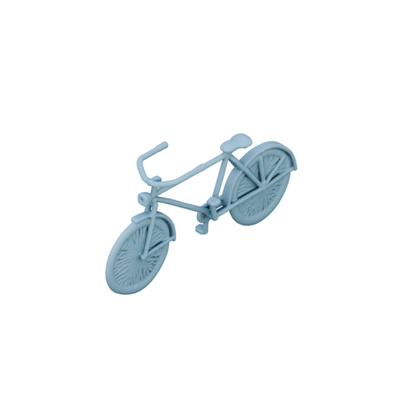} & 
\includegraphics[width=\qcwidth, valign=m]{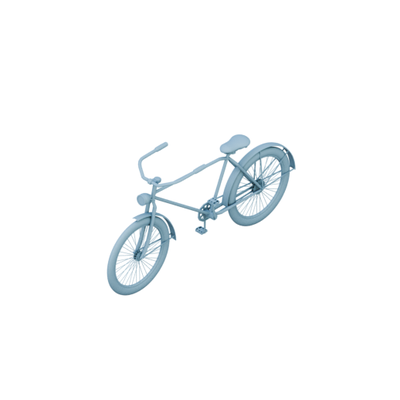} \\

\includegraphics[width=\mvwidth, valign=m]{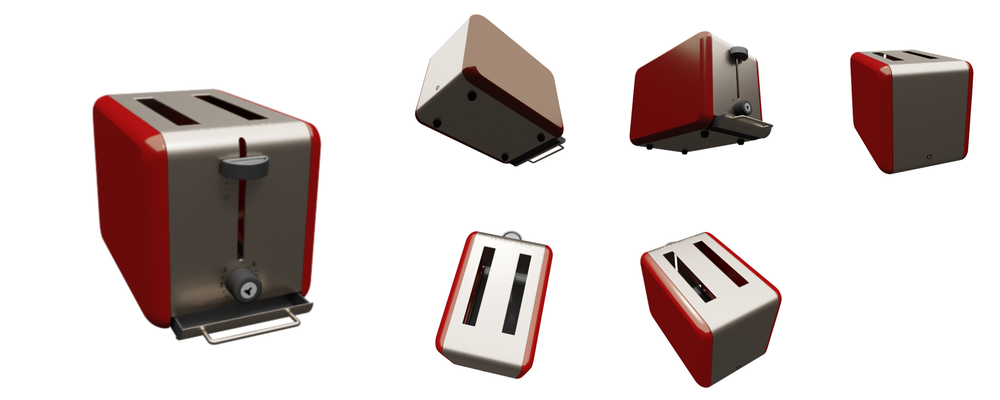} & 
\includegraphics[width=\qcwidth, valign=m]{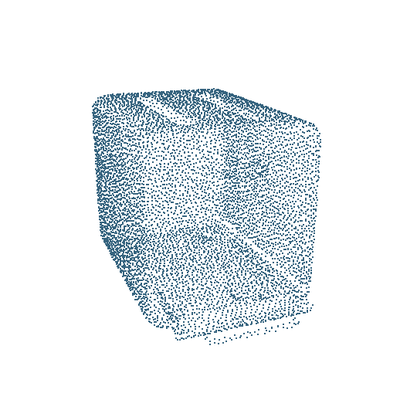} & 
\includegraphics[width=\qcwidth, valign=m]{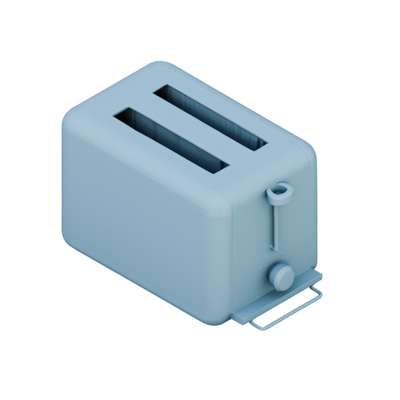} & 
\includegraphics[width=\qcwidth, valign=m]{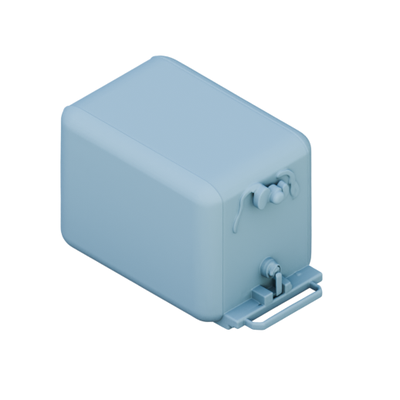} & 
\includegraphics[width=\qcwidth, valign=m]{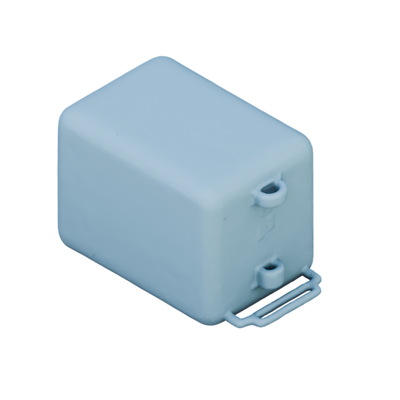} & 
\includegraphics[width=\qcwidth, valign=m]{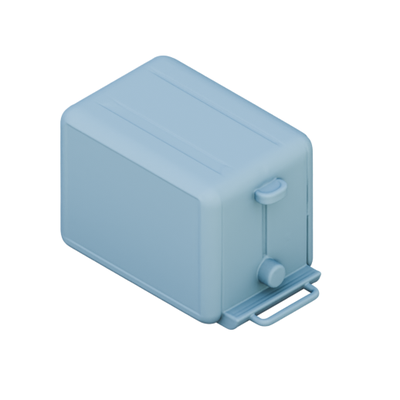} & 
\includegraphics[width=\qcwidth, valign=m]{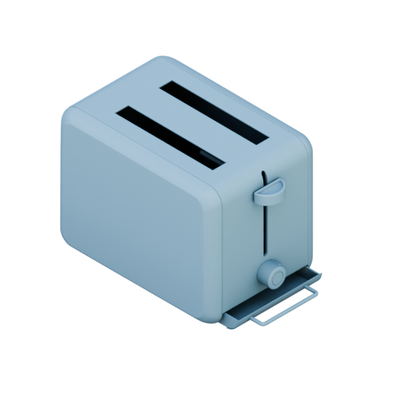} \\

  \end{tabular}
}

    \caption{Multi-view input results}
    \label{fig:supp_toys4k_mv}
  \end{subfigure}

  \caption{\textbf{Qualitative Comparisons on Toys4k}: Single- and multi-view completion results against baselines without occlusions.}
  \label{fig:supp_toys4k}
\end{figure*}
\section{Limitations \& Future Work}\label{supp:limit}

While {\ourmodel} demonstrates strong performance in occlusion handling and downstream applications, there remain several avenues for further improvement that we plan to explore in future work.

\begin{itemize}
    \item \textbf{Camera and view robustness:} Currently, our model is trained with fixed camera distances and intrinsics, which requires the object to be fully in view. Future work could incorporate more diverse camera positioning and image augmentations or real-world captured datasets to improve robustness to camera viewpoints.
    
    \item \textbf{Surface quality and fine details:} While our approach produces high-quality reconstructions overall, extremely thin structures can sometimes be disconnected and fine surface details may be smoothed or thicker than in reality. Leveraging more powerful base models or image encoders could enable better reconstruction of finer geometry.
    
    \item \textbf{Robustness to noise:} Our current model is trained to follow geometric cues faithfully and has not yet been explicitly trained to handle noisy input. Extending the model to handle different noise characteristics is an important direction for more general applicability. Examples of such characteristics include view misalignment in 3D reconstruction models \cite{wang2025vggt, keetha2026mapanything}, irregular surfaces and outliers seen in Gaussian splats \cite{kerbl2023gsplat} or oversmoothed surfaces, perspective-induced distortions, and artifacts near depth discontinuities observed in monocular depth estimation \cite{lin2026depthanything}. 
    
    \item \textbf{Handling diverse point characteristics:} Our method is trained on partial point clouds downsampled to 8192 points using Furthest Point Sampling \cite{qi2017fps}, with padding for smaller inputs. While this allows handling moderately sparse inputs, the model is not explicitly trained for very sparse or sharp-edge points (i.e., from Structure-from-Motion \cite{schonberger2016sfm}). Adapting the model to handle a wider range of point distributions and characteristics would further broaden its applicability to more complex real-world datasets.

    \item \textbf{Texture completion:} {\ourmodel} focuses solely on shape generation and does not predict textures, which could be explored in future work.

\end{itemize}

Overall, these limitations highlight exciting opportunities to further enhance {\ourmodel}, extending its applicability and robustness to a wider range of challenging scenarios.

\end{document}